\pdfoutput=1
\documentclass[11pt]{article}
\usepackage[letterpaper, left=1.2truein, right=1.2truein, top = 1.2truein,bottom = 1.2truein]{geometry}

\usepackage{amsmath, amsfonts, amssymb, amsthm, bm, graphicx, mathtools, enumerate, multirow, pifont, algorithm, algpseudocode, microtype, subfigure, booktabs, url}
\usepackage{hyperref}       
\usepackage{url}            %
\usepackage{booktabs}       %
\usepackage{nicefrac}       %
\usepackage{microtype}      %

\usepackage[round]{natbib}

\usepackage{amsmath, amsfonts, amssymb, amsthm, bm, graphicx, mathtools, enumerate, multirow, pifont, algorithm, algpseudocode}
\theoremstyle{plain} %
\newtheorem{theorem}{Theorem}[section]

\newtheorem{lemma}[theorem]{Lemma}
\newtheorem{corollary}[theorem]{Corollary}
\theoremstyle{definition}

\newtheorem{assumption}[theorem]{Assumption}
\theoremstyle{remark}
\newtheorem{remark}[theorem]{Remark}
\newtheorem{property}{Property}

\newcommand{\SAmeasure}{\nu}

\newcommand{\Bellopt}{\mathcal{T}^\pi}
\newcommand{\Bellhat}{\hat{\mathcal{T}}^{\pi}}
\newcommand{\SAspace}{\mathcal{S\times A}}
\newcommand{\Fspace}{\Theta} %

\newcommand{\Expectbehavior}{\mathbb{E}_{b_\mu}}

\newcommand{\energyone}{\mathcal{E}}
\newcommand{\distL}{\mathcal{L}} 
\newcommand{\energyonebar}{\bar{\mathcal{E}}}
\newcommand{\energyonebarhat}{\hat{\bar{\mathcal{E}}}}
\newcommand{\Upsilonhat}{\Upsilon_{\hat{\theta}}} %
\newcommand{\sumsa}{\sum_{s,a}}
\newcommand{\probsa}{b_\mu(s,a)}
\newcommand{\probsahat}{\hat{b}_\mu(s,a)}
\newcommand{\probsahatone}{\hat{b}_\mu^{(1)}(s,a)}
\newcommand{\probsahattwo}{\hat{b}_\mu^{(2)}(s,a)}
\newcommand{\Saprime}{S_{\alpha}'}

\newcommand{\Aaprime}{A_{\alpha}'}

\newcommand{\Nsa}{N(s,a)}
\newcommand{\sumNsa}{\sum_{i=1}^{\Nsa}} 
\newcommand{\Expect}{\mathbb{E}} 

\newcommand{\supf}{\sup_{\theta \in \Theta}} %
\newcommand{\Upsilonstar}{\Upsilon_{\theta_*}} %
\newcommand{\supsa}{\sup_{s,a}}

\newcommand{\sumij}{\sum_{i\neq j}}
\newcommand{\sumi}{\sum_{i=1}^{N}}
\newcommand{\Yfirst}{Y_{t}^{(1)}}
\newcommand{\Ysecond}{Y_{t}^{(2)}}

\newcommand{\PROB}{\mathbb{P}} 
\newcommand{\probvec}{\mathbf{p}} 
\newcommand{\probvechat}{\hat{\mathbf{p}}} 
\newcommand{\probvechatone}{\hat{\mathbf{p}}_{(1)}} 
\newcommand{\probvechattwo}{\hat{\mathbf{p}}_{(2)}} 

\newcommand{\diamTheta}{{\rm diam}(\Theta;\|\cdot\|)}

\newcommand{\psiN}{\psi_2(\mathbf{N})}
\newcommand{\PROBN}{\PROB^{(\mathbf{N})}}
\newcommand{\ExpectN}{\mathbb{E}^{(\mathbf{N})}}
\newcommand{\Xone}{X_{\theta_1}} %
\newcommand{\Xtwo}{X_{\theta_2}} %
\newcommand{\sumijNsa}{\sum_{i\neq j}^{N(s,a)}}

\newcommand{\Wifzero}{W_{i}^{\theta_0}} %
\newcommand{\Wifone}{W_{i}^{\theta_1}} %
\newcommand{\Wiftwo}{W_{i}^{\theta_2}} %
\newcommand{\Wijfone}{W_{ij}^{\theta_1}} %
\newcommand{\Wijftwo}{W_{ij}^{\theta_2}} %
\newcommand{\Wijfzero}{W_{ij}^{\theta_0}} %

\newcommand{\Wtildeonetwofone}{\tilde{W}_{12}^{\theta_1}} %
\newcommand{\Wtildeonetwoftwo}{\tilde{W}_{12}^{\theta_2}} %
\newcommand{\Wtildeijfone}{\tilde{W}_{ij}^{\theta_1}} %
\newcommand{\Wtildeijftwo}{\tilde{W}_{ij}^{\theta_2}} %
\newcommand{\Wiifone}{W_{ii}^{\theta_1}} %
\newcommand{\Wiiftwo}{W_{ii}^{\theta_2}} %
\newcommand{\Wiifzero}{W_{ii}^{\theta_0}} %
\newcommand{\Xfzero}{X_{\theta_0}} %
\newcommand{\Rsapsisup}{\supsa\|R(s,a)\|_{\psi_2}}
\newcommand{\Rsapsisupsq}{\big(\supsa\|R(s,a)\|_{\psi_2}\big)^2}
\newcommand{\Zsasupsq}{\big( \supsa \Expect\|Z(s,a;\theta_0)\| \big)^2 }  %
\newcommand{\Zsasuptheta}{\supsa \Expect\|Z(s,a;\theta_0)\| } 
\newcommand{\Zsasupsqtheta}{\big( \supsa \Expect\|Z(s,a;\theta_0)\| \big)^2 }

\newcommand{\Bellhatmulti}{(\hat{\mathcal{T}}^\pi)^m}
\newcommand{\Belloptmulti}{(\mathcal{T}^\pi)^m}
\newcommand{\thetamstar}{\theta_{*}^{(m)}}
\newcommand{\thetamhat}{\hat{\theta}^{(m)}}
\newcommand{\thetatilde}{\tilde{\theta}}

\newcommand{\Upsilonmstar}{\Upsilon_{\theta_{*}^{(m)}}}

\newcommand{\Upsilontilde}{\Upsilon_{\tilde{\theta}}}
\newcommand{\Fmhat}{\hat{F}_m}
\newcommand{\suptheta}{\sup_{\theta\in\Theta}}
\newcommand{\GammaNm}{\Gamma_{N,m}}
\newcommand{\Upsilonzero}{\Upsilon_{\theta_0}}
\newcommand{\probabsnorm}{\sumsa \bigg| b_{\mu}(s,a) - \hat{b}_{\mu}(s,a) \bigg|} 
\newcommand{\bellmanestimationerror}{\suptheta \bigg| \Fmhat(\theta)- F_m(\theta) \bigg|} 
\newcommand{\sumstep}{\sum_{t=0}^{m-1}}
\newcommand{\energyzero}{\mathcal{E}_{\theta_0}}
\newcommand{\energytheta}{\bar{\mathcal{E}}_{\theta}}
\newcommand{\gammasum}{\sum_{t=0}^{m-1}\gamma^t}

\newcommand{\probhatmstep}{\hat{p}_{m}}
\newcommand{\probhatbootstrap}{\hat{p}_{m}^{(B)}}
\newcommand{\Bootstrapopt}{\mathcal{B}_{m}}
\newcommand{\FBhat}{\hat{F}_{m}^{(B)}}
\newcommand{\thetabootstrap}{\hat{\theta}_{m}^{(B)}}
\newcommand{\Gammabootstrap}{\tilde{\Gamma}_{B,m}}
\newcommand{\psitwotilde}{\tilde{\psi}_2}
\newcommand{\sumn}{\sum_{i=1}^{n}}
\newcommand{\bigO}{\tilde{O}_p}

\newcommand{\cmark}{\ding{51}}
\newcommand{\xmark}{\ding{55}}

\newcommand{\distancemeasure}{\eta}
\newcommand{\distancemeasurebar}{\bar{\eta}}

\newcommand{\Expectationtilde}{\tilde{\mathbb{E}}} 

\newcommand{\Omegasubsetstageone}{\Omega_\SAspace^{(\epsilon)}}

\newcommand{\Wasserbar}{\overline{\mathbb{W}}_{1}} 
\newcommand{\Wasserone}{\mathbb{W}_1} 
\newcommand{\Wassersup}{\mathbb{W}_{1,\infty}} 

\newcommand{\expectationbounder}{\Wassersup}  
\newcommand{\expectationboundergeneralized}{\tilde{\eta}}  
\newcommand{\metricentropyexpectatonbounderFspace}{\int_{0}^{\infty} \sqrt{\log \mathcal{N}(\Theta, \Wassersup, t)}\mathrm{d}t}
\newcommand{\metricentropyexpectatonbounderTheta}{\int_{0}^{\infty} \sqrt{\log \mathcal{N}(\Theta,\Wassersup, t)}\mathrm{d}t}
\newcommand{\diamF}{{\rm diam}(\Theta;\Wassersup)}

\newcommand{\Cenvmulti}{C_{\rm env}^{(m)}(\Theta)} 
\newcommand{\Benvmulti}{B_{\rm env}^{(m)}(\Theta)}

\newcommand{\That}{\hat{\mathcal{T}}} 
\newcommand{\Thatseq}{\That_{1:m}^\pi} 
\newcommand{\supdelta}{\sup_{\theta:\|\theta - \thetamstar\|\leq\delta}} 
\newcommand{\Omegazero}{\Omega_0} 
\newcommand{\Expectationzero}{\Expect_{0}} 

\newcommand{\Xthetaone}{X_{\theta}^{(1)}} 
\newcommand{\Xthetatwo}{X_{\theta}^{(2)}} 
\newcommand{\Xthetathree}{X_{\theta}^{(3)}} 
\newcommand{\Athetaone}{A_{\theta}^{(1)}} 
\newcommand{\Athetatwo}{A_{\theta}^{(2)}} 
\newcommand{\Athetathree}{A_{\theta}^{(3)}} 

\newcommand{\Xthetaonezero}{X_{\theta_0}^{(1)}} 
\newcommand{\Athetaonezero}{A_{\theta_0}^{(1)}}

\newcommand{\Fhat}{\hat{F}} 
\newcommand{\sumiandj}{\sum_{i,j=1}^{\Nsa}} 
\newcommand{\thetastar}{{\theta_*}} 
\newcommand{\thetaone}{{\theta_1}} 
\newcommand{\thetatwo}{{\theta_2}} 
\newcommand{\thetahat}{{\hat{\theta}}} 

\newcommand{\supdeltawassersup}{{\sup_{\theta\in\Theta^\delta}}}

\begin{document}

\title{Distributional Off-policy Evaluation with Bellman Residual Minimization}
\author{Sungee Hong$^1$ \and Zhengling Qi$^2$ \and Raymond K. W. Wong$^1$  }
\date{%
	$^1$Texas A\&M University\\%
	$^2$George Washington University\\[2ex]%
}

\maketitle

\begin{abstract}
	We study distributional off-policy evaluation (OPE),
	of which the goal is to learn the distribution of the return for a target policy using offline data generated by a different policy. The theoretical foundation of many existing work relies on the supremum-extended statistical distances such as supremum-Wasserstein distance, which are hard to estimate. In contrast, we study the more manageable expectation-extended statistical distances and provide a novel theoretical justification on their validity for learning the return distribution. Based on this attractive property, we propose a new method called Energy Bellman Residual Minimizer (EBRM) for distributional OPE. We provide corresponding in-depth theoretical analyses. We establish a finite-sample error bound for the EBRM estimator under the realizability assumption. Furthermore, we introduce a variant of our method based on a multi-step extension which improves the error bound for non-realizable settings. Notably, unlike prior distributional OPE methods, the theoretical guarantees of our method do not require the completeness assumption.
\end{abstract}

\section{INTRODUCTION} \label{Introduction}

In reinforcement learning (RL), the cumulative sum of (discounted) reward, also known as the return, is a crucial quantity for evaluating the performance of a target policy. While traditional RL focus on the expectation of the return, distributional RL (DRL) relies on the whole distribution.
Since \cite{bellemare2017distributional} first introduced a DRL algorithm
and showed that it outperformed Deep-Q Network \citep{mnih2015human} (based on traditional RL) in Atari games, DRL has gained significant interest and has already been successfully applied in a variety of areas \citep[e.g.,][]{bodnar2019quantile, bellemare2020autonomous, wurman2022outracing, fawzi2022discovering}.

In this paper, we consider the problem of off-policy evaluation (OPE) in DRL under infinite-horizon settings.
The underlying evaluation task is based on offline data generated by a policy that is possibly different from the target policy.
In contrast to traditional OPE, distributional OPE aims to learn the distribution of the return instead of merely its expectation.
Despite its challenging nature,
the whole return distribution offers much richer information than its expectation
and is useful in many ways.
\cite{rowland2023statistical} and \cite{lyle2019comparative} demonstrated that their estimation of distribution can outperform traditional RL even in expectation-related tasks (e.g., estimation and maximization of expected return) under appropriate conditions. Furthermore, by taking into account different distributional attributes, one can study and build risk-sensitive policies \citep[e.g.,][]{dabney2018implicit, ma2021conservative}, and perform risk assessment \citep{huang2022off}. Several recent theoretical studies have been conducted on risk-sensitive distributional learning \citep[e.g.,][]{keramati2020being, liang2024bridging, chen2024provable}.

\subsection{Expectation-extension For Distributional OPE}

Early DRL methods \citep[e.g.,][]{bellemare2017distributional, dabney2018distributional, nguyen2021distributional}
are motivated by
\textit{supremum-extended} distances (see \eqref{supremum_distance} below) due to its contraction property.
However, the supremum-extended distance is based on taking supremum over all state-action pairs, and hence difficult to be controlled well based on data, especially for non-tabular settings %
that have
infinite (and possibly uncountable) number of state-action pairs.
So, in this paper, we resort to the weaker \textit{expectation-extended} distances (see \eqref{expectation_distance}), which allows us to form the objective function in a more tractable way.
As a key result, we provide theoretical justifications of how these expectation-extended objective functions can lead to low inaccuracy (See Theorem \ref{Fundamental_Realizable} of Section \ref{off_policy_evaluation_based_on_multistep}) for OPE.
Our result can also be useful for understanding the practical success of many existing methods, especially in non-tabular settings. Based on expectation-extended distance, we have developed a new method for distributional OPE based on energy distance, which is closely related to maximum mean discrepancy (MMD). We elaborate some key differences between our work and a prior work called MMDRL \citep{nguyen2021distributional} in Section \ref{estimated_bellman_residual}, and demonstrate the superior numerical performance of our method in Section \ref{Simulations}.

\subsection{Statistical Error Bound} \label{statisticalerrorbound_methods}

A key feature of our work is the in-depth analysis of finite-sample error bounds (Sections \ref{Realizable_Scenario} and \ref{Nonrealizable_Scenario}).
To date, there are very few theoretical statistical analysis for infinite-horizontal setting.
Several early papers of DRL \citep[e.g.,][]{bellemare2017distributional, dabney2018distributional} only proved contraction of the (true) distributional Bellman operator with respect to different distances.
\cite{rowland2019statistics} and \cite{marthe2024beyond} showed that a finite number of statistics could be learned accurately through iterations via the true Bellman operator.
However, the nature of these theories does not take into account the estimation errors based on data. 
Only a few prior works in DRL showed statistical convergence for OPE. \cite{rowland2018analysis, rowland2023analysis} established %
almost-sure convergence
of Categorical algorithm and QRTD, but no error bound is provided. \citet{ma2021conservative} derived a statistical error bound for their method, but its theory neglects the fact that practical models cannot encompass all distributions. \cite{peng2024near} has provided a statistical error bound for several models, but they require generative model.
Without these stringent modeling assumptions,
two recent works, FLE \citep{wu2023distributional} and DISCO \citep{wang2023benefits}, also provided error bound analyses, but they rely on another common but strong assumption called ``completeness,'' which will be discussed in Section \ref{sec:completeness} below.
Furthermore, most methods are restricted to tabular settings that have finite number of state-action pairs (Table \ref{Convergence_comparison} of Appendix \ref{Comparison_between_DRL_details}).
In this paper, we will conduct our theoretical analysis with both tabular and non-tabular settings under realizability assumption (i.e., no model-misspecification error), as well as non-realizability.

\subsection{Relaxing Completeness}\label{sec:completeness}

One notable advantage of our analysis (and our estimator) is
that we do not require a completeness assumption, as required in
recent works such as FLE \citep{wu2023distributional} and DISCO \citep{wang2023benefits}.
Simply put, completeness requires that the distributional Bellman operator maps any element in the assumed model back into the model.
Although the completeness assumption has been routinely assumed in many theoretical analyses in traditional RL, it is a strong assumption.
\cite{chen2019information} pointed out that completeness is not monotonic. That is, unlike realizability, enlarging the model to encompass more possible distributions may break completeness. Not only is it tricky to satisfy, but its violation can also lead to convergence toward a fixed point that is distant from the true target \cite[e.g.,][]{kolter2011fixed, tsitsiklis1996analysis, munos2008finite}. (Our simulation illustrates similar issues for DRL in Column 3 of Figure \ref{FLE_wrongconvergence} of Appendix \ref{sec:FLE_wrongconvergence}.)
Given the stringent and nontrivial nature of the completeness assumption, even in traditional RL, recent attempts have been made to relax completeness to realizability \citep[e.g.,][]{perdomo2023complete, zanette2023realizability}. We remark that a series of work based on DICE \citep[e.g.,][]{nachum2019dualdice, zhang2020gendice} do not rely on the completeness assumption, but they are developed for traditional RL.
Table \ref{Comparison_Methods_summary} shows the comparison of important theoretical aspects among different distributional OPE methods.

In Section \ref{Realizable_Scenario}, we analyze our estimator under realizability (but no completeness assumption). In Section \ref{Nonrealizable_Scenario},
we also extent our method to non-realizability settings and propose multi-step extensions to alleviate
the issues caused by misspecification.

\subsection{Summary }

We summarize our contributions as follows. (1) We provide a theoretical foundation for the application of the expectation-extended distance for distributional OPE. See Section \ref{off_policy_evaluation_based_on_multistep}. 
(2) We develop a novel distributional OPE method called EBRM, together with its finite-sample error bound. See Section \ref{Realizable_Scenario}. (3) We further develop a multi-step extension of EBRM for non-realizabile settings in Section \ref{Nonrealizable_Scenario}. 
(4) Our numerical experiments in Section \ref{Simulations} demonstrate strong performance of EBRM compared to some baseline methods.

\begin{table*}[h]
	\caption{Comparison of theoretical aspects of distributional OPE methods. 
		Column ``Without completeness'' indicates whether the corresponding work shows statistical consistency without completeness assumption. Methods without statistical consistency results are marked ``--''. More extensive comparison can be found in Tables \ref{Comparison_Methods_summary_details}--\ref{Convergence_comparison}.} 
	\label{Comparison_Methods_summary}
	\begin{center}
		\begin{tabular}{lcccc}
			\toprule
			\multicolumn{1}{c}{\bf }                    &\multicolumn{1}{c}{\bf Statistical}   &\multicolumn{1}{c}{\bf Without}          & {\bf Multi-dimensional}
			\\ 
			\multicolumn{1}{c}{\bf Method}               &\multicolumn{1}{c}{\bf error bound}   &\multicolumn{1}{c}{\bf completeness}  & {\bf reward}
			\\ \midrule
			Categorical \citep{rowland2018analysis}                      & \xmark~(with only convergence)                               & \cmark                               & \xmark  \\ %
			QRTD \citep{rowland2023analysis}                             & \xmark                               & --                                   & \xmark   \\ %
			MMDRL \citep{nguyen2021distributional}                       & \xmark                               & --                                   & \cmark   \\ %
			EDRL \citep{rowland2019statistics}                           & \xmark                               & --                                   & \xmark   \\ %
			CDE \citep{ma2021conservative}                               & \cmark                               & \cmark                                   & \xmark \\ 
			NTD \citep{peng2024near}                                     & \cmark                               & \cmark                                   & \xmark \\ 
			FLE \citep{wu2023distributional}                             & \cmark                               & \xmark                               & \cmark  \\ %
			{\bf EBRM (our method)}                                      & \cmark                               & \cmark                               & \cmark  \\ %
			\bottomrule
		\end{tabular}
	\end{center}
\end{table*}

\section{OFF-POLICY EVALUATION BASED ON BELLMAN EQUATION} \label{off_policy_evaluation_based_on_multistep}

\subsection{Background}

We consider a distributional off-policy evaluation (OPE) problem within the framework of infinite-horizon Markov Decision Process (MDP), 
which is characterized by a state space $\mathcal{S}$, a discrete action space $\mathcal{A}$, and
a transition kernel $p:\mathcal{S}\times\mathcal{A}\to \mathcal{P}(\mathbb{R}^d\times \mathcal{S})$ with $d\geq1$ and $\mathcal{P}(\mathcal{X})$ denoting the class of probability measures over a generic space $\mathcal{X}$. In other words, $p$ defines a joint distribution of a $d$-dimensional immediate reward and the next state conditioned on a state-action pair.
At each time point, an action is chosen by the agent based on the current state according to some (stochastic) policy, which is a mapping from $\mathcal{S}$ to $\mathcal{P}(\mathcal{A})$.
A trajectory generated by such an MDP can be written as $\{S^{(t)}, A^{(t)}, R^{(t+1)}\}_{t\geq0}$. The return variable is defined as $Z:=\sum_{t=1}^\infty \gamma^{t-1} R^{(t)}$ with $\gamma\in[0,1)$ being a discount factor, based on which we can evaluate the performance of some target policy $\pi$.

OPE is different from on-policy evaluation in that the data are collected using a different policy called behavior policy $b$, other than the target policy $\pi$ that we want to evaluate. Therefore, OPE naturally involves an issue of distributional shift.
Traditional OPE methods are mainly focused on estimating the expectation of the return $Z$ under the target policy $\pi$, while DRL aims to estimate the whole distribution of $Z$.
In DRL,
letting $\distL(X)$ be the probability measure of some random variable (or vector) $X$, our target is to estimate the collection of return distributions conditioned on different initial state-action pairs $(S^{(0)},A^{(0)})=(s,a)$: 
\begin{gather} \label{target_distribution}
	\Upsilon_\pi(s,a) := \distL \bigg( \sum_{t=1}^{\infty} \gamma^{t-1}R^{(t)} \bigg), \text{ where }  \nonumber \\
	(R^{(t+1)},S^{(t+1)})\sim p(\cdot|S^{(t)},A^{(t)}), \ A^{(t+1)}\sim \pi(\cdot|S^{(t+1)}), \nonumber
\end{gather}
collectively written as $\Upsilon_\pi\in\mathcal{P}(\mathbb{R}^d)^{\SAspace}$.
It is analogous to the $Q$-function in traditional RL, whose evaluation at a state-action pair $(s,a)$ is the expectation of the distribution $\Upsilon_\pi(s,a)$.
In this paper, the return variable is allowed to be multivariate (i.e., $d>1$). For distributional evaluation, the multivariate extension is meaningful as the dependence between different dimensions of return can be assessed.

Similar to most existing DRL methods, our proposal is
based on the distributional Bellman equation \citep{bellemare2017distributional}.
Define the distributional Bellman operator by
$\Bellopt: \mathcal{P}(\mathbb{R}^d)^{\SAspace}\to\mathcal{P}(\mathbb{R}^d)^{\SAspace}$
such that, for any $\Upsilon\in\mathcal{P}(\mathbb{R}^d)^{\SAspace}$ and $s,a\in\SAspace$, 
\begin{gather} 
	\big(\Bellopt \Upsilon\big)(s,a):= \nonumber \\
	\int_{\mathbb{R}^d\times \mathcal{S\times A}}  (g_{r,\gamma})_{\#} \Upsilon(s',a') \mathrm{d}\pi(a'|s')\mathrm{d}p(r,s'|s,a), \label{Bellman_operator_definition} 
\end{gather}
where $(g_{r,\gamma})_{\#}:\mathcal{P}(\mathbb{R}^d)\to \mathcal{P}(\mathbb{R}^d)$ maps the distribution of any random vector $X$ to the distribution of $r+\gamma X$.
One can show that $\Upsilon_\pi$ solves the distributional Bellman equation:
\begin{equation}
	\Bellopt \Upsilon = \Upsilon, \label{Bellman_equation}
\end{equation}
with respect to $\Upsilon$. Due to the distributional Bellman equation (\ref{Bellman_equation}), a sensible approach to finding $\Upsilon_\pi$ is based on minimizing the discrepancy between $\Bellopt\Upsilon$ and $\Upsilon$ with respect to $\Upsilon\in\mathcal{P}(\mathbb{R}^d)^{\SAspace}$, which will be called \textit{Bellman residual} hereafter.
To proceed with this approach, two important issues need to be addressed.
First, both $\Bellopt\Upsilon$ and $\Upsilon$ are collections of distributions over $\mathbb{R}^d$, based on which Bellman residual shall be quantified. 
Second, $\Bellopt$ may not be available and therefore needs to be estimated using data.
We will first focus on the quantification of Bellman residual, and defer the proposed estimator of $\Bellopt$ and the formal description of our estimator for $\Upsilon_\pi$ to Section \ref{Realizable_Scenario}.

\subsection{Existing Measures Of Bellman Residuals}
\label{Choice_of_Distance}

To quantify the discrepancy between the two sides of the distributional Bellman equation (\ref{Bellman_equation}), a common strategy is to start by selecting a statistical distance $\distancemeasure(\cdot,\cdot):\mathcal{P}(\mathbb{R}^d)\times \mathcal{P}(\mathbb{R}^d)\rightarrow [0,\infty]$,
and then define an extended-distance over $\mathcal{P}(\mathbb{R}^d)^{\SAspace}$
through combining the statistical distances over all state-action pairs. As shown in Table \ref{Comparison_Methods} in Appendix \ref{Comparison_between_DRL_details}, most existing methods \citep[e.g.,][]{bellemare2017distributional, dabney2018distributional, nguyen2021distributional} 
are theoretically based on some \textit{supremum-extended} distance $\distancemeasure_{\infty}$ defined as
\begin{align}
	\distancemeasure_{\infty}(\Upsilon_1, \Upsilon_2):= \supsa \distancemeasure\bigg\{ \Upsilon_1(s,a), \Upsilon_2(s,a) \bigg\}. \label{supremum_distance}   
\end{align}
Under various choices of $\distancemeasure$ including Wasserstein-$p$ metric with $1\leq p \leq \infty$ \citep{bellemare2017distributional, dabney2018distributional} and maximum mean discrepancy \citep{nguyen2021distributional},
it is shown that $\Bellopt$
is a contraction with respect to $\distancemeasure_{\infty}$.
More specifically, $\distancemeasure_{\infty}(\Bellopt \Upsilon_1, \Bellopt \Upsilon_2) \leq \gamma^{\beta_0} \distancemeasure_{\infty}(\Upsilon_1, \Upsilon_2)$ holds for any $\Upsilon_1,\Upsilon_2\in\mathcal{P}(\mathbb{R}^d)^{\SAspace}$, where the value of $\beta_0>0$ depends on the choice of $\distancemeasure$.
If $\distancemeasure_\infty$ is a metric, then the contractive property implies, for any $\Upsilon\in\mathcal{P}(\mathbb{R}^d)^{\SAspace}$,
\begin{align}
	\distancemeasure_{\infty}(\Upsilon, \Upsilon_\pi) &\leq \sum_{k=1}^{\infty} \distancemeasure_{\infty} \big\{ (\Bellopt)^{k-1} \Upsilon , (\Bellopt)^k\Upsilon \big\} \nonumber \\
	&\leq \frac{1}{1-\gamma^{\beta_0}} \cdot \distancemeasure_{\infty}(\Upsilon, \Bellopt \Upsilon) . \label{supremum_convergence_logic}
\end{align}
As such, minimizing Bellman residual measured by $\distancemeasure_\infty$ would be a sensible approach for finding $\Upsilon_\pi$. However, when $\SAspace$ is very large (e.g., when either state or action is continuous), it is very difficult to estimate or control $\distancemeasure_\infty$ by using offline data and also optimize with respect to $\Upsilon$.

It is more manageable to minimize Bellman residual with \textit{expectation-extended} distance defined by
\begin{gather}
	\distancemeasurebar(\Upsilon_1, \Upsilon_2) := \mathbb{E}_{(S,A)\sim b_\mu} \distancemeasure\bigg\{ \Upsilon_1(S,A), \Upsilon_2(S,A) \bigg\}, \label{expectation_distance}   
\end{gather}
with $(S,A) \sim b_\mu$.
Here $b_\mu = \mu \times b$ refers to the offline data distribution over $\mathcal{S}\times \mathcal{A}$ induced by the behavior policy $b$.
With a slight abuse of the notation, we will overload the notation $b_\mu$ with its density when there is no confusion. %

We remark that \eqref{supremum_convergence_logic} does not hold under $\distancemeasurebar$ because $\distancemeasure_\infty$ and $\distancemeasurebar$ are not necessarily equivalent for
non-tabular state-action spaces (see Appendix \ref{nontriviality_continuous_stateaction}) which necessitates development of new theoretical foundation. Theoretical analysis has already been done on expectation-based objective functions in traditional RL \citep[e.g.,][]{antos2008learning, munos2008finite}, and we hereby extend it to DRL.

\subsection{Expectation-extended Distance}
\label{sec:expectation-extended}

The fundamental question regarding the use of expectation-extended distances is:

\begin{quote}
	\textit{Does minimizing expectation-extended Bellman residual lead to recovery of $\Upsilon_\pi$?}
\end{quote}

In off-policy setting, our target distribution is associated with the target policy $\pi$, which is different from the behavior policy $b$ that generates the data $\mathcal{D}$. Due to this mismatch, the answer to the above question is not trivial. 

To proceed, we shall focus on settings where state-action pairs of interest can be well covered by $b_\mu$, as formally stated in the following assumption. Let $q^\pi(s,a|\tilde{s},\tilde{a})$ be the conditional probability density of the next state-action pair at $(s,a)$ conditioned on the current state-action pair at $(\tilde{s},\tilde{a})$, induced by the transition probability $p$ and the target policy $\pi$.

\begin{assumption} \label{RN_derivative}
	$\sup_{\tilde{s},\tilde{a}, s, a} q^\pi(s,a|\tilde{s},\tilde{a})/\probsa < \infty$. 
\end{assumption}
We note that every OPE method, regardless of traditional (non-distributional) or distributional RL, requires a form of coverage assumption. See Assumption 4.10 of \cite{wu2023distributional} and Section 5.5 of \cite{sutton2018reinforcement} for examples. A trivial example that satisfies our Assumption \ref{RN_derivative} is when the denominator and the numerator are lower bounded and upper bounded, i.e. $p_{\rm min}:=\inf_{s,a}\probsa>0$ and $p_{\rm max}:=\sup_{(s,a),(\tilde{s},\tilde{a})\in\SAspace} q^\pi(s,a|\tilde{s},\tilde{a})<\infty$.

In the following Theorem \ref{Fundamental_Realizable} (proved in Appendix \ref{Proof_Fundamental_Realizable}), we provide a solid ground for Bellman residual minimization based on expectation-extended distances.
\begin{theorem} \label{Fundamental_Realizable}
	Under Assumption \ref{RN_derivative}, if the statistical distance $\distancemeasure$ satisfies translation-invariance, scale-sensitivity, convexity, and relaxed triangular inequality defined in Appendix \ref{properties_of_distance}, then we can bound the expectation-based inaccuracy: for any $\Upsilon\in\mathcal{P}(\mathbb{R}^d)^{\SAspace}$,
	\begin{equation} \label{C_bound}
		\distancemeasurebar(\Upsilon, \Upsilon_\pi) 
		\le
		B(\gamma)\distancemeasurebar(\Upsilon, \Bellopt \Upsilon), 
	\end{equation}
	where $B(\gamma)$ does not depend on $\Upsilon$ and $B(\gamma)<\infty$ for all $0\leq \gamma<1$. The precise bound and the rate of $B(\gamma)$ (as $\gamma\rightarrow 1$) can be found in \eqref{Fundamental_Realizable_specific} and \eqref{Bgamma_rate} of the Appendix.
\end{theorem}
Inequality \eqref{C_bound} provides an analogy to Bound (\ref{supremum_convergence_logic}) for expectation-based distances, answering our prior question positively for multiple expectation-extended distances, including the examples listed in \eqref{suitable_distances} of Appendix \ref{Proof_squared_metric}.
Unlike tabular cases where $\distancemeasurebar$ and $\distancemeasure_{\infty}$ can be viewed as equivalent, this is not a trivial result when applied to general state-action spaces, including the continuous state-action space, as explained in Appendix \ref{nontriviality_continuous_stateaction}.
In its full generality, Theorem \ref{Fundamental_Realizable} provides a foundation for estimators (not only ours) developed under expectation-based distance for settings with \textit{general} (e.g., continuous) state-action spaces.

For solid construction and theoretical analysis, we will focus on a specific statistical distance.
In below we will construct our estimator based on
energy distance \citep{szekely2013energy}, which is defined as
\begin{gather}
	\energyone \{\distL(\mathbf{X}), \distL(\mathbf{Y}) \}:= \label{Energy_Distance} \\
	2\mathbb{E}\|\mathbf{X}-\mathbf{Y}\| - \mathbb{E}\|\mathbf{X}-\mathbf{X}'\| - \mathbb{E}\|\mathbf{Y}-\mathbf{Y}'\| . \nonumber
\end{gather}
Here, $\mathbf{X}'$ and $\mathbf{Y}'$ are independent copies of $\mathbf{X}$ and $\mathbf{Y}$ respectively,
and $\mathbf{X}, \mathbf{X}', \mathbf{Y}, \mathbf{Y}'$ are independent. Energy distance can be computed easily even when we are given with multi-dimensional return $d\geq 2$ and without closed-form density, which makes it more advantageous to other examples in \eqref{suitable_distances}.

\section{ENERGY BELLMAN RESIDUAL MINIMIZER} \label{Realizable_Scenario}

\subsection{Estimated Bellman Residual} \label{estimated_bellman_residual}

Despite the applicability of Theorem \ref{Fundamental_Realizable} to general state-action space, we mainly focus on developing an estimator for the following two specialized settings: (1) tabular case; (2) continuous state-action space with deterministic transition (i.e. each $s,a$ yields a deterministic $r,s'$). Due to space limitation, we hereafter only focus on the tabular setting, leaving the setting with the continuous state-action space in Appendix \ref{sec:continuous_deterministic_realizable}. We will present an in-depth theoretical study under both realizable and non-realizable settings in Sections \ref{theoretical_bound_realizable}, \ref{splitdata_operator}, and \ref{multistep_bellman_operator}.

Our target objective of Bellman residual minimization is
\begin{gather}
	\energyonebar(\Upsilon, \Bellopt \Upsilon) = \sumsa \probsa \cdot \energyone\big\{ \Upsilon(s,a), \Bellopt\Upsilon(s,a) \big\}  \label{Expectation_three_terms}     %
\end{gather}
with the term $\energyone \{ \Upsilon(s,a), \Bellopt\Upsilon(s,a) \}$ being defined as 
\begin{gather}
	2\mathbb{E}\|Z_{\alpha}(s,a) - Z_\beta^{(1)}(s,a)\| - \mathbb{E}\|Z_{\alpha}(s,a) - Z_{\beta}(s,a)\| \nonumber \\
	- \mathbb{E}\| Z_\alpha^{(1)}(s,a) -  Z_\beta^{(1)}(s,a)\|,  \nonumber
\end{gather}
with fixed $s,a$. Here, the four random vectors $Z_{\alpha}(s,a), Z_{\beta}(s,a)\sim \Upsilon(s,a)$ and $Z_{\alpha}^{(1)}(s,a)$, $Z_{\beta}^{(1)}(s,a)  \sim \Bellopt\Upsilon(s,a)$
are all independent. For the tabular case with offline data, we can estimate $b_{\mu}$ and the transition $p$ simply by empirical distributions. That is, given independent and identically distributed (i.i.d.) observations $\mathcal{D}=\{(s_i,a_i,r_i,s_i')\}_{i=1}^{N}$ (with $s_i,a_i\sim b_\mu$ and $r_i,s_i'\sim p(\cdots|s_i,a_i)$), we consider
\begin{gather} 
	\probsahat := \frac{\Nsa}{N} , \ \Nsa:=\sum_{i=1}^{N} \mathbf{1}\big\{ (s_i,a_i)=(s,a) \big\}, \label{probabiliy_stateaction}
\end{gather}
and use the empirical probability measure $\hat{p}(E|s,a)$ defined as follows for any measurable set $E$,
\begin{align}
	\begin{cases}
		\frac{1}{\Nsa}\sum_{i:(s_i,a_i)=(s,a)} 
		\delta_{r_i,s_i'}(E)  &\text{if } \Nsa \geq 1\\
		\delta_{\mathbf{0},s} (E) &\text{if } \Nsa = 0
	\end{cases} \label{empirical_transition}
\end{align}
where $\delta_{r,s'}$ is the Dirac measure at $(r,s')$. Based on this, we can estimate $\Bellopt$ for any $\Upsilon\in\mathcal{P}(\mathbb{R}^d)^{\SAspace}$ by the estimated transition $\hat{p}$ %
and the target policy $\pi$, i.e., replacing $p$ of \eqref{Bellman_operator_definition} with $\hat{p}$.

Denoting the conditional expectation by $\Expectationtilde(\cdots):=\Expect(\cdots|\mathcal{D})$, we can compute $\energyone \{ \Upsilon(s,a), \Bellhat \Upsilon(s,a) \}$ as 
\begin{gather}
	2\Expectationtilde\| Z_{\alpha}(s,a) - \hat{Z}_{\beta}^{(1)}(s,a) \| - \Expectationtilde\|Z_{\alpha}(s,a) - Z_{\beta}(s,a)\| \nonumber \\
	- \Expectationtilde\| \hat{Z}_{\alpha}^{(1)}(s,a) - \hat{Z}_{\beta}^{(1)}(s,a)\|, \label{Biased_Esa}
\end{gather}
where the four $Z_{\alpha}(s,a), Z_{\beta}(s,a)\sim \Upsilon_\theta(s,a)$ and $\hat{Z}_\alpha^{(1)}(s,a), \hat{Z}_\beta^{(1)}(s,a) \sim \Bellhat\Upsilon(s,a)$ are all independent conditioned on the observed data $\mathcal{D}$ that determine $\Bellhat$ via $\hat{p}$. With the above construction, we can estimate the objective function as follows.
\begin{equation}\label{Biased_EStimator_onestep_old}
	\energyonebarhat(\Upsilon, \Bellhat\Upsilon) = \sumsa \probsahat\cdot \energyone \big\{ \Upsilon(s,a) , \Bellhat \Upsilon(s,a) \big\}
\end{equation}
Now letting $\{\Upsilon_\theta : \theta\in\Fspace \}\subseteq \mathcal{P}(\mathbb{R}^d)^\SAspace$ be the hypothesis class of $\Upsilon_\pi$,
where each distribution $\Upsilon_\theta$ is indexed by an element of candidate space $\Fspace$, a special case of which is the parametric case $\Fspace\subseteq\mathbb{R}^p$. Then the proposed estimator of $\Upsilon_\pi$ is
$\Upsilon_{\hat{\theta}}$ where
\begin{equation} \label{Biased_EStimator_onestep}
	\hat{\theta} \in \arg\min_{\theta\in\Fspace} \energyonebarhat(\Upsilon_\theta, \Bellhat\Upsilon_\theta).
\end{equation}
We call our method the \textit{Energy Bellman Residual Minimizer} (EBRM) and summarize it in Algorithm \ref{EBRM_singlestep_algorithm} of Appendix \ref{sec:EBRM_singlestep_algorithm}. We will refer to the approach here as EBRM-single-step, as opposed to the multi-step extension EBRM-multi-step in Section \ref{multistep_bellman_operator}.

In passing, we would like to mention that energy distances have been used in the construction of a prior DRL method named MMDRL \citep{nguyen2021distributional}, but
there are significant differences between MMDRL and the proposed EBRM.
An important difference is that
EBRM minimizes with respect to $\theta$ in \textit{both} entries of \eqref{Biased_EStimator_onestep} that correspond to LHS and RHS of the distributional Bellman equation, whereas MMDRL fixes the RHS with the previous parameter update in every iteration.
As shown in numerical experiments (Table \ref{Atari_maintext}), EBRM performed significantly better than MMDRL.
Moreover, MMDRL does not provide any theoretical guarantee for its statistical performance, while we derived the statistical error bound for EBRM (Theorem \ref{Realizable_Final_Theorem}), as well as its multi-step extension (Section \ref{Nonrealizable_Scenario}).

\subsection{Statistical Error Bound} \label{theoretical_bound_realizable}

In this subsection, we will provide a statistical error bound for EBRM-single-step.
We will first focus on the realizable setting and defer the analysis for the non-realizable case to Section \ref{Nonrealizable_Scenario}.
\begin{assumption} \label{realizable}
	The target $\Upsilon_\pi$ can be realized by our candidate space $\Theta$. That is, $\Upsilon_\pi \in \mathcal{P}_\Theta:=\{\Upsilon_\theta\in \mathcal{P}(\mathbb{R}^d)^{\SAspace} \ | \ \theta\in\Theta \}$.
\end{assumption}
This realizability assumption is already weaker than the completeness assumption, which is commonly used in theoretical analyses for distributional RL \citep[e.g.,][]{wu2023distributional, wang2023benefits} and traditional RL \citep[e.g.,][]{chen2019information, fan2020theoretical}. Under completeness, 
for all $\theta\in\Fspace$, there exists a $\theta'\in\Fspace$ such that $\Bellopt\Upsilon_\theta=\Upsilon_{\theta'}$. This implies realizability due to $\Upsilon_\pi = \lim_{T\rightarrow\infty}(\Bellopt)^T \Upsilon_\theta$ under mild conditions.

Additionally, we make several mild assumptions regarding the transition probability $p$ and the candidate space $\Fspace$, including the sub-Gaussian rewards. 
A random variable (vector) $\mathbf{X}$ being sub-Gaussian implies its tail probability decaying as fast as Gaussian distribution
, quantified with finite sub-Gaussian norm $\|\mathbf{X}\|_{\psi_2} < \infty$.

\begin{assumption} \label{bounded} For any $\theta\in\Theta$, the random element $Z(s,a;\theta)$, which follows $\Upsilon_\theta(s,a)$, has finite expectation of norms, and the reward distributions conditioned on $s,a$ (namely, $R(s,a)$) are sub-Gaussian, i.e.,
	\begin{gather*}
		\supf \supsa \Expect\|Z(s,a;\theta)\| < \infty \mbox{ and } \supsa\|R(s,a)\|_{\psi_2} < \infty.
	\end{gather*}
\end{assumption}
We further explain the sub-Gaussian norm $\|\cdot\|_{\psi_2}$ in Appendix \ref{explanation_subgaussian}. Note that the bounded reward assumptions that are widely adopted (Table \ref{Convergence_comparison} of Appendix \ref{Comparison_between_DRL_details}) are stronger than Assumption \ref{bounded}.

Then we can obtain the convergence rate $O(\sqrt{\log(N/\delta)/N})$ as follows. Its proof can be found in Appendix \ref{Proof_Realizable_Final_Theorem}, and its special case for $\Theta\subseteq \mathbb{R}^p$ (under Assumption \ref{Lipschitz}) is covered in Corollary \ref{Realizable_Final_corollary}.

\begin{theorem} \label{Realizable_Final_Theorem}
	{\bf (Inaccuracy for realizable scenario)} Under Assumptions \ref{RN_derivative}, \ref{realizable}, \ref{bounded},
	for any $\delta\in(0,1)$, given large enough sample size $N\geq N(\delta)$, our estimator $\hat{\theta}\in\Fspace$ given by (\ref{Biased_EStimator_onestep}) satisfies the following bound with probability at least $1-\delta$,
	\begin{align}
		\energyonebar(\Upsilonhat, \Upsilon_\pi) \lesssim  %
		\sqrt{\frac{1}{N} \log \left( \frac{ (|\SAspace| + N)}{\delta} \right) }, \nonumber %
	\end{align}
	where $N(\delta)$ depends on the complexity of $\Theta$ (details in \eqref{finite_bound_realizable_final} Appendix \ref{realizable_finalizing_the_bound}) and $\lesssim$ means smaller than or equal to the given bound (RHS) multiplied by a positive number that does not depend on $N$.
\end{theorem}
The effects of all other factors (e.g. $d$, $\gamma$, model complexity, sub-Gaussianity, etc.) can be seen from the finite-sample error bound of Appendix \ref{realizable_finalizing_the_bound}. We also provide the closed-form estimator for the continuous state-action space with deterministic transition in \ref{continuous_estimator}, along with its theoretical analysis (Theorem \ref{continuous_deterministic_realizable}).

\begin{remark}
	FLE \citep{wu2023distributional} also provided error bounds in their Corollary 4.14. However, their inaccuracy measure is for $(S,A)$-marginalized distribution. Theorem \ref{Realizable_Final_Theorem} focuses on expectation-extended inaccuracy, taking into account inaccuracy levels for every state-action pair, which is more informative than that in FLE.
\end{remark}

\begin{remark} \label{remark:comparison_FLE}
	It is not easy to form a direct theoretical comparison between EBRM and FLE, since our results for EBRM is based on weaker assumptions (e.g., realizability instead of completeness, subgaussianity instead of bounded rewards). However, if we assume bounded 1-dimensional ($d=1$) tabular setting, we can derive a faster guarantee of convergence rate in Wasserstein-$p$ metric for all $p\geq 1$ for EBRM. See \eqref{ineq:EBRM_vs_FLE} of Appendix \ref{proof:comparison_FLE}. 
\end{remark}

\section{NON-REALIZABLE SETTINGS} \label{Nonrealizable_Scenario}

Without abundant data, we normally resort to a simpler model (e.g. linear MDP by \cite{wang2023benefits}), where non-realizability can arise as a serious issue.
Even in traditional RL, theoretical analysis under non-realizability (e.g., \cite{perdomo2023complete} and \cite{miyaguchi2021asymptotically}) is of significant interest.
In DRL, as our target is to estimate the conditional distribution of return given any state-action pair, which is an infinite-dimensional object, non-realizability could be serious and challenging even for tabular settings, not to say non-tabular settings.
In this section, we investigate non-realizability.

\subsection{Combating Non-realizability With Multi-step Extensions} \label{Advantage_of_multistep}

First, we note that, even under non-realizability, Theorem \ref{Fundamental_Realizable} still holds. Then, as long as the violation of Assumption \ref{realizable} is minuscule and so $\inf_{\theta\in\Theta}\energyonebar(\Upsilon_\theta, \Bellopt\Upsilon_\theta)\approx0$, we can still derive a tight error bound (shown in Appendix \ref{singlestep_nonrealizable_tabular}). However, if the violation is severe (i.e., large extent of non-realizability), the minimal Bellman residual $\inf_{\theta\in\Theta}\energyonebar(\Upsilon_\theta, \Bellopt\Upsilon_\theta)$ can be large.
This can lead to a possibly serious mismatch between the ``best approximation'' $\thetatilde$ (the target) and the Bellman residual minimizer $\theta_*$:
\begin{equation} 
	\tilde{\theta} := \arg\min_{\theta\in\Fspace} \energyonebar( \Upsilon_\theta, \Upsilon_\pi ) \neq  \arg\min_{\theta\in\Fspace} \energyonebar(\Upsilon_\theta, \Bellopt \Upsilon_\theta)=: \theta_*.  \label{pseudotrue}
\end{equation}
Clearly, this mismatch is not due to sample variability, so it is unrealistic to hope that $\hat{\theta}$ defined by \eqref{Biased_EStimator_onestep} would necessarily converge in probability to $\tilde{\theta}$ as $N\rightarrow \infty$.

To solve this issue, we propose a new approach. Temporarily ignoring mathematical rigor, the most important insight is that we can approximate $(\Bellopt)^m \Upsilon\approx \Upsilon_\pi$ with sufficiently large step level $m\in\mathbb{N}$. 
Thanks to the properties of energy distance,
we have the following for some constant $C>0$:
\begin{align} \label{objectiveinaccuracy_difference_boundedby_G}
	\suptheta|\energyonebar(\Upsilon_\theta, (\Bellopt)^m \Upsilon_\theta) - \energyonebar(\Upsilon_\theta, \Upsilon_\pi)| \leq C \gamma^m ,
\end{align}
as shown in Appendix \ref{Proof_Fact1to2}.
As $m\rightarrow \infty$, the RHS of \eqref{objectiveinaccuracy_difference_boundedby_G} shrinks to zero, making $m$-step Bellman residual $\energyonebar(\Upsilon_\theta, (\Bellopt)^m \Upsilon_\theta)$ approximate the inaccuracy function $\energyonebar(\Upsilon_\theta, \Upsilon_\pi)$. This leads the two minimizers to be close, as illustrated schematically in Figure \ref{mstep_approximating} and exemplified in Appendix \ref{model_misspecification_rho}:
\begin{align} \label{mstep_minimizers} 
	\theta_*^{(m)}&:=\arg\min_{\theta\in\Fspace} \energyonebar(\Upsilon_\theta, (\Bellopt)^m \Upsilon_\theta) \\
	&\approx \arg\min_{\theta\in\Fspace} \energyonebar(\Upsilon_\theta, \Upsilon_\pi) =: \tilde{\theta} \ \ \text{for large enough } m . \nonumber
\end{align}
One can intuitively guess that larger step level $m$ is required when the extent of non-realizability is large. 

Multi-step has been employed to improve sample efficiency in traditional RL \citep[e.g.,][]{munos2016safe, chen2021finite} and distributional RL \citep[e.g.,][]{tang2022nature}.
However, to the best of our knowledge, it has never been used to address non-realizability, as in \eqref{mstep_minimizers}.

\subsection{Splitting-based Estimator} \label{splitdata_operator}

Given a value of step level $m\in\mathbb{N}$, our task is to estimate the multi-step Bellman residual $\energyonebar(\Upsilon_\theta, (\Bellopt)^m \Upsilon_\theta)$. One natural way is to use the whole data $\mathcal{D}$ to estimate $\Bellhat$ and re-use it for $m$ consecutive times $\Bellhatmulti$. However, this causes two issues: (1) This generally requires the computation of $N^m$ trajectories of $s,a,r^{(1)},s^{(1)},a^{(1)},\cdots r^{(m)},s^{(m)},a^{(m)}$ (as discussed in Appendix \ref{exponential_trajectories}), leading to a heavy computational burden. (2) Re-using the same data $\mathcal{D}$ for every step of transition causes dependence between our samples, leading to difficulty in deriving theoretical guarantee.

We introduce a splitting-based estimator in this subsection for the sake of strong theoretical guarantee, but will also explain a more practical estimator based on bootstrap in the following subsection. In the splitting-based method, we divide the data into $m$ equally-sized subdatasets $\mathcal{D}=\mathcal{D}_1\cup \cdots \cup \mathcal{D}_m$, and use each subdataset to estimate the Bellman operator at $j$-th step, say $\hat{\mathcal{T}}_j^\pi$. By estimating $(\Bellopt)^m$ via $\Thatseq := \hat{\mathcal{T}}_1^\pi \cdots \hat{\mathcal{T}}_m^\pi$, we can define the following objective functions:
\begin{gather}
	F_m(\theta):= \bar{\mathcal{E}}\big( \Upsilon_\theta, (\mathcal{T}^\pi)^m \Upsilon_\theta \big) \ \ \& \ \ \thetamstar:=\arg\min_{\theta\in\Theta}F_m(\theta) \nonumber \\ 
	F(\theta ):= \bar{\mathcal{E}}\big( \Upsilon_\theta, \Upsilon_\pi \big) \ \ \& \ \   \thetatilde:=\arg\min_{\theta\in\Theta} F(\theta) \label{objective_functions} \\
	\Fmhat(\theta) := \energyonebarhat( \Upsilon_\theta, \Thatseq \Upsilon_\theta) \ \ \& \ \   \thetamhat:=\arg\min_{\theta\in\Theta}\Fmhat(\theta).  \nonumber
\end{gather}

For the theoretical investigation, we make the following two assumptions.
\begin{assumption} \label{Lipschitz}
	The candidate space $\Theta\subset\mathbb{R}^p$ is compact. There exists $L>0$ such that 
	\begin{align*}
		\expectationbounder(\Upsilon_{\theta_1}, \Upsilon_{\theta_2}) \leq L \|\theta_1-\theta_2\| \quad \mbox{for} \quad \forall \theta_1,\theta_2\in\Theta,
	\end{align*}
	where $\Wassersup$ is the supremum-extended \eqref{supremum_distance} Wasserstein-1 metric $\Wasserone$.
\end{assumption}

We can replace $\Wassersup$ of Assumption \ref{Lipschitz} with another metric, as long as it satisfies the properties mentioned in Appendix \ref{Wassersup_replacing}. Lipschitz continuity is involved in bounding the entropy with respect to $\Wassersup$.
(See Remark \ref{metricentropy_Lipschitz}.) 

\begin{remark}
	Assumption \ref{Lipschitz} is used for bounding the model complexity. Of course, it may be strong for complex models (e.g., deep neural network). However, we expect that such complex models can almost satisfy realizability, so we can apply single-step estimation \eqref{Expectation_three_terms}.
\end{remark}

\begin{assumption} \label{mstep_strongconvex_all}
	There exists a constant $c>0$ such that for all $m\in\mathbb{N}$ (or sufficiently large $m$) we have
	$F_m(\theta) \geq F_m(\thetamstar) + c\cdot \| \theta - \thetamstar \|^2$ for their unique minimizers $\thetamstar$.
\end{assumption}

This assumption is imposed to obtain the convergence rate of our estimator by adopting M-estimation theory \citep[][Chapter 5]{van2000asymptotic}. The quadratic bound can be relaxed to general $q\geq 1$ (e.g. Assumption \ref{uniqueness_generalized}). Albeit not necessary, we shall assume uniqueness of $\thetamstar$ and $\thetatilde$ for convenience. Although Assumption \ref{mstep_strongconvex_all} seems to be strong, we relax this to Assumption \ref{uniqueness_generalized} for the more practical bootstrap estimator (Section \ref{multistep_bellman_operator}). Moreover, we anticipate that Assumption \ref{uniqueness_generalized} with $q=2$ shall imply Assumption \ref{mstep_strongconvex_all} under mild conditions due to the uniform convergence of $F_m(\cdot)$ to $F(\cdot)$ shown in \eqref{objectiveinaccuracy_difference_boundedby_G}. Although Assumption \ref{uniqueness_generalized} with $q=2$ seems similar to a strong convexity assumption, this can accommodate non-convex functions (see Appendix \ref{example_strongly_convex}).
Proof is given in Appendix \ref{sec:degeneracy}.

\begin{theorem}
	\label{Degeneracy_increasingstep}
	{\bf (Inaccuracy for non-realizable scenario)} Under Assumptions \ref{RN_derivative}, \ref{bounded}, \ref{Lipschitz}, \ref{mstep_strongconvex_all}, if we set as $m=\lfloor \log_{\frac{1}{\gamma}}(C\cdot N)\rfloor$ for an arbitrary $C>0$, then the estimator $\thetamhat$ \eqref{objective_functions} satisfies the following convergence rate,
	\begin{align*}
		\energyonebar(\Upsilon_{\thetamhat} , \Upsilontilde) \leq \frac{L \big( L\sqrt{p} + \sqrt{\frac{d}{1-\gamma}} \big)}{\sqrt{p_{\rm min}}}  O_p \bigg( \sqrt{\frac{(\log_{\frac{1}{\gamma}}N)^3}{N}} \bigg).
	\end{align*}
\end{theorem}
Under realizability (Assumption \ref{realizable}), where we have $\Upsilontilde=\Upsilon_\pi$, Theorem \ref{Degeneracy_increasingstep} implies convergence towards $\Upsilon_\pi$. Its
convergence rate degenerates into that of Theorem \ref{Realizable_Final_Theorem} up to logarithmic difference. Without realizability, we still have convergence towards the best approximation $\tilde{\theta}$ \eqref{pseudotrue} in the same rate. This multi-step idea can be extended towards non-tabular setting. However, this requires the data to be collected as trajectories, as assumed by \cite{huang2022off}.

\subsection{Bootstrap-based Estimator} \label{multistep_bellman_operator}
Although splitting-based estimator solves the dependence issue, it requires computation of $(N/m)^m$ many trajectories by the same logic as Appendix \ref{exponential_trajectories}. This is still very large. Therefore, we propose another estimator that is based on bootstrap, which can be more practical.
The idea of bootstrap-based estimator is to approximate $\Bellhatmulti$ by bootstrapping from the $N^m$ many possible trajectories. Denoting the bootstrap-based estimator as $\Bootstrapopt$ and letting $M$ be the number of resampled trajectories, we can approximate $\Bellhatmulti$ with $\Bootstrapopt$ as $M\rightarrow\infty$. Here, we define the objective functions, including re-definition of $\Fmhat$ which is different from that of \eqref{objective_functions}. We will refer to this method as EBRM-multi-step. Its exact procedure is summarized in Algorithm \ref{EBRM_multistep_algorithm} of Appendix \ref{bootstrap_estimation_procedure}.
\begin{gather}  
	\Fmhat(\theta):= \hat{\bar{\mathcal{E}}}\big( \Upsilon_\theta, (\hat{\mathcal{T}}^\pi)^m \Upsilon_\theta \big)  \ \ \& \ \ \thetamhat:=\arg\min_{\theta\in\Theta}\hat{F}_m(\theta), \nonumber \\
	\FBhat(\theta):=\energyonebarhat\big( \Upsilon_\theta, \Bootstrapopt \Upsilon_\theta \big)  \ \ \& \ \ \thetabootstrap:=\arg\min_{\theta\in\Theta}\FBhat(\theta). \label{Bootstrap_objective_function}
\end{gather}

\begin{table*}[h] 
	\caption{Mean inaccuracy with standard deviation in parenthesis over 100 simulations.
		Smallest value in boldface. The top and bottom tables are simulation results for Sections \ref{atari_simulation} and \ref{noncomplete_simulations}, respectively.}  \label{Atari_maintext}
	\begin{center}
		\begin{tabular}{lccc|ccc}
			\toprule
			\multicolumn{1}{c}{\bf } & \multicolumn{3}{c}{ \bf $\energyone$-inaccuracy}  & \multicolumn{3}{c}{\bf $\Wasserone$-inaccuracy}  \\  \midrule
			\multicolumn{1}{c}{Game}  &\multicolumn{1}{c}{Acrobot} &\multicolumn{1}{c}{Cartpole} &\multicolumn{1}{c}{Mountain}   &\multicolumn{1}{c}{Acrobot} &\multicolumn{1}{c}{Cartpole} &\multicolumn{1}{c}{Mountain}
			\\ \hline 
			EBRM    & {\bf 0.5205}  &  {\bf 0.3419}  & {\bf 1.2904} & {\bf 0.9551}  & {\bf 0.5738}  &  {\bf 2.1918}    \\ 
			& (0.4246)  &  (0.3877)  &  (0.1277)                   &  (0.7051)  &  (0.3715) &  (0.1435)    \\ \hline  
			QRDQN    & 6.5878  &  8.7865  &  8.9607                         & 4.1412  &  5.2342   &  4.8682    \\ 
			& (1.9709)  &  (1.8717)  &  (1.9169)                   &  (0.8044)  &  (0.9123)   &  (0.8810)    \\ \hline  
			MMDRL    & 6.1661  &  9.1530  &  9.0803                     &  4.1805  &  5.6362    &  5.2500    \\ 
			&  (1.7239)  &  (1.7039)  &  (1.7074)               &  (0.7368)  &  (0.7489)  &  (0.7707)    \\ \bottomrule  
		\end{tabular} 
	\end{center}
	\begin{center}
		\begin{tabular}{lccc|ccc}
			\toprule
			\multicolumn{1}{c}{\bf } & \multicolumn{3}{c}{ \bf $\energyonebar$-inaccuracy}  & \multicolumn{3}{c}{\bf $\Wasserbar$-inaccuracy}  \\  \midrule
			\multicolumn{1}{c}{Sample size}  &\multicolumn{1}{c}{2000} &\multicolumn{1}{c}{5000} &\multicolumn{1}{c}{10000}   &\multicolumn{1}{c}{2000} &\multicolumn{1}{c}{5000} &\multicolumn{1}{c}{10000}
			\\ \hline 
			EBRM    &{\bf 107.464}  &  {\bf 81.097}  & {\bf 70.572} & {\bf 138.195}  & {\bf 89.389}  &  {\bf 82.698}    \\ 
			& (40.038)  &  (37.623) &  (9.791)                   &  (23.291)  &   (17.663)&  (5.661)    \\ \hline  
			FLE    & 453.045  &  620.049  &  789.298                     &  260.564  &  348.004   &  437.266     \\ 
			&  (41.320)  &  (37.668)  &  (39.350)               &  (20.451)  & (19.126)  &  (20.810)   \\  \bottomrule
		\end{tabular} 
	\end{center}
\end{table*}

\begin{figure*}[!ht] 
	\vspace{.3in}
	\begin{center}
		\includegraphics[width=1.00\linewidth]{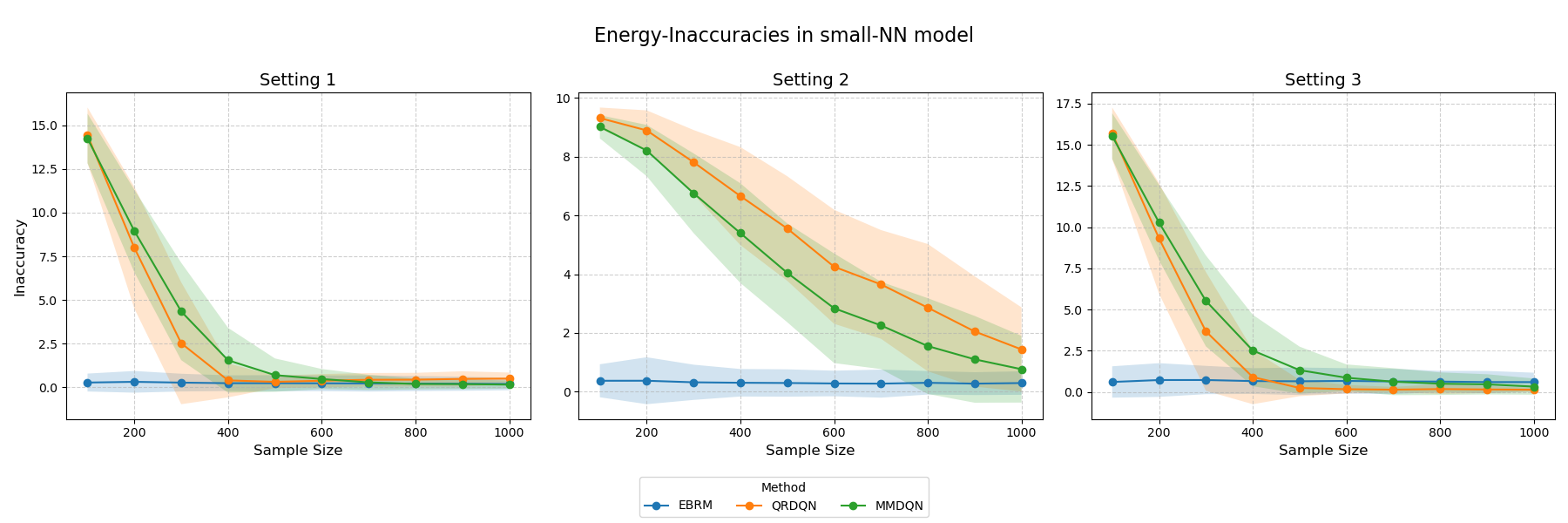}
		\caption{$\energyone$-Inaccuracy for three different settings of Cartpole games based on small neural network model. Lines represent mean inaccuracy values and shaded regions represent the interval $(\text{Mean}\pm 2 \cdot \text{STD})$ (blue: EBRM, orange: QRDQN, green: MMDQN).}
		\label{fig:cartpole_energy_smallNN_energy} 
	\end{center}
	\vspace{.3in}
\end{figure*}

Since we can choose the value of $M$ (number of resampled trajectories), this algorithm is indeed more practical than the splitting-based estimator suggested in the previous section. We could also develop a theoretical guarantee for convergence towards the best approximation $\Upsilon_{\thetatilde}$ \eqref{objective_functions} that we can achieve under the non-realizability. However, since we cannot be free from the dependence issue, our proof technique ends up with a slower convergence guarantee (Theorem \ref{thetaBhat_convergence_generalized} of Appendix \ref{Proof_Nonrealizable_bootstrap_convergencerate}), which is in fact based on Assumption \ref{uniqueness_generalized} (weaker than Assumption \ref{mstep_strongconvex_all}). Its finite-sample error bound is given in Appendix \ref{nonrealizable_finite_sample_error_bound}.

\section{EXPERIMENTS} \label{Simulations}

In this section, we conduct a numerical study of the proposed method. We fix a target policy $\pi$ and then estimate the target $\Upsilon_\pi \in \mathcal{P}(\mathbb{R}^d)^{\SAspace}$ with offline data. We compared its accuracy with other DRL methods based on Energy distance and Wasserstein metric.
Its main purpose lies in confirming the theoretical advantage of EBRM that we proved in Sections \ref{Realizable_Scenario} and \ref{Nonrealizable_Scenario}.
All simulation codes can be found in \url{https://github.com/hse1223/Distributional-OPE-EBRM.git}.

\subsection{OpenAI-gym Games} \label{atari_simulation}

We applied our method in three OpenAI-gym classical games: acrobot, cartpole, and mountaincar, the second of which was used in comparing various OPE methods in traditional (non-distributional) RL \citep[e.g.,][]{wang2023projected,narita2021debiased}. Under the discount rate $\gamma=0.90$, we have formed our model with neural network and estimated the true distributions $Z_\pi(s,a)\sim \Upsilon_\pi(s,a)$ with particles (or Dirac delta's), which is one of the most intuitive ways of expressing a distribution. For fair comparison, we compared our method EBRM with other particle (or quantile) based methods, QRDQN \citep{dabney2018distributional} and MMDRL \citep{nguyen2021distributional} with the same neural network structure (Appendix \ref{atari_games_simulation_details}). FLE \citep{wu2023distributional} is not included, as it requires conditional densities which cannot be directly constructed based on particles.

It is difficult to compute the true expectation-extended distance \eqref{expectation_distance} for assessment purposes under these games, due to its large cardinality $|\SAspace|$.
Therefore, we adopted a simpler way of measuring the inaccuracy, as used in Corollary 4.14 of \cite{wu2023distributional}, which is the statistical distance between the marginalized distributions $\Upsilon^{marginal}\in\mathcal{P}(\mathbb{R}^d)$, i.e., the mixture of $\{\Upsilon(s,a): s,a\in\SAspace\}$ with weights $\{\pi_\mu(s,a):s,a\in\SAspace\}$.
Here, $\pi_\mu$ is combination of $S\sim \mu$ and $A\sim \pi(\cdot|A)$. In Table \ref{Atari_maintext} (top), we compare the inaccuracy of the estimated marginal distribution based on both energy distance and Wasserstein-1 metric. Our method EBRM showed the most accurate results in all cases.
We also conducted more experiments based on the Cartpole games by varying the neural network model size and the sample size. Figure \ref{fig:cartpole_energy_smallNN_energy} shows the superior performance of EBRM for small neural network model. 
We see similar comparison results for the large neural network model (see Appendix \ref{sec:cartpole_extensive}).

\subsection{Experiments Under Non-completeness} \label{noncomplete_simulations}

In addition to OpenAI-gym games, we have also performed simulations to demonstrate the theoretical guarantee of EBRM when completeness is not satisfied (Theorems \ref{Realizable_Final_Theorem} and \ref{thetaBhat_convergence_generalized}). Toward that end, we will compare our bootstrap-based EBRM \eqref{Bootstrap_objective_function} with a method that is based on the completeness assumption. We assumed a simple tabular setting with $\mathcal{S}=\{1,2,\dots,30\}$ and $\mathcal{A}=\{-1,1\}$ (details in Appendix \ref{given_environment}). 
A practical algorithm for choosing the step level is explained in Appendix \ref{Tuning_parameters_in_simulations_EBRM}.

We have tried both realizable settings (big and small variances) and non-realizable settings (big and small $\gamma$), whose details are explained in Appendices \ref{given_environment}--\ref{nonrealizable_model_plot}. EBRM outperformed its competitors in most settings (Tables \ref{realizable99_energy_inaccuracy}--\ref{nonrealizable_wassermix_inaccuracy}). In particular, when completeness is severely violated (non-realizable and $\gamma=0.99$ in Table \ref{Atari_maintext} (bottom)), EBRM successfully approximated the minimum inaccuracy $\min_{\theta\in\Theta}\energyonebar(\Upsilon_\theta , \Upsilon_\pi) \approx 63.216$. The worse performance of FLE is not surprising, since it requires the completeness assumption (details in Appendix \ref{sec:FLE_wrongconvergence}).
Besides, we have conducted an ablation study for the step level of EBRM, and 
demonstrated that the step level plays a key role in combating non-realizability.
See Appendix \ref{sec:ablation} for more details.

\subsubsection*{Acknowledgements}

Portions of this research were conducted with the advanced computing resources provided by Texas A\&M High Performance Research Computing.

\bibliographystyle{chicago} %
\bibliography{References}       %

\clearpage

\appendix

\section{PROOFS FOR SECTIONS \ref{off_policy_evaluation_based_on_multistep} AND \ref{Realizable_Scenario}}

\subsection{Bounding Radon-Nikodym Derivative} \label{Proof_Csup}

Define $q_{b_\mu}^{\pi:t}$ to be the density of the probability of $S^{(t)},A^{(t)}$ where $A^{(j)}\sim \pi(\cdot|S^{(j)})$ and $(R^{(j)}, S^{(j)}) \sim p(\cdots|S^{(j-1)}, A^{(j-1)})$ for $1\leq j \leq t$ and $(S^{(0)}, A^{(0)})\sim b_\mu$. Let $t\in\mathbb{N}$ be arbitrary. Then we have the following for all $s,a\in\SAspace$ by Assumption \ref{RN_derivative}, with $\SAmeasure$ being the underlying measure of $\SAspace$, 
\begin{align*}
	\frac{q_{b_\mu}^{\pi:t}(s,a)}{\probsa}  \leq \int_{\SAspace} q^\pi(s,a|\tilde{s},\tilde{a}) \cdot q_{b_\mu}^{\pi:t-1}(\tilde{s},\tilde{a}) \mathrm{d}\nu(\tilde{s},\tilde{a}) \cdot \frac{1}{\probsa} \leq \sup_{\tilde{s},\tilde{a}} \supsa \frac{q^\pi(s,a|\tilde{s},\tilde{a})}{\probsa} <\infty ,
\end{align*}
where $ q_{b_\mu}^{\pi:0}=b_\mu$. Since $t\in\mathbb{N}$ was arbitrary, this implies existence of $C(t), C_{\rm sup}\in(0,\infty)$ such that 
\begin{align} \label{Ct_and_Csup}
	\supsa \bigg\{\frac{q_{b_\mu}^{\pi:t}(s,a)}{b_\mu(s,a)} \bigg\} \leq C(t) \leq C_{{\rm sup}} < \infty \ \ \text{for } \forall \ t\in\mathbb{N}.
\end{align}

\subsection{Proof Of Theorem \ref{Fundamental_Realizable}} \label{Proof_Fundamental_Realizable}

\subsubsection{Properties Of Distance}\label{properties_of_distance}

\begin{property} \label{translation_scale_sensitivity} 
	$\distancemeasure$ satisfies translation-invariance and scale-sensitivity of order $\beta_0>0$. That is, with $z$ being an arbitrary (nonrandom) constant of computable size and $c\in\mathbb{R}$, 
	\begin{gather*}
		\distancemeasure\{\mathcal{L}(z+X), \mathcal{L}(z+Y)\}\leq \distancemeasure\{\mathcal{L}(X),\mathcal{L}(Y)\}  \ \ \& \ \ 
		\distancemeasure\{\distL(cX),\distL(cY)\} \leq |c|^{\beta_0}\distancemeasure\{\distL(X),\distL(Y)\}. 
	\end{gather*}
\end{property}
\begin{property} \label{mixture_inequality} 
	Letting $\mu_1(\cdot),\mu_2(\cdot): \mathcal{Z}\rightarrow \mathcal{P}(\mathbb{R}^d)$ have different probability measures depending on the index random variable $Z\in\mathcal{Z}$ that follows a distribution $P(\cdot)$, the distance between probability-mixtures $\int_{\mathcal{Z}}\mu_1(z)\mathrm{d}P(z)$ and $\int_{\mathcal{Z}}\mu_2(z)\mathrm{d}P(z)$ satisfies convexity, that is
	\begin{align*}
		\distancemeasure\bigg\{ \int_{\mathcal{Z}}\mu_1(z)\mathrm{d}P(z),   \int_{\mathcal{Z}}\mu_2(z)\mathrm{d}P(z) \bigg\} \leq \int_{\mathcal{Z}}\distancemeasure\{ \mu_1(z), \mu_2(z)	\} \mathrm{d} P(z) =: \mathbb{E}_{Z\sim P}[ \distancemeasure \{\mu_1(Z),\mu_2(Z)\}] .
	\end{align*}
\end{property}

\begin{property} \label{squared_metric} 
	There exists some value $q\geq 0$ such that following relaxed triangular inequality holds for all integers $K\geq 2$,
	\begin{align} \label{relaxed_triangle_p}
		\distancemeasure(\distL(X_0),\distL(X_K)) &\leq K^q \sum_{i=0}^{K-1} \distancemeasure(\distL(X_i),\distL(X_{i+1})).
	\end{align}
	This is satisfied by all $(q+1)$-powered metric (with nonnegative integer $q\geq 0$), that is $\distancemeasure(P,Q)=\rho^{q+1}(P,Q)$ for some probability metric $\rho$. Using convexity of $f(x)=x^{q+1}$ ($x\geq 0$), we can derive $(x_1+\cdots x_K)^{q+1}\leq K^q \cdot (x_1^{q+1} +\cdots + x_K^{q+1})$, which directly implies that $\eta=\rho^{q+1}$ satisfies \eqref{relaxed_triangle_p}. 
\end{property}

\subsubsection{Proof} 

Let $\Upsilon\in\mathcal{P}(\mathbb{R}^d)^{\SAspace}$ be arbitrarily chosen. Starting with Relaxed Triangular Inequality (\ref{relaxed_triangle_p}) with $K=2$, we obtain the following for an arbitrary $t\in\mathbb{N}$,
\begin{align*} %
	\distancemeasure\bigg\{\Upsilon(s,a), \Upsilon_\pi(s,a) \bigg\}  &\leq 2^q \bigg[ \underbrace{ \distancemeasure \bigg\{ \Upsilon(s,a), (\Bellopt)^t \Upsilon(s,a)  \bigg\}}_{(a)} + \underbrace{\distancemeasure \bigg\{ (\Bellopt)^{t} \Upsilon(s,a), \Upsilon_\pi(s,a) 	\bigg\}}_{(b)}  \bigg].
\end{align*}
Let us first deal with $(b)$. Define $P_t^\pi(\cdots|s,a)$ to be the probability measure of $(\sum_{i=1}^t \gamma^{i-1} R^{(i)},S^{(t)},A^{(t)})$ after $t$ steps starting from the initial state-action pair $(S,A)=(s,a)$ under the given transition probability $(R^{(t)},S^{(t)}) \sim p(\cdots| S^{(t-1)}, A^{(t-1)})$ and the target policy $A^{(t)}\sim\pi(\cdot|S^{(t)})$. Further denoting the probability measure of $y + \gamma^t Z(s^{(t)},a^{(t)})$ with $Z(s,a)\sim \Upsilon(s,a)$ as $(g_{y,\gamma^t})_{\#}\Upsilon(s^{(t)},a^{(t)})$ for the fixed value of $s^{(t)},a^{(t)}$, $y=\sum_{i=1}^t \gamma^{i-1}r^{(i)}$ (which aligns with the notation $(g_{r,\gamma})_{\#}$ in \eqref{Bellman_operator_definition}), we can obtain the following by Properties \ref{translation_scale_sensitivity} and \ref{mixture_inequality},
\begin{align} 
	(b) &= \distancemeasure \bigg\{ (\Bellopt)^t \Upsilon(s,a) , (\Bellopt)^t\Upsilon_\pi(s,a) 	\bigg\}  \ \ \because (\Bellopt)^t \Upsilon_\pi (s,a) = \Upsilon_\pi(s,a)   \nonumber \\ 
	&= \distancemeasure\bigg\{ \int (g_{y,\gamma^t})_{\#}\Upsilon(s^{(t)},a^{(t)}) \mathrm{d}P_t^\pi (y,s^{(t)},a^{(t)}|s,a) ,  \int (g_{y,\gamma^t})_{\#}\Upsilon_\pi (s^{(t)},a^{(t)}) \mathrm{d}P_t^\pi(y,s^{(t)},a^{(t)}|s,a) \bigg\}   \nonumber  \\
	&\leq \gamma^{t\beta_0} \mathbb{E}_\pi \bigg[ \distancemeasure\bigg\{  \Upsilon(S^{(t)},A^{(t)}) , \Upsilon_\pi (S^{(t)},A^{(t)}) 	\bigg\} \bigg| S=s,A=a \bigg]  .   \nonumber
\end{align}

From now on, we will use $\Expectbehavior$ and $\mathbb{E}_{q_{b_\mu}^{\pi:t}}$ to denote the expectation with respect to the probability $(S,A)\sim b_\mu$ and $(S,A)\sim q_{b_\mu}^{\pi:t}$. Treating $(S,A)$ as random, we can obtain the following,
\begin{align*} %
	\Expectbehavior \bigg[(b)\bigg] & \leq \gamma^{t\beta_0} \mathbb{E}\bigg[ \distancemeasure\bigg\{\Upsilon(S^{(t)},A^{(t)}), \Upsilon_\pi(S^{(t)},A^{(t)})) \bigg\} \bigg| (S,A)\sim b_\mu, \  A^{(i)}\sim\pi(\cdot|S^{(i)}) \ \forall i\geq 1  \bigg] \nonumber  \\ 
	&= \gamma^{t\beta_0} \mathbb{E}_{q_{b_\mu}^{\pi:t}} \bigg[\distancemeasure\bigg\{ \Upsilon(S,A), \Upsilon_\pi(S,A)	\bigg\} \bigg] 
	\leq C(t) \cdot \gamma^{t\beta_0}\cdot \distancemeasurebar ( \Upsilon, \Upsilon_\pi)  \ \ \text{by \eqref{Ct_and_Csup}}.
\end{align*}
Now let us deal with $(a)$ using Property \ref{squared_metric}. Let $\Upsilon_k(s,a)=(\Bellopt)^{k}\Upsilon(s,a)$. For sufficiently large $t\in\mathbb{N}$, we can bound $\distancemeasure(\Upsilon(s,a), (\Bellopt)^{t} \Upsilon(s,a))$ with following by relaxed triangle inequality \eqref{relaxed_triangle_p} with $K=2$,
\begin{align*}
	2^q\cdot  \distancemeasure(\Upsilon_0(s,a), \Upsilon_1(s,a)) + (2^q)^2 \cdot  \distancemeasure(\Upsilon_1(s,a), \Upsilon_3(s,a)) + (2^q)^3\cdot \distancemeasure(\Upsilon_3(s,a), \Upsilon_7(s,a)) + \cdots 
\end{align*}
Further applying relaxed triangle inequality \eqref{relaxed_triangle_p} with general $K\geq 2$ to each term with $\distancemeasure$, we can formalize it into
\begin{align*}
	(a) & \leq \sum_{k=1}^{\infty} (2^q)^{2k-1} \cdot \sum_{j=0}^{s_1(k)} \distancemeasure\bigg( \Upsilon_{s_1(k) + j}(s,a), \Upsilon_{s_1(k) + j + 1}(s,a) \bigg) \ \ \text{where } s_1(k)=2^{k-1}-1.
\end{align*}
Therefore we can further obtain following using similar logic as bounding $(b)$ with $C_{\rm sup}$ defined in \eqref{Ct_and_Csup},
\begin{align*} %
	&\Expectbehavior\bigg[(a)\bigg]\leq \sum_{k=1}^{\infty} (2^q)^{2k-1} \cdot \sum_{j=0}^{s_1(k) } \Expectbehavior\bigg[ \distancemeasure\bigg\{ \Upsilon_{s_1(k)  + j}(S,A), \Upsilon_{s_1(k)  + j+1}(S,A) \bigg\} \bigg] \leq C_{{\rm sup}} \cdot B(\gamma;\beta_0,q) \cdot \distancemeasurebar (\Upsilon, \Bellopt \Upsilon)
\end{align*} 
where we have 
\begin{gather}
	B(\gamma;\beta_0,q) := \sum_{k=1}^{\infty} (2^q)^{2k-1} \cdot \sum_{j=0}^{2^{k-1}-1} \gamma^{(2^{k-1}-1+j)\beta_0} = \sum_{k=1}^{\infty} (2^q)^{2k-1} \cdot \gamma^{(2^{k-1}-1)\beta_0} \cdot \sum_{j=0}^{2^{k-1}-1} (\gamma^{\beta_0})^j \nonumber \\ 
	\leq \bigg(\sum_{k=1}^{\infty} (2^q)^{2k-1} \cdot \gamma^{(2^{k-1}-1)\beta_0} \bigg) \cdot \bigg( \sum_{j=0}^{\infty} (\gamma^{\beta_0})^j \bigg) \leq \frac{1}{2^q(1-\gamma^{\beta_0})}\sum_{k=1}^{\infty} (4^{qk} \cdot \gamma^{(2^{k-1}-1)\beta_0}) < \infty. \nonumber %
\end{gather}
Then, we finally obtain the first result for all $t\in\mathbb{N}$, and the second result by letting $t\rightarrow\infty$,
\begin{gather}
	\distancemeasurebar(\Upsilon, \Upsilon_\pi)  \leq 2^q C_{{\rm sup}} \cdot \bigg\{ B(\gamma;\beta_0,q)\cdot \distancemeasurebar(\Upsilon, \Bellopt \Upsilon)  + \gamma^{t\beta_0} \distancemeasurebar(\Upsilon, \Upsilon_\pi) \bigg\} , \nonumber \\
	\therefore \ \distancemeasurebar(\Upsilon, \Upsilon_\pi) \leq C_{{\rm sup}} B_1(\gamma;\beta_0,q)\cdot \distancemeasurebar(\Upsilon, \Bellopt \Upsilon) \quad \text{where} \quad B_1(\gamma;\beta_0,q):=\frac{1}{(1-\gamma^{\beta_0})}\sum_{k=1}^{\infty} (4^{qk} \cdot \gamma^{(2^{k-1}-1)\beta_0}). \label{Fundamental_Realizable_specific}
\end{gather}

\subsubsection{Asymptotic Bound With The Discount Rate} \label{sec:B_gamma_asymptotic}

One may wonder how large the term $B_1(\gamma ; \beta_0, q)$ is, with respect to the discount rate $\gamma\in(0,1)$ becomes larger, that is $\gamma \rightarrow 1$. For regular metric that satisfies $q=0$ (refer to \ref{properties_of_distance}), it is trivial to derive
\begin{align*}
	B_1(\gamma; \beta_0, q) \leq \bigg(\frac{1}{1-\gamma^{\beta_0}}\bigg)^2 \quad \text{for} \quad q=0.
\end{align*}
Therefore, let us assume $q>0$. Then we have
\begin{align*}
	B_1(\gamma; \beta_0,q) = \frac{1}{1-\gamma^{\beta_0}} \cdot \bigg\{ \sum_{k=1}^{K_0} 4^{qk}\cdot (\gamma^{\beta_0})^{2^k} + \sum_{k=K_0+1}^{\infty} 4^{qk}\cdot (\gamma^{\beta_0})^{2^k} \bigg\},
\end{align*}
where $K_0\in\mathbb{N}$ is sufficiently large so that $k\geq K_0$ implies
\begin{align*}
	4^{qk}\cdot (\gamma^{\beta_0})^{2^k} \leq (\gamma^{\beta_0})^k, \quad \text{or equivalently,} \quad k \geq \log_2 k + \log_2\log_{\frac{1}{\gamma^{\beta_0}}}\bigg( \frac{4^q}{\gamma^{\beta_0}} \bigg).
\end{align*}
Let $c_0\in(0,1)$ be the maximum value that satisfies $(1-c_0)x\geq \log_2 x$ for all $x>0$. Numerical experiments tell us $c_0\approx 0.47$. Then, it suffices to define $K_0 \buildrel let \over = \lfloor \frac{1}{c_0}\cdot \log_2 \log_{\frac{1}{\gamma^{\beta_0}}} (2\cdot 4^q) \rfloor$.
Then, letting $C(q)=4^q/(4^q-1)$, we have 
\begin{gather*}
	\sum_{k=1}^{K_0} 4^{qk} \cdot (\gamma^{\beta_0})^{2^k} \leq \frac{4^q}{4^q-1} \cdot (4^{qK_0}-1) \leq C(q) \cdot \bigg\{ \log_{\frac{1}{\gamma^{\beta_0}}}(2\cdot 4^q) \bigg\}^{\frac{2q}{c_0}}, \\
	\sum_{k=K_0+1}^{\infty} 4^{qk}\cdot (\gamma^{\beta_0})^{2^k} \leq \sum_{k=K_0+1}^{\infty} (\gamma^{\beta_0})^k \leq \frac{1}{1-\gamma^{\beta_0}}.
\end{gather*}
Assuming increasing discount rate $\gamma\rightarrow 1$, by letting $x=\log(1/\gamma^{\beta_0})$, we can apply l'Hospital's rule to obtain $\lim_{\gamma\rightarrow1}\log_{1/\gamma^{\beta_0}}(2\cdot 4^q) \cdot (1-\gamma^{\beta_0}) = \log(2\cdot 4^q)$.
Thus, we eventually obtain the following with the value of $\beta_0$ and $q$ depending on $\distancemeasure$ (Section \ref{properties_of_distance}),
\begin{align} \label{Bgamma_rate}
	B_1(\gamma; \beta_0, q) \lesssim \bigg(\frac{1}{1-\gamma^{\beta_0}}\bigg)^{\frac{2q}{c_0}+1} \quad \text{as} \quad \gamma\rightarrow 1 \qquad \text{with } c_0\approx 0.47.
\end{align}

\subsection{Examples Of Valid Probability Distances} \label{Proof_squared_metric}

Here are examples of probability distances that satisfy properties \ref{properties_of_distance}, which are of course not necessarily metrics, but mostly are powered metrics. Given probability measures $\mu_1,\mu_2$, $J(\mu_1, \mu_2)$ indicate the class of joint distributions of two marginals $\mu_1$ and $\mu_2$. $\mathbf{X},\mathbf{X}'\sim \mu_1$ and $\mathbf{Y},\mathbf{Y}'\sim \mu_2$ are all independent. $F_{\mu}$ is cdf of $\mu$.
\begin{align}
	\mathbb{W}_p^k \ \text{ for } k\geq p,& \ \text{ where } \mathbb{W}_p^p(\mu_1,\mu_2):=\inf_{J(\mu_1,\mu_2)}\Expect_{X\sim \mu_1, Y\sim \mu_2}\|\mathbf{X}-\mathbf{Y}\|^p, \nonumber \\
	{\rm MMD}_\kappa^k \ (0<\kappa<2) \ \text{ for } k \geq 1,& \ \text{ where } {\rm MMD}_\kappa^2(\mu_1,\mu_2):=2\mathbb{E}\{\|\mathbf{X}-\mathbf{Y}\|\}^\kappa - \mathbb{E}\{\|\mathbf{X}-\mathbf{X}'\|\}^\kappa - \mathbb{E}\{\|\mathbf{Y}-\mathbf{Y}'\|\}^\kappa, \nonumber \\
	l_p^k \ \text{ for } k \geq 1,& \ \text{ where } l_p^p(\mu_1, \mu_2):= \int_{-\infty}^{\infty}\big|  F_{\mu_1}(z) - F_{\mu_2}(z) \big|^p \mathrm{d}z  \quad (d=1). \label{suitable_distances}
\end{align} 
Note that energy distance that we used in \eqref{Energy_Distance} is a special case of ${\rm MMD}_\kappa^k$ with $k=2$ and $\kappa=1$.

First, let us justify powered Wasserstein metric. Property \ref{translation_scale_sensitivity} can be verified from Section 3.2 of \cite{bellemare2017distributional}. Property \ref{squared_metric} holds, since this is a powered metric. As we mentioned beneath \eqref{relaxed_triangle_p}, we have $q=p-1$. Property \ref{mixture_inequality} can be verified by Kantorovich duality mentioned in Lemma C.5 of \cite{wu2023distributional}. Denoting Wasserstein-p metric as $\mathbb{W}_p$ and letting $\Gamma:=\{ (\psi, \phi): \psi, \phi:\mathbb{R}^d \rightarrow \mathbb{R}, \ \psi(\mathbf{x}) - \phi(\mathbf{y}) \leq  \|\mathbf{x} - \mathbf{y}\|^p\}$, we have 
\begin{gather*}
	\mathbb{W}_p^p \bigg\{ \int \mu(z) \mathrm{d}P(z), \int \nu(z) \mathrm{d}P(z) \bigg\} = \sup_{\psi, \phi \in \Gamma} \bigg\{ \Expect_{Y\sim \int \mu(z) \mathrm{d}P(z)} \psi(Y) - \Expect_{Y\sim \int \nu(z) \mathrm{d}P(z)} \phi(Y) \bigg\} \\
	= \sup_{\psi, \phi \in \Gamma} \Expect_{Z\sim P} \bigg\{ \Expect_{Y\sim \mu(Z)} \psi(Y) - \Expect_{Y\sim \nu(Z)} \phi(Y) \bigg\} 
	\leq \Expect_{Z\sim P} \mathbb{W}_p^p \big\{ \mu(Z), \nu(Z) \big\}.
\end{gather*}

Second, let us justify ${\rm MMD}_\kappa^k$. Property \ref{translation_scale_sensitivity} holds with $\beta_0=\kappa/2$. It is straightforward to show, by using the definition \eqref{suitable_distances}. Property \ref{squared_metric} holds with $q=0$, since ${\rm MMD}_\kappa$ is a metric. Property \ref{mixture_inequality} can be shown by using the fact that maximum mean discrepancy is in fact a norm between two mean embedding in RKHS $(\mathcal{H}_\kappa, \|\cdot\|_\kappa)$ \citep{gretton2012kernel} with kernel $k(\mathbf{x},\mathbf{y}) = \|\mathbf{x}\|^\kappa + \|\mathbf{y}\|^\kappa - \|\mathbf{x}-\mathbf{y}\|^\kappa$. Denote the mean embedding of $X\sim \mu_1(z)$ as $m_1(z)$ and $Y\sim \mu_2(z)$ as $m_2(z)$, we can derive the following by using the convexity of norm,
\begin{gather*}
	{\rm MMD}_\kappa \bigg\{ \int \mu_1(z) \mathrm{d}P(z), \int \mu_2(z) \mathrm{d}P(z) \bigg\} = \bigg\| \int m_1(z) \mathrm{d}P(z) - \int m_2(z) \mathrm{d}P(z) \bigg\|_\kappa  \leq \int \|m_1(z) - m_2(z)\|_\kappa \mathrm{d}P(z) \\
	= \int {\rm MMD}_\kappa \big\{ \mu_1(z), \mu_2(z) \big\}\mathrm{d}P(z).
\end{gather*}
Based on these properties, it is straightforward to see that $\energyone$ that we used in the main text satisfies all the properties with $\beta_0=1$ and $q=1$. Note that plugging in $K=2$ gives us the following special case,
\begin{align} \label{relaxed_triangle_pis2}
	\energyone\{\distL(X_0), \distL(X_2)\} \leq 2 \cdot[ \energyone \{  \distL(X_0) , \distL(X_1) \} + \energyone \{  \distL(X_1) , \distL(X_2) \} ].
\end{align}

Third, let us justify powered norm of cdf. Property \ref{translation_scale_sensitivity} is shown by Proposition 3.2 of \cite{odin2020dynamic} with $\beta_0=1/p$. Property \ref{mixture_inequality} can be shown in the same way with MMD, since it is based on a functional norm,
Property \ref{squared_metric} holds with $q=p-1$.

Finally, we can extend this into higher degree. Given that a powered metric $\eta=m^p$ (with $m$ being a probability metric) satisfies properties in \ref{properties_of_distance} with $\beta_0$ and $q=p-1$, we can also show that $m^k$ with $k\geq p$ is also a valid distance that satisfies the properties. Property \ref{translation_scale_sensitivity} can be shown to hold, by replacing $\beta_0$ with $\beta_0^{k/p}$. Property \ref{squared_metric} holds with $q=k-1$. Property \ref{mixture_inequality} can be shown as follows,
\begin{gather*}
	m^k\bigg( \int \mu_1(z)\mathrm{d}P(z), \int \mu_2(z)\mathrm{d}P(z) \bigg) = \bigg\{ m^p \bigg( \int \mu_1(z)\mathrm{d}P(z), \int \mu_2(z)\mathrm{d}P(z) \bigg) \bigg\}^{k/p} \\
	\leq \bigg(\Expect_{Z\sim P} \big\{ m^p(\mu_1(Z), \mu_2(Z)) \big\} \bigg)^{k/p} \leq \Expect_{Z\sim P} \big\{ m^k(\mu_1(Z), \mu_2(Z)) \big\}.
\end{gather*}

\subsection{Algorithm Of Single-step EBRM} \label{sec:EBRM_singlestep_algorithm}

Here is the algorithm for single-step EBRM \eqref{Biased_EStimator_onestep_old}.

\begin{algorithm}
	\caption{EBRM-single-step} \label{EBRM_singlestep_algorithm}
	\begin{algorithmic}
		\\
		\textbf{Input: } $\Fspace$, $\mathcal{D}=\{(s_i,a_i,r_i,s_i')\}_{i=1}^{N}$ \\
		\textbf{Output: } $\hat{\theta}$
		\State Estimate $\hat{b}_\mu$ and $\hat{p}$. \Comment{Refer to Equation (\ref{probabiliy_stateaction}).}
		\State Compute $\hat{\theta}=\arg\min_{\theta\in\Fspace}\energyonebarhat(\Upsilon_\theta, \Bellhat\Upsilon_\theta)$. \Comment{Refer to Equations (\ref{Biased_Esa}) and (\ref{Biased_EStimator_onestep_old}).}
	\end{algorithmic}
\end{algorithm}

\subsection{Explanation Of Sub-Gaussian Norm} \label{explanation_subgaussian}

Sub-Gaussianity can be quantified with sub-Gaussian norm $\|\cdot\|_{\psi_2}:\mathcal{P}(\mathbb{R})\rightarrow\mathbb{R} \ \text{or } \mathcal{P}(\mathbb{R}^d)\rightarrow \mathbb{R}$ (Definitions 2.5.6 and 3.4.1 by \citet{vershynin2018high}. The left and right are for random variable and random vector, respectively.
\begin{align} \label{subgaussian_norm_definition}
	\|X\|_{\psi_2} := \inf \bigg\{ t>0: \ \mathbb{E} \big\{ \exp(X^2/t^2) \big\} \leq 2 \bigg\} \ (d=1) \quad \& \quad \|\mathbf{X}\|_{\psi_2} := \sup_{\mathbf{x}\in\mathbb{R}^d:\|\mathbf{x}\|=1} \|\langle \mathbf{X},\mathbf{x}\rangle \|_{\psi_2}.
\end{align}
Sub-Gaussian norm is verified to be a valid norm in Exercise 2.5.7 suggested by \citet{vershynin2018high}. Random variable (vector) $\mathbf{X}$ is called sub-Gaussian if it satisfies $\|\mathbf{X}\|_{\psi_2} < \infty$. A lot of useful inequalilties, such as Dudley's integral inequality and and Hoeffding's inequality (Theorems 8.1.6, 2.6.2 of \cite{vershynin2018high}) are based on sub-Gaussianity assumption.

\subsection{Bounding Expectation-difference With Wasserstein-1 Metric} \label{Wassersup_diamF_satisfication}

Let $c\in\mathbb{R}^d$, $\gamma_1,\gamma_2\in[0,1]$, and $(s,a),(\tilde{s},\tilde{a})\in\SAspace$ be arbitrary, $Z_i(s,a)\sim \Upsilon_i(s,a)$ with $Z_1(s,a)$, $Z_2(\tilde{s},\tilde{a})$ and $Z_3(s,a)$, $Z_4(\tilde{s},\tilde{a})$ being pairwise independent. Letting $J_{13}$ be the possible dependence structures (or joint distributions) between marginal distributions of $Z_1(s,a)$ and $Z_2(s,a)$, and $J_{24}$ be that between $Z_3(\tilde{s},\tilde{a})$ and $Z_4(\tilde{s},\tilde{a})$, we have
\begin{align*}
	\bigg| \Expect\|c+\gamma_1 &Z_1(s,a) - \gamma_2 Z_2 (\tilde{s},\tilde{a}) \| - \Expect\|c+\gamma_1 Z_3(s,a) - \gamma_2 Z_4 (\tilde{s},\tilde{a}) \| \bigg| \\
	&= \inf_{J_{13},J_{24}} \bigg| \Expect\|c + \gamma_1 Z_1(s,a) - \gamma_2 Z_2(\tilde{s}, \tilde{a})\| - \Expect\|c + \gamma_1 Z_3(s,a) - \gamma_2 Z_4(\tilde{s}, \tilde{a})\| \bigg| \nonumber \\
	&\leq \inf_{J_{13},J_{24}} \Expect\|c + \gamma_1 Z_1(s,a) - \gamma_2 Z_2(\tilde{s}, \tilde{a}) - c - \gamma_1 Z_3(s,a) + \gamma_2 Z_4(\tilde{s}, \tilde{a}) \|\\
	&= \inf_{J_{13}} \Expect \| \gamma_1  Z_1(s,a) - \gamma_1 Z_3(s,a)\| + \inf_{J_{24}} \Expect \| \gamma_2 Z_2(\tilde{s}, \tilde{a}) - \gamma_2 Z_4(\tilde{s}, \tilde{a})\|    \nonumber \\
	&\leq \gamma_1\cdot \Wassersup(\Upsilon_1,\Upsilon_3) + \gamma_2\cdot \Wassersup(\Upsilon_2,\Upsilon_4),
\end{align*}
where the second last line holds by the definition of Wasserstein-1 metric.

\subsection{Proof Of Theorem \ref{Realizable_Final_Theorem}} \label{Proof_Realizable_Final_Theorem}

Throughout the proof, we will use $C>0$, $C_k>0$ ($k\in\mathbb{N}$) to denote appropriate universal constants.

In addition, we define the minimizer of Bellman residual $\theta_*:\in \arg\min_{\theta\in\Fspace} \energyonebar(\Upsilon_\theta, \Bellopt \Upsilon_\theta)$.
Since $\Upsilon_\pi=\Upsilon_{\theta'}$ for some $\theta'\in\Fspace$ by Assumption \ref{realizable}, we have $\energyonebar(\Upsilon_{\theta'},\Bellopt\Upsilon_{\theta'})= \energyonebar(\Upsilon_\pi, \Bellopt\Upsilon_\pi)=0$, thereby becoming the minimizer of Bellman residual. Then we can let $\theta_*=\theta'$, and have $\energyonebar(\Upsilonstar,\Bellopt\Upsilonstar)=0$, that is $\Upsilonstar=\Bellopt \Upsilonstar$.

Lastly, we would like to allow abuse of notation $\Wassersup(\theta_1,\theta_2):=\Wassersup(\Upsilon_{\theta_1}, \Upsilon_{\theta_2})$, with which we will define the diameter $\diamF:=\sup_{\theta_1,\theta_2\in\Theta}\{\Wassersup(\theta_1,\theta_2)\}$. Based on the metric $\expectationbounder$, we will quantify the model complexity with \textit{covering number} (Definition 4.2.2 by \citet{vershynin2018high}). With $N_{\expectationbounder}(\theta,t):=\{\theta'\in\Fspace: \expectationbounder(\theta',\theta)<t \}$ being $t$-neighborhood of $\theta\in\Fspace$, we define the covering number as $\mathcal{N}(\Fspace, \expectationbounder, t) :=\min \{ \tilde{M} \in \mathbb{N}: \exists \theta_1,\cdots, \theta_{\tilde{M}} \ \text{s.t. } \Fspace  \subset \cup_{i=1}^{\tilde{M}} N_{\expectationbounder}(\theta_i,t) \}$.

However, note that $\Wassersup$ in the previous paragraph can be replaced with a general distance measure $\expectationboundergeneralized$, as long as it satisfies the first two properties suggested in Appendix \ref{Wassersup_replacing}.

\subsubsection{Decomposition Into Two Discrepancies}  \label{gamma_and_delta}

Defining $\Gamma_N$ and $\Delta_N$ as 
\begin{gather*}
	\Gamma_N := \supf \energyonebar (\Bellopt \Upsilon_\theta, \Bellhat \Upsilon_\theta)  \ \ \& \ \ \Delta_N := \supf \bigg| \energyonebar ( \Upsilon_\theta, \Bellhat \Upsilon_\theta  ) -  \energyonebarhat ( \Upsilon_\theta, \Bellhat \Upsilon_\theta  ) \bigg| ,
\end{gather*}
we can decompose the term $\energyonebar(\Upsilonhat, \Bellopt \Upsilonhat)$ as follows. 
\begin{align}
	\energyonebar(\Upsilonhat, \Bellopt \Upsilonhat) & \leq 2\cdot \bigg\{ \energyonebar(\Upsilonhat, \Bellhat \Upsilonhat) + \energyonebar(\Bellhat \Upsilonhat, \Bellopt \Upsilonhat)  \bigg\} \nonumber \\
	& \leq 2\cdot \bigg\{ \energyonebarhat(\Upsilonhat, \Bellhat \Upsilonhat) + \bigg| \energyonebarhat(\Upsilonhat, \Bellhat \Upsilonhat) - \energyonebar(\Upsilonhat, \Bellhat \Upsilonhat) \bigg|  
	+ \energyonebar(\Bellhat \Upsilonhat, \Bellopt \Upsilonhat) \bigg\} \nonumber \\
	& \leq 2\cdot \bigg\{ \energyonebarhat(\Upsilonstar, \Bellhat \Upsilonstar) + \Delta_N + \Gamma_N  \bigg\}  \nonumber \\
	& \leq 2\cdot \bigg\{ \energyonebar(\Upsilonstar, \Bellhat \Upsilonstar)  + \bigg| \energyonebarhat(\Upsilonstar, \Bellhat \Upsilonstar) - \energyonebar(\Upsilonstar, \Bellhat \Upsilonstar)  \bigg| + \Delta_N + \Gamma_N \bigg\} \label{relaxing_point_singlestep} \\
	& \leq 2\cdot \bigg\{ \energyonebar(\Bellopt \Upsilonstar, \Bellhat \Upsilonstar) + 2\Gamma_N + \Delta_N  \bigg\} \qquad \text{ by Assumption \ref{realizable}}  \nonumber \\
	& \leq 4\cdot (\Gamma_N + \Delta_N).  \nonumber 
\end{align}

Combined with the result \eqref{Fundamental_Realizable_specific} of Theorem \ref{Fundamental_Realizable} that requires Assumption \ref{RN_derivative}, it leads to the following bound,
\begin{gather}
	\energyonebar(\Upsilonhat, \Upsilon_\pi) \leq  8 C_{{\rm sup}} B_1(\gamma)\cdot (\Gamma_N + \Delta_N), \quad \mbox{where}  \label{Decomposition}\\ 
	B_1(\gamma):=B_1(\gamma;1,1)=\frac{1}{2(1-\gamma)}\sum_{k=1}^{\infty} 4^k \gamma^{2^{k-1}-1} \quad \mbox{by definition above \eqref{Fundamental_Realizable_specific}} \nonumber
\end{gather}
since we have verified $\beta_0=1$ in \ref{Proof_squared_metric}. Now it suffices to bound $\Gamma_N$ and $\Delta_N$, which will be referred to as \textit{Bellman discrepancy} and \textit{state-action discrepancy} to indicate the sources of error, $\Bellopt$ and $\probsa$, respectively. Before we proceed, we list several properties of sub-Gaussian norm (\ref{subgaussian_norm_definition}) that we will utilize in our analysis. Corresponding proofs can be found in Section \ref{Proof_psi2_properties}.
\begin{remark} \label{psi2_properties}
	(Properties of sub-Gaussian norm) We have the following properties regarding sub-Gaussian norm, 
	\begin{enumerate}%
		\item For $X\sim Ber(p)$, we have $\|X\|_{\psi_2}\leq 1/\sqrt{\log2}$.
		\item For a constant $c\in \mathbb{R}$, $\|c\|_{\psi_2}=c/\sqrt{\log2}$.
		\item For a random variable $X\in\mathbb{R}$, $\|\mathbb{E}(X)\|_{\psi_2} \leq \|X\|_{\psi_2}$ holds.
		\item For a random vector $\mathbf{X}\in \mathbb{R}^d$, $\big\| \|\mathbf{X}\| \big\|_{\psi_2} \leq d \big\|\mathbf{X}\big\|_{\psi_2}$ holds.
		\item For a random variable $X\in\mathbb{R}$, $\|X-\mathbb{E}(X)\|_{\psi_2} \leq C \|X\|_{\psi_2}$ holds.
		\item For iid mean-zero random variables $X_1,\cdots, X_n$, we have $\| \frac{1}{n} \sum_{i=1}^{n} X_i \|_{\psi_2}\leq \frac{C}{\sqrt{n}}\cdot \|X_1\|_{\psi_2}.$
	\end{enumerate}
\end{remark}

\subsubsection{Conditioning On Sufficient Sample Size For Each State-action Pair} \label{Conditioning_Event}

Prior to bounding $\Gamma_N$ and $\Delta_N$ of \eqref{Decomposition}, we will first condition upon an event where each state-action pair is observed \textit{sufficiently many times}. 

Before we proceed, we should note that the given probability space $(\Omega,\Sigma, \PROB)$ can be factorized into two stages. Letting $\mathbf{N}=(\Nsa)_{s,a\in\SAspace}\in\mathbb{R}^{\SAspace}$ to be a random vector that indicates the observed number of samples for each state-action pair, we can see that $(\Omega,\Sigma, \PROB)$ consists of two consecutive probability events denoted as follows,
\begin{align}
	\text{Stage 1:}& \ (\Omega_{\SAspace}, \Sigma_{\SAspace}, \PROB_{\SAspace}) \Rightarrow \text{determines sampling of state-action pairs } S_i,A_i\sim b_\mu,  \label{probability_stage1_2} \\
	\text{Stage 2:}& \ (\Omega^{(\mathbf{N})}, \Sigma^{(\mathbf{N})}, \PROBN) \Rightarrow \text{conditioned on }(S_i,A_i), \text{ determines } R_i,S_i' \sim p(\cdots|S_i,A_i). \nonumber
\end{align}
This implies that having sufficiently many observations for each $s,a$ is solely associated with probability space of Stage 1. Now let us discuss how ``sufficiently large'' $\Nsa$ is characterized \eqref{Conditional_Omega}.

Temporarily assuming $N\geq 2$, we can divide the data $\mathcal{D}=\{(S_i,A_i,R_i,S_i')\}_{i=1}^{N}$ into two halves, 
\begin{align*}
	\mathcal{D}_1=\{(S_i,A_i,R_i,S_i')\}_{i=1}^{\lfloor N/2 \rfloor} \quad \mbox{and} \quad \mathcal{D}_2=\{(S_i,A_i,R_i,S_i')\}_{i=\lfloor N/2 \rfloor +1}^{N}.
\end{align*}
Note that we denoted observations $(S_i,A_i,R_i,S_i')$ in capital letters, so as to indicate that they are random objects. Based on this, we define the following notations based on \eqref{probabiliy_stateaction},
\begin{align*}
	\probvec = \big(\probsa\big)_{s,a\in\SAspace}\in [0,1]^{\SAspace} \ \ &\& \ \ \probvechat = \big( \probsahat \big)_{s,a\in\SAspace}\in [0,1]^{\SAspace} , %
\end{align*}
and it is straightforward to see $\probvechat = \lfloor N/2 \rfloor / N \cdot \probvechatone + (N-\lfloor N/2 \rfloor)/ N \cdot \probvechattwo$, 
where each term in the RHS is sample mean based on $\mathcal{D}_1$ and $\mathcal{D}_2$, i.e. $\probvechatone = \frac{1}{N/2}\sum_{i=1}^{N/2} \mathbf{y}_i^{(1)}$ and $\probvechattwo = \frac{1}{N-\lfloor N/2 \rfloor}\sum_{i=1}^{N-\lfloor N/2 \rfloor} \mathbf{y}_i^{(2)}$
with $\mathbf{y}_i^{(j)}\in\{0,1\}^{\SAspace}$ being indicators having 1 only at the state-action pair that $(S_i,A_i)$ correspond to in $\mathcal{D}_j$ ($j=1,2$). Within Stage 1 probability space \eqref{probability_stage1_2}, we define the following subset with given $\epsilon\in(0,1]$,
\begin{align} \label{Conditional_Omega}
	\Omegasubsetstageone:= \bigg\{ \omega\in\Omega_{\SAspace} \ \bigg| \  \| \probvechatone - \probvec \| < \frac{1}{2}p_{{\rm min}}\cdot \epsilon \quad \mbox{and} \quad \| \probvechattwo - \probvec \| < \frac{1}{2}p_{{\rm min}}\cdot \epsilon \bigg\},
\end{align}
under which we can verify that following holds (proofs in \ref{Proof_Fact1to3}), 
\begin{align} \label{Fact1to3}
	\text{Fact 1: }&\probsahat =\frac{\Nsa}{N} \in \bigg[ \frac{1}{2}\probsa, \frac{3}{2}\probsa \bigg] \text{ for } \forall s,a\in\SAspace,\\
	\text{Fact 2: }&\Nsa\geq 2 \text{ for } \forall s,a\in\SAspace, \nonumber \\
	\text{Fact 3: }&\|\hat{\mathbf{p}} - \mathbf{p}\| < \frac{1}{2}p_{{\rm min}}\cdot \epsilon. \nonumber
\end{align}

Throughout the following subsections \ref{Gamma_bounding_realizable} and \ref{Delta_bounding_realizable} where we shall bound $\Gamma_N$ and $\Delta_N$, we will resort to conditional probability measure $\PROBN(\cdots):=\PROB(\cdots|\mathbf{N})$ along with its corresponding sub-Gaussian norm $\|\cdot\|_{\psiN}$. In other words, we will consider $\Nsa$ to be fixed (non-random), assuming that Facts (\ref{Fact1to3}) are satisfied, and later calculate its unconditional probability with $\PROB$ in \ref{realizable_finalizing_the_bound} by Inequality (\ref{prob_conditional_bound}).

\subsubsection{Bounding Bellman Discrepancy} \label{Gamma_bounding_realizable}

Once more, we would like to emphasize that $\Nsa$ are fixed, and Facts (\ref{Fact1to3}) hold. The probability space we are dealing with in this subsection is Stage 2 probability space \eqref{probability_stage1_2}.

Let us define the following stochastic process that can be used in bounding Bellman discrepancy $\Gamma_N$:
\begin{gather} 
	X_\theta:= \energyonebar\big( \Bellhat \Upsilon_\theta, \Bellopt \Upsilon_\theta \big) \quad \mbox{and} \quad X_\theta(s,a):=\mathcal{E}\bigg\{ \Bellhat\Upsilon_\theta(s,a), \Bellopt \Upsilon_\theta(s,a) \bigg\}, \nonumber \\
	\therefore \ \Gamma_N = \supf X_\theta \leq \supf \bigg|X_\theta -X_{\theta_0}\bigg| + X_{\theta_0}, \label{Gamma_sup_and_f0}
\end{gather}
where $\theta_0\in\Fspace$ is a fixed value that will be chosen at the later in the proof. 

First, let us handle the supremum term $\supf |X_\theta- \Xfzero|$ of Decomposition (\ref{Gamma_sup_and_f0}) with Dudley's integral inequality (Theorem 8.1.6 of \cite{vershynin2018high}). Due to $X_\theta=\sumsa \probsa\cdot X_\theta(s,a)$,
we first need to bound the term $\|\Xone(s,a) - \Xtwo(s,a)\|_{\psiN}$. Towards that end, we can simplify it as follows,
\begin{align} %
	X_\theta (s,a) = \frac{2}{\Nsa}\sumNsa W_i^\theta - \frac{1}{\Nsa^2}\sumNsa\sum_{j=1}^{\Nsa} W_{ij}^\theta, \nonumber
\end{align}
where $W_i^\theta$ and $W_{ij}^\theta$ are the random variables that have following realizations,
\begin{gather}
	w_{i}^\theta := \Expect \| R_{\alpha} + \gamma Z_{\alpha} (\Saprime, \Aaprime; \theta) -r_i -\gamma Z_{\beta}(s_i',A_i';\theta)  \| - \Expect \| R_{\alpha} + \gamma Z_{\alpha} (\Saprime, \Aaprime; \theta) - R_{\beta} -\gamma Z_{\beta}(S_{\beta}',A_{\beta}';\theta)  \|, \nonumber \\ %
	w_{ij}^\theta := \Expect \|r_i + \gamma Z_{\alpha}(s_i',A_i'; \theta) - r_j - \gamma Z_{\beta}(s_j',A_j';\theta)\| - \Expect \| R_{\alpha} + \gamma Z_{\alpha} (\Saprime, \Aaprime; \theta) - R_{\beta} -\gamma Z_{\beta}(S_{\beta}',A_{\beta}';\theta)  \| . \nonumber
\end{gather}
where $Z(s,a;\theta)\sim\Upsilon_\theta(s,a)$,  $(R,S')\sim p(\cdots|s,a)$, $(\hat{R},\hat{S}')\sim \hat{p}(\cdots|s,a)$, $A'\sim \pi(\cdot|S')$, $\hat{A}'\sim \pi(\cdot|\hat{S}')$, and having different subscripts ($\alpha$ or $\beta$) means they are independent, although they may follow the same distribution(s). Since we have
\begin{align*}
	\Expect(W_i^{\theta})=0, \ \ \Expect(W_{ij}^{\theta})=0 \ \text{if }i\neq j, \ \ \Expect(W_{ii}^{\theta})\neq 0,
\end{align*}
we should further obtain
\begin{align}
	\|\Xone&(s,a) - \Xtwo(s,a)\|_{\psiN} \leq \frac{\Nsa - 1}{\Nsa}\cdot \bigg\| \frac{1}{\Nsa\cdot (\Nsa-1)} \sumijNsa (\Wijfone - \Wijftwo) \bigg\|_{\psiN} \label{psi2_decomposition_Realizable} \\
	& \qquad + \bigg\| \frac{2}{\Nsa} \sumNsa (\Wifone - \Wiftwo) \bigg\|_{\psiN} + \bigg\| \frac{1}{\Nsa^2} \sumNsa (\Wiifone - \Wiiftwo) \bigg\|_{\psiN}, \nonumber
\end{align}
and we will bound each term one by one. Before we begin with the first term, we would like to introduce a useful trick that will be used repetitively throughout the proof. First, it is easy to see $|\Wijfone - \Wijftwo| \leq 4\gamma \cdot \expectationbounder(\theta_1,\theta_2)$ by Appendix \ref{Wassersup_diamF_satisfication}.
Defining another random variable that satisfies following based on Remark \ref{psi2_properties},
\begin{gather}
	\tilde{W}_{ij}^{\theta}:= \frac{1}{2}\cdot (W_{ij}^{\theta} + W_{ji}^{\theta}) \ \ \text{where} \ \ 1\leq i < j \leq \Nsa, \quad 
	\therefore \ \|\Wtildeonetwofone - \Wtildeonetwoftwo\|_{\psiN} \leq  \frac{4\gamma}{\sqrt{\log 2}}\cdot \expectationbounder(\theta_1,\theta_2). \nonumber  %
\end{gather}
Note that we can rewrite $\sumijNsa (\Wijfone - \Wijftwo)=2\cdot\sum_{i<j}^{\Nsa} (\Wtildeijfone - \Wtildeijftwo)$, and the terms that are being added are not independent. Towards that end, we divide our cases into two, when $\Nsa\geq2$ is an even number or an odd number. When $\Nsa$ is even, we can directly use Lemma S4 of \citet{wang2022low} to group $\big\{(i,j) : 1\leq i < j \leq \Nsa \big\}$ into $(\Nsa-1)$ groups $G_k \ (1\leq k \leq \Nsa-1)$, each of which contains $|G_k|=\Nsa/2$ pairs of $(i,j)$, with no pair overlapping in any component. Then we have the following by Remark \ref{psi2_properties},
\begin{gather}
	\bigg\| \frac{1}{\Nsa\cdot (\Nsa-1)} \sumijNsa (\Wijfone - \Wijftwo) \bigg\|_{\psiN} \leq \bigg\| \frac{1}{\Nsa-1} \sum_{k=1}^{\Nsa-1} \frac{1}{\Nsa/2} \sum_{(i,j)\in G_k} (\Wtildeijfone - \Wtildeijftwo) \bigg\|_{\psiN} \nonumber \\
	\leq \bigg\| \frac{1}{\Nsa/2} \sum_{(i,j)\in G_1} (\Wtildeijfone - \Wtildeijftwo) \bigg\|_{\psiN} \leq \frac{C_1}{\sqrt{\Nsa}}\cdot \|\Wtildeonetwofone - \Wtildeonetwoftwo\|_{\psiN} . \label{Wtilde_psi2_odd}
\end{gather}
Now let us assume that $\Nsa\geq 2$ is an odd number, which automatically gives us $\Nsa\geq 3$. Then the term can be bounded as
\begin{gather}
	\leq \frac{\Nsa-2}{\Nsa}\cdot \bigg\|  \frac{2}{(\Nsa-1)\cdot (\Nsa-2)} \sum_{i<j}^{\Nsa-1} (\Wtildeijfone - \Wtildeijftwo) \bigg\|_{\psiN} + \frac{2}{\Nsa}\cdot \|\Wtildeonetwofone - \Wtildeonetwoftwo\|_{\psiN} \nonumber \\
	\leq \frac{C_2}{\sqrt{\Nsa}}\cdot\|\Wtildeonetwofone - \Wtildeonetwoftwo\|_{\psiN} \ \ \ \text{by (\ref{Wtilde_psi2_odd}) since } \Nsa-1 \text{ is an even number.} \nonumber
\end{gather}
where we used $\Nsa-1\geq \frac{\Nsa}{2}$ for $\Nsa\geq 3$ in the last line. Then, applying the same logic to the second and third terms of Inequality (\ref{psi2_decomposition_Realizable}), we can apply the same trick to obtain the following based on Fact \eqref{Fact1to3},
\begin{align*}
	\|\Xone &- \Xtwo\|_{\psiN} \leq \frac{C_9\gamma}{\sqrt{N}}\cdot \sumsa \sqrt{\probsa}\cdot \expectationbounder(\theta_1,\theta_2) .
\end{align*}
Without loss of generality, we can assume that separability holds. In addition, $\Wassersup$ is proved to be a metric by \citet{bellemare2017distributional} in their Lemma 2. Therefore, by Assumption \ref{bounded}, we can apply Dudley's Integral Inequality (Theorem 8.1.6 of \cite{vershynin2018high}) to obtain the following for $\forall u>0$,
\begin{gather}
	\PROBN\bigg[ \supf \bigg|X_\theta - X_{\theta_0}\bigg| \leq \frac{C_{10}\gamma}{\sqrt{N}} \cdot \sumsa\sqrt{\probsa}\cdot \bigg\{ \metricentropyexpectatonbounderFspace \nonumber \\
	+ u\cdot \diamF \bigg\} \bigg] \geq 1-2\exp(-u^2). \label{Gamma_supremum_part}
\end{gather}

The next part is bounding the term $X_{\theta_0}$ of Decomposition (\ref{Gamma_sup_and_f0}). We first fix a state-action pair $s,a\in\SAspace$, we can obtain the following decomposition,
\begin{align*}
	\Xfzero(s,a) &\leq 2\cdot \bigg| \frac{1}{\Nsa}\sumNsa \Wifzero \bigg| + \frac{\Nsa-1}{\Nsa}\cdot \bigg| \frac{1}{\Nsa\cdot (\Nsa-1)} \sumijNsa \Wijfzero \bigg| + \frac{1}{\Nsa^2}\cdot \bigg| \sumNsa \Wiifzero \bigg|. %
\end{align*}

We use the following
\begin{align}
	\big\| \Expect\|R(s,a)\|\big\|_{\psi_2} \leq   \big\| \|R(s,a)\| \big\|_{\psi_2}\leq d\cdot \|R(s,a)\|_{\psi_2} \ \ \text{by Remark \ref{psi2_properties}}, \label{reward_psi_bounding}
\end{align}
and the following lemma
that is proved in \ref{Proof_Hoeffding_twotuple}.
\begin{lemma}\label{Hoeffding_twotuple} Given $X_1,\cdots,X_N\sim iid \ (N\geq 2)$, let $X_{ij}:=h(X_i,X_j)$ for some bivariate function $h(\cdot,\cdot)$, and assume that $\|X_{12}\|_{\psi_2} < \infty$ holds. Then we have the following inequality for $\forall \epsilon>0$,
	\begin{align*}
		\PROB\bigg\{ \bigg| \frac{1}{N(N-1)}\sumij X_{ij} - \Expect(X_{12}) \bigg| \geq \epsilon \bigg\} \leq 4N\cdot \exp\bigg\{ \frac{-C \cdot N\cdot \epsilon^2}{\|X_{12}-\Expect(X_{12})\|_{\psi_2}^2} \bigg\}.
	\end{align*}
\end{lemma}

By applying Theorem 2.6.2 of \cite{vershynin2018high} and Lemma \ref{Hoeffding_twotuple}, we obtain the following, skipping all technical details. For arbitrary $\epsilon_1>0$, 
\begin{align}
	&\PROBN\bigg\{ X_{\theta_0}\leq 2\epsilon_1 + \frac{1}{N}\cdot\frac{8}{p_{{\rm min}}}\cdot \bigg( \supsa \Expect\|R(s,a)\| + \gamma\cdot \supsa \Expect\|Z(s,a;\theta_0)\| \bigg) \bigg\} \nonumber \\
	&\geq\PROBN\bigg\{ \supsa X_{\theta_0}(s,a)\leq 2\epsilon_1 + \frac{1}{N}\cdot\frac{8}{p_{{\rm min}}}\cdot \bigg( \supsa \Expect\|R(s,a)\| + \gamma\cdot \supsa \Expect\|Z(s,a;\theta_0)\| \bigg) \bigg\} \nonumber \\
	&\geq 1-(2|\SAspace|+6N)\cdot \exp\bigg\{ \frac{-C_{20}\cdot p_{{\rm min}}\cdot N \cdot \epsilon_1^2}{\big(\supsa \big\| \|R(s,a)\| \big\|_{\psi_2} + \gamma \cdot \supsa \Expect\|Z(s,a;\theta_0)\|\big)^2} \bigg\}. \label{Gamma_fzero_part}  
\end{align}
Note that we obtained a bound for $\supsa X_{\theta_0}(s,a)$, which is one step further than $X_{\theta_0}$. This shall be later used in proof of Lemma \ref{Fmhat_convergence} for non-realizable scenario, which is suggested in \ref{Proof_Fmhat_convergence}.

Now we can combine the bounds (\ref{Gamma_supremum_part}) and (\ref{Gamma_fzero_part}) to take up Decomposition (\ref{Gamma_sup_and_f0}) as following for $\forall u>0, \ \epsilon_1>0, \ N\geq2$, 
\begin{gather}
	\PROBN\bigg\{ \Gamma_N \leq \frac{C_{10}\gamma}{\sqrt{N}} \cdot \sumsa\sqrt{\probsa}\cdot \bigg( \metricentropyexpectatonbounderFspace + u\cdot \diamF \bigg) \nonumber \\
	+ 2\epsilon_1 + \frac{1}{N}\cdot\frac{8}{p_{{\rm min}}}\cdot \bigg( \supsa \Expect\|R(s,a)\| + \gamma\cdot \supsa \Expect\|Z(s,a;\theta_0)\| \bigg) \bigg\} \nonumber \\
	\geq 1-2\exp(-u^2) - (2|\SAspace|+6N) \cdot \exp\bigg\{ \frac{-C_{21}\cdot p_{{\rm min}}\cdot N \cdot \epsilon_1^2}{d^2\cdot \Rsapsisupsq + \gamma^2\cdot \Zsasupsq} \bigg\}, \label{Gamma_Realizable}
\end{gather}
where the second last inequality holds by Inequality (\ref{reward_psi_bounding}) and the technique $(a+b)^2\leq 2(a^2+b^2)$. We should always remember that we are conditioning on the event $\Omegasubsetstageone$ defined in Definition (\ref{Conditional_Omega}).

\subsubsection{Bounding State-action Discrepancy} \label{Delta_bounding_realizable}

This time, we will bound state-action discrepancy $\Delta_N$ that occurs due to the estimation error of $\probsa$. As warned in the last paragraph of \ref{Conditioning_Event}, we are still assuming $\Nsa$ to be fixed, satisfying Facts \eqref{Fact1to3}. Accordingly, we only deal with Stage 2 probability space \eqref{probability_stage1_2} with the conditional probability measure $\PROBN$.

With $\mathbf{p}$ and $\hat{\mathbf{p}}$ defined in \ref{Conditioning_Event} and $\epsilon>0$ being the value specified in Definition (\ref{Conditional_Omega}), and defining $\|\mathbf{x}\|_1:=\sum_{i=1}^{p_0} |x_i|$ for $\forall\mathbf{x}=(x_1,\cdots,x_{p_0})^\intercal$, we can derive the following based on $(x_1+\cdots + x_{p_0})^2\leq p_0\cdot (x_1^2+\cdots x_{p_0}^2)$,
\begin{align*}
	\sumsa \bigg| \probsahat - \probsa \bigg| = \|\mathbf{p}-\hat{\mathbf{p}}\|_1 \leq\sqrt{|\SAspace|}\cdot \|\mathbf{p}-\hat{\mathbf{p}}\| \leq \frac{1}{\sqrt{p_{\rm min}}}\cdot \|\probvec - \probvechat\|,
\end{align*}
where we used $p_{\rm min}\leq 1/|\SAspace|$. Then we have the following extension for $\Delta_N$ defined in Section \ref{gamma_and_delta},
\begin{align*}
	\Delta_N 
	&\leq \frac{\sqrt{p_{{\rm min}}}}{2}\cdot \epsilon\cdot  \supf\supsa \energyone\bigg\{ \Upsilon_\theta(s,a), \Bellhat \Upsilon_\theta(s,a) \bigg\} \ \ \text{by Facts (\ref{Fact1to3})} . %
\end{align*}
Now let us handle the supremum term. Letting $R_i(s,a) \ (1\leq i \leq \Nsa)$ be the reward vectors observed conditioned on $s,a$, we have the following hold based on the notations introduced in \eqref{Biased_Esa},
\begin{align}
	\supf\supsa \energyone\bigg\{ \Upsilon_\theta(s,a), \Bellhat \Upsilon_\theta(s,a) \bigg\}
	\leq 4(1+\gamma)\cdot \supf \supsa \Expect\|Z(s,a;\theta)\| + 4\cdot \supsa \bigg\{ \frac{1}{\Nsa}\cdot \sumNsa \|R_i(s,a)\| \bigg\}. \nonumber %
\end{align}
The first term can be bounded as follows, using the property of $\expectationbounder$ introduced in Appendix \ref{Wassersup_diamF_satisfication},
\begin{gather}
	\supf \supsa \Expect\|Z(s,a;\theta)\|= \supf \bigg\{ \supsa \Expect\|Z(s,a;\theta)\| - \supsa\Expect\|Z(s,a;\theta_0)\| \bigg\} + \supsa \Expect\|Z(s,a;\theta_0)\| \nonumber \\
	\leq \supf \expectationbounder(\theta,\theta_0) + \supsa \Expect\|Z(s,a;\theta_0)\| \leq \diamF + \supsa \Expect\|Z(s,a;\theta_0)\|. \nonumber %
\end{gather}
Now let $\epsilon_2>0$ be arbitrary. For a fixed $s,a\in\SAspace$, we can derive the following based on previous techniques and \eqref{reward_psi_bounding},
\begin{align}
	\PROBN\bigg[ \supsa \bigg\{& \frac{1}{\Nsa}\sumNsa\|R_i(s,a)\| \bigg\} \geq \supsa\Expect\|R(s,a)\| + \epsilon_2 \bigg] \leq 2|\SAspace|\cdot \exp\bigg\{ \frac{-C_{23} \cdot p_{{\rm min}} \cdot N\cdot \epsilon_2^2}{ d^2\cdot\Rsapsisupsq } \bigg\}. \label{supRsa_bounding}
\end{align}
Then this gives us
\begin{gather}
	\PROBN\bigg[ \Delta_N \leq 2\sqrt{p_{{\rm min}}}\cdot \epsilon \cdot \bigg\{ (1+\gamma)\cdot \bigg( \diamF + \supsa \Expect\|Z(s,a;\theta_0)\| \bigg) \nonumber \\
	+ \supsa \Expect\|R(s,a)\| + \epsilon_2 \bigg\} \bigg] \geq 1- 2|\SAspace|\cdot \exp\bigg\{ \frac{-C_{23} \cdot p_{{\rm min}} \cdot N\cdot \epsilon_2^2}{ d^2\cdot\Rsapsisupsq } \bigg\}. \label{Delta_Realizable}
\end{gather}

\subsubsection{Finalizing The Bound} \label{realizable_finalizing_the_bound}

Recall that we defined $\Omegasubsetstageone$ \eqref{Conditional_Omega} in \ref{Conditioning_Event} where the samples are collected sufficiently many for each $s,a$. Assuming this, we have bounded $\Gamma_N$ and $\Delta_N$ throughout \ref{Gamma_bounding_realizable} and \ref{Delta_bounding_realizable}, each in \eqref{Gamma_Realizable} and \eqref{Delta_Realizable}. Simply put, letting $E\subset\Omega$ be the event where $\Gamma_N$ and $\Delta_N$ simultaneously achieve the specified bounds \eqref{Gamma_Realizable} and \eqref{Delta_Realizable} can be understood as $\PROB(E|\Omegasubsetstageone)$. Skipping detailed derivations, we get
\begin{align}
	\PROB(E)&\geq \PROB(E\cap \Omegasubsetstageone) \geq 1-(1-\PROB(\Omegasubsetstageone)) - (1-\PROB(E|\Omegasubsetstageone)). \label{prob_conditional_bound}
\end{align}
According to \eqref{probability_stage1_2}, it may be more rigorous to denote $\PROB_{\SAspace}(\Omegasubsetstageone)$ instead of $\PROB(\Omegasubsetstageone)$ in \eqref{prob_conditional_bound}, but we allowed using $\PROB(\Omegasubsetstageone)$ since $\PROB$ is an integrated probability measure of both $\PROB_{\SAspace}$ and $\PROBN$.

Term $\PROB(E|\Omegasubsetstageone)$ can be derived by aggregating two bounds \eqref{Gamma_Realizable} and \eqref{Delta_Realizable}. That is, for $\forall \epsilon_1>0, \ \epsilon_2>0, \ u>0$,
\begin{align*}
	\PROB(E|\Omegasubsetstageone) &\geq 1-2|\SAspace|\cdot \exp\bigg\{ \frac{-C_{23} \cdot p_{{\rm min}} \cdot N\cdot \epsilon_2^2}{ d^2\cdot\Rsapsisupsq } \bigg\} -2\exp(-u^2) \\
	& - (2|\SAspace|+6N) \cdot \exp\bigg\{ \frac{-C_{21}\cdot p_{{\rm min}}\cdot N \cdot \epsilon_1^2}{d^2\cdot \Rsapsisupsq + \gamma^2\cdot \Zsasupsq} \bigg\}.
\end{align*}

Then it remains for us to calculate $\PROB(\Omegasubsetstageone)$ in \eqref{prob_conditional_bound}, and we should assume $\epsilon\in(0,1]$ as mentioned in Definition (\ref{Conditional_Omega}). Towards that end, we use the following lemma that we proved in \ref{Proof_Vector_Bernstein_multinomial},
\begin{lemma}\label{Vector_Bernstein_multinomial}
	For $\mathbf{X}_i \buildrel iid \over \sim {\rm Multinomial}(n, \probvec)$ with $\probvec = (p_1,\cdots, p_H)^\intercal$ with $\sum_{h=1}^{H}p_h=1$, we have the following for $\forall \epsilon>0$,
	\begin{align*}
		\PROB\big( \|\probvechat - \probvec\| \geq \epsilon \big) \leq \exp\bigg(\frac{1}{4}\bigg)\cdot \exp\bigg( \frac{-n\cdot \epsilon^2}{32} \bigg) \ \ \text{where } \  \probvechat = \frac{1}{n}\sum_{i=1}^{n} \mathbf{X}_i.
	\end{align*}
\end{lemma}
If $N\geq 2$ holds (as we assumed in \ref{Conditioning_Event}), we have $\lfloor N/2 \rfloor \geq N/6$ and $(N-\lfloor N/2 \rfloor) \geq N/6$. Then applying the above lemma leads to 
\begin{align}
	\PROB(\Omegasubsetstageone) & \geq 1-C_{24}\cdot \exp(-C_{25}\cdot p_{{\rm min}}^2\cdot N\cdot \epsilon^2). \label{Omega_probability}
\end{align}

We adjust the existing variables as $\epsilon_1=\sqrt{p_{{\rm min}}}\cdot \epsilon, \ \epsilon_2=\sqrt{p_{{\rm min}}}\cdot \epsilon, \ u=\sqrt{N}\cdot p_{{\rm min}}\cdot \epsilon$.
Based on following, which holds due to Cauchy-Schwartz Inequality,
\begin{align} \label{sqrt_prob_bounding}
	\sumsa \sqrt{\probsa} \leq \bigg( \sumsa \probsa \bigg)^{1/2} \cdot \bigg( \sumsa 1 \bigg)^{1/2} = \sqrt{|\SAspace|} \leq \frac{1}{\sqrt{p_{{\rm min}}}}, 
\end{align}
we can rewrite the probability bound as follows, skipping technical details. With probability larger than $1-\delta$,  
\begin{align}
	&\energyonebar(\Upsilonhat, \Upsilon_\pi) \leq  \frac{32 C_{29}}{\sqrt{p_{{\rm min}}}} \cdot C_{{\rm sup}} \cdot B_1(\gamma)\cdot C_{\rm env}(\Theta) \cdot \sqrt{\frac{1}{N} \log \bigg( \frac{C_{27}\cdot (|\SAspace| + N)}{\delta} \bigg) } , \label{finite_bound_realizable_final}
\end{align}
under $N\geq N_{(1)}(\delta)$ and $N\geq N_{(2)}(\delta)$. $N\geq N_{(1)}(\delta)$ implies
\begin{align*} %
	&\frac{1}{N}\log\bigg( \frac{C_{27}\cdot (|\SAspace| + N)}{\delta} \bigg) \leq \frac{C_{28}\cdot p_{{\rm min}}^2}{C_{\rm den}}.
\end{align*}
$N\geq N_{(2)}(\delta)$ implies 
\begin{gather*} \label{N2delta_definition}
	\frac{1}{N}\cdot \frac{8}{p_{{\rm min}}}\cdot \bigg( \supsa \Expect\|R(s,a)\| + \gamma\cdot \supsa \Expect\|Z(s,a;\theta_0)\| \bigg)  + \frac{C_{10}\gamma}{\sqrt{N}}\cdot \sqrt{|\SAspace|} \nonumber \\
	\times \metricentropyexpectatonbounderFspace \leq \frac{C_{29}\cdot (1+\gamma)}{\sqrt{p_{{\rm min}}}}  \cdot C_{\rm env}(\Theta)  \cdot \sqrt{\frac{1}{N} \log \bigg( \frac{C_{27}\cdot (|\SAspace| + N)}{\delta} \bigg) } .   \nonumber
\end{gather*}
Here are definitions of terms.
\begin{gather} \label{def:realizability_terms}
	C_{\rm den}:=d^2\cdot \Rsapsisupsq + \gamma^2\cdot \Zsasupsq + 1, \quad \theta_0:\in\arg\min_{\theta\in\Fspace} \supsa \Expect\|Z(s,a;\theta)\|, \\
	C_{{\rm env}}(\Fspace):= \diamF^2 + \big( \supsa \Expect\|Z(s,a;\theta_0)\|\big)^2 + d^2\cdot \big( \supsa \|R(s,a)\|_{\psi_2} \big)^2 + 1. \nonumber %
\end{gather}

\subsubsection{Under Slight Extent Of Non-realizability} \label{singlestep_nonrealizable_tabular}

As we have warned in Section \ref{theoretical_bound_realizable}, having perfect realizability may be difficult in real practice. Therefore, we will slightly adjust the proof, in order to analyze the error bound of our estimator \eqref{Biased_EStimator_onestep} under non-realizability. 

We can take up from \eqref{relaxing_point_singlestep} as follows,
\begin{gather*}
	\energyonebar(\Upsilonhat, \Bellopt \Upsilonhat) \leq 2\cdot \bigg\{ 2\cdot \energyonebar(\Upsilonstar, \Bellopt \Upsilonstar) + 2\cdot \energyonebar(\Bellopt\Upsilonstar, \Bellhat\Upsilonstar) + 2\Delta_N + \Gamma_N \bigg\} \\
	= 4\cdot \energyonebar(\Upsilonstar, \Bellopt \Upsilonstar) + 4\Delta_N + 6 \Gamma_N.
\end{gather*}
Then, we can repeat the subsequent proofs of bounding $\Delta_N$ and $\Gamma_N$, and thereby obtain the following with probability larger than $1-\delta$,
\begin{gather} 
	\energyonebar(\Upsilonhat, \Upsilon_\pi) \leq  \frac{C}{\sqrt{p_{{\rm min}}}} \cdot C_{{\rm sup}} \cdot B_1(\gamma)\cdot  C_{{\rm env}}(\Fspace)\cdot \bigg\{ \underbrace{\inf_{\theta\in\Theta} \energyonebar(\Upsilon_\theta , \Bellopt \Upsilon_\theta)}_{\text{due to non-realizability}} + \sqrt{\frac{1}{N} \log \bigg( \frac{C_2 \cdot (|\SAspace| + N)}{\delta} \bigg) } \bigg\}  .  \nonumber %
\end{gather}
If the term $\inf_{\theta\in\Theta}\energyonebar(\Upsilon_\theta , \Bellopt \Upsilon_\theta)$ is very small, then the inaccuracy bound of $\energyonebar(\Upsilonhat, \Upsilon_\pi)$ has only a small value that does not vanish as sample size grows $N\rightarrow\infty$.

\subsection{Special Case For Parametric Models Under Realizability} \label{Proof_Realizable_Final_corollary}

In the case of $\Theta\subset \mathbb{R}^p$, by further assuming Assumption \ref{Lipschitz}, we can simplify the metric entropy term $\metricentropyexpectatonbounderFspace$ in \eqref{N2delta_definition}, leading to the following corollary. If another metric $\expectationboundergeneralized$ is used instead of $\Wassersup$, then Assumption \ref{Lipschitz} should be modified accordingly. 
See Appendix \ref{Wassersup_replacing}.

\begin{corollary}
	\label{Realizable_Final_corollary}
	{\bf (Inaccuracy for parametric model in realizable scenario)} Under Assumptions \ref{RN_derivative}, \ref{realizable}, \ref{bounded}, \ref{Lipschitz}, for arbitrary $\delta\in(0,1)$, given large enough sample size $N\geq N(\delta):= \max\{2,N_{(1)}(\delta),N_{(2)}(\delta)\}$ (defined before \eqref{def:realizability_terms}), we have the following with probability larger than $1-\delta$,
	\begin{align}
		\energyonebar(\Upsilonhat, \Upsilon_\pi) \leq  \frac{C_1}{\sqrt{p_{{\rm min}}}} \cdot C_{{\rm sup}} \cdot B_1(\gamma)\cdot C_{{\rm env}}'(\Fspace)\cdot \sqrt{\frac{1}{N} \log \bigg( \frac{C_2 \cdot (|\SAspace| + N)}{\delta} \bigg) }, \nonumber
	\end{align}
	where $B_1(\gamma)>0$, $C_{{\rm sup}}>0$, $C_{\rm env}(\Theta)'$ are defined in Equations \eqref{Decomposition}, \eqref{Ct_and_Csup} and \eqref{Cenv_prime}, and $p_{{\rm min}}>0$ under Assumption \ref{RN_derivative}.
\end{corollary}

\subsubsection{Proof}

We inherit the result of Appendix \ref{realizable_finalizing_the_bound}, except several changes. Applying Assumption \ref{Lipschitz}, we have $\diamF \leq L \cdot \diamTheta$, which allows us to replace $C_{\rm env}(\Theta)$ with 
\begin{align}
	C_{\rm env}'(\Theta) := L^2\cdot \diamTheta^2 + \big( \supsa \Expect\|Z(s,a;\theta_0)\|\big)^2 + d^2\cdot \big( \supsa \|R(s,a)\|_{\psi_2} \big)^2 + 1  \label{Cenv_prime}
\end{align}
where $\diamTheta<\infty$ is ensured by compactness (Assumption \ref{Lipschitz}). In addition, we can make use of the following fact (proof in \ref{Proof_metricentropy_Lipschitz}) to further simplify $N_{(2)}(\delta)$ based on following lemma. 
\begin{remark} \label{metricentropy_Lipschitz} Under Assumption \ref{Lipschitz}, we have the following,
	\begin{align*}
		\metricentropyexpectatonbounderTheta \leq 6\sqrt{2\pi}\cdot L \sqrt{p}\cdot \diamTheta.
	\end{align*}
\end{remark}

\subsection{Continuous State-action Space With Deterministic Transition} \label{sec:continuous_deterministic_realizable}

\subsubsection{Estimator And Its Convergence} \label{continuous_estimator}

Assume the case where $\SAspace$ is continuous (more generally speaking, there is zero probability of observing the same state-action pair twice). However, the transition $p(r,s'|s,a)$ is deterministic. That is, given $s,a$, there is no randomness in $(r,s')$. In this case, our estimator \eqref{Biased_EStimator_onestep_old} can achieve a $1/\sqrt{N}$ convergence rate. Note that our estimator \eqref{Biased_EStimator_onestep_old} becomes the following:
\begin{align} \label{Xtheta_definition}
	X_\theta:=2 \Xthetaone - \Xthetatwo - \Xthetathree \quad \& \quad \hat{\theta} := \arg\min_{\theta\in\Theta} X_\theta, 
\end{align}
where the realizations of $\Xthetaone$, $\Xthetatwo$, $\Xthetathree$ based on observed data $\mathcal{D}=\{(s_i,a_i,r_i,s_i')\}_{i=1}^{N}$ are as follows,
\begin{align*}
	x_\theta^{(1)} &:= \frac{1}{N} \sumi \Expect\| Z_\alpha(s_i,a_i;\theta) - r_i - \gamma Z_\beta(s_i',A_i';\theta) \| \quad \text{with } A_i'\sim \pi(\cdot|s_i'), \\
	x_\theta^{(2)} &:= \frac{1}{N} \sumi \Expect\| Z_\alpha(s_i,a_i;\theta) - Z_\beta (s_i,a_i;\theta) \| \\
	x_\theta^{(3)} &:= \frac{1}{N} \sumi \Expect\| Z_\alpha(s_i',A_i^{\alpha};\theta) - Z_\beta (s_i',A_i^{\beta};\theta) \|  \quad \text{with } A_i^{\alpha},A_i^{\beta} \buildrel iid \over \sim \pi(\cdot|s_i'),
\end{align*}
where $\alpha$ and $\beta$ are used to indicate mutually independent distributions like we did in \eqref{Biased_Esa}. We have $O_p(1/\sqrt{N})$ convergence as follows.

\begin{assumption}  \label{RN_derivative_new}
	We have $\sup_{t\in\mathbb{N}}\supsa \{q_{b_\mu}^{\pi:t}(s,a)/ b_\mu(s,a) \} \leq C_{{\rm sup}} < \infty$, where $q_{b_\mu}^{\pi:t}(s,a)$ is defined in Section \ref{Proof_Csup}. 
\end{assumption}

\begin{theorem} \label{continuous_deterministic_realizable}
	Under Assumptions \ref{RN_derivative_new}, \ref{bounded}, with probability larger than $1-\delta$, we have the following bounds. 
	
	For realizabilty (under Assumption \ref{realizable}), we have
	\begin{align*}
		&\energyonebar(\Upsilonhat, \Upsilon_\pi) \leq  \frac{C\cdot  C_{{\rm sup}}}{\sqrt{N}}  \cdot B_1(\gamma;1,1) \cdot  \bigg\{ \metricentropyexpectatonbounderFspace + C_{mod} \cdot \{ \diamF + 1 \} \cdot \sqrt{\log \bigg( \frac{12}{\delta} \bigg)} \bigg\} .
	\end{align*}
	For non-realizabilty (without Assumption \ref{realizable}), we have
	\begin{align*}
		&\energyonebar(\Upsilonhat, \Upsilon_\pi) \leq C_{{\rm sup}} B_1(\gamma; 1,1) \cdot \inf_{\theta\in\Theta}\energyonebar(\Upsilonstar, \Bellopt\Upsilonstar) + \frac{C}{\sqrt{N}} \cdot  C_{{\rm sup}} \cdot B_1(\gamma;1,1) \\
		&\qquad\qquad \times  \bigg\{ \metricentropyexpectatonbounderFspace + C_{mod} \cdot \{ \diamF + 1 \} \cdot \sqrt{\log \bigg( \frac{12}{\delta} \bigg)} \bigg\} .
	\end{align*}
	with $C_{mod}:= d\cdot \supsa\|R(s,a)\|_{\psi_2} + \supsa \Expect\|Z(s,a;\theta_0)\| + 1$ and $B_1(\gamma;1,1):=\frac{1}{(1-\gamma)}\sum_{k=1}^{\infty} (4^{k} \cdot \gamma^{(2^{k-1}-1)})$. The rate of $B_1(\gamma;1,1)$ with respect to $\gamma\in(0,1)$ is explained in \eqref{Bgamma_rate}.
\end{theorem}

If the realizability is satisfied, then we have $\inf_{\theta\in\Theta}\energyonebar(\Upsilonstar, \Bellopt\Upsilonstar)=0$, which leads to $\energyonebar(\Upsilonhat, \Upsilon_\pi) = O_p(1/\sqrt{N})$. On the other hand, under non-realizability, we have irreducible term $C_{{\rm sup}} B_1(\gamma; 1,1) \cdot \inf_{\theta\in\Theta}\energyonebar(\Upsilonstar, \Bellopt\Upsilonstar) >0$ that does not shrink with sample size.

\subsubsection{Proof}

Even when we replace Assumption \ref{RN_derivative} with Assumption \ref{RN_derivative_new}, Theorem \ref{Fundamental_Realizable} still holds. This can be easily seen, since Assumption \ref{RN_derivative} is used only to show the statement of Assumption \ref{RN_derivative_new}. It is actually the statement of Assumption \ref{RN_derivative_new} that we used in proof.

By Theorem \ref{Fundamental_Realizable} (or \eqref{Fundamental_Realizable_specific}), we have $\energyonebar(\Upsilonhat , \Upsilon_\pi) \leq B(\gamma;\beta_0, q)\cdot \energyonebar(\Upsilonhat, \Bellopt \Upsilonhat)$. So, it suffices to show the convergence of $\energyonebar(\Upsilonhat, \Bellopt \Upsilonhat)\rightarrow^P 0$. With $\theta_*:=\arg\min_{\theta\in\Theta} \energyonebar(\Upsilon_\theta, \Bellopt \Upsilon_\theta)$, we have $\Upsilon_{\theta_*}=\Upsilon_\pi$ by Assumption \ref{realizable}. This leads to following,
\begin{align*}
	\energyonebar(\Upsilonhat, \Bellopt \Upsilonhat) &\leq X_{\hat{\theta}} + | X_{\hat{\theta}} - \energyonebar(\Upsilonhat, \Bellopt \Upsilonhat)|  \\
	&\leq X_{\theta_*} +  | X_{\hat{\theta}} - \energyonebar(\Upsilonhat, \Bellopt \Upsilonhat)| \quad \text{by the definition of } \hat{\theta} \text{ in \eqref{Xtheta_definition}.} \\
	&\leq \energyonebar(\Upsilonstar, \Bellopt\Upsilonstar) + |X_{\theta_*} - \energyonebar(\Upsilonstar, \Bellopt\Upsilonstar)| + | X_{\hat{\theta}} - \energyonebar(\Upsilonhat, \Bellopt \Upsilonhat)|   \\
	&\leq \energyonebar(\Upsilonstar, \Bellopt\Upsilonstar) + 2\cdot \suptheta |X_\theta - \energyonebar(\Upsilon_\theta , \Bellopt \Upsilon_\theta)|.
\end{align*}
Thus, we have the following by \eqref{Fundamental_Realizable_specific} with $q=1$ and $\beta_0=1$ (see Section \ref{Proof_squared_metric}),
\begin{gather} \label{inaccuracy_finalbefore_continuousdeterministic}
	\energyonebar(\Upsilonhat, \Upsilon_\pi) \leq  C_{{\rm sup}} B_1(\gamma; 1,1) \cdot \bigg\{ \inf_{\theta\in\Theta}\energyonebar(\Upsilon_\theta, \Bellopt\Upsilon_\theta,) + 2\cdot \suptheta |X_\theta - \energyonebar(\Upsilon_\theta , \Bellopt \Upsilon_\theta)| \bigg\}. 
\end{gather}

We can rewrite the population objective function as follows. As mentioned below \eqref{expectation_distance}, Using the notation $b_\mu$ as a probability measure,
\begin{gather*}
	\energyonebar(\Upsilon_\theta, \Bellopt\Upsilon_\theta) = \int_{\SAspace} \energyone\bigg\{ \Upsilon_\theta(s,a) , \Bellopt\Upsilon_\theta (s,a) \bigg\} \mathrm{d}\probsa = 2 \Athetaone - \Athetatwo - \Athetathree , \\
	\Athetaone := \Expect\| Z_\alpha (S,A;\theta) - R - \gamma Z_\beta(S',A';\theta) \|, \quad \Athetatwo := \Expect\| Z_\alpha (S,A;\theta) - Z_\beta (S,A;\theta) \| ,\\
	\Athetathree := \Expect\| \gamma Z_\alpha(S',A_\alpha';\theta) - \gamma Z_\beta (S', A_\beta';\theta ) \|,
\end{gather*}
where $(S,A,R,S')$ is iid generated from composition of $S,A\sim b_\mu$ followed by $R,S'\sim p(\cdots|S,A)$ (no randomness) 
and $A_\alpha',A_\beta'\buildrel iid \over \sim \pi(\cdot|S')$. Then, we have 
\begin{align*}
	\suptheta |X_\theta - \energyonebar(\Upsilon_\theta , \Bellopt \Upsilon_\theta)| \leq 2\cdot \suptheta |\Xthetaone - \Athetaone| + \suptheta |\Xthetatwo - \Athetatwo| + \suptheta |\Xthetathree - \Athetathree|.
\end{align*}
Let us handle the first term.
\begin{align*}
	\suptheta |\Xthetaone - \Athetaone| \leq \suptheta | (\Xthetaone - \Athetaone) - (\Xthetaonezero - \Athetaonezero) | + |\Xthetaonezero - \Athetaonezero| .
\end{align*}
We can express $\Xthetaone - \Athetaone = \frac{1}{N}\sumi W_i^\theta$ where the realization of $W_i^\theta$ is written as 
\begin{align*}
	w_i^\theta := \Expect\| Z_\alpha(s_i,a_i;\theta) - r_i - \gamma Z_\beta (s_i', A_i';\theta) \| - \Expect\| Z_\alpha (S,A;\theta) - R - \gamma Z_\beta(S',A';\theta)\|.
\end{align*}
Now, we repeat the logic that we used in Section \ref{Gamma_bounding_realizable}. We have $\Expect(W_i^\theta)=0$ and $|w_i^{\theta_1} - w_i^{\theta_2}| \leq 4\cdot\Wassersup(\theta_1, \theta_2)$ by the same logic with Appendix \ref{Wassersup_diamF_satisfication}. Then we apply Dudley's inequality (Theorem 8.1.6 of \cite{vershynin2018high}) to obtain the following: with probability larger than $1-2\exp(-u^2)$, 
\begin{align*}
	\suptheta | (\Xthetaone - \Athetaone) &- (\Xthetaonezero - \Athetaonezero) | \leq \frac{C_1}{\sqrt{N}} \cdot \bigg\{ \metricentropyexpectatonbounderFspace + u\cdot \diamF \bigg\}.
\end{align*}
We can also apply Theorem 2.6.2 of \cite{vershynin2018high} to obtain the following by the same logic as \eqref{reward_psi_bounding},
\begin{align*}
	\PROB(|\Xthetaonezero - \Athetaonezero| \geq \epsilon) &\leq 2\cdot \exp \bigg\{ \frac{-C_1\cdot N\cdot \epsilon^2}{\| W_1^{\theta_0} - \Expect(W_1^{\theta_0}) \|_{\psi_2}^2} \bigg\} \leq 2\cdot \exp (-C_2\cdot M \cdot \epsilon^2 / C_{mod}^2).
\end{align*}

We can apply the same logic for $\suptheta |\Xthetatwo - \Athetatwo|$ and $\suptheta |\Xthetathree - \Athetathree|$. Then, we eventually obtain 
the following with probability larger than $1-\delta$
\begin{align*}
	\suptheta|&X_\theta - \energyonebar(\Upsilon_\theta , \Bellopt \Upsilon_\theta)| \leq \frac{C_5}{\sqrt{N}} \cdot \bigg\{ \metricentropyexpectatonbounderFspace + C_{mod} \cdot \{ \diamF + 1 \} \cdot \sqrt{\log \bigg( \frac{12}{\delta} \bigg)} \bigg\} .
\end{align*}

Combined with \eqref{inaccuracy_finalbefore_continuousdeterministic}, we have the desired result.

\subsection{Comparison With FLE} \label{proof:comparison_FLE}

As we mentioned in Remark \ref{remark:comparison_FLE}, we will assume $d=1$, tabular setting ($|\SAspace|<\infty$), and bounded rewards. 
\begin{align}
	\text{EBRM : } \mathbb{W}_p(\Upsilonhat^{marginal}, \Upsilon_\pi^{marginal}) \leq O_p(N^{\frac{-1/p}{2+\alpha}}) \quad \text{VS} \quad 
	\text{FLE : } \mathbb{W}_p(\Upsilonhat^{marginal}, \Upsilon_\pi) \leq O_p(N^{\frac{-1}{2p}(1-\alpha)}). \label{ineq:EBRM_vs_FLE}
\end{align}

Following is the proof. We first define the following two terms,
\begin{align*}
	F(\theta)=\energyonebar(\Upsilon_\theta, \Bellopt, \Upsilon_\theta) \quad \& \quad \Fhat(\theta) = \energyonebarhat(\Upsilon_\theta , \Bellhat \Upsilon_\theta ) .
\end{align*}
Then we have following. The data $\{s_i,a_i,r_i,s_i'\}_{i=1}^N$ are the observations that we are given with. Conditioned on a fixed $s,a$, $\{r_i(s,a), s_i'(s,a)\}_{i=1}^{\Nsa}$ are the observations that we observed for that particular $s,a$.
\begin{align*}
	\Fhat(\theta) &= \frac{1}{N} \sumi \bigg\{ \underbrace{2\cdot \Expect\| Z_\alpha(s_i,a_i;\theta) - r_i - \gamma Z_\beta(s_i',A_i';\theta) \| - \Expect\|Z_\alpha(s_i,a_i;\theta) - Z_\beta(s_i,a_i;\theta)\| }_{\buildrel let \over = A_i^\theta} \bigg\} \\
	& - \sumsa \frac{\Nsa}{N} \cdot \frac{1}{\Nsa^2} \sumiandj \underbrace{\Expect\| r_i(s,a) + \gamma Z_\alpha(s_i'(s,a) , A_i'(s,a) ;\theta ) - r_j(s,a) - \gamma Z_\beta(s_j'(s,a), A_j'(s,a); \theta )  \|}_{\buildrel let \over = C_{ij}^\theta(s,a)}.
\end{align*}
Then we have 
\begin{align*}
	\Delta_N(\theta)&:=\big( \Fhat(\theta) - F(\theta) \big) - \big( \Fhat(\thetastar) - F(\thetastar) \big) \\
	&= \bigg\{ \frac{1}{N}\sumi (A_i^\theta - A_i^\theta) - \Expect(A_1^\theta - A_1^\thetastar) \bigg\} \\
	& \quad - \bigg\{ \sumsa \frac{\Nsa}{N}\cdot \frac{1}{\Nsa^2} \cdot \sumiandj\big( C_{ij}^\theta(s,a) - C_{ij}^\thetastar(s,a) \big) - \sumsa \probsa \cdot \Expect\big( C_{12}^\theta (s,a) - C_{12}^\thetastar(s,a) \big) \bigg\} \\
	&= \underbrace{\bigg\{ \frac{1}{N}\sumi (A_i^\theta - A_i^\thetastar) - \Expect(A_i^\theta - A_i^\thetastar) \bigg\}}_{\buildrel let \over = \Delta_N^{(1)}}  \\
	& \quad - \underbrace{\bigg\{ \sumsa \bigg( \frac{\Nsa}{N} - \probsa  \bigg) \cdot \frac{1}{\Nsa^2} \cdot \sumiandj \big( C_{ij}^\theta (s,a) - C_{ij}^\thetastar(s,a) \big) \bigg\}}_{\buildrel let \over = \Delta_N^{(2)}(\theta)} \\
	& \quad + \sumsa \probsa \cdot \underbrace{\bigg\{ \frac{1}{\Nsa^2} \sumiandj \big( C_{ij}^\theta (s,a) - C_{ij}^\thetastar(s,a) \big) - \Expect\big( C_{12}^\theta (s,a) - C_{12}^\thetastar(s,a) \big)  \bigg\} }_{\buildrel let \over = \Delta_N^{(3)}(\theta;s,a)}
\end{align*}
Denote the last term above as $\Delta_N^{(3)}(\theta)= \sumsa \probsa\cdot \Delta_N^{(3)}(\theta;s,a)$. Let us bound the following three terms respectively.
\begin{align*}
	\Expect\bigg\{ \sup_{\Wassersup(\theta, \theta_*)\leq \delta} |\Delta_N(\theta)|  \bigg\} \leq \Expect\bigg\{ \sup_{\Wassersup(\theta, \theta_*)\leq \delta} |\Delta_N^{(1)}(\theta)|  \bigg\}  + \Expect\bigg\{ \sup_{\Wassersup(\theta, \theta_*)\leq \delta} |\Delta_N^{(2)}(\theta)|  \bigg\}  + \Expect\bigg\{ \sup_{\Wassersup(\theta, \theta_*)\leq \delta} |\Delta_N^{(3)}(\theta)|  \bigg\}  
\end{align*}

\noindent Let us start with the first one. With a universal constant $C>0$, we can apply Lemma 2.6.8 of \cite{vershynin2018high} to obtain the following, 
\begin{align*}
	\|\Delta_N^{(1)}(\thetaone) - \Delta_N^{(2)}(\thetatwo)\|_{\psi_2} = \bigg\| \frac{1}{N}\sumi (A_i^\thetaone - A_i^\thetatwo) - \Expect(A_1^\theta - A_1^\thetatwo)  \bigg\|_{\psi_2} \leq C \cdot \bigg\| \frac{1}{N} (A_i^\thetaone - A_i^\thetatwo) \bigg\|_{\psi_2}.
\end{align*}
Since we have $|A_i^\thetaone - A_i^\thetatwo|\leq 4 \cdot \Wassersup(\thetaone, \thetatwo)$ (shown in the proof of previous sections), this leads to following by Proposition 2.6.1, Example 2.5.8, Theorem 8.1.3 of \cite{vershynin2018high}. With $\Theta^\delta:=\{\theta\in\Theta: \Wassersup(\theta, \thetastar) \leq \delta\}$, we have
\begin{gather*}
	\| \Delta_N^{(1)}(\thetaone)- \Delta_N^{(2)}(\thetatwo) \|_{\psi_2} \leq \frac{C}{\sqrt{N}} \cdot \Wassersup(\thetaone, \thetatwo) ,\\
	\therefore \ \Expect\bigg\{\supdeltawassersup |\Delta_N^{(1)}(\theta)|  \bigg\} \leq  \frac{C}{\sqrt{N}} \cdot \int_0^\infty \sqrt{\log \mathcal{N}(\Theta^\delta , \Wassersup, \epsilon)}\mathrm{d}\epsilon \leq  \frac{C}{\sqrt{N}} \cdot \int_0^\infty \sqrt{\log \mathcal{N}(\Theta , \Wassersup, \epsilon)}\mathrm{d}\epsilon .
\end{gather*}

\noindent Now, let us assume $\log \mathcal{N}(\Theta , \Wassersup , \epsilon) \leq C_{met}\cdot \epsilon^{-\alpha}$ for some $\alpha \in (0,2)$. Then, we have 
\begin{align*}
	\Expect\bigg\{\supdeltawassersup |\Delta_N^{(1)}(\theta)|  \bigg\} \leq \frac{C}{\sqrt{N}} \cdot \frac{C_{met}^{1/2}}{1-\alpha/2} \cdot \delta^{1-\alpha/2}.   
\end{align*}

\noindent Now let us begin with the second term,
\begin{align*}
	\Expect\bigg\{ \supdeltawassersup |\Delta_N^{(2)}(\theta) | \bigg\} &= \Expect \bigg\{ \supdeltawassersup \bigg| \sumsa \bigg( \frac{\Nsa}{N} - \probsa \bigg) \cdot \frac{1}{\Nsa^2} \sumiandj \big( C_{ij}^\theta (s,a) - C_{ij}^\thetastar(s,a) \big)  \bigg| \bigg\}  \\
	&\leq  \Expect \bigg\{ \supdeltawassersup  \sumsa \bigg| \frac{\Nsa}{N} - \probsa \bigg| \cdot \frac{1}{\Nsa^2} \sumiandj \underbrace{\bigg| C_{ij}^\theta (s,a) - C_{ij}^\thetastar(s,a) \bigg|}_{\leq 2 \cdot \Wassersup(\theta, \thetastar)} \bigg\} \\
	&\leq 2\delta \cdot \sumsa \Expect \bigg| \frac{\Nsa}{N} - \probsa \bigg| \leq C\cdot \delta \cdot \frac{1}{\sqrt{N}} \cdot |\SAspace|.
\end{align*}

\noindent Now let us bound the third term,
\begin{align*}
	\Expect\bigg\{ \supdeltawassersup |\Delta_N^{(3)}(\theta)| \bigg\} = \Expect\bigg\{ \supdeltawassersup |\sumsa \probsa \cdot \Delta_N^{(3)}(\theta;s,a)| \bigg\} \leq \sumsa \probsa \cdot \Expect\bigg\{ \supdeltawassersup |\Delta_N^{(3)}(\theta;s,a)| \bigg\}.
\end{align*}

\noindent For notational convenience, let $n=\Nsa$ and ignore simplify $C_{ij}^\theta=C_{ij}^\theta (s,a)$. 
\begin{align*}
	\Expect\bigg\{ \Delta_N^{(3)}(\theta;s,a) \bigg\} &= \frac{n-1}{n} \cdot \Expect\bigg\{ \supdeltawassersup \bigg| \frac{1}{n(n-1)} \sumiandj \big( (C_{ij}^\theta - C_{ij}^\thetastar) - \Expect(C_{12}^\theta - C_{12}^\thetastar) \big)  \bigg| \bigg\} \\
	& = \frac{n-1}{n} \cdot \Expect\bigg\{ \supdeltawassersup \bigg| \underbrace{\frac{1}{n(n-1)} \sumij \big( (C_{ij}^\theta - C_{ij}^\thetastar) - \Expect(C_{12}^\theta - C_{12}^\thetastar) \big)}_{\buildrel let \over = \tilde{\Delta}_n^{(3)}(\theta;s,a)} \bigg| \bigg\} \\
	&\quad + \frac{1}{n} \cdot \underbrace{\Expect\bigg\{ \supdeltawassersup \bigg| \frac{1}{n}\sum_{i=1}^n \big( (C_{ii}^\theta - C_{ii}^\thetastar) - \Expect(C_{12}^\theta - C_{12}^\thetastar) \big) \bigg| \bigg\}}_{\leq 2\delta} .
\end{align*}

\noindent Now let us bound the term $\Expect\{\supdeltawassersup|\tilde{\Delta}_n^{(3)}(\theta;s,a)|\}$ as follows with Lemma 2.6.8, Proposition 2.6.1 of \cite{vershynin2018high},
\begin{gather*}
	\| \tilde{\Delta}_n^{(3)}(\thetaone;s,a) - \tilde{\Delta}_n^{(3)}(\thetatwo;s,a) \|_{\psi_2} = \bigg\| \frac{1}{n(n-1)}\sumij \big( (C_{ij}^\thetaone - C_{ij}^\thetatwo) - \Expect(C_{12}^\thetaone - C_{12}^\thetatwo) \big) \bigg\|_{\psi_2} \\
	\leq C \cdot \bigg\| \frac{1}{n(n-1)}\sumij (C_{ij}^\thetaone - C_{ij}^\thetatwo) \bigg\|_{\psi_2} \leq \frac{C}{\sqrt{n}}\cdot \Wassersup(\thetaone, \thetatwo).
\end{gather*}
In the last inequality above, we have used Lemma S4 of \citet{wang2022low} as we have done in previous proofs. Without loss of generality, we can assume $0<\delta<1$. Conditioned under $\Omegasubsetstageone$ with $\epsilon=1$, we have $n\geq \frac{1}{2}p_{\rm min} \cdot N$. 
\begin{align*}
	\Expect\bigg\{ \supdeltawassersup |\Delta_N^{(3)}(\theta;s,a)| \ \bigg| \ \Omegasubsetstageone \bigg\} &\leq \frac{C}{\sqrt{n}} \cdot \int_0^\infty \sqrt{\log \mathcal{N}(\Theta, \Wassersup, \epsilon)}\mathrm{d}\epsilon + C \cdot \frac{1}{n}  \\
	&\leq \frac{C}{\sqrt{p_{\rm min}}} \cdot \frac{1}{\sqrt{N}} \cdot \frac{C_{met}^{1/2}}{1-\alpha/2}\cdot \delta^{1-\alpha/2} + \frac{C}{N\cdot p_{\rm min}}, \\
	\mathbb{P}\bigg\{\big(\Omegasubsetstageone\big)^c\bigg\} \leq C_{24}\cdot &\exp(-C_{25}\cdot p_{{\rm min}}^2\cdot N ) \quad \text{by \eqref{Omega_probability}} .\end{align*}
Combined with $\supsa \suptheta |\Delta_N^{(3)}(\theta;s,a)| < \infty$ (under bounded rewards assumption), we obtain following,
\begin{align*}
	\therefore \ \Expect\bigg\{ \supdeltawassersup |\Delta_N^{(3)}(\theta)| \bigg\} & \leq \frac{C}{\sqrt{p_{\rm min}}} \cdot \frac{C_{met}^{1/2}}{1-\alpha/2} \cdot \frac{\delta^{1-\alpha/2}}{\sqrt{N}} + \frac{C}{N\cdot p_{\rm min}} \quad \text{for sufficiently large } N .
\end{align*}
Putting all three terms together, we finally obtain
\begin{align*}
	\Expect\bigg\{ \supdeltawassersup |\Delta_N(\theta)| \bigg\} \leq \frac{C}{\sqrt{N}} \cdot \max\bigg\{ \frac{C_{met}^{1/2}}{1-\alpha/2} \cdot \frac{1}{\sqrt{p_{\rm min}}} , |\SAspace|  \bigg\} \cdot \delta^{1-\alpha/2} + \frac{C}{p_{\rm min}} \cdot \frac{1}{N} .
\end{align*}

\noindent Then, we assume $\Upsilon_\theta (s,a)$ are all 1-dimensional and have bounded support in $[a,b]$ for all $s,a\in\SAspace$. By Theorem \ref{Fundamental_Realizable}, we obtain the following,
\begin{gather*}
	F(\theta)- F(\thetastar) = \energyonebar(\Upsilon_\theta , \Bellopt \Upsilon_\theta) \geq \frac{1}{B(\gamma)} \cdot \energyonebar(\Upsilon_\theta, \Upsilon_{\thetastar}) \geq \frac{1}{B(\gamma)} \cdot \frac{2}{b-a} \cdot \overline{\Wasserone^2} \geq \frac{1}{B(\gamma)} \cdot \frac{2}{b-a} \cdot \overline{\mathbb{W}}_1^2(\theta, \thetastar) \\
	\geq \frac{1}{B(\gamma)} \cdot \frac{2}{b-a} \cdot p_{\rm min} \cdot \Wassersup(\theta, \thetastar) .
\end{gather*}
\noindent Here, we have used the following two facts: With $l_p$ defined in \eqref{suitable_distances},
\begin{gather*}
	\energyone(P,Q) \geq 2 \cdot l_2^2(P,Q) \geq \frac{2}{b-a} \cdot l_1^2(P,Q) = \frac{2}{b-a}\cdot \Wasserone^2(P,Q), \\
	\frac{1}{p_{\rm min}} \cdot \overline{\mathbb{W}}_1 (\theta, \thetastar) = \sumsa \frac{\probsa}{p_{\rm min}} \cdot \Wasserone\big\{ \Upsilon_\theta(s,a) , \Upsilon_{\thetastar}(s,a) \big\} \geq \Wassersup(\theta, \thetastar).
\end{gather*}

Based on Theorem \ref{peeling_argument}, we obtain the following,
\begin{gather*}
	\Wassersup(\thetahat, \thetastar) \leq C_A^{\frac{-1}{1+\alpha/2}} \cdot O_p(N^{\frac{1}{2+\alpha}}) + C_B^{1/2} \cdot O_p(N^{-1/2}) \leq \max\bigg\{ C_A^{\frac{1}{1+\alpha/2}}, C_B^{1/2} \bigg\} \cdot O_p(N^{\frac{-1}{2+\alpha}}) , \quad \text{where} \\
	C_A = C \cdot \max \bigg\{ \frac{C_{met}^{1/2}}{1-\alpha/2} \cdot \frac{1}{\sqrt{p_{\rm min}}} , |\SAspace| \bigg\} \quad \& \quad C_B = \frac{C}{\sqrt{p_{\rm min}}}.
\end{gather*}
Using the fact $\mathbb{W}_p \leq (b-a)^{1-\frac{1}{p}}\cdot \Wasserone^{1/p}$ (from Section 2.3 of \cite{panaretos2019statistical}), we obtain 
\begin{align*}
	\mathbb{W}_{p,\infty} \leq (b-a)^{1-\frac{1}{p}} \cdot \max\bigg\{ C_A^{\frac{1}{1+\alpha/2}}, C_B^{1/2} \bigg\} \cdot O_p(N^{\frac{-1}{2+\alpha}}).
\end{align*}

Now, let us compare the convergence rate between EBRM and FLE \citep{wu2023distributional}. Let us assume the same model-complexity that they used in Corollary 4.14:
\begin{align*}
	\log \mathcal{N}_{[ \ ]}(\tilde{\mathcal{F}}_\Theta, \|\cdot\|_{\infty,1}, \epsilon) \lesssim \epsilon^{-\alpha} \quad \text{for some } \alpha\in(0,2), 
\end{align*}
where $\tilde{\mathcal{F}}_\Theta:=\{ f_\theta(\cdot|\cdots) : s,a-\text{conditioned densities of }\Upsilon_\theta\}$ and $\|f_\thetaone - f_\thetatwo\|_{\infty,1}:= \supsa \int_{-\infty}^{\infty}|f_\thetaone(z|s,a) - f_\thetatwo(z|s,a)|\mathrm{d}z$. Temporarily ignoring all factors other than $N$, FLE achieves the following rate according to Corollary 4.14 of \cite{wu2023distributional}:
\begin{align*}
	\mathbb{W}_p(\Upsilonhat^{marginal}, \Upsilon_\pi) \leq O_p(N^{\frac{-1}{2p}}) \cdot \big\{ \log \mathcal{N}_{[ \ ]}(\mathcal{F}_\Theta, \|\cdot\|_{\infty,1}, 1/N) \big\}^{\frac{1}{2p}} \leq O_p(N^{\frac{-1}{2p}(1-\alpha)}).
\end{align*}

For EBRM, we let $l(\cdot|s,a)$ and $u(\cdot|s,a)$ positive-valued functions (like conditional density functions), and we have
\begin{gather*}
	\|L(\cdot|\cdots) - U(\cdot|\cdots)\|_{\infty,1} \leq (b-a) \cdot \|l(\cdot|\cdots) - u(\cdot|\cdots)\|_{\infty,1}\\
	\text{where } L(y|s,a):=\int_a^y l(z|\cdots)\mathrm{d}z , \quad U(y|\cdots) := \int_a^y u(z|\cdots)\mathrm{d}z .
\end{gather*}
Then we obtain the following,
\begin{gather*}
	\log \mathcal{N}(\Theta, \Wassersup, \epsilon) \leq \log \mathcal{N}_{[ \ ]}(\mathcal{F}_\Theta, \|\cdot\|_{\infty,1}, \epsilon/2) \leq \log \mathcal{N}_{[ \ ]}(\tilde{\mathcal{F}}_\Theta, \|\cdot\|_{\infty,1}, \frac{\epsilon}{2(b-a)}) \lesssim \epsilon^{-\alpha}, \\
	\therefore \  \Wasserone(\Upsilonhat^{marginal}, \Upsilon_\pi^{marginal}) \leq \Wassersup(\thetahat, \thetastar) = O_p(N^{\frac{-1}{2+\alpha}}) \quad \& \quad \mathbb{W}_p(\Upsilonhat^{marginal}, \Upsilon_\pi^{marginal}) \leq O_p(N^{\frac{-1/p}{2+\alpha}}). 
\end{gather*}

For $\forall \alpha\in(0,2)$ and $\forall p \geq 1$, EBRM has faster convergence rate than FLE under tabular, 1-dimensional, bounded setting.

\clearpage

\section{PROOFS FOR SECTION \ref{Nonrealizable_Scenario}} \label{nonrealizable_proof}

As in Appendix \ref{Realizable_Scenario}, we will use $C_k>0$ ($k\in\mathbb{N}$) to denote appropriate universal constants throughout the proof.

\subsection{Exponential Increasing Rate Of Trajectories} \label{exponential_trajectories}

Based on how we defined $\Bellhat$ based on $\hat{p}$ \eqref{probabiliy_stateaction}, $\Bellhatmulti\Upsilon_\theta(s,a)$ utilizes the trajectories of tuples $(s,a,r,s')$ that can occur consecutively under the estimated probability measure $\hat{p}(\cdots|s,a)$ and the target policy $\pi(a|s)$, 
\begin{gather}
	\big(s,a, r_{i}^{(1)},s_{i}^{(1)}, a_i^{(1)}, r_{i}^{(2)},s_{i}^{(2)}, a_i^{(2)}, r_{i}^{(3)},s_{i}^{(3)}, a_i^{(3)}, \cdots , r_{i}^{(m)},s_{i}^{(m)}\big), \label{m_trajectory} \\
	\text{where } r_i^{(t)},s_i^{(t)}\sim \hat{p}(\cdots|s_i^{(t-1)},a_i^{(t-1)}), \ a_i^{(t)}\sim \pi(\cdot|s_i^{(t)}) \ \text{for }\forall t\geq1, \ \text{with } (s_i^{(0)},a_i^{(0)})=(s,a). \nonumber
\end{gather}

Let us first verify how many such trajectories (\ref{m_trajectory}) can amount to, which start from a common state-action pair $s,a$ with length $m=2$. 

First there are $\Nsa$ many tuples that can occur in the first step,
\begin{align*}
	(s,a,r_i^{(1)},s_i^{(1)}) \ \ \ \big(1\leq i \leq \Nsa\big).
\end{align*} 
Now fix one observation with index $i$, and then there can be $|\mathcal{A}|$ many actions at most that can follow $s_i^{(1)}$, giving us the following tuples,
\begin{align*}
	(s,a,r_i^{(1)}, s_i^{(1)}, a_1), \ (s,a,r_i^{(1)}, s_i^{(1)}, a_2), \ \cdots, \  (s,a,r_i^{(1)}, s_i^{(1)}, a_{|\mathcal{A}|}) \ \ \ \big(1\leq i \leq \Nsa\big).
\end{align*}
Now we are given with $|\mathcal{A}|$ different state-action pairs, $(s_i^{(1)},a_1),\cdots,(s_i^{(1)},a_{|\mathcal{A}|})$, and then the following observations of $(r_i^{(2)}, s_i^{(2)})$ can be as many as $\sum_{k=1}^{|\mathcal{A}|} N(s_i^{(1)}, a_{k}) \leq \sum_{s,a}N(s,a) = N$.
This eventually gives us at most $\Nsa\times N$ trajectories of length $m=2$ starting from the given state-action pair $s,a$,
\begin{align*}
	(s,a,r_i^{(1)},s_i^{(1)}, a_i^{(1)}, r_i^{(2)}, s_i^{(2)}) \ \ \ \big( 1\leq i \leq \Nsa\times N \big).
\end{align*}
Then we can add up for all state-action pairs that we can begin with, which leads to $\sumsa \Nsa\times N = N^2$ many trajectories at most.
We can generalize this result for an arbitrary value of $m\in\mathbb{N}$, which gives us $\Nsa \times N^{m-1}$ many trajectories for a given state-action pair $s,a$, leading to
$N^m$ many trajectories \eqref{m_trajectory} of length $m$ if we sum them all up for all state-action pairs as the initial point.

\subsection{How Supremum-extended Wasserstein-1 Metric Can Be Replaced} \label{Wassersup_replacing}

One may wonder why supremum-extended \eqref{supremum_distance} Wasserstein-1 metric $\Wassersup$ appears in Assumption \ref{Lipschitz}. In fact, $\Wassersup$ can be replaced with a general statistical distance measure, say $\expectationboundergeneralized:\mathcal{P}(\mathbb{R}^d)\times\mathcal{P}(\mathbb{R}^d)\rightarrow\mathbb{R}$, if it satisfies the following properties.

\begin{enumerate}
	\item $\expectationboundergeneralized$ is a metric. It is a contraction with respect to $\expectationboundergeneralized$, i.e. $\expectationboundergeneralized(\Bellopt\Upsilon_1, \Bellopt\Upsilon_2)\leq \gamma^{\beta_0}\cdot \expectationboundergeneralized(\Upsilon_1,\Upsilon_2)$ for some $\beta_0>0$, for any $\Bellopt$ where the corresponding transition probability $p$ \eqref{Bellman_operator_definition} can be arbitrary.
	\item For arbitrary $c\in\mathbb{R}^d$, $\gamma_1,\gamma_2\in[0,1]$, $(s,a),(\tilde{s},\tilde{a})\in\SAspace$, letting $Z_i(s,a)\sim \Upsilon_i(s,a)$ be such that $(Z_1(s,a),Z_3(s,a))\in\mathbb{R}^d\times\mathbb{R}^d$ and $(Z_2(\tilde{s},\tilde{a}),Z_4(\tilde{s},\tilde{a}))\in\mathbb{R}^d\times \mathbb{R}^d$ are mutually independent, $\expectationboundergeneralized$ should satisfy 
	\begin{align}
		\bigg| \Expect\|c+\gamma_1 Z_1(s,a) - \gamma_2 Z_2 (\tilde{s},\tilde{a}) \| - \Expect\|c+&\gamma_1 Z_3(s,a) - \gamma_2 Z_4 (\tilde{s},\tilde{a}) \| \bigg| \leq \gamma_1\cdot \expectationboundergeneralized(\Upsilon_1, \Upsilon_3) + \gamma_2\cdot \expectationboundergeneralized(\Upsilon_2,\Upsilon_4). \nonumber %
	\end{align}
\end{enumerate}

Note that $\Wassersup$ satisfies all three properties. It is shown to be a metric in by \citet{bellemare2017distributional} in their Lemma 2. In Appendix \ref{Wassersup_diamF_satisfication}, we have shown that $\Wassersup$ satisfies the second property. Lastly, \citet{bellemare2017distributional} showed that it makes $\Bellopt$ a contraction in their Lemma 3. If we could find another distance $\expectationboundergeneralized$ that satisfies all three properties, then we can replace $\Wassersup$ with $\expectationboundergeneralized$ in all proofs in the following subsections (Appendices \ref{Proof_Fmhat_convergence}, \ref{bootstrap_objective_function_converging}, \ref{Proof_thetaBhat_convergence}). Of course, if we are to proceed with $\expectationboundergeneralized$, we will need to replace Assumption \ref{Lipschitz} with $\expectationboundergeneralized(\Upsilon_{\theta_1}, \Upsilon_{\theta_2}) \leq L \|\theta_1-\theta_2\|$.

\subsection{Bootstrap-based Estimation} \label{bootstrap_estimation_procedure}

Let us first discuss how $\Bellhatmulti$ is implemented. Generalizing from the definition of $\Bellhat$ based on \eqref{empirical_transition}, we consider $\hat{Z}^{(m)}(s,a;\theta)\sim\Bellhatmulti\Upsilon_\theta(s,a)$ as the distribution of an $m$-length trajectories of tuples $(s,a,r,s')$ that is generated under the estimated transition $\hat{p}$ and the target policy $\pi$:
\begin{gather} \label{multistep_Estimated_Bellman_Operator}
	\hat{Z}^{(m)}(s,a;\theta) \buildrel D \over = \sum_{t=1}^{m} \gamma^{t-1} \hat{R}^{(t)} + \gamma^m Z(\hat{S}^{(m)},\hat{A}^{(m)};\theta),
\end{gather}
where $(\hat{R}^{(t)},\hat{S}^{(t)})\sim \hat{p}(\cdots|\hat{S}^{(t-1)},\hat{A}^{(t-1)})$, $\hat{A}^{(t)}\sim\pi(\cdot|\hat{S}^{(t)})$ for all $t\geq 1$, $(\hat{S}^{(0)},\hat{A}^{(0)})=(s,a)$. 

However, the estimation of the $\Bellhatmulti \Upsilon_\theta$ (\ref{multistep_Estimated_Bellman_Operator}) generally requires the computation of $N^m$ trajectories (as discussed in Appendix \ref{exponential_trajectories}), which amounts to a heavy computational burden. 

To alleviate such burden, we will instead bootstrap $M\ll N^m$ many trajectories by first sampling the initial state-action pairs $(s_i^{(0)},a_i^{(0)}) \ (1\leq i \leq M)$ from $\hat{b}_{\mu}$ and then resampling the subsequent $r_i^{(t+1)}, s_i^{(t+1)} \sim \hat{p}(\cdots|s_i^{(t)},a_i^{(t)})$ and $a_i^{(t+1)}\sim \pi(\cdot|s_i^{(t+1)})$ for $m$ steps. Let $\probhatbootstrap (\cdots|s,a)$ be the empirical probability measure of $(\sum_{t=1}^{m}\gamma^{t-1}r_i^{(t)}, s_i^{(m)})$ conditioning on $(s_i^{(0)},a_i^{(0)})=(s,a)$. We define the \textit{bootstrap operator} as follows, with an abuse of notation $\Bootstrapopt Z(s,a;\theta)\sim \Bootstrapopt \Upsilon_\theta(s,a)$:
\begin{gather} 
	\Bootstrapopt Z(s,a;\theta):=\sum_{t=1}^{m} \gamma^{t-1} \hat{R}_{b}^{(t)} + \gamma^m Z(\hat{S}_{b}^{(m)},\hat{A}_{b}^{(m)};\theta), \label{Bootstrap_operator} 
\end{gather}
where we have $(\sum_{t=1}^{m} \gamma^{t-1} \hat{R}_{b}^{(t)} , \hat{S}_{b}^{(m)} ) \sim \probhatbootstrap (\cdots|s,a)$ and $\hat{A}_{b}^{(m)}\sim \pi(\cdot|\hat{S}_{b}^{(m)})$. Then we can compute our objective function and derive the bootstrap-based multi-step estimator:
\begin{gather*}
	\FBhat(\theta):=\energyonebarhat\big( \Upsilon_\theta, \Bootstrapopt \Upsilon_\theta \big), \ \thetabootstrap:=\arg\min_{\theta\in\Theta}\FBhat(\theta). 
\end{gather*}

Our final algorithm can be summarized as follows, in Algorithm \ref{EBRM_multistep_algorithm}.

\begin{algorithm}
	\caption{EBRM-multi-step} \label{EBRM_multistep_algorithm}
	\begin{algorithmic}
		\\
		\textbf{Input: } $\Theta$, $\mathcal{D}=\{(s_i,a_i,r_i,s_i')\}_{i=1}^{N}$, $m$, $M$ \\
		\textbf{Output: } $\thetabootstrap$
		\State Estimate $\hat{b}_\mu$ and $\hat{p}$. \Comment{Refer to Equation (\ref{probabiliy_stateaction}).}
		\State Randomly generate $M$ tuples of $(\sum_{t=1}^{m}\gamma^{t-1}r_i^{(t)}, s_i^{(m)}) \ (1\leq i \leq M)$. \Comment{Refer to Equation \eqref{Bootstrap_operator}.}
		\State $\thetabootstrap=\arg\min_{\theta\in\Theta}\energyonebarhat(\Upsilon_\theta, \Bootstrapopt\Upsilon_\theta)$. 
	\end{algorithmic}
\end{algorithm}

\subsection{Estimation Error Of Multi-step Bellman Residual} \label{Proof_Fmhat_convergence}

\begin{lemma} \label{Fmhat_convergence}
	{\bf (Convergence of estimated Bellman residual)}  Under Assumptions \ref{RN_derivative}, \ref{bounded}, and \ref{Lipschitz}, for arbitrary $\epsilon\in(0,1]$, we have
	\begin{align}
		&\bellmanestimationerror \leq C_{1}\cdot m^2\cdot \frac{1}{p_{{\rm min}}^2}\cdot \Cenvmulti \cdot \bigg( \frac{1}{N^{1/4}}+ \sqrt{\epsilon} \bigg)  \nonumber
	\end{align}
	with probability larger than 
	\begin{align}
		&1-C_{2}\cdot m\cdot (|\SAspace| + N) \cdot \exp\big\{ -C_{3}\cdot p_{{\rm min}}^2 \cdot N \cdot \epsilon^2 / C_{\rm den}(m) \big\}  \nonumber
	\end{align}
	with $p_{{\rm min}}>0$ defined in Assumption \ref{RN_derivative}, $\theta_0\in\Theta$ in \eqref{def:realizability_terms}, $C_{\rm den}(m)$ and  $\Cenvmulti$ in \eqref{C_den_multi}.
\end{lemma}
Although larger values of step level $m$ both increase and decrease some terms, the decreasing parts have a non-zero lower bounds $\gamma^m \sqrt{p}+1$ and $1+\gamma^m$ of \eqref{C_den_multi}. Thus it can be seen that increased values of step level $m$ eventually leads to looser bound, necessitating larger sample size $N$.

\subsubsection{Decomposition}

We start with the following decomposition based on definition of $\Fmhat$ in \eqref{Bootstrap_objective_function},
\begin{align}
	&\bellmanestimationerror = \suptheta\bigg| \energyonebar\big\{ \Upsilon_\theta , \Belloptmulti\Upsilon_\theta \big\} - \energyonebarhat\big\{ \Upsilon_\theta , \Bellhatmulti\Upsilon_\theta \big\} \bigg|\leq \nonumber \\
	& \underbrace{\suptheta\bigg| \energyonebar\big\{ \Upsilon_\theta , \Belloptmulti\Upsilon_\theta \big\} - \energyonebar\big\{ \Upsilon_\theta , \Bellhatmulti\Upsilon_\theta \big\} \bigg|}_{(Term \ 1)}  + \underbrace{\suptheta\bigg| \energyonebar\big\{ \Upsilon_\theta , \Bellhatmulti\Upsilon_\theta \big\} - \energyonebarhat\big\{ \Upsilon_\theta , \Bellhatmulti\Upsilon_\theta \big\} \bigg| }_{(Term \ 2)}. \label{Term1_and_Term2_nonrealizable_step1}
\end{align}
Unfortunately, we cannot bound $(Term \ 1)$ with $\suptheta\energyonebar(\Belloptmulti\Upsilon_\theta,\Bellhatmulti\Upsilon_\theta)$ by applying triangular inequality, since $\energyonebar$ is not a metric. Instead, we can devise an alternative (Lemma \ref{energy_difference}), based on the fact that $\energyone$ is in fact a squared metric (Property \ref{squared_metric}), yet we need to pay price by having square-root. Refer to \ref{Proof_energy_difference} for its proof.
\begin{lemma}\label{energy_difference}
	For arbitrary $\Upsilon_0, \Upsilon_1,\Upsilon_2\in \mathcal{P}(\mathbb{R}^d)^{\SAspace}$, we have 
	\begin{align*}
		&\big| \energyonebar(\Upsilon_0,\Upsilon_1) - \energyonebar(\Upsilon_0,\Upsilon_2) \big| \leq 4\cdot \energyonebar(\Upsilon_1,\Upsilon_2)^{1/2} \times \bigg[ \max\bigg\{ \energyonebar(\Upsilon_0,\Upsilon_1), \energyonebar(\Upsilon_0, \Upsilon_2) \bigg\} + \energyonebar(\Upsilon_1,\Upsilon_2) \bigg]^{1/2}.
	\end{align*}
\end{lemma}
By applying Lemma \ref{energy_difference}, we obtain the following,
\begin{gather} 
	\GammaNm := \suptheta\energyonebar\big\{\Belloptmulti\Upsilon_\theta, \Bellhatmulti\Upsilon_\theta \big\}, \nonumber \\ %
	(Term \ 1) \leq 12\cdot \GammaNm^{1/2}\cdot \bigg\{ \GammaNm^{1/2} + \suptheta \energyonebar\big\{ \Upsilon_\theta, \Belloptmulti\Upsilon_\theta \big\}^{1/2} \bigg\}. \nonumber
\end{gather}

Now let us deal with $(Term \ 2)$ that can be decomposed as follows,
\begin{gather}
	(Term \ 2) \leq \suptheta \bigg| \sumsa \big\{\probsa - \probsahat\big\} \cdot \bigg[ \energyone\bigg\{ \Upsilon_\theta(s,a), \Bellhatmulti\Upsilon_\theta(s,a) \bigg\} - \energyone\bigg\{ \Upsilonzero(s,a), \Bellhatmulti\Upsilonzero(s,a) \bigg\}  \bigg] \bigg| \nonumber \\
	+ \bigg| \sumsa \big\{ \probsa - \probsahat \big\} \cdot \energyone\bigg\{ \Upsilonzero(s,a), \Bellhatmulti\Upsilonzero(s,a) \bigg\} \bigg|. \nonumber %
\end{gather}
To handle the first line of above decomposition, we can first obtain the following bound by \ref{Wassersup_diamF_satisfication} and contraction of $\Bellhat$ w.r.t. $\expectationbounder$, where we will use an abuse of notation $\Bellhat \theta:= \Bellhat \Upsilon_\theta$, 
\begin{gather}
	\bigg|\energyone\bigg\{ \Upsilon_\theta (s,a) , \Bellhatmulti \Upsilon_\theta(s,a) \bigg\} - \energyone\bigg\{ \Upsilonzero (s,a) , \Bellhatmulti \Upsilonzero(s,a) \bigg\} \bigg| \leq 4\cdot \expectationbounder(\theta,\theta_0) + 4\cdot  \expectationbounder\big\{ \Bellhatmulti \theta , \Bellhatmulti\theta_0 \big\} \nonumber \\
	\leq 4(1+\gamma^m) \cdot \Wassersup(\theta, \theta_0) \leq 4(1+\gamma^m)\cdot \diamF \label{energy_difference_trick}
\end{gather}
where the second last line holds, since a Bellman operator is a contraction with respect to $\expectationbounder$, as we mentioned under \eqref{objectiveinaccuracy_difference_boundedby_G} (proof by \citet{bellemare2017distributional} in their Lemma 3). Further applying \eqref{relaxed_triangle_pis2} on the second line of $(Term 2)$ decomposition, 
we can take up Decomposition (\ref{Term1_and_Term2_nonrealizable_step1}) as follows,
\begin{gather}
	\bellmanestimationerror \leq 12\cdot \GammaNm^{1/2}\cdot \bigg\{ \GammaNm^{1/2} + \suptheta \energyonebar\big\{ \Upsilon_\theta, \Belloptmulti\Upsilon_\theta \big\}^{1/2} \bigg\}  \nonumber \\
	+ 4\cdot \probabsnorm \cdot \bigg[ (1+\gamma^m)\cdot \diamF  \nonumber \\ %
	+ \supsa \energyone\bigg\{ \Upsilonzero(s,a), \Belloptmulti \Upsilonzero(s,a) \bigg\} + \supsa \energyone\bigg\{ \Belloptmulti\Upsilonzero(s,a), \Bellhatmulti\Upsilonzero(s,a) \bigg\} \bigg]. \nonumber
\end{gather}
Here, we can further simplify two terms, $\supsa \energyone\big\{ \Belloptmulti\Upsilonzero(s,a), \Bellhatmulti\Upsilonzero(s,a) \big\}$ and $\GammaNm$. First let $\big(\hat{S}^{(t)}(s,a), \hat{A}^{(t)}(s,a) \big)$ be the $t$-th state-action pair that follows the distribution (\ref{multistep_Estimated_Bellman_Operator}), that is the random state-action pair which can be reached by consecutively simulating from the estimated probability $\hat{p}(\cdots|s,a)$ and the target policy $\pi(a|s)$ starting from the initial state-action pair $s,a$. Furthermore, let us denote such probability (density) as $\hat{q}_{b_\mu}^{\pi:t}(\cdots|s,a)$ that is conditioned on a fixed initial state-action pair $s,a$, and denote the marginalized probability as $\hat{q}_{b_\mu}^{\pi:t}(\cdots)$ that treats the initial state-action pair $S,A\sim\probsa$ as random. This aligns with the notation $q_{b_\mu}^{\pi:t}(\cdots)$ defined below Assumption \ref{RN_derivative}. Then we have the following bound,
\begin{align}
	&\supsa \energyone\bigg\{ \Belloptmulti\Upsilonzero(s,a), \Bellhatmulti\Upsilonzero(s,a) \bigg\} \nonumber \\
	&\leq \supsa \bigg[ m\cdot \sumstep \energyone\bigg\{ (\Bellhat)^{m-t}(\Bellopt)^{t} \Upsilonzero(s,a), (\Bellhat)^{m-t-1}(\Bellopt)^{t+1} \Upsilonzero(s,a) \bigg\} \bigg] \ \ \text{by \eqref{relaxed_triangle_p}}  \nonumber \\
	&\leq m\cdot \sumstep \gamma^{m-t-1}\cdot \supsa \energyone \bigg\{ \Bellhat (\Bellopt)^t \Upsilonzero(s,a), \Bellopt (\Bellopt)^t \Upsilonzero(s,a) \bigg\}. \nonumber %
\end{align}
We also have following by the same logic of Appendix \ref{Proof_Fundamental_Realizable}, where the subscripts of $\Expectationtilde_{\hat{q}_{b_\mu}^{\pi:{m-t-1}}}$ and $\Expectationtilde_{b_\mu}$ indicate the distribution of $S,A$ (based on definition of $\Expectationtilde$ \eqref{Biased_Esa} that considers $\Bellhat$ as nonrandom),
\begin{gather}
	\GammaNm 
	\leq\frac{m}{p_{{\rm min}}} \cdot \sumstep \gamma^{m-t-1} \cdot \bigg[ \suptheta \bigg|\energyonebar\big\{ \Bellhat (\Bellopt)^t \Upsilon_\theta, \Bellopt(\Bellopt)^{t} \Upsilon_\theta \big\} - \energyonebar\big\{ \Bellhat (\Bellopt)^t \Upsilonzero, \Bellopt(\Bellopt)^{t} \Upsilonzero \big\} \bigg| \nonumber \\ 
	+ \supsa \energyone \bigg\{ \Bellhat (\Bellopt)^t \Upsilonzero(s,a), \Bellopt(\Bellopt)^{t} \Upsilonzero(s,a) \bigg\} \bigg], \label{GammaNm_decomposition} 
\end{gather}
which is based on $\supsa \{\hat{q}_{b_\mu}^{\pi:{m-t-1}}(s,a) / b_\mu(s,a) \} \leq 1/p_{{\rm min}}$.
Then we can bound $\suptheta|\Fmhat(\theta)- F_m(\theta)|$ by 
\begin{gather}
	12\cdot \bigg\{ \frac{m}{p_{{\rm min}}}\cdot \sumstep \gamma^{m-t-1}\cdot \big( \Yfirst + \Ysecond \big) \bigg\}^{1/2} 
	\times \bigg[ \suptheta \energytheta^{1/2} + \bigg\{ \frac{m}{p_{{\rm min}}}\cdot \sumstep \gamma^{m-t-1}\cdot (\Yfirst + \Ysecond) \bigg\}^{1/2} \bigg] \label{bellman_error_decomposition} \\ %
	+ 4\cdot\probabsnorm \times \bigg\{ (1+\gamma^m)\cdot \diamF + \supsa \energyzero(s,a)+ m\cdot \sumstep\gamma^{m-t-1}\cdot \Yfirst \bigg\} \nonumber %
\end{gather}
where each term is defined as 
\begin{gather*}
	\Yfirst := \supsa \energyone \bigg\{ \Bellhat (\Bellopt)^t \Upsilonzero(s,a), \Bellopt(\Bellopt)^{t} \Upsilonzero(s,a) \bigg\},\\
	\Ysecond := \suptheta \bigg|\energyonebar\big\{ \Bellhat (\Bellopt)^t \Upsilon_\theta, \Bellopt(\Bellopt)^{t} \Upsilon_\theta \big\} - \energyonebar\big\{ \Bellhat (\Bellopt)^t \Upsilonzero, \Bellopt(\Bellopt)^{t} \Upsilonzero \big\} \bigg| , \\
	\energytheta :=\energyonebar\big\{ \Upsilon_\theta, \Belloptmulti\Upsilon_\theta \big\} \ \ \ \& \ \ \ \energyzero(s,a):=\energyone\bigg\{ \Upsilonzero(s,a), \Belloptmulti\Upsilonzero(s,a) \bigg\}.
\end{gather*}

\subsubsection{Derivation Of Bound} \label{Proof_Fmhat_convergence_boundderivation}

Now we can see that there exist three random quantities: $\sumsa|\probsa - \probsahat|$, $\Yfirst$, $\Ysecond$ for each $t\in\{0,1,\cdots , m-1\}$.
We can employ similar techniques that we used in Appendix \ref{Proof_Realizable_Final_Theorem}. We will again assume $\Omegasubsetstageone$ of Definition (\ref{Conditional_Omega}), and utilize the conditional probability $\PROBN(\cdots)$. Under $\Omegasubsetstageone \ (\epsilon\in(0,1])$, whose probability is larger than 
\begin{align*}
	\PROB(\Omegasubsetstageone)\geq 1-C_{1}\cdot \exp(-C_{2}\cdot p_{{\rm min}}^2\cdot N\cdot \epsilon^2),
\end{align*}
we can verify by Appendix \ref{Delta_bounding_realizable} and (\ref{Omega_probability}),
\begin{align*}
	\probabsnorm \leq \frac{\sqrt{p_{{\rm min}}}}{2}\cdot \epsilon \ \ \ \text{for } \epsilon\in(0,1].
\end{align*}

The remaining two terms $\Yfirst$ and $\Ysecond$ are merely repetitions of what we showed in \ref{Gamma_bounding_realizable} that required Assumption \ref{bounded} and \ref{Lipschitz}, since they are in fact
\begin{gather}
	\Yfirst=\supsa X_{\theta_0}^{(t)}(s,a), \ \text{where } X_\theta^{(t)}(s,a):= \energyone\bigg\{ \Bellhat \Upsilon_\theta^{(t)}(s,a), \Bellopt \Upsilon_\theta^{(t)}(s,a) \bigg\}, \ \ \Upsilon_\theta^{(t)}:=(\Bellopt)^t \Upsilon_\theta, \nonumber \\
	\Ysecond=\suptheta \bigg| X_\theta^{(t)} - X_{\theta_0}^{(t)} \bigg|, \ \ \ \text{where } \ X_\theta^{(t)}:=  \energyonebar\big( \Bellhat \Upsilon_\theta^{(t)}, \Bellopt \Upsilon_\theta^{(t)} \big) = \sumsa \probsa \cdot X_{\theta}^{(t)}(s,a) , \nonumber %
\end{gather}
where the notations align with Definition (\ref{Gamma_sup_and_f0}) of $X_\theta$ and $X_\theta(s,a)$. The proofs are exactly the same except that $Z$ is replaced with $Z^{(t)}$ in definitions of $W_i^{\theta}$, $W_{ij}^\theta$, where $Z^{(t)}(s,a;\theta)\sim (\Bellopt)^t\Upsilon_\theta(s,a)$.
We can utilize the probability bounds (\ref{Gamma_supremum_part}) and (\ref{Gamma_fzero_part}). Let us first allow the following abuse of notation $(\Bellopt)^t\Theta$, which we will define as  
\begin{align} \label{abuse_Bellopt_parameter}
	(\Bellopt)^t\Theta := \bigg\{ (\Bellopt)^t \theta \ : \ \theta\in\Theta \bigg\} \ \ \text{for } \forall t\in\{0,1,2,\cdots, m-1\}, \ \ \text{where }\Bellopt\theta:=\Bellopt\Upsilon_\theta.
\end{align}
In order to further bound the metric entropy and the diameter based on \eqref{abuse_Bellopt_parameter}, we can develop a new metric,
\begin{gather*}
	\expectationbounder^{(t)}(\theta_1,\theta_2):= \expectationbounder\big\{ (\Bellopt)^t \theta_1, (\Bellopt)^t \theta_2  \big\}\leq \gamma^t \cdot \expectationbounder(\theta_1,\theta_2), \\
	\therefore \ \mathcal{N}\big((\Bellopt)^t \Theta , \expectationbounder, z\big) =  \mathcal{N}\big(\Theta , \expectationbounder^{(t)}, z\big).
\end{gather*}
Since it satisfies $\gamma^t$-Lipschitz continuity w.r.t. $\expectationbounder$, we can apply the logic in Appendix \ref{Proof_metricentropy_Lipschitz} to obtain
\begin{gather*}
	\int_0^\infty \sqrt{\log \mathcal{N}\big((\Bellopt)^t \Theta , \expectationbounder, z\big)}\mathrm{d}z \leq \gamma^t \cdot \metricentropyexpectatonbounderTheta, \\
	{\rm diam}\big( (\Bellopt)^t\Theta; \expectationbounder \big) \leq \gamma^t\cdot \diamF.
\end{gather*}
Applying the same logic that we used in Appendix \ref{Proof_Realizable_Final_Theorem}, conditioned under $\Omegasubsetstageone$ \eqref{Conditional_Omega}, we can derive the following for arbitrary $\epsilon\in(0,1]$, $\epsilon_1>0$, $u>0$, for each $t\in\{0,\cdots, m-1\}$,
\begin{gather}
	\probabsnorm \leq \frac{\sqrt{p_{{\rm min}}}}{2}\cdot \epsilon, \ \  \nonumber \\ 
	\Yfirst \leq 2\epsilon_1 + \frac{1}{N}\cdot\frac{8}{p_{{\rm min}}}\cdot \bigg\{ \sum_{k=0}^{t} \gamma^k\cdot \supsa \Expect\|R(s,a)\| + \gamma^{t+1} \cdot \supsa \Expect\|Z(s,a;\theta_0)\| \bigg\},  \nonumber \\
	\Ysecond \leq \frac{C_3\gamma^{t+1}}{\sqrt{N}} \cdot \sumsa\sqrt{\probsa}\cdot \bigg\{ \int_0^\infty \sqrt{\log \mathcal{N}\big(\Theta , \expectationbounder, z\big)}\mathrm{d}z + u\cdot {\rm diam}\big(\Theta; \expectationbounder \big) \bigg\}, \label{Yone_Ytwo} 
\end{gather}
with probability larger than the following.
\begin{align*}
	1&-2\exp(-u^2)  - (2|\SAspace|+6N) \cdot  \exp\bigg\{ \frac{-C_5\cdot p_{{\rm min}}\cdot N \cdot \epsilon_1^2}{ d^2\cdot \big(\gammasum\big)^2 \cdot \Rsapsisupsq + \gamma^2  \Zsasupsqtheta } \bigg\} %
\end{align*}

Before taking up Decomposition (\ref{bellman_error_decomposition}), we can further derive the following for $\forall t\in\{0,1,2,\cdots,m-1\}$, based on Inequality (\ref{Yone_Ytwo}),

This further leads \eqref{GammaNm_decomposition} to the following, by using $\sumsa\sqrt{\probsa}\leq 1/\sqrt{p_{{\rm min}}}$ \eqref{sqrt_prob_bounding} and $m\gammasum\leq m^2$, 
\begin{align}
	\GammaNm &\leq  \frac{m}{p_{{\rm min}}}\cdot \sumstep\gamma^{m-t-1}\cdot (\Yfirst + \Ysecond) \leq E_1(m)\cdot \epsilon_1 + \frac{1}{\sqrt{N}}\cdot E_2(m)\cdot u + \frac{1}{\sqrt{N}}\cdot E(m), \label{Yone_plus_Ytwo}
\end{align}
where
\begin{align} 
	E_1(m)&:= \frac{2}{p_{{\rm min}}}\cdot m\gammasum \quad \& \quad E_2(m):= C_4\cdot m^2\gamma^m \cdot \frac{1}{p_{{\rm min}}} \sumsa \sqrt{\probsa}\cdot \diamF,  \nonumber \\
	E(m)&:= C_4\cdot  m^2\cdot \frac{1}{p_{{\rm min}}^2}\cdot \bigg\{ \gamma^m\cdot \metricentropyexpectatonbounderTheta + \frac{1}{\sqrt{N}}\cdot \sumstep\gamma^t \cdot \supsa\Expect\|R(s,a)\| \nonumber \\ 
	&\qquad \qquad \qquad \qquad \qquad  + \frac{1}{\sqrt{N}}\cdot \gamma \cdot \supsa \Expect\|Z(s,a;\theta_0)\| \bigg\}.  \nonumber %
\end{align}
Now let us get back to Decomposition (\ref{bellman_error_decomposition}) by further simplifying each line. Towards that end, we will first adjust the variables as follows for arbitrary $\epsilon_0\in(0,1]$,
\begin{gather}
	u=\sqrt{N}\cdot p_{{\rm min}}\cdot \epsilon_0, \ \ \& \ \ \epsilon_1=\sqrt{p_{{\rm min}}}\cdot \epsilon_0, \ \ \& \ \ \nonumber \\
	\epsilon=\frac{\epsilon_0}{ d\cdot \gammasum \cdot \Rsapsisup + \Zsasuptheta + 1 } \in(0,1] \nonumber %
\end{gather}

The first line of \eqref{bellman_error_decomposition} can be bounded as follows, based on (\ref{Yone_plus_Ytwo}), where $G(m):= E_1(m)\cdot \sqrt{p_{{\rm min}}} + E_2(m)\cdot p_{{\rm min}}$,
\begin{align*}
	\text{Line-1 of \eqref{bellman_error_decomposition}} &\leq 12\cdot \bigg\{ G(m)\cdot \epsilon_0 + \frac{1}{\sqrt{N}}\cdot E(m) \bigg\}^{1/2} \cdot \bigg[ \suptheta\energytheta^{1/2} + \bigg\{ G(m)\cdot \epsilon_0 + \frac{1}{\sqrt{N}}\cdot E(m) \bigg\}^{1/2} \bigg]\\
	&\leq 12\cdot \bigg\{ \suptheta \energytheta^{1/2} + \sqrt{E(m)} + \sqrt{G(m)} \bigg\}\cdot \bigg\{ \frac{\sqrt{E(m)}}{N^{1/4}} + \sqrt{G(m)}\cdot \sqrt{\epsilon_0} \bigg\},
\end{align*}
where the last line used $\epsilon_0\in(0,1]$ and $\sqrt{x+y}\leq \sqrt{x}+\sqrt{y}$ for $x,y\geq 0$. Next we can also obtain the following,
\begin{align*}
	&\text{Line-2 of \eqref{bellman_error_decomposition}} \leq C_7\cdot \sqrt{p_{{\rm min}}} \cdot \bigg\{ \diamF + m\gammasum\cdot \bigg( 1+ \frac{1}{N}\cdot \frac{1}{p_{{\rm min}}} \bigg) \bigg\}\cdot \epsilon_0.
\end{align*}
Switching the notation $\epsilon_0$ into $\epsilon$, combining these eventually allows us to further rewrite Decomposition (\ref{bellman_error_decomposition}). 
By using $m\gammasum\leq m^2$ and \eqref{sqrt_prob_bounding}, for $\forall \epsilon \in (0,1]$, we have 
\begin{align}
	\bellmanestimationerror 
	&\leq C_{16}\cdot m^2\cdot \frac{1}{p_{{\rm min}}^2}\cdot \Cenvmulti \cdot \bigg( \frac{1}{N^{1/4}}+ \sqrt{\epsilon} \bigg) \nonumber %
\end{align}
The probability bound for \eqref{Yone_Ytwo} can be integrated for all $t\in\{0,1,2,\cdots,m-1\}$, combined with the same trick that we used in (\ref{prob_conditional_bound}), to obtain the following lower bound,
\begin{gather}
	1-C_{10}\cdot m\cdot (|\SAspace| + N) \cdot \exp\bigg\{-C_{11}\cdot p_{{\rm min}}^2 \cdot N \cdot \epsilon^2 / C_{\rm den}(m) \bigg\}, \nonumber %
\end{gather}
where the terms are defined as 
\begin{gather}
	C_{\rm den}(m):=d^2\cdot \big(\gammasum\big)^2 \cdot \Rsapsisupsq + \gamma^2 \cdot \Zsasupsqtheta + 1 , \label{C_den_multi} \\
	\Cenvmulti:=L \cdot (\gamma^m  \sqrt{p}+1)\cdot \diamTheta+ \gammasum \cdot \supsa \Expect\|R(s,a)\| + (1+\gamma^m)\cdot \supsa \Expect\|Z(s,a;\theta_0)\| + 1. \nonumber %
\end{gather}
This gives us the desired result of Lemma \ref{Fmhat_convergence}.

\subsection{Obtaining The Bound Of Bootstrap-based Objective Function} \label{bootstrap_objective_function_converging}

Our final estimator of the objective function $\FBhat$ \eqref{Bootstrap_objective_function} is based on bootstrap, not $\Fmhat$ \eqref{Bootstrap_objective_function} covered in Lemma \ref{Fmhat_convergence}. So we shall develop it into following.

\begin{lemma} \label{FBhat_convergence}
	Under same assumptions of Lemma \ref{Fmhat_convergence}, for any $m\in\mathbb{N}$ and arbitrary $\epsilon,\epsilon'\in(0,1]$,
	\begin{align*}
		\suptheta\bigg| \FBhat(\theta) - F_m(\theta) \bigg|\leq \frac{C_{32}}{p_{{\rm min}}^2}\cdot \Benvmulti\cdot \bigg\{ m^2\cdot \bigg( \frac{1}{N^{1/4}} + \sqrt{\epsilon} \bigg) + m\cdot \bigg(\frac{1}{M^{1/4}} + \sqrt{\epsilon'}\bigg)  \bigg\} 
	\end{align*}
	holds with probability larger than 
	\begin{align*}
		& 1- \mathcal{D}(N)-C_{1}\cdot m\cdot (|\SAspace| + N) \cdot \exp(-C_{2}\cdot p_{{\rm min}}^2 \cdot N \cdot \epsilon^2 / C_{\rm den}(m)) \nonumber \\
		& \ \ \ - C_{1}\cdot (|\SAspace| + M)\cdot \exp(-C_{2}\cdot p_{{\rm min}}^2 \cdot M \cdot {\epsilon'}^2 / B_{\rm den}(m)),
	\end{align*}
	where $C_{\rm den}(m)$, $B_{\rm den}(m)$, $\Benvmulti$ are defined in \eqref{C_den_multi}, \eqref{B_env_multi}, and $\mathcal{D}(N)\rightarrow 0$ as in \eqref{New_Terms_Created2}. 
\end{lemma}

\subsubsection{Three Stages Of Probability Space}

We can decompose the term $\suptheta | \FBhat(\theta) - F_m(\theta)|$ as follows,
\begin{align}
	\suptheta\bigg| \FBhat(\theta) - F_m(\theta) \bigg| \leq \suptheta\bigg| \FBhat(\theta) - \Fmhat(\theta) \bigg| + \bellmanestimationerror. \label{bootstrap_first_decomposition}
\end{align}
At this point, we should recognize that our probability space \eqref{probability_stage1_2} is expanded due to bootstrapping procedure reflected in $\FBhat$. Now our probability space $(\Omega, \Sigma, \PROB)$ can be factorized into three stages,
\begin{align}
	\text{Stage 1:}& \ (\Omega_{\SAspace}, \Sigma_{\SAspace}, \PROB_{\SAspace}) \Rightarrow \text{determines which state-action pairs } S_i,A_i \text{ are sampled,}  \label{probability_stage1_3} \\
	\text{Stage 2:}& \ (\Omega^{(\mathbf{N})}, \Sigma^{(\mathbf{N})}, \PROBN) \Rightarrow \text{conditioned on }(S_i,A_i), \text{ determines } R_i,S_i' \sim p(\cdots|S_i,A_i), \nonumber \\
	\text{Stage 3:}& \ (\Omega_B^{(\mathcal{D})}, \Sigma_B^{(\mathcal{D})}, \PROB_B^{(\mathcal{D})}) \Rightarrow \text{conditioned on }\mathcal{D}, \text{ determines the bootstrapped trajectories in \eqref{Bootstrap_operator}}. \nonumber
\end{align}
We have already bounded $\suptheta | \Fmhat(\theta) - F_m(\theta)|$ of \eqref{bootstrap_first_decomposition} in Lemma \ref{Fmhat_convergence}, which is controlled by Stage 1 and 2 probability spaces \eqref{probability_stage1_3}. Now the remaining term $\suptheta | \FBhat(\theta) - \Fmhat(\theta)|$ of \eqref{bootstrap_first_decomposition} is solely based on Stage 3 probability space \eqref{probability_stage1_3}, conditioned on the observed data $\mathcal{D}=\{(s_i,a_i,r_i,s_i')\}_{i=1}^{N}$. However, since the bootstrapped probability space (Stage 3) is affected by what was observed in the previous two stages, we will assume some nice properties are satisfied in Stage 1 and Stage 2 probability spaces, which are already mentioned within the proof of Lemma \ref{Fmhat_convergence} in \ref{Proof_Fmhat_convergence}.

\subsubsection{Inherited Results From Lemma \ref{Fmhat_convergence}} \label{Inherited_Results_from_Lemma_Fmhat_convegence}

Here we will define two events. The first event $E_{1,a}$ is equivalent event to $\Omegasubsetstageone$ \eqref{Conditional_Omega}, under which Facts \eqref{Fact1to3} are satisfied.
Next, we will define the second event $E_{1,b}$ where the following two statements are satisfied, which we inherit from Lemma \ref{Fmhat_convergence},
\begin{gather}
	\GammaNm\leq E_1(m)\cdot \sqrt{p_{{\rm min}}}\cdot \epsilon + E_2(m)\cdot p_{{\rm min}}\cdot \epsilon + \frac{1}{\sqrt{N}}\cdot E(m) , \label{GammaNm_decomposition_temp} \\
	\bellmanestimationerror \leq C_{1}\cdot m^2\cdot \frac{1}{p_{{\rm min}}^2}\cdot \Cenvmulti \cdot \bigg( \frac{1}{N^{1/4}}+ \sqrt{\epsilon} \bigg) \nonumber
\end{gather}
We have derived in the last part of Appendix \ref{Proof_Fmhat_convergence_boundderivation} that both events hold with probability larger than $1-C_{2}\cdot m\cdot (|\SAspace| + N) \cdot \exp\big\{ -C_{3}\cdot p_{{\rm min}}^2 \cdot N \cdot \epsilon^2 / C_{\rm den}(m) \big\}$.

Let us assume that the events $E_{1,a}$ and $E_{1,b}$ both hold, and then bound the term $\suptheta | \FBhat(\theta) - \Fmhat(\theta) |$ of \eqref{bootstrap_first_decomposition}. We need to emphasize that at this moment, $\mathcal{D}$ is given, and Stage 3 probability space \eqref{probability_stage1_3} is the only source of probability. In other words, we can consider $\Fmhat$ as our population objective function, which is based upon $\hat{b}_\mu$ (\ref{probabiliy_stateaction}) and $\probhatmstep(\cdots|s,a)$. $\probhatmstep(\cdots|s,a)$ represents the empirical measure of $(\sumstep\gamma^t \hat{R}^{(t)}, \hat{S}^{(m)})$ conditioned on initial state-action pair $s,a$ that can occur by applying $\hat{p}(r,s'|s,a)$ and $\pi(a|s)$ for $m$ consecutive times (\ref{multistep_Estimated_Bellman_Operator}). In other words, by treating $\Bellhatmulti$ as the population operator and $\Bootstrapopt$ as its approximation, we can obtain
\begin{align}
	\suptheta\bigg| \FBhat(\theta) - \Fmhat(\theta) \bigg| &\leq 12\cdot \Gammabootstrap^{1/2}\cdot \bigg\{ \Gammabootstrap^{1/2} + \suptheta \energyonebarhat \big\{ \Upsilon_\theta , \Bellhatmulti\Upsilon_\theta \big\}^{1/2} \bigg\} \ \ \ \text{by same logic in Appendix \ref{Proof_Fmhat_convergence}}, \nonumber \\ 
	&\leq C_4\cdot \Gammabootstrap^{1/2}\cdot \bigg\{ \Gammabootstrap^{1/2} + \suptheta \energytheta^{1/2} + \GammaNm^{1/2} \bigg\}. \label{Three_Terms_bootstrap}
\end{align}
where we have a new term that we will refer to as \textit{bootstrap discrepancy} 
\begin{align} %
	\Gammabootstrap:= \suptheta \energyonebarhat\big\{ \Bellhatmulti\Upsilon_\theta , \Bootstrapopt\Upsilon_\theta \big\}. \nonumber
\end{align}
Let us bound the three terms one by one. First, we can bound the supremum term $\suptheta\energytheta^{1/2}$ by following, based on Assumption \ref{Lipschitz},
\begin{align}
	2\cdot \bigg\{ \sqrt{1+\gamma^m} \cdot \sqrt{L} \cdot \sqrt{\diamTheta} + \bigg(\gammasum\cdot\supsa \Expect\|R(s,a)\|\bigg)^{1/2}  + \bigg((1+\gamma^m)\cdot\supsa \Expect\|Z(s,a;\theta_0)\|\bigg)^{1/2} \bigg\} \nonumber %
\end{align}
Based on what we have mentioned in the beginning of \ref{Inherited_Results_from_Lemma_Fmhat_convegence}, we can further bound Bellman discrepancy as follows by using \eqref{GammaNm_decomposition_temp},
\begin{gather}
	\GammaNm^{1/2} 
	\leq C_8\cdot \frac{m}{p_{{\rm min}}^{1/4}} \cdot \bigg\{ 1+\gamma^{m/2}\cdot \sqrt{L}\cdot \sqrt{\diamTheta} \bigg\}\cdot \sqrt{\epsilon} + \frac{C_8}{N^{1/4}} \cdot \frac{m}{p_{{\rm min}}}\cdot \sqrt{L}\cdot p^{1/4}\cdot \gamma^{m/2}\cdot \sqrt{\diamTheta}  \nonumber \\
	+ \frac{C_8}{\sqrt{N}}\cdot \frac{m}{p_{{\rm min}}} \cdot \bigg\{ \bigg( \gammasum \cdot \supsa \Expect\|R(s,a)\| \bigg)^{1/2} + \bigg(\gamma\cdot \supsa \Expect\|Z(s,a;\theta_0)\| \bigg)^{1/2} \bigg\}, \nonumber %
\end{gather}
where the second last inequality can be derived by putting together Assumption \ref{Lipschitz}, Inequality (\ref{sqrt_prob_bounding}), and Remark \ref{metricentropy_Lipschitz}. Since both of these hold under $E_{1,a}$ and $E_{1,b}$, the probability is not reduced.

\subsubsection{Bounding Bootstrap Discrepancy}

In further bounding \eqref{Three_Terms_bootstrap}, bootstrap discrepancy $\Gammabootstrap$ is the only term is probabilistic due to Stage 3 probability space \eqref{probability_stage1_3}. Comparing \eqref{multistep_Estimated_Bellman_Operator} and (\ref{Bootstrap_operator}), we can see that $\Bootstrapopt$ is in fact the single-step estimator of $\Bellhatmulti$ that can be viewed as a new population operator in the new probability space generated by bootstrapping from the already-observed data $\mathcal{D}$. In this regard, the relationship between $\Bootstrapopt$ and $\Bellhatmulti$ aligns with that between $\Bellhat$ and $\Bellopt$, only with a few differences. The reward $R$ is replaced by discounted sum $\sum_{t=1}^{m}\gamma^{t-1}\hat{R}^{(t)}$, $S'$ is replaced by $\hat{S}^{(m)}$, $A'$ is replaced by $\hat{A}^{(m)}$, and the discount rate $\gamma$ is replaced by $\gamma^m$.  In addition, several other quantities are also replaced as follows,
\begin{gather}
	\probsa \leftarrow \probsahat, \ \ \ \Expect(\cdots) \leftarrow \Expectationtilde(\cdots), \ \ \ \|\cdot\|_{\psi_2}\leftarrow \|\cdot\|_{\psitwotilde}, \ \ \ N\leftarrow M, \nonumber \\
	p_{{\rm min}}\leftarrow \hat{p}_{\rm min}:=\min\{ \probsahat \ : \ \probsahat>0\}=\min\{\probsahat\} \ \  \text{by Facts (\ref{Fact1to3})}. \nonumber %
\end{gather}
where $\Expectationtilde(\cdots)$ \eqref{Biased_Esa} and $\|\cdot\|_{\psitwotilde}$ are the expectation and sub-Gaussian norms corresponding to the conditional probability measure $\PROB(\cdots|\mathcal{D})$. With the replacements by the estimated quantities (that will now be regarded as a new population quantity in Stage 3 probability space), we can replicate the proofs of \ref{Gamma_bounding_realizable}. Skipping the calculation details, we obtain the following with 
an arbitrary $\epsilon'\in(0,1]$,
\begin{align}
	\Gammabootstrap&\leq \frac{1}{M}\cdot\frac{8}{\hat{p}_{\rm min}}\cdot \bigg( \sum_{t=1}^{m}\gamma^{t-1}\cdot \supsa \Expectationtilde\|\hat{R}(s,a)\| + \gamma^m \cdot \supsa \Expect\|Z(s,a;\theta_0)\| \bigg) \label{Gamma_Bootstrap_valuebound} \\
	& \quad + \frac{C_{13}}{\sqrt{M}}\cdot \gamma^m\cdot \sumsa\sqrt{\probsahat}\cdot L\sqrt{p}\cdot \diamTheta + C_{14}\cdot (1+\gamma^m)\cdot \sqrt{\hat{p}_{\rm min}}\cdot \epsilon', \nonumber
\end{align}
with probability larger than
\begin{align}
	&1-C_{15}\cdot(|\SAspace|+M) \times \exp\bigg\{ \frac{-C_{16}\cdot \hat{p}_{\rm min}^2\cdot M \cdot {\epsilon'}^2}{\big( \sum_{t=1}^{m}\gamma^{t-1}\cdot \supsa \big\| \| \hat{R}^{(t)}(s,a) \| \big\|_{\psitwotilde} + \gamma^{m}\cdot \supsa \Expect\|Z(s,a;\theta_0)\| + 1 \big)^2 } \bigg\}. \nonumber  %
\end{align}
Note that this is probability bound conditioned under $E_{1,a}\cap E_{1,b}$.

\subsubsection{Finalizing The Bound}

The bound \eqref{Gamma_Bootstrap_valuebound} are not yet useful though, since they are not fully represented with population quantities. This is because we are caring about Stage 3 probability space \eqref{probability_stage1_3} conditioned upon the observed data $\mathcal{D}$ (that is associated with Stage 1 and 2 probability spaces). So we hope to bound the following terms with the corresponding population quantities,
\begin{align}
	\sumsa \sqrt{\probsahat} \ \ \& \ \ \hat{p}_{\rm min} \ \ \& \ \ \supsa \Expectationtilde\|\hat{R}(s,a)\| \ \ \& \ \ \supsa \big\| \|\hat{R}(s,a)\| \big\|_{\psitwotilde}, \label{estimated_quantities_to_bound}
\end{align}
but it comes with a price, that is subtraction of probability.

Let us first condition upon $E_{1,a}$. It is straightforward to derive $\sumsa \sqrt{\probsahat}\leq\sqrt{\frac{3}{2}}\cdot \sumsa \sqrt{\probsa}$ and $\frac{1}{2} p_{{\rm min}}\leq \hat{p}_{\rm min} \leq \frac{3}{2}p_{{\rm min}}$.
Unlike the first two terms of (\ref{estimated_quantities_to_bound}) that could be bounded solely $E_{1,a}$, the remaining two terms cannot be deterministically bounded, necessitating the derivation of probabilistic bound.

Since we are conditioning on $E_{1,a}$, we should deal with Stage 2 probability space \eqref{probability_stage1_3}, using the conditional probability $\PROBN(\cdots)$ introduced below Definition (\ref{Conditional_Omega}). Based on Derivation (\ref{supRsa_bounding}) for an arbitrary $\epsilon_3'\in(0,1]$,
\begin{align}
	&\PROBN\bigg[ \supsa \Expectationtilde\|\hat{R}(s,a)\| \leq \supsa \Expect\|R(s,a)\| + \sqrt{p_{{\rm min}}}\cdot\epsilon_3' \bigg]  \geq 1 - 2|\SAspace|\cdot \exp\bigg\{ \frac{-C_{17} \cdot p_{{\rm min}}^2 \cdot N\cdot {\epsilon_3'}^2}{ d^2\cdot \big( \supsa  \|R(s,a)\|_{\psi_2} \big)^2 } \bigg\}, \nonumber
\end{align}
Now let us bound the forth term of (\ref{estimated_quantities_to_bound}). First, let $s,a\in\SAspace$ be arbitrary. Define two random variables $U(s,a):=\|R(s,a)\|$ and $\hat{U}(s,a):=\|\hat{R}(s,a)\|$, along with the following functions,
\begin{gather*}
	A_{s,a}(t):=\Expect\bigg\{ \exp\bigg( \frac{U(s,a)^2}{t^2} \bigg) \bigg\}, \quad 
	\hat{A}_{s,a}(t):= \Expectationtilde \bigg\{ \exp\bigg( \frac{\hat{U}(s,a)^2}{t^2} \bigg) \bigg\} = \frac{1}{\Nsa}\sumNsa \exp\bigg( \frac{U_i(s,a)^2}{t^2} \bigg),
\end{gather*}
where $U_i(s,a)=\|R_i(s,a)\| \ (1\leq i \leq \Nsa)$ represents the samples. Let $t_0(s,a)>0$ be the value such that $A_{s,a}\big(t_0(s,a)\big) = 1$.
It is obvious to see $t_0(s,a) > \|\|R(s,a)\|\|_{\psi_2}$, based on that $A_{s,a}(\|\|R(s,a)\|\|_{\psi_2})=2$ holds and $A_{s,a}(t)$ is a strictly decreasing function. We can bound its probability term as follows,
\begin{align}
	\PROBN\big( \big\| \|\hat{R}(s,a)\| &\big\|_{\psitwotilde} \leq t_0(s,a) \big) = \PROBN\big\{ \hat{A}_{s,a}\big(t_0(s,a)\big) \leq 2 \big\} \ \ \ \text{by Definition (\ref{subgaussian_norm_definition})} \nonumber \\
	&\geq \PROBN\bigg\{ \bigg|\hat{A}_{s,a}\big(t_0(s,a)\big) - A_{s,a}\big(t_0(s,a)\big) \bigg| \leq 2-A_{s,a}\big(t_0(s,a)\big) \bigg\}  \nonumber \\
	&\geq 1-\ExpectN\big| \hat{A}_{s,a}\big(t_0(s,a)\big) - A_{s,a}\big(t_0(s,a)\big) \big| \ \ \ \text{by Markov's Inequality}. \nonumber %
\end{align}
Note that we could apply Markov's Inequality in the last line since $\Expect|A_{s,a}(t)|<\infty$ by sub-Gaussianity assumption \ref{bounded} that implies $ \|U(s,a)\|= \| \|R(s,a)\| \|_{\psi_2}<\infty$. 
Now let us shrink the above term
with the following lemma that is proved in \ref{Proof_expectation_shrinkage_finite_firstmoment},
\begin{lemma} \label{expectation_shrinkage_finite_firstmoment}
	If $X_i \ (1\leq i \leq n)$ are iid with $\Expect(X_1)=0$, $\Expect|X_1|<\infty$, then the expectation of the sample mean shrinks to zero as follows,
	\begin{align*}
		\Expect\big| \bar{X}_n \big| \leq \inf_{z>0}\bigg[ \frac{1}{\sqrt{n}}\cdot \bigg\{ \Expect\big\{ X_1^2\cdot \mathbf{1} (|X_1|\leq z ) \big\} \bigg\}^{1/2} + \Expect\big\{ |X_1|\cdot \mathbf{1}(|X_1|>z) \big\} \bigg]  \rightarrow 0 \ \ \ \text{as }n\rightarrow\infty.
	\end{align*}
	Note that we have a deterministic sequence its convergence to zero is guaranteed, however its speed depends on the tail of the distribution $X_1$.
\end{lemma}
By defining $V(s,a) := \exp( U(s,a)^2 / t_0(s,a)^2 )$, we can apply Lemma \ref{expectation_shrinkage_finite_firstmoment} to bound $\ExpectN | \hat{A}_{s,a} (t_0(s,a)) - A_{s,a}(t_0(s,a) ) |$ by
\begin{align}
	\sqrt{\frac{2}{p_{{\rm min}}}} \cdot \inf_{z>0} \bigg[ \frac{1}{\sqrt{N}}\cdot \bigg\{ \Expect\big\{ V(s,a)^2\cdot \mathbf{1}\big( V(s,a)\leq z \big) \big\} \bigg\}^{1/2} + \Expect\big\{ V(s,a)\cdot \mathbf{1}\big( V(s,a)>z \big) \big\}  \bigg], \nonumber %
\end{align}
where the last line holds by Facts (\ref{Fact1to3}). Now defining the following new variable $t_0^*:=\supsa t_0(s,a)$ which implies $A_{s,a}(t_0^*)\leq 1$ for all $s,a\in\SAspace$,
we have the following (see \eqref{New_Terms_Created2} for definition of $\mathcal{D}(N)$), 
\begin{align}
	\PROBN\bigg\{ \supsa \big\| \|\hat{R}(s,a)\| \big\|_{\psitwotilde}  \leq t_0^* \bigg\} \
	&\geq 1- \mathcal{D}(N), \nonumber %
\end{align}

By letting $\epsilon_3'=\epsilon'\in(0,1]$, we can bound all four estimated quantities (\ref{estimated_quantities_to_bound}) at the same time as follows,
\begin{gather}
	\supsa \Expectationtilde\|\hat{R}(s,a)\| \leq \supsa\Expect\|R(s,a)\| + \epsilon' \quad \& \quad \supsa \big\| \|\hat{R}(s,a)\| \big\|_{\psitwotilde} \leq t_0^*, \nonumber %
\end{gather}
with probability larger than
\begin{align} 
	1-2|\SAspace|\cdot \exp\bigg\{ \frac{-C_{17} \cdot p_{{\rm min}}^2 \cdot N\cdot {\epsilon'}^2}{ d^2\cdot\big(\supsa \|R(s,a)\|_{\psi_2}\big)^2 } \bigg\} - \mathcal{D}(N). \nonumber %
\end{align}
Let us define a new event called $E_2$ where all these four quantities \eqref{estimated_quantities_to_bound} are bounded as we mentioned.

Now that we have bounded the estimated quantities with its population counterparts, we can rewrite the bound of $\Gammabootstrap$ \eqref{Gamma_Bootstrap_valuebound} accordingly, under $E_{1,a}, E_{1,b}, E_2$. Then, skipping the technical details, we can aggregate all aforementioned results to obtain 
\begin{gather}
	\suptheta\bigg| \FBhat(\theta) - \Fmhat(\theta) \bigg| 
	\leq \frac{C_{31}}{p_{{\rm min}}^{3/2}}\cdot m \cdot \Benvmulti\cdot \bigg( \frac{1}{M^{1/4}}+ \sqrt{\epsilon'} \bigg) \label{B_env_multi} \\
	\text{where}\quad\Benvmulti:= L\sqrt{p}\cdot \diamTheta + \gammasum\cdot \supsa\Expect\|R(s,a)\| + \supsa \Expect\|Z(s,a;\theta_0)\| + 1, \nonumber \\
	\text{and}\quad B_{\rm den}(m):= \big(\gammasum\big)^2 \cdot {t_0^*}^2 + \gamma^{2m} \cdot \Zsasupsqtheta + 1, \nonumber 
\end{gather}
with probability larger than 
\begin{align}
	&1- \mathcal{D}(N)-C_{22}\cdot m\cdot (|\SAspace| + N)  \cdot \exp( -C_{23}\cdot p_{{\rm min}}^2 \cdot N \cdot \epsilon^2 / C_{\rm den}(m) ) \nonumber \\
	& \quad - C_{22}\cdot (|\SAspace| + M)\cdot  \exp(-C_{23}\cdot p_{{\rm min}}^2 \cdot M \cdot {\epsilon'}^2 / B_{\rm den}(m)). \nonumber %
\end{align}

Now the final task is to incorporate this with the bound of Lemma \ref{Fmhat_convergence}, based on Decomposition (\ref{bootstrap_first_decomposition}). As we mentioned in the beginning of \ref{Inherited_Results_from_Lemma_Fmhat_convegence}, we already inherited the bound, so we do not have to subtract any additional probability from the current bound \eqref{B_env_multi}. Then we have
\begin{align}
	\suptheta\bigg| \FBhat(\theta) - F_m(\theta) \bigg|
	&\leq C_{1}\cdot \frac{1}{p_{{\rm min}}^2}\cdot m^2 \cdot \Cenvmulti \cdot \bigg( \frac{1}{N^{1/4}}+ \sqrt{\epsilon} \bigg) + \frac{C_{31}}{p_{{\rm min}}^{3/2}}\cdot m \cdot \Benvmulti \cdot \bigg( \frac{1}{M^{1/4}}+ \sqrt{\epsilon'} \bigg) \nonumber \\   %
	&\leq \frac{C_{32}}{p_{{\rm min}}^2}\cdot \Benvmulti\cdot \bigg\{ m^2\cdot \bigg( \frac{1}{N^{1/4}} + \sqrt{\epsilon} \bigg) + m\cdot \bigg(\frac{1}{M^{1/4}} + \sqrt{\epsilon'}\bigg)  \bigg\} \nonumber %
\end{align}
with probability larger than 
\begin{align*}
	& 1- \mathcal{D}(N)-C_{22}\cdot m\cdot (|\SAspace| + N) \cdot \exp(-C_{23}\cdot p_{{\rm min}}^2 \cdot N \cdot \epsilon^2 / C_{\rm den}(m)) \nonumber \\
	& \ \ \ - C_{22}\cdot (|\SAspace| + M)\cdot \exp(-C_{23}\cdot p_{{\rm min}}^2 \cdot M \cdot {\epsilon'}^2 / B_{\rm den}(m)). 
\end{align*}

\subsection{Final Convergence} \label{Proof_thetaBhat_convergence}

We start from the following assumption, which is weaker than Assumption \ref{mstep_strongconvex_all} that we stated for splitting-based estimator (Section \ref{splitdata_operator}).

\begin{assumption} \label{uniqueness_generalized}
	For some $q\geq 1$, the inaccuracy function $F(\cdot):\Theta\subset \mathbb{R}^p\rightarrow\mathbb{R}$ \eqref{objective_functions} satisfies $F(\theta)\geq F(\thetatilde) + c_q\cdot \|\theta - \thetatilde\|^q$ for some $c_q>0$.
\end{assumption}

\subsubsection{Inaccuracy Of Parameter Estimation} \label{inaccuracy_of_parameter_estimation}

Our idea is that larger $N,M,m$ will lead to tighter (probabilistic) bound of $\suptheta| \FBhat(\theta) - F(\theta) |$, which can be decomposed as follows,
\begin{align*}
	\suptheta\bigg| \FBhat(\theta) - F(\theta) \bigg| \leq \suptheta\bigg| \FBhat(\theta) - F_m(\theta) \bigg| + \suptheta\bigg| F_m(\theta) - F(\theta) \bigg|.
\end{align*}

Note that the first term of RHS is the probabilistic term that we bounded in Lemma \ref{FBhat_convergence}, and the second term is a deterministic term that can be bounded based on following (proof in \ref{Proof_Fact1to2}), with $C_{\rm bias}$ defined in \eqref{New_Terms_Created2},
\begin{align}
	\suptheta\bigg| F_m(\theta) - F(\theta) \bigg| \leq 4\gamma^m\cdot C_{\rm bias} \quad %
	&\text{with} \quad \expectationbounder(\theta,\pi):=\expectationbounder \big\{\Upsilon_\theta, \Upsilon_\pi \big\} \label{Fact1to2}.
\end{align}

Now in order to relate this to the estimation inaccuracy of $\thetabootstrap$ \eqref{Bootstrap_objective_function}, we shall use the function $\psi(\cdot)$ introduced by Example 1.3 suggested by \citet{sen2018gentle},
\begin{align}
	\psi(\delta):=\inf_{\theta\in\Theta:\| \theta - \thetatilde \| \geq \delta}F(\theta) - F(\thetatilde) \quad \mbox{and} \quad \psi^{-1}(y):=\inf\bigg\{ \delta>0: \psi(\delta)\geq y \bigg\}.  \label{psibar_psibarinverse_definition} 
\end{align}
Depending on whether $\Theta$ includes any element in the outermost boundary, $\psi(\cdot)$ can be defined either for $0\leq \delta < \sup_{\theta\in\Theta}\|\theta - \thetatilde\|$ or $0\leq \delta \leq \sup_{\theta\in\Theta}\|\theta - \thetatilde\|$. We will extend the function in the following trivial way of extending horizontally from the rightmost point, 
\begin{align*}
	\psi(\delta):=\begin{cases}
		\psi(\suptheta\|\theta - \thetatilde\|) & \text{if } \psi(\suptheta\|\theta - \thetatilde\|) \text{ is defined,} \\
		\sup_{0\leq \delta'<\suptheta\|\theta - \thetatilde\|} \psi(\delta') & \text{otherwise.}
	\end{cases}
\end{align*}
There are several important properties of $\psi(\cdot)$ that are proved in \ref{Proof_psifunction_properties} based on Assumption \ref{uniqueness_generalized}.
\begin{remark} \label{psifunction_properties}
	The function $\psi$ \eqref{psibar_psibarinverse_definition} satisfies the following properties.
	\begin{enumerate}
		\item $\psi^{-1}(\cdot)$ is an increasing function such that $\psi^{-1}(y)\rightarrow 0$ as $y\rightarrow0$ and $\psi^{-1}(y)=\infty$ for $y>\sup_{\delta>0}\psi(\delta)$.
		\item $\lim_{\epsilon\rightarrow 0+} \psi \big\{ \psi^{-1}(y) + \epsilon \big\}\geq y$ holds for all $y\in[0, \psi(\suptheta\|\theta - \thetatilde\|)]$.
		\item Let $\hat{F}:\Theta \subset \mathbb{R}^p \rightarrow \mathbb{R}$ be an arbitrary estimate of $F$ \eqref{objective_functions} that have minimizer(s) within $\Theta$. For an arbitrary value $\delta>0$, if there exists a minimizer $\exists\hat{\theta}\in\arg\min_{\theta\in\Theta}\hat{F}(\theta)$ such that $\|\hat{\theta} - \thetatilde\|>\delta$, then $\suptheta | \hat{F}(\theta)-F(\theta)|\geq \frac{1}{2}\lim_{\delta'\rightarrow\delta+} \psi(\delta')$ holds.
	\end{enumerate}
\end{remark}

Based on the Remark \ref{psifunction_properties} (3rd statement), we have the following bound for $\psi_+(\delta):=\lim_{\delta'\rightarrow\delta+}\psi(\delta')$,
\begin{align*}
	\PROB\bigg\{ \exists \thetabootstrap\in\Theta \text{ such that } \|\thetabootstrap - &\thetatilde\| \geq \delta \bigg\} \leq \PROB\bigg\{ \suptheta\bigg|\FBhat(\theta) - F(\theta)\bigg| \geq \frac{1}{2}  \psi_+(\delta) \bigg\}.
\end{align*}
Now letting $\delta=\psi^{-1}\{2\cdot A(m,N,M,\epsilon,\epsilon') + 8\gamma^m\cdot C_{\rm bias}\}$ where 
\begin{gather}
	A(m,N,M,\epsilon,\epsilon'):= \frac{C_3}{p_{{\rm min}}^2}\cdot \Benvmulti  \times \bigg\{ m^2\cdot \bigg( \frac{1}{N^{1/4}} + \sqrt{\epsilon} \bigg) + m\cdot \bigg(\frac{1}{M^{1/4}} + \sqrt{\epsilon'}\bigg)  \bigg\}, \label{A_and_Bs}
\end{gather}
we have 
\begin{align}
	&\PROB\bigg\{ \exists \thetabootstrap\in\Theta \text{ such that }  \|\thetabootstrap - \thetatilde\| \geq \psi^{-1}\bigg(2\cdot A(m,N,M,\epsilon,\epsilon') + 8\gamma^m\cdot C_{\rm bias}\bigg) \bigg\} \nonumber \\
	&\leq \PROB\bigg\{\suptheta\bigg|\FBhat(\theta) - F(\theta)\bigg| \geq A(m,N,M,\epsilon,\epsilon') + 4\gamma^m\cdot C_{\rm bias} \bigg\} \quad \text{by Remark \ref{psifunction_properties} (2nd Point)} \nonumber \\
	&\leq \mathcal{D}(N)+C_{1}\cdot m\cdot (|\SAspace| + N) \cdot \exp(-C_{2}\cdot p_{{\rm min}}^2 \cdot N \cdot \epsilon^2 / C_{\rm den}(m)) \nonumber \\
	& \qquad \qquad + C_{1}\cdot (|\SAspace| + M)\cdot \exp(-C_{2}\cdot p_{{\rm min}}^2 \cdot M \cdot {\epsilon'}^2 / B_{\rm den}(m))  \label{thetaBhat_convergence_complicated} %
\end{align}

Now let us simplify the bound.
Let $\delta_1,\delta_2\in(0,1)$ be arbitrary. Now let us simplify the result \eqref{thetaBhat_convergence_complicated} by letting 
\begin{gather}
	\epsilon = \sqrt{\frac{C_{\rm den}(m)}{C_{2}\cdot p_{{\rm min}}^2}} \cdot \sqrt{\frac{1}{N}\cdot \log\bigg( \frac{C_{1}\cdot m\cdot (|\SAspace| + N)}{\delta_1} \bigg)  }, \nonumber \\
	\epsilon'  = \sqrt{\frac{B_{\rm den}(m)}{C_{2}\cdot p_{{\rm min}}^2}} \cdot \sqrt{\frac{1}{M}\cdot \log\bigg( \frac{C_{1}\cdot (|\SAspace| + M)}{\delta_2} \bigg)  }. \label{epsilon_and_epsilonprime} 
\end{gather}
Recall that we have to ensure $\epsilon,\epsilon'\in(0,1]$. Furthermore, since $A(m,N,M,\epsilon,\epsilon')$ in \eqref{A_and_Bs} contains $(N^{1/4}+\sqrt{\epsilon})$ and $(M^{1/4} + \sqrt{\epsilon'})$. Since $\epsilon$ and $\epsilon'$ \eqref{epsilon_and_epsilonprime} decays in a slower rate than $N^{1/4}$ and $M^{1/4}$, we can ignore them when the sample size is sufficiently large. To ensure these, we will assume $N\geq N_m(\delta_1), \ M\geq M_m(\delta_2)$ where $N_m(\delta_1)$ and $M_m(\delta_2)$ are the smallest integers such that $N\geq N_m(\delta_1), \ M\geq M_m(\delta_2)$ implies the following, with $C_{\rm den}(m)$ \eqref{C_den_multi} and $B_{\rm den}(m)$ \eqref{B_env_multi},
\begin{gather} \label{Nm_Mm_definition}
	{\frac{1}{N}} \leq {\frac{C_{\rm den}(m)}{C_{2}\cdot p_{{\rm min}}^2}} \cdot {\frac{1}{N}\cdot \log\bigg( \frac{C_{1}\cdot m\cdot (|\SAspace| + N)}{\delta_1} \bigg)  } \leq 1, \nonumber \\
	{\frac{1}{M}} \leq {\frac{B_{\rm den}(m)}{C_{2}\cdot p_{{\rm min}}^2}} \cdot {\frac{1}{M}\cdot \log\bigg( \frac{C_{1}\cdot (|\SAspace| + M)}{\delta_2} \bigg)  } \leq 1. \nonumber 
\end{gather}
Then let us bound the term $A(m,N,M,\epsilon,\epsilon')$ of Equation (\ref{A_and_Bs}) as follows, using the values of $\epsilon,\epsilon'$ specified in Equation (\ref{epsilon_and_epsilonprime}) and assuming \eqref{Nm_Mm_definition}. Skipping the calculation details, we can derive 
\begin{align}
	&A(m,N,M,\epsilon,\epsilon')\leq  A_1(m,N,M,\delta_1,\delta_2) \nonumber \\
	&:= C_{\rm model} \cdot \bigg[ m^2\cdot\bigg\{ \frac{1}{N}\cdot \log\bigg( \frac{C_{1}\cdot m\cdot (|\SAspace| + N)}{\delta_1} \bigg)  \bigg\}^{\frac{1}{4}}  +  m\cdot\bigg\{ \frac{1}{M}\cdot \log\bigg( \frac{C_{1}\cdot (|\SAspace| + M)}{\delta_2} \bigg)  \bigg\}^{\frac{1}{4}} \bigg] \label{Aone_definition} 
\end{align}
where $C_{{\rm model}}$ is defined in \eqref{New_Terms_Created2}.
Now we can rewrite Bound (\ref{thetaBhat_convergence_complicated}) as follows by Remark \ref{psifunction_properties} (1st Statement),
\begin{align*}
	&\PROB\bigg\{ \|\thetabootstrap - \thetatilde\| \leq \psi^{-1}\bigg( 2\cdot A_1(m,N,M,\delta_1,\delta_2) + 8\gamma^m\cdot C_{\rm bias} \bigg) \bigg\} \geq 1 - \mathcal{D}(N) - \delta_1 - \delta_2. %
\end{align*}
Next, we can analyze the convergence rate of our estimated distribution towards the best approximation $\Upsilon_{\tilde{\theta}}$ in Energy Distance, based on the same idea that we employed in \eqref{energy_difference_trick} and Assumption \ref{Lipschitz},
\begin{align}
	\energyone\bigg\{ \Upsilon_{\theta_1}(s,a), \Upsilon_{\theta_2}(s,a) \bigg\}  
	&\leq 2\cdot\expectationbounder(\theta_1,\theta_1) + 2\cdot\expectationbounder(\theta_1,\theta_2) \leq 2L\cdot \|\theta_1-\theta_2\|,  \label{energy_euclid_lipschitz}
\end{align}
which leads to $\energyonebar ( \Upsilon_{\thetabootstrap}, \Upsilon_{\thetatilde} ) \leq 2L\cdot \|\thetabootstrap-\thetatilde\|$.

\subsubsection{Finite Sample Error Bound} \label{nonrealizable_finite_sample_error_bound}

Under Assumptions \ref{RN_derivative}, \ref{bounded}, \ref{Lipschitz}, for any step level $m\in\mathbb{N}$ and arbitrary $\delta_1,\delta_2 \in(0,1)$, given that $N\geq \max\{N_m(\delta_1),2\}, \ M \geq \max\{M_m(\delta_2),2\}$ defined in \eqref{Nm_Mm_definition}, we have the following bound with probability larger than $1-\mathcal{D}(N) - \delta_1 - \delta_2$, all values of $\thetabootstrap\in\arg\min_{\theta\in\Theta} \FBhat(\theta)$ satisfy
\begin{align}
	&\|\thetabootstrap  - \thetatilde\|\leq \psi^{-1}\bigg( 2C_{{\rm model}}\cdot \bigg[ m^2\cdot \bigg\{ \frac{1}{N}\cdot \log\bigg( \frac{C_{1}\cdot m\cdot (|\SAspace| + N)}{\delta_1} \bigg)  \bigg\}^{\frac{1}{4}} \label{bootstrap_meaningful_bound}  \\
	& \qquad\qquad\qquad\qquad\qquad\qquad + m\cdot \bigg\{ \frac{1}{M}\cdot \log\bigg( \frac{C_{1}\cdot (|\SAspace| + M)}{\delta_2} \bigg)  \bigg\}^{\frac{1}{4}} \bigg] + 4\gamma^m \cdot C_{{\rm bias}}   \bigg)  , \nonumber
\end{align}
along with 
\begin{align}
	\energyonebar\big( \Upsilon_{\thetabootstrap}, \Upsilon_{\thetatilde} \big) \leq 2L\cdot \|\thetabootstrap - \thetatilde\|,   \label{energy_lipschitz_parameter}
\end{align}
where $\mathcal{D}(N)\rightarrow0$ as $N\rightarrow\infty$ (\ref{New_Terms_Created2}), and $\psi^{-1}$ (\ref{psibar_psibarinverse_definition}) is an increasing function that ensures $\psi^{-1}(y)\rightarrow 0$ as $y\rightarrow 0$, as stated in Remark \ref{psifunction_properties} (1st statement).

Here is the recap of definitions of the terms that we used, along with $V(s,a):= \exp( \|R(s,a)\|^2/t_0(s,a)^2 )$,
\begin{align}
	&\mathcal{D}(N):=|\SAspace|\cdot \frac{\sqrt{2}}{\sqrt{p_{{\rm min}}}}\cdot \inf_{r>2} \bigg\{N^{\frac{1}{r}-\frac{1}{2}} + \supsa \Expect\big\{ V(s,a) \cdot \mathbf{1}\big( V(s,a)>N^{1/r} \big) \big\}  \bigg\}  \rightarrow 0 \text{ as }N\rightarrow\infty  \nonumber \\
	&C_{{\rm bias}}:= \expectationbounder(\thetatilde,\pi) + L\cdot \diamTheta , \nonumber \\ 
	&C_{{\rm model}}:=\frac{C_{4}}{p_{{\rm min}}^{5/2}}  \bigg\{ L^2 p  \diamTheta^2 + \bigg(\frac{1}{1-\gamma}\bigg)^2  \max\big\{d\cdot \supsa \|R(s,a)\|_{\psi_2} , t_0^*\big\}^2  + \big(\supsa \Expect\|Z(s,a;\theta_0)\|\big)^2  + 1 \bigg\}^{\frac{3}{4}},  \nonumber \\ 
	&t_0^*:=\supsa t_0(s,a) \ \ \text{where } t_0(s,a)>0 \ \text{are values such that } \ \Expect\bigg\{ \exp\bigg( \frac{\|R(s,a)\|^2}{t_0(s,a)^2} \bigg) \bigg\}=1. \label{New_Terms_Created2} 
\end{align}
Just for a brief note, minimizer(s) of the estimated objective function $\thetabootstrap\in\arg\min_{\theta\in\Theta} \FBhat(\theta)$ always exists due to continuity of $\FBhat$ (proof in \ref{guarantee_boostrap_minimizer}) and closedness of $\Theta$ mentioned in Assumption \ref{Lipschitz}, which also implies $\diamTheta<\infty$.

\subsubsection{Asymptotic Setting} \label{psibarinverse_upperbound}

Based on the finite-sample error bound provided in \ref{nonrealizable_finite_sample_error_bound}, we will now assume $N,M,m$ are large enough to satisfy the assumptions $N\geq \max\{N_m(\delta_1),2\}, \ M \geq \max\{M_m(\delta_2),2\}$. We should also assume the following holds as $N,M,m\rightarrow\infty$ (terms defined in \eqref{Aone_definition} and \eqref{Fact1to2}),
\begin{align}
	2\cdot A_1(m,N,M,\delta_1,\delta_2) + 8\gamma^m\cdot C_{\rm bias} \rightarrow 0 \label{AB1B2_psic0_condition}
\end{align}
where the LHS is exactly the term inside $\psi^{-1}(\cdot)$ in \eqref{bootstrap_meaningful_bound}. This condition is necessary to ensure that the RHS of Bound \eqref{bootstrap_meaningful_bound} to have a finite value by Remark \ref{psifunction_properties} (1st statement). It should be verified that these conditions hold in the asymptotic sense, which we will discuss in the following section \ref{Proof_Nonrealizable_bootstrap_convergencerate} where we choose the actual growing speed of $M$ and $m$ with respect to $N$.

By Assumption \ref{uniqueness_generalized}, we could derive the following within the proof of Remark \ref{psifunction_properties} in \ref{Proof_psifunction_properties} (1st statement),
\begin{align*}
	\psi^{-1}(y) &\leq \frac{1}{{c_q}^{1/q}}\cdot y^{1/q} \ \ \ \text{for } \forall y \in [0,\sup_{\delta>0}\psi(\delta)) \quad \text{by \eqref{psiinverse_upperbound_technique}}.
\end{align*}
Based on this, by letting $\delta_1=\delta_2=\delta/2$ in \eqref{bootstrap_meaningful_bound}, we have following with probability larger than $1-\mathcal{D}(N)-\delta$,
\begin{align}
	\|\thetabootstrap - \thetatilde\| \lesssim   \bigg[\underbrace{m^2\cdot\bigg\{\frac{1}{N}\cdot \log \bigg( \frac{2mN}{\delta} \bigg) \bigg\}^{\frac{1}{4}}}_{\mathrm{data}} +  \underbrace{m\cdot\bigg\{\frac{1}{M}\log\bigg(\frac{2M}{\delta}\bigg)  \bigg\}^{\frac{1}{4}}}_{\mathrm{bootstrap}} + \underbrace{\gamma^m}_{\mathrm{bias}} \bigg]^{\frac{1}{q}}, \label{shrinking_NMm}
\end{align}
where $\lesssim$ means bounded by the given bound (RHS) multiplied by a positive number that does not depend on $N,M,m$. 
Each of the three terms \eqref{shrinking_NMm} corresponds to the inaccuracy associated with the observed data $\{(s_i,a_i,r_i,s_i')\}_{i=1}^{N}$, the resampled trajectories \eqref{Bootstrap_operator}, and the extent of non-realizability. Each can be reduced by increasing $N$, $M$, and $m$, however larger $m$ makes the first two terms more challenging for to shrink, so it resembles bias-variance trade-off.   

Accordingly to \eqref{shrinking_NMm}, the conditions of $N,M,m$ \eqref{AB1B2_psic0_condition} can be rewritten as following, 
\begin{gather}
	N\geq \max\{N_m(\delta/2),2\}, \quad M\geq \max\{M_m(\delta/2),2\}, \label{NMm_condition} 
\end{gather}
based on Definitions \eqref{Nm_Mm_definition}, along with the following based on Definition \eqref{Aone_definition},
\begin{align}
	&2C_{{\rm model}}\cdot \bigg[m^2\cdot \bigg\{ \frac{1}{N}\cdot \log\bigg( \frac{2C_{1}\cdot m\cdot (|\SAspace| + N)}{\delta} \bigg)  \bigg\}^{\frac{1}{4}} \nonumber \\
	& \qquad\qquad\qquad +m\cdot \bigg\{ \frac{1}{M}\cdot \log\bigg( \frac{2C_{1}\cdot (|\SAspace| + M)}{\delta} \bigg)  \bigg\}^{\frac{1}{4}} \bigg]  + 8\gamma^m \cdot C_{{\rm bias}} \rightarrow 0. \label{AB1B2_psic0_condition_new}
\end{align}

\subsubsection{Optimal Convergence Rate} \label{Proof_Nonrealizable_bootstrap_convergencerate}

With general $q\geq 1$ (instead of $q=2$), we can increase $M$ and $m$ with the sample size $N$ to obtain the following convergence rate towards the best-approximation represented by $\thetatilde$. The following theorem suggests the optimal way of choosing $M$ and $m$ so that it converges fastest as $N$ grows.

\begin{theorem} \label{thetaBhat_convergence_generalized}
	Under Assumptions \ref{RN_derivative}, \ref{bounded}, \ref{Lipschitz}, \ref{uniqueness_generalized}, letting $M=\lfloor C_1\cdot N\rfloor$ and $m = \lfloor \frac{1}{4} \log_{(1/\gamma)} ( C_2N / \log N ) \rfloor$ for any positive
	constants $C_1,C_2>0$, we have the following convergence rate of the upper bound
	\begin{align*}
		\energyonebar\big( \Upsilon_{\thetabootstrap}, \Upsilon_{\thetatilde} \big) \leq \bigO\bigg[ \frac{1}{N^{1/(4q)}}\cdot \bigg\{ \log_{\frac{1}{\gamma}} \bigg( \frac{N}{\log N} \bigg) \bigg\}^{2/q} \bigg],
	\end{align*}
	where $\bigO$ indicates the rate of convergence up to logarithmic order. Letting $q=2$ gives us the convergence rate of $\bigO(1/N^{1/8})$.
\end{theorem}

Now we will assume an asymptotic case where the sample size grows to infinity $N\rightarrow \infty$, and $M,m$ grow accordingly with a chosen rate. Towards that end, we have to achieve two goals. First, we have to ensure that \eqref{NMm_condition} and \eqref{AB1B2_psic0_condition_new} are satisfied as $N\rightarrow \infty$. Second, we have to make \eqref{shrinking_NMm} shrink in the fastest possible rate. As long as $M$ (which we can let to be arbitrarily large) grows in the same rate with (or faster than) $N$, 
\begin{align}
	M= \lfloor C_5\cdot N \rfloor \quad \text{for arbitrary} \quad C_5>0, \label{M_condition}
\end{align}
the second term of \eqref{shrinking_NMm} becomes ignorable in the asymptotic sense.

As noted below \eqref{shrinking_NMm}, increasing $m$ has trade-off effect, so we shall derive an appropriate speed of $m$, and then verify that it can satisfy \eqref{NMm_condition} with large enough $N$. Using \eqref{M_condition}, we can take up \eqref{shrinking_NMm} as follows for sufficiently large $N$,
\begin{align*}
	\|\thetabootstrap - \thetatilde\| \lesssim  \bigg[ m^2\cdot \bigg\{\frac{1}{N}\cdot \log \bigg( \frac{mN}{\delta} \bigg) \bigg\}^{\frac{1}{4}} + \gamma^m \bigg]^{\frac{1}{q}}.
\end{align*}
Now we choose the optimal level of $m$ that makes the two terms converge in the same rate, 
\begin{align*}
	\gamma^m \approx  m^2\cdot \frac{1}{N^{1/4}} \cdot \big(\log(mN)\big)^{1/4} .
\end{align*}
However, this relationship is very intricate, so we could not calculate $m$ that makes both sides perfectly match. So we alternatively solved an easier equation that gives us the following relationship,
\begin{align}
	\gamma^m\approx C_6 \cdot \bigg(\frac{\log N}{N}\bigg)^{1/4}  , \quad \therefore \  m\buildrel let \over = \bigg\lfloor \frac{1}{4} \cdot \log_{\frac{1}{\gamma}} \bigg(\frac{C_7 \cdot N}{\log N}\bigg) \bigg\rfloor .  \label{m_condition}
\end{align}
Note that the values of $C_6,C_7>0$ can be arbitrary, as long as $C_7=C_6^{-4}$ holds. Skipping the calculation details, it can be ascertained that the orders \eqref{M_condition} and \eqref{m_condition} ensures \eqref{NMm_condition} and \eqref{AB1B2_psic0_condition_new} to hold as $N\rightarrow \infty$. This can be verified based on the fact that $\sup_{m\in\mathbb{N}}C_{\rm den}(m)<\infty$ and $\sup_{m\in\mathbb{N}}B_{\rm den}(m)<\infty$, which are defined in \eqref{C_den_multi} and \eqref{B_env_multi}.

Furthermore, it allows us to achieve 
\begin{align*}
	\bigg\{\frac{m^8}{N}\cdot \log \bigg(& \frac{mN}{\delta} \bigg) \bigg\}^{\frac{1}{4}}  \lesssim  \frac{1}{N^{\frac{1}{4}}} \cdot \bigg\{ \log_{\frac{1}{\gamma}} \bigg( \frac{C_{8}  N}{\log N} \bigg)^{\frac{1}{4}} \bigg\}^2 \cdot \bigg( \log \bigg[ N \cdot \bigg\{ \log_{\frac{1}{\gamma}} \bigg(  \frac{C_{8} N}{\log N} \bigg)^{\frac{1}{4}} \bigg\} \bigg/ \delta \bigg] \bigg)^{\frac{1}{4}} ,
\end{align*}
which eventually leads to following when combined with \eqref{energy_lipschitz_parameter},
\begin{align*}
	\energyonebar\big( \Upsilon_{\thetabootstrap}, \Upsilon_{\thetatilde} \big) \leq \bigO\bigg[ \frac{1}{N^{1/(4q)}}\cdot \bigg\{ \log_{\frac{1}{\gamma}} \bigg( \frac{N}{\log N} \bigg) \bigg\}^{2/q} \cdot \bigg[ \log\bigg\{ N\cdot \log_{\frac{1}{\gamma}} \bigg( \frac{N}{\log N} \bigg) \bigg\} \bigg]^{1/(4q)} \bigg],
\end{align*}
which gives us desired result of Theorem \ref{thetaBhat_convergence_generalized}.

\subsection{Splitting-based Estimator} \label{sec:degeneracy}

Then a natural question arises for the bootstrap-based estimator. ``In case of using $m=1$ and assuming realizability, does Theorem \ref{thetaBhat_convergence_generalized} degenerate to the convergence rate $\bigO(\sqrt{\log N / N})$ shown in Theorem \ref{Realizable_Final_Theorem}?'' Superficially, it seems not. If we put $q=2$ (which is generally assumed in theoretical studies), we obtain a much slower rate of convergence $\bigO(N^{1/8}\cdot\log(N/\log N))$. However, this is mainly since we resorted to a different proof structure that can be applied when samples are re-used (bootstrap).

Assume that we split the $|\mathcal{D}|=N$ observations into $m$ equally-sliced subdatasets, say $|\mathcal{D}_1|=\cdots = |\mathcal{D}_m|=N/m$ (assuming $N/m$ to be an integer for convenience). Instead of bootstrapping, we will use each $\mathcal{D}_j$ ($j=1,\cdots, m$) separately to estimate Bellman operator at the $j$-th step, denoting them as $\hat{\mathcal{T}}_j^\pi$. Of course the population Bellman operator is the same for every step $j=1,\cdots,m$, however their estimate will be different in our case. With any step level $m$, the population objective function $F_m(\cdot)$ stays the same as \eqref{objective_functions}. On the contrary, the estimated objective function should be modified into
\begin{align} \label{split_m_estimator}
	\Fmhat(\theta) := \energyonebarhat( \Upsilon_\theta, \Thatseq \Upsilon_\theta) \quad \text{where } \quad \Thatseq = \That_1^\pi \cdots \That_m^\pi,
\end{align}
where each $\That_j^\pi:\mathcal{P}^{\SAspace} \rightarrow \mathcal{P}^{\SAspace}$ is based on $\hat{p}_j$ estimated from $|\mathcal{D}_j|$ ($j=1,\cdots , m$). 

\subsubsection{Convergence With An Arbitrary Step Level}

We will first fix our step level and then analyze the convergence rate of our estimator. In addition, we have to assume strong convexity of our population objective function $F_m(\cdot)$.

\begin{assumption} \label{mstep_strongconvex} We have
	$F_m(\theta) \geq F_m(\thetamstar) + c\cdot \| \theta - \thetamstar \|^2$ for the unique minimizer $\thetamstar$.
\end{assumption}

Then, we obtain the following result for realizability and non-realizability, respectively.

\begin{lemma} \label{Degeneracy}
	Under Assumptions \ref{RN_derivative}, \ref{bounded}, \ref{Lipschitz}, \ref{mstep_strongconvex}, for any level $m$, the estimator \eqref{split_m_estimator} satisfies the following result regardless of realizability,
	\begin{align*}
		\energyonebar(\Upsilon_{\hat{\theta}^{(m)}}, \Upsilonmstar) \leq \frac{|\SAspace|}{\sqrt{p_{\rm min}}}\cdot L (L\sqrt{p} + \sqrt{d}) \cdot O_p\bigg( \sqrt{\frac{m^3}{N}} \bigg).
	\end{align*}
\end{lemma}

We can see that the asymptotic convergence rate under realizability with single-step estimator ($m=1$) becomes $O_p(1/\sqrt{N})$, which can be seen as successful degeneration into Theorem \ref{Realizable_Final_Theorem}. In fact, it is slightly faster than $O_p(\sqrt{\log N/ N})$ (claimed by Theorem \ref{Realizable_Final_Theorem}) by logarithmic scale. This is not strange, since we have made additional assumptions. In addition, $\gamma$ also plays a role in the realizable case via $1/(1-\gamma)$, and it is included in the $O_p$ term.

\subsubsection{Proof Of Lemma \ref{Degeneracy}}

We will resort to the modified version of peeling argument (Theorem \ref{peeling_argument}). Its proof follows the same structure with Theorems 5.1 and 5.2 stated by \cite{sen2018gentle}. Then the term $W_N(\delta)$ is defined as follows,
\begin{align}
	&W_N(\delta):=\supdelta \bigg| (\hat{F}_m(\theta) - F_m(\theta)) - (\hat{F}_m(\thetamstar) - F_m(\thetamstar)) \bigg| \nonumber \\
	&\leq \supdelta \bigg| \energyonebarhat (\Upsilon_\theta , \Thatseq \Upsilon_\theta) - \energyonebar(\Upsilon_\theta , \Belloptmulti \Upsilon_\theta )  -   \energyonebarhat (\Upsilonmstar , \Thatseq \Upsilonmstar) + \energyonebar(\Upsilonmstar , \Belloptmulti \Upsilonmstar ) \bigg| \nonumber \\
	&\leq \supdelta \bigg| \energyonebarhat(\Upsilon_\theta , \Thatseq \Upsilon_\theta) - \energyonebarhat(\Upsilonmstar, \Thatseq \Upsilonmstar) - \energyonebar(\Upsilon_\theta , \Thatseq \Upsilon_\theta) +  \energyonebar(\Upsilonmstar, \Thatseq \Upsilonmstar) \bigg| \label{WNdelta_SD} \\
	& \quad  + \supdelta \bigg| \energyonebar(\Upsilon_\theta , \Thatseq \Upsilon_\theta) - \energyonebar(\Upsilonmstar, \Thatseq \Upsilonmstar) -\energyonebar(\Upsilon_\theta, \Belloptmulti\Upsilon_\theta) + \energyonebar(\Upsilonmstar, \Belloptmulti\Upsilonmstar)  \bigg|. \label{WNdelta_BD}
\end{align}
Our goal is to bound the expectation of two terms \eqref{WNdelta_SD} and \eqref{WNdelta_BD} as follows. But prior to that, let us split the cases into two:
\begin{enumerate}
	\item For $\forall j\in\{1,\cdots, m\}$, $\forall s,a\in\SAspace$, we have $|\frac{N_j(s,a)}{N/m} -\probsa | \leq \frac{1}{2}p_{\rm min}$, where $N_j(s,a)$ is the number of observations of $s,a$ within $|\mathcal{D}_j|$. 
	\item Otherwise. 
\end{enumerate}
We will denote the first event as $\Omegazero$ and its conditioned expectation as $\Expectationzero(\cdots):=\Expect(\cdots|\Omegazero)$. For a fixed $j\in\{1,\cdots , m\}$, letting $\probvechat^{(j)}$ be defined accordingly to Section \ref{Conditioning_Event} based on the $j$-th subdataset $\mathcal{D}_j$, we have the following based on Lemma \ref{Vector_Bernstein_multinomial},
\begin{align}
	\PROB\bigg\{ \Omega_0^c \bigg\} &\leq \sum_{j=1}^m \PROB\bigg\{ \| \probvechat^{(j)} - \probvec  \| \geq \frac{1}{2}p_{\rm min}  \bigg\} \leq m\cdot \exp\bigg( \frac{1}{4} \bigg) \cdot \exp\bigg( \frac{-\frac{N}{m}\cdot \frac{1}{4} p_{\rm min}^2}{32} \bigg)  \leq m \cdot O\bigg( \exp\big( \frac{-N}{m} \big) \bigg). \label{omegazero_complement}
\end{align}

Now let us bound the expectation of \eqref{WNdelta_BD}. First, let us condition upon $\Omegazero$.
\begin{gather*}
	\eqref{WNdelta_BD}\leq \sumsa \probsa \sum_{j=0}^{m-1} B_j(s,a), 
\end{gather*}
where $B_j(s,a)$ represents the following term
\begin{align*}
	&\supdelta \bigg| \energyone\bigg\{ \Upsilon_\theta (s,a) , \That_{1:j}^\pi (\Bellopt)^{m-j} \Upsilon_\theta(s,a) \bigg\} - \energyone\bigg\{ \Upsilonmstar (s,a) , \That_{1:j}^\pi (\Bellopt)^{m-j} \Upsilonmstar(s,a) \bigg\} \\
	& \ \ \ - \energyone\bigg\{ \Upsilon_\theta(s,a) , \That_{1:j+1}^\pi (\Bellopt)^{m-j-1} \Upsilon_\theta (s,a) \bigg\} + \energyone\bigg\{ \Upsilonmstar(s,a), \That_{1:j+1}^\pi (\Bellopt)^{m-j-1} \Upsilonmstar(s,a) \bigg\} \bigg| .
\end{align*}
Here, for $j=0$, we observe $\That_{1:0}$, and this can be understood as the identity operator. One important fact is that $\mathcal{D}_1,\cdots \mathcal{D}_{j-1}$, which $\That_1^\pi, \cdots , \That_{j-1}^\pi$ are based upon, are independent from $\mathcal{D}_j$. By applying the same logic that we used in \ref{Gamma_bounding_realizable} (i.e. employing the trick \ref{Wassersup_diamF_satisfication}), it comes down to bounding the following term with $\Expectationtilde_0$ being the conditional expectation upon $\mathcal{D}_1,\cdots \mathcal{D}_{j-1}$:
\begin{align*}
	\supdelta \bigg| \frac{1}{N_j(s,a)} \sum_{i=1}^{N_j(s,a)} f_i^\theta - \Expectationtilde_0(f_i^\theta) \bigg| + \supdelta \bigg| \frac{1}{N_j(s,a)^2} \sum_{i_1=1}^{N_j(s,a)} \sum_{i_2=1}^{N_j(s,a)}  f_{i_1 i_2}^\theta - \Expectationtilde_0(f_{12}^\theta) \bigg| .
\end{align*}
for some $f_i^\theta$ and $f_{i_1i_2}^\theta$ such that
\begin{gather*}
	|f_i^{\theta_1} - f_i^{\theta_2}|, |f_{i_1i_2}^{\theta_1} - f_{i_1i_2}^{\theta_2}| \leq 4(1+\gamma^m) \cdot \Wassersup(\theta_1,\theta_2) \ \text{for } i_1\neq i_2, \\
	\suptheta |f_{ii}^{\theta} - \Expectationtilde_0(f_{12}^{\theta}) | = O(1). 
\end{gather*}
Then, by Theorem 8.1.3 of \cite{vershynin2018high}, which is expectation-version of Dudley's integral inequality, we obtain
\begin{align*}
	\Expectationtilde_0 \big\{B_j(s,a)\big\} &\leq \frac{C_1}{\sqrt{\min_{s,a}N_j(s,a)}} \cdot \int_0^\infty \sqrt{ \log \mathcal{N}(N_{\|\cdot\|}(\thetamstar, \delta), \Wassersup, \epsilon) } \mathrm{d}\epsilon + \frac{d}{1-\gamma}\cdot \mathcal{O}\bigg( \frac{1}{\min_{s,a}N_j(s,a)} \bigg) \\
	&\leq \frac{L\sqrt{p}}{\sqrt{p_{\rm min}}} \cdot \delta \cdot O\bigg( \frac{1}{\sqrt{N/m}} \bigg) + \frac{1}{p_{\rm min}} \cdot \frac{d}{1-\gamma}\cdot O\bigg( \frac{1}{N/m} \bigg)
\end{align*}
where the second term in the first line is due to the diagonal terms $f_{ii} - \Expectationtilde(f_{12}^\theta)$, and the second line is based on Remark \ref{metricentropy_Lipschitz} (based on Assumption \ref{Lipschitz}). This further leads to 
\begin{align*}
	\Expectationzero\big\{ \eqref{WNdelta_BD} \big\} \leq \frac{|\SAspace|}{\sqrt{p_{\rm min}}}\cdot L\sqrt{p}\cdot \delta \cdot O\bigg( \sqrt{\frac{m^3}{N}} \bigg) + \frac{1}{p_{\rm min}} \cdot \frac{d}{1-\gamma} \cdot O\bigg( \frac{m^2}{N} \bigg).
\end{align*}
Due to Assumption \ref{bounded}, it can be bounded under $\Omegazero^c$, that is $\eqref{WNdelta_BD} \leq O(1)$. Then, combined with \eqref{omegazero_complement}, we obtain 
\begin{align*}
	\Expect\big\{ \eqref{WNdelta_BD} \big\} &\leq \frac{1}{\sqrt{p_{\rm min}}}\cdot L\sqrt{p}\cdot \delta \cdot O\bigg( \sqrt{\frac{m^3}{N}} \bigg) + \frac{1}{p_{\rm min}} \cdot \frac{d}{1-\gamma} \cdot O\bigg( \frac{m^2}{N} \bigg) + O\bigg( \exp\big( \frac{-N}{m} \big) \bigg)\\
	&=\frac{1}{\sqrt{p_{\rm min}}}\cdot L\sqrt{p}\cdot \delta \cdot O\bigg( \sqrt{\frac{m^3}{N}} \bigg) + \frac{1}{p_{\rm min}} \cdot \frac{d}{1-\gamma}\cdot O\bigg( \frac{m^2}{N} \bigg).
\end{align*}

Now let us bound the expectation of \eqref{WNdelta_SD}. 
\begin{align*}
	& \eqref{WNdelta_SD} \leq \supdelta \bigg| \sumsa (\probsahat - \probsa) \cdot \energyone\bigg\{ \Upsilon_\theta(s,a) , \Thatseq \Upsilon_\theta(s,a) \bigg\} -  \energyone\bigg\{ \Upsilonzero(s,a) , \Thatseq \Upsilonzero(s,a) \bigg\} \bigg| \\
	&\leq \probabsnorm \cdot \supdelta \supsa \bigg|  \energyone\bigg\{ \Upsilon_\theta(s,a) , \Thatseq \Upsilon_\theta(s,a) \bigg\} -  \energyone\bigg\{ \Upsilonzero(s,a) , \Thatseq \Upsilonzero(s,a) \bigg\}  \bigg| \\
	&\leq \probabsnorm \cdot \supdelta 4\cdot(1+\gamma^m)\cdot \Wassersup(\theta, \thetamstar) \quad \text{by technique \eqref{energy_difference_trick}} \\
	&\leq \probabsnorm \cdot 8L\cdot \delta. 
\end{align*}
Using the terms $\probvec$ and $\probvechat$ introduced in Section \ref{Conditioning_Event}, we can apply the technique that we used in Section \ref{Delta_bounding_realizable} as follows,
\begin{gather}
	\Expect\bigg\{  \sumsa \bigg| \probsa - \probsahat \bigg| \bigg\} =\Expect\big(\| \probvechat - \probvec \|_1\big) \leq \sqrt{|\SAspace|} \cdot \Expect \big( \|\probvechat - \probvec\| \big) \nonumber \\ %
	\leq \sqrt{|\SAspace|} \cdot \Expect \big( \|\probvechat - \probvec\|^2 \big)^{1/2} \leq \sqrt{|\SAspace|} \cdot \bigg\{ \int_0^1 \PROB \big( \|\probvechat - \probvec\|^2\geq t \big) \mathrm{d}t \bigg\}^{\frac{1}{2}} \leq O\bigg(\sqrt{\frac{|\SAspace|}{N}}\bigg). \nonumber
\end{gather}
Then, we finally have
\begin{align*}
	&\Expect\big\{ \eqref{WNdelta_SD}\big\} \leq 8L\cdot \sqrt{|\SAspace|} \cdot \delta\cdot  O\bigg( \frac{1}{\sqrt{N}} \bigg).
\end{align*}
Putting together \eqref{WNdelta_SD} and \eqref{WNdelta_BD}, and using $p_{\rm min}\leq 1/|\SAspace|$, we obtain
\begin{align*}
	\Expect\big\{W_N(\delta)\big\} \leq \frac{1}{\sqrt{p_{\rm min}}}\cdot L\sqrt{p}\cdot \delta \cdot O\bigg( \sqrt{\frac{m^3}{N}} \bigg) + \frac{1}{p_{\rm min}} \cdot \frac{d}{1-\gamma}\cdot O\bigg( \frac{m^2}{N} \bigg).
\end{align*}

Now we can apply the peeling argument (Theorem \ref{peeling_argument}). We will skip the proof for $\hat{\theta}^{(m)} \rightarrow^P \thetamstar$, since this is analogous to the proof structure in \eqref{bootstrap_meaningful_bound}. Since we have $\alpha=1$ and $q=2$ by Assumption \ref{mstep_strongconvex}, we have 
\begin{align*}
	\|\hat{\theta}^{(m)} - \thetamstar\| \leq \frac{1}{\sqrt{p_{\rm min}}}\cdot \bigg( L\sqrt{p} + \sqrt{\frac{d}{1-\gamma}} \bigg) \cdot O_p\bigg( \sqrt{\frac{m^3}{N}} \bigg).
\end{align*}
Applying the logic of \eqref{energy_euclid_lipschitz}, we obtain the desired result. The quantities $\supsa\|R(s,a)\|_{\psi_2}$ and $\supsa\Expect\|Z(s,a;\thetamstar)\|$ is hidden in the term $O_p(1)$ of Lemma \ref{Degeneracy}.

\subsubsection{Proof Of Theorem \ref{Degeneracy_increasingstep}} \label{splitestimator_finalstage}

Its proof is simple. By \eqref{relaxed_triangle_pis2}, we have $\energyonebar(\Upsilon_{\hat{\theta}^{(m)}}, \Upsilontilde) \leq 2\cdot \{  \energyonebar(\Upsilon_{\hat{\theta}^{(m)}}, \Upsilonmstar) + \energyonebar(\Upsilonmstar, \Upsilontilde )  \}$. By \eqref{energy_euclid_lipschitz}, we have $\energyonebar(\Upsilonmstar, \Upsilon_{\thetatilde}) \leq 2L\cdot \|\thetamstar - \thetatilde\|$. We can employ the same technique \eqref{shrinking_NMm} to obtain 
\begin{align*}
	\|\thetamstar - \thetatilde\| \leq \bigg( \suptheta \bigg| F_m(\theta) - F(\theta) \bigg| \bigg)^{\frac{1}{2}}  \leq O(\gamma^{\frac{m}{2}}) \quad \text{by \eqref{Fact1to2}}.
\end{align*}
Now letting $m=\lfloor\log_{\frac{1}{\gamma}}(C\cdot N)\rfloor$ for an arbitrary $C>0$, we obtain the desired rate.

\clearpage

\section{SUPPORTING RESULTS}

\subsection{Supporting Theoretical Results}

\subsubsection{Modified Version Of Peeling Argument} \label{Borrowed_results}

We can slightly modify the standard proof structure of peeling argument (e.g. see the proof of Theorem 5.1 of \cite{sen2018gentle}) to obtain the following.

\begin{theorem} \label{peeling_argument}
	(Peeling argument modified from Theorem 5.1 of \cite{sen2018gentle}) For a metric space $(\Theta, d)$, assuming that $\theta_0\in\Theta$ is the unique minimizer of $M(\theta)$ that is constructed based on a loss function $l_\theta(x)$, we define
	\begin{gather*}
		M(\theta) = \mathbb{E}l_\theta(X), \quad \theta_0 = \arg\min_{\theta\in\Theta} M(\theta), \quad M_n(\theta) = \frac{1}{n}l_\theta (X_i), \quad \hat{\theta}_n = \arg\min_{\theta\in\Theta} M(\theta_0), \\
		\Delta_n(\theta) = (M_n(\theta) - M(\theta)) - (M_n(\theta_0) - M(\theta_0)), \quad W_n(\delta) = \sup_{d(\theta , \theta_0)\leq \delta} |\Delta_n(\theta)|.
	\end{gather*}
	Assume that there exists some $q\geq 1$ that satisfies
	\begin{align*}
		M(\theta) \geq M(\theta_0) + \lambda d^q(\theta , \theta_0), \quad \mathbb{E}\big\{W_n(\delta)\big\}\leq \frac{c_1\cdot\delta^\alpha}{\sqrt{n}} + c_2\cdot d_n \quad \text{for some } \alpha\in(0,q) .
	\end{align*}
	Given that $\hat{\theta}_n\rightarrow^P \theta_0$, we have $d(\hat{\theta}_n, \theta_0) = c_1^{\frac{1}{q-\alpha}} \cdot O_p\{ (1 / n)^{\frac{1}{2(q-\alpha)}} \} + c_2^{\frac{1}{q}} \cdot O_p( d_n^{\frac{1}{q}} )$.
\end{theorem}

\subsubsection{Proof Of Remark \ref{psi2_properties}} \label{Proof_psi2_properties}

Properties 1 and 2 are mentioned in Example 2.5.8 suggested by \citet{vershynin2018high}. Property 3 follows directly by using ${\Expect(X)^2}/{t^2} \leq \Expect ( {X^2}/{t^2} )$.
Property 4 can be verified as follows. Let $\mathbf{x}=(x_1,\cdots,x_d)^\intercal\in\mathbb{R}^d$ with unit norm $\|\mathbf{x}\|=1$ be arbitrary. Denoting the canonical vectors as $\mathbf{e}_1,\cdots, \mathbf{e}_d$, we have the following by \eqref{subgaussian_norm_definition} and $|x_j|\leq 1$,
\begin{align*}
	\big| \langle \mathbf{X}, \mathbf{x} \rangle \big| &= \big| \langle \mathbf{X}, x_1 \mathbf{e}_1 \rangle  + \cdots + \langle \mathbf{X}, x_d \mathbf{e}_d \rangle  \big|  \leq |x_1|\cdot \|\mathbf{X}\|_{\psi_2} + \cdots +  |x_d|\cdot \|\mathbf{X}\|_{\psi_2}\leq d \|\mathbf{X}\|_{\psi_2}
\end{align*} 
Property 5 is verified in Exercise 2.7.10 suggested by \citet{vershynin2018high}. Property 6 is straightforward from Proposition 2.6.1 of \citet{vershynin2018high}.

\subsubsection{Proof Of Facts (\ref{Fact1to3})} \label{Proof_Fact1to3}

Letting $\probsahatone$ and $\probsahattwo$ be the components of $\probvechatone$ and $\probvechattwo$ (Appendix \ref{Conditioning_Event}) corresponding to $(s,a)$, we have 
\begin{align*}
	\probsahatone = \frac{N_1(s,a)}{\lfloor N/2 \rfloor} \quad &\& \quad \probsahattwo = \frac{N_2(s,a)}{N - \lfloor N/2 \rfloor }, \\
	\text{with } N_1(s,a)=\sum_{i=1}^{\lfloor N/2\rfloor}\mathbf{1}\big\{ (S_i,A_i)=(s,a) \big\} \quad &\& \quad N_2(s,a)=\sum_{i=\lfloor N/2\rfloor+1}^{N}\mathbf{1}\big\{ (S_i,A_i)=(s,a) \big\}.
\end{align*}
Let us first prove Fact 1. Letting $s,a\in\SAspace$ be arbitrary, we have the following,
\begin{align*} 
	\bigg| \probsahatone - \probsa \bigg| &\leq \supsa \bigg| \probsahatone - \probsa \bigg|  = \|\probvechatone - \probvec\|_{\infty}\leq \| \probvechatone - \probvechat \| \\
	& \leq \frac{1}{2}p_{{\rm min}}\cdot \epsilon \leq \frac{1}{2}\probsa \ \ (\because \epsilon<1). \label{proberror_epsilon}
\end{align*}
With the same logic, we also have $| \probsahattwo - \probsa | \leq \frac{1}{2}\probsa$, and thereby $\probsahatone, \probsahattwo \in [ \frac{1}{2} \probsa, \frac{3}{2}\probsa ]$
which validates Fact 1.
Fact 2 can be validated, since we have $N_1(s,a),N_2(s,a)\geq 1$ and $p_{\rm min}>0$.
Showing Fact 3 is straightforward from \eqref{Conditional_Omega}.

\subsubsection{Proof Of Lemma \ref{Hoeffding_twotuple}} \label{Proof_Hoeffding_twotuple}

Let us temporarily assume that $N\in\mathbb{N}$ is an even number. Newly define $Y_{ij}:=(X_{ij} + X_{ji})/2$ for $1\leq i < j \leq N$. Then we can apply the trick that we used in \eqref{Wtilde_psi2_odd} of defining $N-1$ groups $G_1,\cdots,G_{N-1}$ and applying Lemma S4 of \citet{wang2022low}, which leads to
\begin{align} %
	\PROB\bigg\{ \bigg| \frac{1}{N(N-1)} \sumij X_{ij} - \Expect(X_{12}) \bigg| \geq \epsilon \bigg\} %
	&\leq (N-1)\cdot \PROB \bigg\{ \bigg| \frac{1}{N/2} \sum_{(i,j)\in G_1} Y_{ij} - \Expect(Y_{12}) \bigg| \geq \epsilon \bigg\}  \nonumber \\
	&\leq 2N\cdot \exp \bigg\{ \frac{-C_2\cdot N \cdot \epsilon^2}{\|X_{12}-\Expect(X_{12})\|_{\psi_2}^2} \bigg\}. \nonumber %
\end{align}
Then we can apply the idea used in Appendix \ref{Gamma_bounding_realizable} to expand ourselves into odd numbers $N\in\mathbb{N}$. Eventually we obtain the desired result for an arbitrary integer $N\in\mathbb{N}$.

\subsubsection{Proof For Lemma \ref{Vector_Bernstein_multinomial}} \label{Proof_Vector_Bernstein_multinomial}

It is trivial for $\epsilon>1$ since the probability term in the LHS shall be 0, so we will assume $\epsilon\in(0,1)$. Let $\mathbf{Y}_i:=\mathbf{X}_i - \probvec$, and we can see $\Expect(\mathbf{Y}_1)=\mathbf{0}\in\mathbb{R}^H$, along with 
\begin{gather*}
	\|\mathbf{Y}_1\| = \|\mathbf{X_1} - \probvec \| \leq \|\mathbf{X}_1\| + \|\probvec\| \leq 2 \ \ \big( \because \ \|\probvec\| = \sqrt{\sum_{h=1}^{H} p_h^2}\leq \sqrt{\sum_{h=1}^{H} p_h}=1 \big).
\end{gather*}
Therefore we can let $\mu=2$ and $\sigma^2=4$, and applying Lemma 18 of \citet{kohler2017sub} gives us the desired result for $\epsilon\in(0,2)$, so it validates the result for $\epsilon\in(0,1)$.

\subsubsection{Proof Of Remark \ref{metricentropy_Lipschitz}} \label{Proof_metricentropy_Lipschitz}

Let $t>0$ be arbitrarily chosen. Under Assumption \ref{Lipschitz}, $\|\theta_1-\theta_2\| \leq t/L$ implies $\expectationbounder(\theta_1,\theta_2)\leq t$. Letting $M_0=\mathcal{N}(\Theta,\|\cdot\|, t/L)$ defined under \ref{Proof_Realizable_Final_Theorem}, and $\theta_1,\cdots, \theta_{M_0}$ to be such centers, we have 
\begin{align*}
	\Theta \subset \bigcup_{i=1}^{M_0} N_{\|\cdot\|}(\theta_i,t/L)\subset \bigcup_{i=1}^{M_0} N_{\expectationbounder}(\theta_i,t), \ \ \ \  \therefore \ \mathcal{N}(\Theta, \expectationbounder, t) \leq M_0=\mathcal{N}(\Theta,\|\cdot\|, t/L),
\end{align*}
which leads to 
\begin{align} 
	\metricentropyexpectatonbounderTheta &
	\leq \int_0^\infty \sqrt{\log \mathcal{N}(\Theta, \|\cdot\|, t/L)}\mathrm{d}t  %
	\leq L\cdot \int_0^{\diamTheta} \sqrt{\log \mathcal{N}(\Theta, \|\cdot\|, t)}\mathrm{d}t, \nonumber %
\end{align}
where the last line holds since we have $\mathcal{N}(\Theta,\|\cdot\|,t)=1$ for $\forall t\geq\diamTheta$. Here we can use the well-known Volume Comparison Lemma 
that enables us to take up as follows,
\begin{align} 
	&\metricentropyexpectatonbounderTheta  \leq  L\cdot \int_0^{\diamTheta} \sqrt{p\cdot \log \bigg\{ 1+\frac{2\cdot \diamTheta}{t} \bigg\}}\mathrm{d}t \nonumber \\
	& \leq  L\cdot \int_0^{\diamTheta} \sqrt{p\cdot \log \bigg\{ \frac{3\cdot \diamTheta}{t} \bigg\}}\mathrm{d}t 
	\leq 6\sqrt{2\pi} \cdot L\sqrt{p}\cdot {\rm diam}(\Theta;\|\cdot\|). \nonumber %
\end{align}

\subsubsection{Proof Of Lemma \ref{energy_difference}} \label{Proof_energy_difference}

Letting $\mu_0, \mu_1, \mu_2\in\mathcal{P}$ be arbitrary, we have $\energyonebar(\mu_0,\mu_1) \leq (\text{MMD}(\mu_0,\mu_2) + \text{MMD}(\mu_2,\mu_1))^2$ that leads to following,
\begin{align*}
	\energyone(\mu_0,\mu_1) 
	&=\energyone(\mu_0,\mu_2) + \sqrt{\energyone(\mu_1,\mu_2)}\cdot \big\{2\sqrt{\energyone(\mu_0, \mu_2)} + \sqrt{\energyone(\mu_1,\mu_2)} \big\}.
\end{align*}
Now let us extend this result towards $\energyonebar$ as follows,
\begin{align*}
	&\energyonebar(\Upsilon_0,\Upsilon_1) \leq 
	\energyonebar(\Upsilon_0,\Upsilon_2) + \sumsa \bigg[ \sqrt{\probsa\cdot \energyone\{\Upsilon_1(s,a), \Upsilon_2(s,a)\}}  \\ 
	&\qquad \qquad \qquad \qquad \qquad \times \sqrt{\probsa}\cdot \bigg\{ 2\sqrt{\energyone\{\Upsilon_0(s,a), \Upsilon_2(s,a)\}} + \sqrt{\energyone\{\Upsilon_1(s,a), \Upsilon_2(s,a)\}} \bigg\} \bigg] \\
	&\leq  \energyonebar(\Upsilon_0,\Upsilon_2) + \bigg( \sumsa \probsa\cdot \energyone\{\Upsilon_1(s,a), \Upsilon_2(s,a)\} \bigg)^{1/2}   \\
	&\qquad \qquad \qquad \times \bigg[ \sumsa \probsa\cdot \bigg\{ 2\sqrt{\energyone\{\Upsilon_0(s,a), \Upsilon_2(s,a)\}} + \sqrt{\energyone\{\Upsilon_1(s,a), \Upsilon_2(s,a)\}} \bigg\}^{2} \bigg]^{1/2}\\
	&\leq \energyonebar(\Upsilon_0, \Upsilon_2) +  4\cdot 
	\energyonebar(\Upsilon_1,\Upsilon_2)^{1/2}\cdot \big\{\energyonebar(\Upsilon_0, \Upsilon_2) + \energyonebar(\Upsilon_1,\Upsilon_2) \big\}^{1/2} ,
\end{align*}
where the second inequality is based on Cauchy-Schwartz inequality, the last line is based on $(x+y)^2\leq 2(x^2+y^2)$. 
Similarly, we have $\energyonebar(\Upsilon_0,\Upsilon_2) \leq \energyonebar(\Upsilon_0, \Upsilon_1) + 4\cdot \energyonebar(\Upsilon_1,\Upsilon_2)^{1/2}\cdot \{\energyonebar(\Upsilon_0, \Upsilon_1) + \energyonebar(\Upsilon_1,\Upsilon_2) \}^{1/2}.$
This eventually leads to 
\begin{align*}
	\bigg|\energyonebar(\Upsilon_0,\Upsilon_1) -  \energyonebar(\Upsilon_0, \Upsilon_2)\bigg| 
	&\leq  4\cdot \energyonebar(\Upsilon_1,\Upsilon_2)^{1/2} \cdot \bigg[ \max\bigg\{\energyonebar(\Upsilon_0, \Upsilon_1), \energyonebar(\Upsilon_0, \Upsilon_2)\bigg\} + \energyonebar(\Upsilon_1,\Upsilon_2) \bigg]^{1/2}.
\end{align*}

\subsubsection{Proof Of Lemma \ref{expectation_shrinkage_finite_firstmoment}} \label{Proof_expectation_shrinkage_finite_firstmoment}

Let $z>0$ be arbitrary, and let $E_i=X_i\cdot \mathbf{1}\big\{ |X_i|\leq z \big\}$ and $F_i=X_i\cdot \mathbf{1}\big\{ |X_i|>z \big\}$. Since we have $X_i = E_i + F_i$, we have $\Expect| \sumn X_i | \leq \Expect| \sumn E_i | + \Expect| \sumn F_i |$.
Note that each term satisfies the following due to $\Expect(E_1)=0$,
\begin{gather*}
	\Expect\bigg| \sumn E_i \bigg| \leq \Expect\bigg\{ \big( \sumn E_i \big)^2 \bigg\}^{1/2} = \sqrt{\mathbb{V} \bigg( \sumn E_i \bigg)} \leq \sqrt{n}\cdot \mathbb{V}(E_1)^{1/2} = \sqrt{n}\cdot \big[\Expect\big\{ X_1^2\cdot \mathbf{1}\big( |X_1|\leq z \big) \big\}\big]^{1/2} \\
	\Expect\bigg| \sumn V_i \bigg| \leq n\cdot \Expect|V_1| = n \cdot \Expect\big\{ |X_1|\cdot \mathbf{1}\big(|X_1|>z\big) \big\}, \\ 
	\therefore \ \Expect\big| \bar{X}_n \big| \leq \inf_{z>0}\bigg[ \frac{1}{\sqrt{n}}\cdot \bigg\{ \Expect\big\{ X_1^2\cdot \mathbf{1} (|X_1|\leq z ) \big\} \bigg\}^{1/2} + \Expect\big\{ |X_1|\cdot \mathbf{1}(|X_1|>z) \big\} \bigg] \rightarrow 0 \ \ \text{as }n\rightarrow\infty .
\end{gather*}

\subsubsection{Proof Of Equation (\ref{Fact1to2})} \label{Proof_Fact1to2}

With $F_m(\cdot)$ and $F(\cdot)$ defined in \eqref{objective_functions}, we we have 
\begin{align*}
	\big| F_m(\theta) - F(\theta) \big| %
	&\leq \sumsa \probsa \cdot \bigg| \energyone\bigg\{ \Upsilon_\theta(s,a), \Belloptmulti \Upsilon_\theta (s,a) \bigg\} - \energyone\bigg\{ \Upsilon_\theta(s,a), \Upsilon_\pi (s,a) \bigg\} \bigg|.
\end{align*}
We can further bound it by using the technique shown in Line \eqref{energy_difference_trick}. With abuse of notation $\Belloptmulti\theta$ introduced in Definition (\ref{abuse_Bellopt_parameter}), we can derive the following by the trick used in \eqref{energy_difference_trick} and Assumption \ref{Lipschitz},
\begin{align}
	\bigg| \energyone\bigg\{ \Upsilon_\theta(s,a), \Belloptmulti \Upsilon_\theta(s,a) \bigg\} - \energyone\bigg\{ \Upsilon_\theta(s,a), \Upsilon_\pi(s,a) \bigg\} \bigg| %
	&\leq 4\gamma^m\cdot \big\{ \expectationbounder(\thetatilde, \pi) + \sup_\theta\expectationbounder(\theta, \thetatilde) \big\} \nonumber \\
	&\leq 4\gamma^m\cdot \big\{ \expectationbounder(\thetatilde, \pi) + L\cdot \diamTheta \big\}. \nonumber %
\end{align}

\subsubsection{Proof Of Remark \ref{psifunction_properties}} \label{Proof_psifunction_properties}

Let us show the first fact. $\psi^{-1}(\cdot)$ is an increasing function, since the following holds for arbitrary $y_1, y_2$ ($y_1\leq y_2$),
\begin{align*}
	\psi^{-1}(y_1)=\inf_{\delta>0}\big\{ \psi(\delta) \geq y_1 \big\}\leq \inf_{\delta>0}\big\{ \psi(\delta) \geq y_2 \big\} = \psi^{-1}(y_2). 
\end{align*}
If $y>\sup_{\delta>0}\psi(\delta)$, then $\psi^{-1}(y)=\inf(\varnothing):=\infty$ by definition of infimum. To prove $\lim_{y\rightarrow 0}\psi^{-1}(y)=0$, it suffices to show right-side convergence. Towards that end, we let $\delta>0$ be sufficiently small, that is $\delta<\sup_{\theta\in\Theta}\|\theta - \thetatilde\|$, which leads to following by Definition \eqref{psibar_psibarinverse_definition} and Assumption \ref{uniqueness_generalized} (with general $q\geq 1$),
\begin{gather}
	\psi(\delta) = \inf_{ \theta\in\Theta: \|\theta - \thetatilde\|\geq \delta} F(\theta) - F(\thetatilde) \geq c_q \cdot \delta^q, \quad 
	\therefore \ \psi^{-1}(y) \leq \frac{1}{{c_q}^{1/q}}\cdot y^{1/q} \ \ \ \text{for } \forall y \in [0,\sup_{\delta>0}\psi(\delta)) . \label{psiinverse_upperbound_technique}
\end{gather}
This gives us $\psi^{-1}(y)\rightarrow0$ as $y\rightarrow0$.

The second fact can be shown as follows. Let $y\in[0,\psi(\suptheta\|\theta - \thetatilde\|)]$ be arbitrary ($\psi^{-1}(y)<\infty$), and let $\epsilon>0$ be arbitrarily small. Letting $\delta_0:=\inf_{\delta>0} \big\{\psi(\delta) \geq y\big\}$, we have following,
\begin{gather*}
	\psi\big\{ \psi^{-1}(y) + \epsilon \big\} = \psi\bigg[ \inf_{\delta>0} \bigg\{ \psi(\delta) \geq y \bigg\} + \epsilon \bigg] = \psi(\delta_0+\epsilon)\geq y, \quad 
	\therefore \ \lim_{\epsilon\rightarrow 0+} \psi \big\{ \psi^{-1}(y) + \epsilon \big\}\geq y.
\end{gather*}
The third fact can be validated by extending the proof of Example 1.3 suggested by \citet{sen2018gentle}. Suppose that there exists a minimizer $\hat{\theta}\in\arg\min_{\theta\in\Theta}\hat{F}(\theta)$ such that $\|\hat{\theta} - \theta_0\|\geq \delta$. Now we temporarily make new notations $G=-F$ and $\hat{G}=-\hat{F}$, which leads to $\hat{\theta}\in\arg\max_{\theta\in\Theta}\hat{G}(\theta)$ and $\psi(\delta)= G(\thetatilde) - \sup_{\theta\in\Theta:\|\theta - \thetatilde\|\geq \delta} G(\theta)$. Then we have $\hat{G}(\thetatilde)\leq \sup_{\theta\in\Theta:\|\theta - \thetatilde\|\geq \delta}\hat{G}(\theta)$, which leads to 
\begin{gather*}
	\sup_{\theta\in\Theta:\|\theta - \thetatilde\|\geq \delta} \hat{G}(\theta) - \hat{G}(\thetatilde) + \psi(\delta) \geq \psi(\delta), \quad
	\therefore \ \sup_{\theta\in\Theta:\|\theta - \thetatilde\|\geq \delta} \big\{ (\hat{G}(\theta) - G(\theta)) - (\hat{G}(\theta) - G(\thetatilde)) \big\} \geq \psi(\delta),
\end{gather*}
from which we can derive $\suptheta|\hat{F}(\theta) - F(\theta)|\geq \frac{1}{2}\psi(\delta)$. Up to this point we have derived 
\begin{align} %
	\exists \hat{\theta}\in\arg\min_{\theta\in\Theta}\hat{F}(\theta) \ \text{such that } \|\hat{\theta} - \theta_0\|\geq \delta  \quad \rightarrow \quad \suptheta|\hat{F}(\theta) - F(\theta)|\geq \frac{1}{2}\psi(\delta). \nonumber
\end{align}
Now let us assume that there exists a value $\hat{\theta}\in\Theta$ such that $\|\hat{\theta} - \thetatilde\| > \delta$. Then we have $\|\hat{\theta}-\thetatilde\|=\delta + \epsilon_0$ for some $\epsilon_0>0$. 
By above statement,
we have $\suptheta|\hat{F}(\theta) - F(\theta)|\geq \frac{1}{2}\psi(\delta+\epsilon_0)\geq \frac{1}{2} \lim_{\epsilon\rightarrow 0+} \psi(\delta+\epsilon)$, by using the fact that $\psi(\cdot)$ is an increasing function. This gives us the desired result.

\subsubsection{Continuity Of Bootstrap-based Objective Function}\label{guarantee_boostrap_minimizer}

Note that $\Bootstrapopt$ can also be viewed as a Bellman operator corresponding to a different transition probability $\probhatbootstrap$ \eqref{Bootstrap_operator}. That being said, by replicating the trick used in \eqref{energy_difference_trick} and Assumption \ref{Lipschitz}, we have 
\begin{align*}
	\bigg|\energyone\bigg\{& \Upsilon_{\theta_1} (s,a) , \Bootstrapopt \Upsilon_{\theta_1}(s,a) \bigg\} - \energyone\bigg\{ \Upsilon_{\theta_2} (s,a) , \Bootstrapopt \Upsilon_{\theta_2}(s,a) \bigg\} \bigg| 
	\leq 8L\cdot \|\theta_1-\theta_2\|,
\end{align*}
which further leads to $|\FBhat(\theta_1) - \FBhat(\theta_2)|\leq 8L\cdot \|\theta_1-\theta_2\|$, implying Lipschitz continuity of the objective function \eqref{Bootstrap_objective_function}.

\subsection{Further Discussion}

\subsubsection{Theorem \ref{Fundamental_Realizable} Is Not Trivial For General State-action Space}  \label{nontriviality_continuous_stateaction}

As we warned below Theorem \ref{Fundamental_Realizable}, Theorem \ref{Fundamental_Realizable} can be trivial in tabular setting $|\SAspace|<\infty$, but this is not trivial for non-tabular setting $|\SAspace|\geq\infty$. This is analogous to the fact that $L_2$-norm and $L_\infty$-norm are not equivalent for infinite-dimensional objects (e.g. functional objects).
For a function $g:\SAspace\rightarrow \mathbb{R}$, let us define $\|g\|_2:=\sqrt{E(g(S,A)^2)}$ and $\|g\|_\infty := \sup_{s,a} |g|$. Assuming $g(s,a) = e^{-s}$, $(S,A)\sim Uniform( [0,1]^2)$ leads to bounded $L_2$-norm $\|g\|_2=\sqrt{1-e^{-1}}<\infty$, yet we still have unbounded $L_{\infty}$-norm $\|g\|_\infty = \infty$.

First, let us assume tabular setting. Let $p_{min}$ be the minimum (or infimum) value of all $b_\mu(s,a)$ where $b_\mu$ is the density value at the given state-action pair. If we are given with a tabular setting where the underlying measure of the state-action space is the count measure, we have $b_\mu(s,a)=\mathbb{P}(S,A=s,a)$. Due to $\probsa/p_{min}\geq1$, the following holds for $\eta_\infty(\Upsilon_1, \Upsilon_2):=\supsa\eta(\Upsilon_1(s,a), \Upsilon_2(s,a))$,
\begin{align*}
	\eta_{\infty}(\Upsilon_1, \Upsilon_2) &\leq \sumsa\eta\{\Upsilon_1(s,a), \Upsilon_2(s,a)\} \leq \sum_{s,a} \frac{b_\mu(s,a)}{p_{min}} \cdot \eta\{\Upsilon_1(s,a), \Upsilon_2(s,a)\} = \frac{1}{p_{min}} \cdot \bar{\eta}(\Upsilon_1, \Upsilon_2). 
\end{align*}
Then, we can make use of this to bound the inaccuracy with Bellman residual as follows, 
\begin{align*}
	\bar{\eta}(\Upsilon, \Upsilon_\pi) \leq \eta_{\infty}(\Upsilon, \Upsilon_\pi) \leq \frac{1}{1-\gamma^{\beta_0}} \cdot \eta_{\infty}(\Upsilon, \mathcal{T}^\pi \Upsilon) \leq \frac{1}{1-\gamma^{\beta_0}} \cdot  \frac{1}{p_{min}} \cdot \bar{\eta}(\Upsilon, \mathcal{T}^\pi \Upsilon) ,
\end{align*}
where the second inequality is due to Equation \eqref{supremum_convergence_logic}. 

However, we cannot apply the same logic to state-action space with infinite cardinality $|\SAspace|\geq \infty$. Assume continuous state-action space, for example. In this case, it is tempting to switch all the summation $\sum_{s,a}$ into integral $\int_{\mathcal{S}\times \mathcal{A}}$. Unfortunately, however, we cannot show $\eta_\infty (\Upsilon_1, \Upsilon_2) \leq \int_{\mathcal{S}\times \mathcal{A}} \eta\{\Upsilon_1(s,a), \Upsilon_2(s,a)\} \mathrm{d}m(s,a)$, where $m$ is the underlying measure of the state-action space (say, Lesbesgue measure). Even when the space is countable (which allows us to use discrete summation $\sum_{s,a}$), we still have an issue since the infimum value of the density will be $p_{min}=0$, which leads to infinite-bound on the RHS.

In conclusion, the expectation-extended distance and supremum-extended distance are not equivalent when state-action space has infinite cardinality (i.e. non-tabular state-action space). This tells us that Theorem \ref{Fundamental_Realizable} is not trivial, since this can be applied for non-tabular state-action space (with either countable or uncountable cardinality). The well-known Atari games is nearly a non-tabular example with extremely many state-action pairs (state space alone having huge cardinality $|\mathcal{S}| = 256\times 250\times 160 \times 3$ which has little difference with $\infty$ at least in the practical sense). The OpenAI-gym classical games of Section \ref{atari_simulation} is an example which has continuous state space (with uncountably infinite cardinality). These two examples can be supported by our Theorem \ref{continuous_deterministic_realizable}, which was built upon Theorem \ref{Fundamental_Realizable}, since they have deterministic transition.

\subsubsection{Model Misspecification Example} \label{model_misspecification_rho}

Equation \eqref{mstep_minimizers} can be illustrated schematically as follows.

\begin{figure}[ht] 
	\vspace{.3in}
	\begin{center}    
		\includegraphics[width=0.5\linewidth]{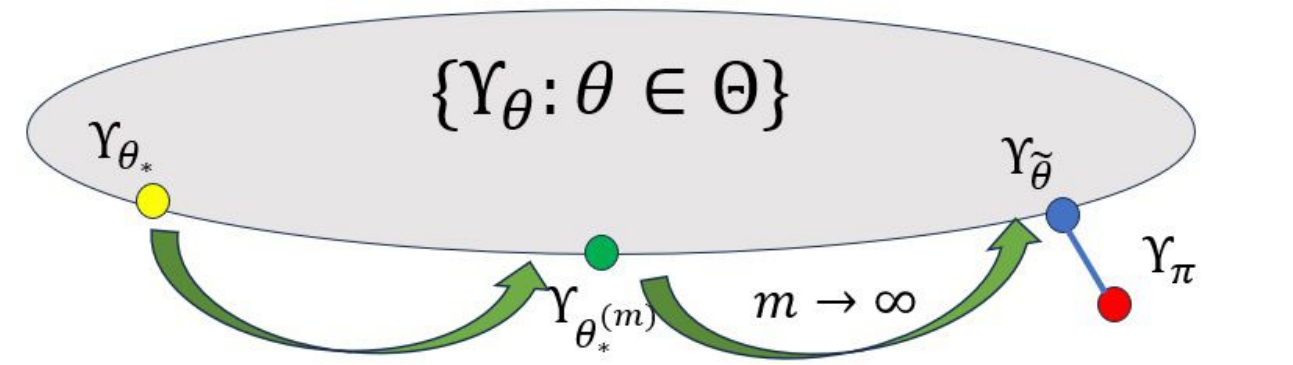}
		\caption{Larger $m$ makes $(\Bellopt)^m\Upsilon_\theta\approx \Upsilon_{\pi}$ in expected energy distance, and thereby leads to $\theta_*^{(m)}\approx \tilde{\theta}$.}
		\label{mstep_approximating} 
	\end{center}
	\vspace{.3in}
\end{figure} 

Here is an example. Letting $\rho=\theta$, we assume an arbitrarily large state-action space $|\SAspace|\in\mathbb{N}$, with an arbitrary behavior policy $b$ and target policy $\pi$. We assume that rewards are 2-dimensional ($d=2$), and follow the same distribution conditioned on any state-action pair, 
\begin{align*}
	R(s,a)\sim N(\mathbf{0}_2, \Sigma(\rho_0)) \quad \text{where} \quad \Sigma(\rho) := \sigma_0^2 \cdot \left( \begin{array}{cc}
		1 & \rho \\
		\rho &  1
	\end{array} \right), \quad \therefore \ Z_\pi(s,a)\sim N(\mathbf{0}_2, \Sigma(\rho_0) / (1-\gamma^2) ).
\end{align*}

\begin{figure}[ht] 
	\vspace{.3in}
	\begin{center}
		\includegraphics[width=1.0\linewidth]{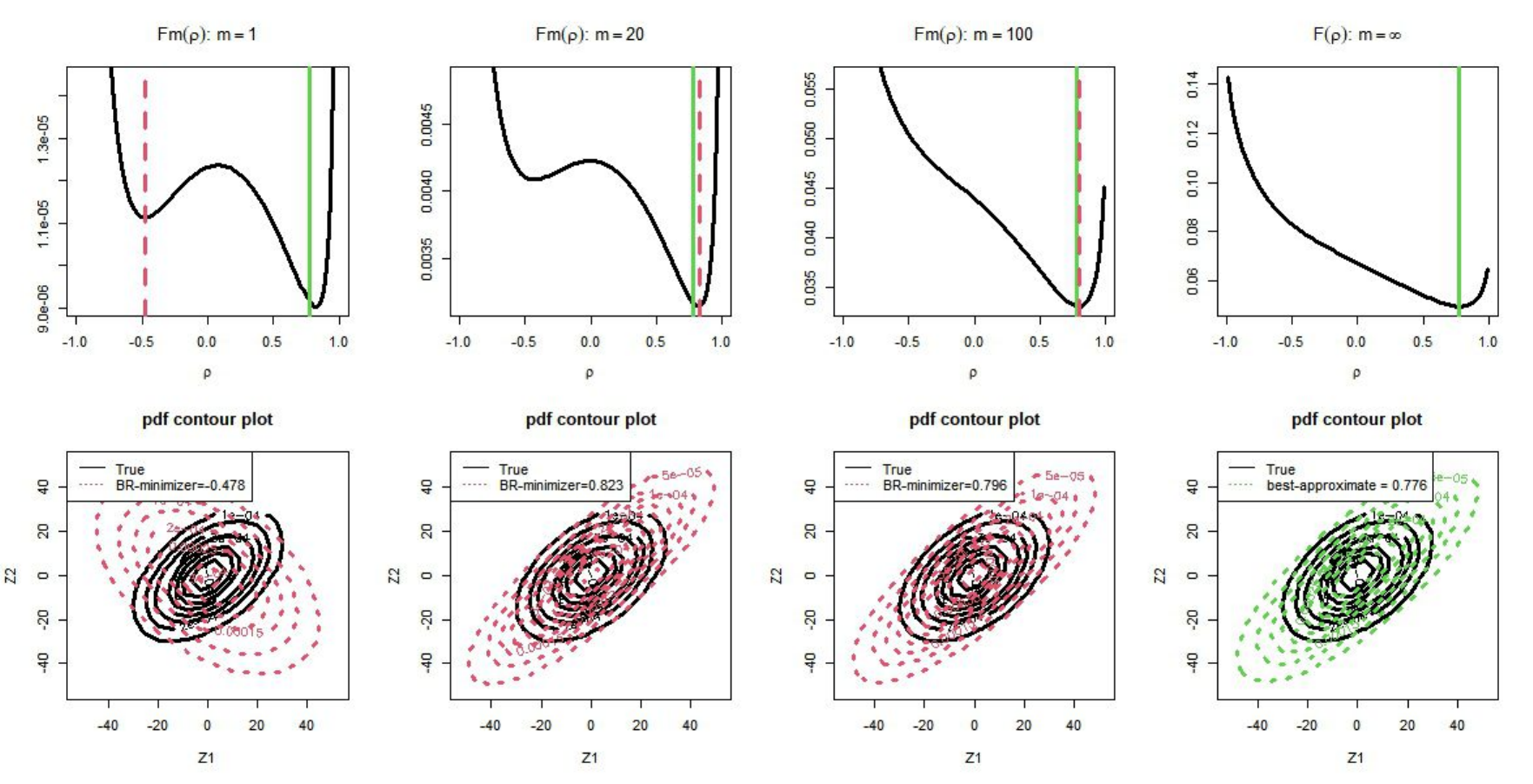}
		\label{mstep_approximating_appendix} 
		\caption{At the top, red line represents the selected minimizer of each $F_m$ by R function \texttt{optimize}, which may be a non-global local minimizer, whereas the green line represents the best $\energyonebar$-approximation. The bottom plots represent the corresponding pdf contour plots.}
	\end{center}
	\vspace{.3in}
\end{figure}

With $\gamma=0.99$, we assume that the true values are $\sigma_0^2=4$ and $\rho_0=0.5$, but we misspecified the variance with $\sigma_1^2=10$, and then we let $\rho\in[-1,1]$ to be the only parameter of interest. In this case, $\tilde{\rho}\approx0.776$ gives us the best $\energyonebar$-approximation, being the minimizer of the inaccuracy function $F(\rho)$ \eqref{objective_functions}. We can see in Figure \ref{mstep_approximating_appendix} that the minimizers of the population objective functions $F_m(\rho)$ approaches $\tilde{\rho}$ as we increase $m\rightarrow\infty$.  Moreover, in this example, as $F_m$ uniformly converges to $F$ \eqref{objectiveinaccuracy_difference_boundedby_G}, the risk of selecting the non-global local minimizer is reduced. However, we can also see in the rightmost bottom plot that there is irreducible error from the true distribution, due to misspecification of the variance value $\sigma_1^2>\sigma_0^2$.

\subsubsection{Examples That Satisfies Assumption \ref{uniqueness_generalized} With Quadratic Bound} \label{example_strongly_convex}

We will suggest two examples: strongly-convex and non-convex functions $F(\theta)$, which satisfy Assumption \ref{uniqueness_generalized} with $q=2$.

Here is a strongly convex example. We can arbitrarily choose the cardinality $|\SAspace|>0$ and discount rate $\gamma\in(0,1)$. We can also assume any behavior and target policies $b$ and $\pi$. Conditioned on $s,a$, its reward follows $R(s,a)\sim N(0,\thetatilde)$ with $\thetatilde>0$, which leads the target distribution $\Upsilon_\pi(s,a)$ to follow $N(0,\thetatilde/(1-\gamma^2))$ at all $s,a\in\SAspace$. We select a model so that $\Upsilon_\theta(s,a)$ is the probability measure of $N(0,\theta/(1-\gamma^2))$ with $\theta\in\Theta\subset[0,\infty)$ for all $s,a\in\SAspace$. Then we have 
\begin{align*}
	F(\theta)&=\energyonebar(\Upsilon_\theta , \Upsilon_\pi) = \sumsa \probsa \cdot \energyone\{ \Upsilon_\theta (s,a) , \Upsilon_\pi (s,a) \} 
	= \frac{2}{\sqrt{\pi}}\cdot \frac{1}{\sqrt{1-\gamma^2}} \cdot (\sqrt{2(\theta + \thetatilde)} - \sqrt{\theta} - \sqrt{\thetatilde}).
\end{align*}
For simplicity, let us assume $\thetatilde=1$ and $\Theta=[0,2]$. Since we have $F'(1)=0$ and $F''(1)=\frac{1}{4}\cdot \sqrt{\frac{1}{\pi\cdot (1-\gamma^2)}}>0$, this satisfies Assumption \ref{uniqueness_generalized} with $q=2$.

Next, we will suggest a non-convex example of $F(\mu)$ that may have multiple local minimizers. We will skip detailed calculations. Assume that the reward follows $R(s,a)\sim N(\mu_*, \sigma_0^2)$ with known value of $\sigma_0^2$ at all $s,a$, which leads to the true return distribution $Z_\pi(s,a)\sim N(\mu_* / (1-\gamma), \sigma_0^2 / (1-\gamma^2))$. Let us model our candidate distributions with $\mu\in[a,b]$ so that $Z(s,a;\mu)\sim N(g(\mu) / (1-\gamma), \sigma_0^2 / (1-\gamma^2) )$ with a pre-determined function $g$. Although letting $g(\mu)=\mu$ makes $F(\cdot)$ strongly convex, we may want to include various $g$. Relaxing $g$ into a non-monotonic function will lead to non-convex $F$ (possibly with local minimizers), however it satisfies Assumption 4.2 as long as following hold:

(i) There exists a unique value (say $\mu_0$) that satisfies $g(\mu_0)=\mu_*$, \\
(ii) $g(\cdot)$ is a straight line segment in a neighborhood of $\mu_0$. 

Note that (ii) can be relaxed into a polynomial curve instead of straight line segment, which can lead us to resort to more generalized Assumption \ref{uniqueness_generalized} that yields slower convergence rate (Theorem \ref{thetaBhat_convergence_generalized}). \\

\clearpage

\section{OTHER MATERIALS} \label{other_materials}

\subsection{Comparison Between Distributional OPE Methods In Details} \label{Comparison_between_DRL_details}

Here is a more detailed version of comparison of Table \ref{Comparison_Methods_summary}.

\begin{table*}[h]
	\caption{Comparison of theoretical aspects of distributional OPE methods. }
	\label{Comparison_Methods_summary_details}
	\begin{center}
		\begin{tabular}{lcccc}
			\toprule
			\multicolumn{1}{c}{\bf }                    &\multicolumn{1}{c}{\bf Statistical}   &\multicolumn{1}{c}{\bf Without}          & {\bf Multi-dimensional}
			\\ 
			\multicolumn{1}{c}{\bf Method}               &\multicolumn{1}{c}{\bf error bound}   &\multicolumn{1}{c}{\bf completeness}  & {\bf reward}
			\\ \midrule
			Categorical \citep{rowland2018analysis}                      & \xmark                               & \cmark                               & \xmark  \\ %
			QRTD \citep{rowland2023analysis}                             & \xmark                               & -                                   & \xmark   \\ %
			IQN \citep{dabney2018implicit}                               & \xmark                               & -                                   & \xmark   \\ %
			FQF \citep{yang2019fully}                                    & \xmark                               & -                                   & \xmark   \\ %
			QEMRL \citep{kuang2023variance}                              & \xmark                               & -                                   & \xmark   \\ %
			MMDRL \citep{nguyen2021distributional}                       & \xmark                               & -                                   & \cmark   \\ %
			SinkhornDRL \citep{sun2022distributional}                    & \xmark                               & -                                   & \cmark   \\ %
			MD3QN \citep{zhang2021distributional}                        & \xmark                               & -                                   & \cmark   \\ %
			EDRL \citep{rowland2019statistics}                           & \xmark                               & -                                   & \xmark   \\ %
			CDE \citep{ma2021conservative}                               & \cmark                               & \cmark                                   & \xmark \\ 
			NTD \citep{peng2024near}                                     & \cmark                               & \cmark                                   & \xmark \\ 
			FLE \citep{wu2023distributional}                             & \cmark                               & \xmark                               & \cmark  \\ %
			{\bf EBRM (our method)}                                      & \cmark                               & \cmark                               & \cmark  \\ %
			\bottomrule
		\end{tabular}
	\end{center}
\end{table*}

While early TD (temporal difference) methods of DRL demonstrated outperformance in Atari games over traditional deep-Q learning \citep{mnih2015human}, they do not have sound theoretical foundation. Instead of proving convergence or deriving statistical error bound, they mostly end up showing contraction of operators with respect to some supremum-extended \eqref{supremum_distance} probability distances. These operators are mostly Bellman operators or Bellman operators followed by some projection (e.g. $\Pi_\mathcal{C}, \Pi_{\Wasserone}, \mathcal{O}_c: \mathcal{P}(\mathbb{R}^d)^{\SAspace}\rightarrow  \mathcal{P}(\mathbb{R}^d)^{\SAspace}$). Refer to \cite{rowland2018analysis}, \cite{dabney2018distributional}, \cite{ma2021conservative} for their exact definitions. 

Of course, showing this contraction is not sufficient for showing probabilistic convergence. Exact Bellman-update is not even possible, due to random transition and tremendously many state-action pairs. This is why they use functional approximation (e.g. deep neural network) and minimize an objective function instead. In order to theoretically justify their theoretical convergence, by \eqref{supremum_convergence_logic}, they should show that their algorithm minimizes supremum-extended Bellman residual $\distancemeasure_{\infty}(\Upsilon, \Bellopt \Upsilon)$. Here, $\distancemeasure_{\infty}$ is the supremum-extended metric that they showed contraction of their operator. However, as can be seen from Table \ref{Comparison_Methods}, the objective functions that they have used do not seem to have any relationship with $\eta_\infty$ that can lead to minimizing the term $\distancemeasure_{\infty}(\Upsilon, \Bellopt \Upsilon)$.

\begin{table}[ht]  
	\caption{Contractive distances and objective functions of TD-based methods in distributional OPE} 
	\label{Comparison_Methods}
	\begin{center}
		\begin{tabular}{lll}
			\toprule
			\multicolumn{1}{c}{\bf Method}  &\multicolumn{1}{c}{\bf Operator \& Contractive Distance} &\multicolumn{1}{c}{\bf Objective Function} 
			\\ \midrule 
			Categorical         & $\Bellopt$: Wasserstein-$p$ ($p\geq 1$)     & Cross Entropy \\
			algorithm & (supremum-extended)  &  \\  
			\citep{bellemare2017distributional}   &   $\Pi_{\mathcal{C}} \Bellopt$: Cramer distance    &            \\ 
			\citep{rowland2018analysis} & (supremum-extended) &  \\ \hline
			\textbf{Quantile-based algorithms}        &      &   \\
			&      &   \\
			QRTD \citep{dabney2018distributional} & $\Pi_{\Wasserone} \Bellopt$: Wasserstein-$\infty$  & Huber Loss \\
			IQN \citep{dabney2018implicit}       & (supremum-extended) &  \\ 
			FQF \citep{yang2019fully}   &      &   \\
			QEMRL \citep{kuang2023variance}           &  &  \\  \hline
			\textbf{Particle-based algorithms}          &  &  \\ 
			&  &  \\ 
			MMDRL        & $\Bellopt$: supremum-extended $\text{MMD}_k$ & $\text{MMD}_k^2$  \\
			\citep{nguyen2021distributional}        & (unrectified kernel)   & (unrectified / Gaussian kernel)  \\
			MD3QN \citep{zhang2021distributional} & (supremum-extended) &   \\ 
			SinkhornDRL  & $\Bellopt$: Sinkhorn Divergence & Sinkhorn Divergence  \\
			\citep{sun2022distributional}   & (supremum-extended)  &    \\ \hline
			EDRL         & {no additional result} & Expectile Regression Loss  \\ 
			\citep{rowland2019statistics}         & about contraction  &  \\ \bottomrule 
		\end{tabular}
	\end{center}
\end{table}

Following the aforementioned methods, further works are done to verify statistical convergence in infinite-horizontal settings (Table \ref{Convergence_comparison}). However, many of them are restricted in many aspects. For instance, most of them only showed consistency without any statistical error bound (or probability convergence rate). Even in such cases, many are based on strong restrictions. These include (i) tabular setting ($|\SAspace|<\infty$), (ii) bounded rewards, (iii) availability of generative model (i.e. known transition probability $p(r,s'|s,a)$). In this regard, FLE \citep{wu2023distributional} is a great achievement in that it was free from all these restrictions. However, it was still based on a very strong condition called ``completeness'' that we have explained in Section \ref{Introduction}.

\begin{table}[t]
	\caption{Theoretical convergence results in distributional OPE} 
	\label{Convergence_comparison}
	\begin{center}
		\begin{tabular}{ll}
			\toprule
			\multicolumn{1}{c}{\bf Method}  &\multicolumn{1}{c}{\bf Type of convergence and underlying assumptions} 
			\\ \midrule 
			Categorical         &  Almost-sure convergence towards $\Upsilon_C \approx \Upsilon_\pi$ (Theorem 1, Proposition 4)  \\ 
			algorithm    &  (requires sampling from all state-action pairs, no convergence rate,    \\ 
			\citep{rowland2018analysis}    &  bounded reward)    \\ \hline
			QRTD         &  Almost-sure convergence towards a set (Theorem 8)   \\ 
			\citep{rowland2023analysis} & (Tabular case, no convergence rate) \\ \hline 
			CDE          & Shows finite-sample error bound (Theorem 3.6)  \\ 
			\citep{ma2021conservative} &  (Tabular case, bounded reward, ignores model restriction) \\ \hline
			NTD        &  shows finite-sample error bound (Theorem 3.1)   \\ 
			\citep{peng2024near} & (tabular case, requires generative model) \\ \hline
			FLE          &  Convergence rate suggested (Corollary 4.14) \\ 
			\citep{wu2023distributional}    &   (bounded reward, requires completeness)  \\  \hline
			{\bf EBRM}         &  shows finite-sample error bound (Theorems \ref{Realizable_Final_Theorem}, Appendix \ref{nonrealizable_finite_sample_error_bound} )  \\ 
			{\bf (our method)}   &   (no aforementioned restrictions)  \\ \bottomrule
		\end{tabular}
	\end{center}
\end{table}

\clearpage

\subsection{OpenAI-gym Classical Games Simulation} \label{sec:openai}

\subsubsection{Simulation Settings} \label{atari_games_simulation_details}

In all three methods (EBRM, QRDQN, MDMQN), we used the following network structure and tuning parameters. In three games, the state spaces are $\mathcal{S}\subsetneq \mathbb{R}^p$ with $p=6,4,2$ and the numbers of actions are $|\mathcal{A}|=3,2,3$ for Acrobot, Cartpole and Mountaincar, respectively. We are using 5 particles (or quantiles) to estimate the conditional distributions. 

\begin{table}[ht]
	\caption{Algorithmic details} 
	\begin{center}
		\begin{tabular}{lc}
			\toprule
			\multicolumn{1}{c}{\bf Tuning parameters}  &\multicolumn{1}{c}{ Value }
			\\ \midrule 
			layers and nodes  & FC ($\text{dim}(\mathcal{S}) \rightarrow 10$), ReLU, FC ($10 \rightarrow |\mathcal{A}|\times5$)  \\ 
			Number of particles  & 5 \\ 
			Learning Rate  & 0.01 \\ 
			optimizer  & Adam \\ \bottomrule  
		\end{tabular}
	\end{center}
\end{table}

OpenAI-gym games are originally designed as finite-horizontal settings where the game ends once the agent reaches the terminal state. However, we adjusted the game into infinite-horizontal setting by letting the agent return to a fixed point once it reaches the terminal state. We let the reward at the terminal state to be different from other rewards. Sample size is fixed as 100, where samples are collected based on a trajectory of $(s,a,r,s')$ through uniform policy $b(a|s)=1/|\mathcal{A}|$ and some initial state distribution $s\sim\mu$.

Since we are estimating the distribution of true return variable $Z_\pi(s,a)\sim \Upsilon_\pi(s,a)$ that is ``random,'' we let our target policy to be random. Our target policy was epsilon-greedy policy of the optimal policy that we trained with DQN \citep{mnih2015human}. That is, with probability $1-\epsilon$, the agent selects the optimal action learned by DQN, and with probability $\epsilon$, it randomly selects an action with equal probability.

However, $\epsilon$-greedy policy alone does not make $Z_\pi(s,a)$ random. Since it takes a long while to reach the terminal state and the rewards in non-terminal states are the same, the return becomes non-random for most of the times. Therefore, we let the agent repeat the selected action for multiple times, and consider this as a single transition. The settings are as follows for each game.

\begin{table}[ht]
	\caption{Details of experimental setting (OpenAI-gym games)} 
	\begin{center}
		\begin{tabular}{lccc}
			\toprule
			\multicolumn{1}{c}{\bf }  &\multicolumn{1}{c}{ Acrobot} &\multicolumn{1}{c}{ Cartpole } &\multicolumn{1}{c}{ Mountaincar}
			\\ \midrule 
			reward at non-terminal state  & -1 & 1 & -1   \\ 
			reward at terminal state & 5 & -5 & 2 \\ 
			epsilon  & 0.1 & 0.9 & 0.3  \\ 
			repetition of each action  & 10 & 1 & 10 \\ \bottomrule  
		\end{tabular}
	\end{center}
\end{table}

We used energy distance and Wasserstein metric to assess the inaccuracy from the true distribution that we approximated with Monte Carlo simulations. However, since there are too many state-action pairs, it is computationally impossible to approximate $\Upsilon_\pi(s,a)$ for all $s,a$. Therefore, we evaluated the marginal distributions instead, according to Corollary 4.14 of \cite{wu2023distributional}, as we have mentioned in Section \ref{atari_simulation}.

\subsubsection{More Extensive Simulations On Cartpole Game} \label{sec:cartpole_extensive}

Cartpole games seemed to best satisfy realizability assumption among all three games and have been tried by many other papers in reinforcement learning \citep[e.g.,][]{wang2023projected,narita2021debiased}. Therefore, we have done more extensive simulations on Cartpole games, based on various sample sizes and larger neural networks. In all three methods (EBRM, QRDQN, MDMQN), we used the following network structures (small and big) and tuning parameters (Table \ref{table:networks_cartpole}). In Cartpole, we have $\mathcal{S}\subsetneq \mathbb{R}^4$ (which means $\text{dim}(\mathcal{S})=4$) and 2 actions ($|\mathcal{A}|=2$). 

\begin{figure}[ht] 
	\vspace{.3in}
	\begin{center}
		\includegraphics[width=1.00\linewidth]{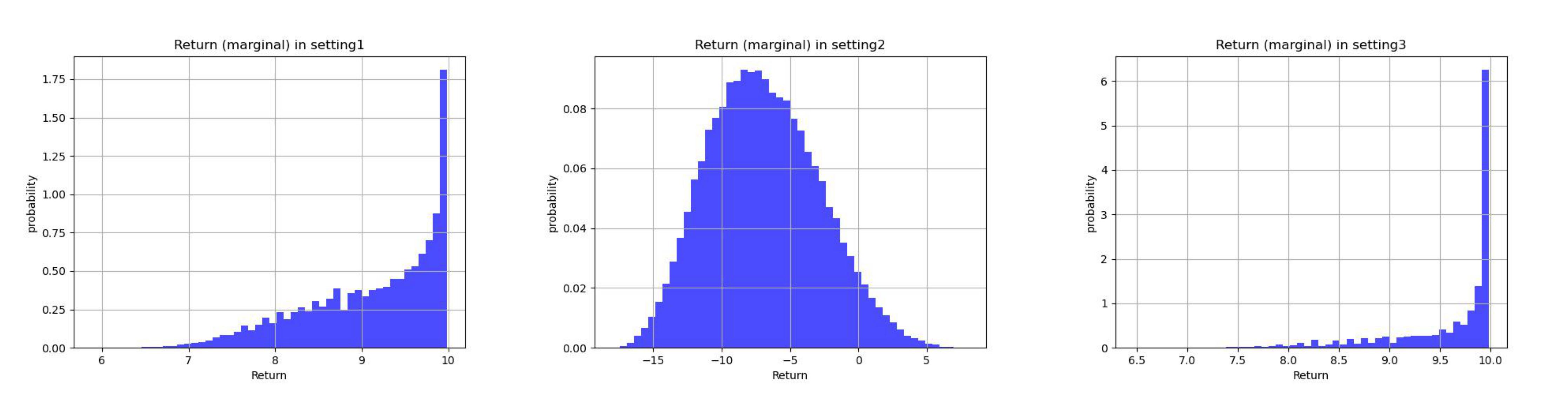}
		\caption{Histogram of Monte-Carlo approximated return distribution (marginalized) of Settings 1, 2, 3. Simulated with 100000 samples.}
		\label{fig:cartpole_returnplots} 
	\end{center}
	\vspace{.3in}
\end{figure}

The way we have constructed the settings is basically the same with that of Appendix \ref{atari_games_simulation_details}. However, we have tried more various sample sizes (ranging from 100 to 1000) and made further adjustments in the settings. We have tried three different settings (Table \ref{table:cartpole_three_settings}), where each setting gives us different types of marginal return distributions: left-skewed (Setting 1), bell-shaped (Setting 2), and highly-concentrated (Setting 3) marginal distributions. These are visualized in Figure \ref{fig:cartpole_returnplots}.

\begin{figure}[ht] 
	\vspace{.3in}
	\begin{center}
		\includegraphics[width=0.90\linewidth]{Figures/Energy_NNsmall.png}
		\includegraphics[width=0.90\linewidth]{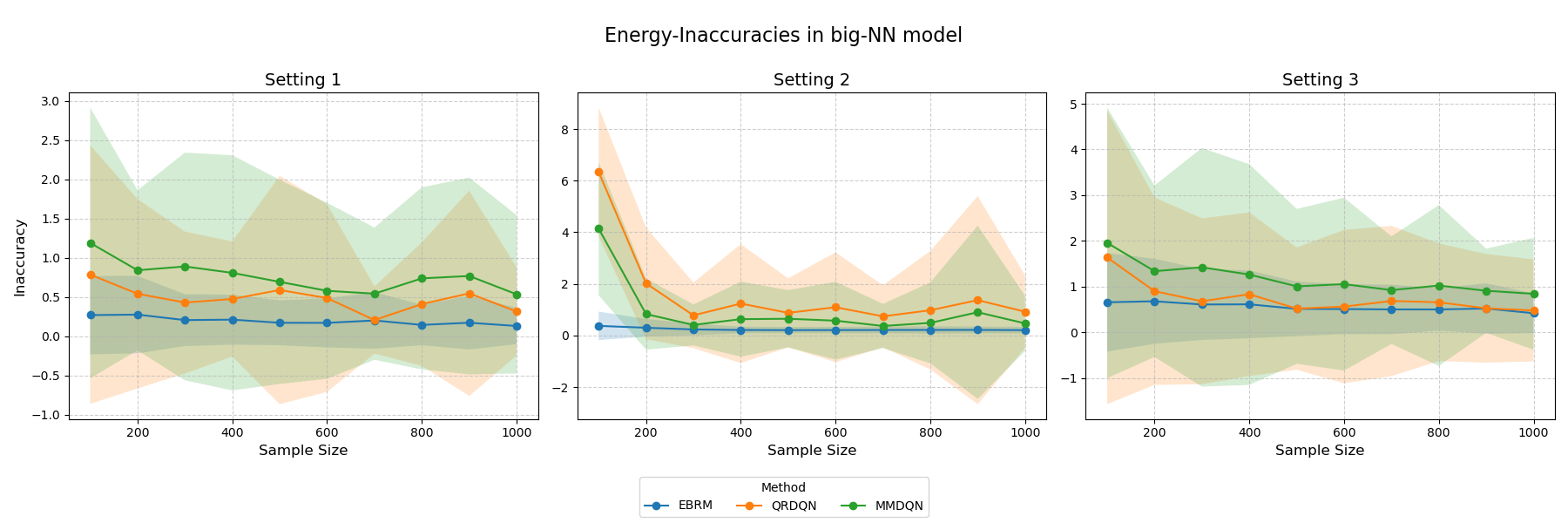}
		\includegraphics[width=0.90\linewidth]{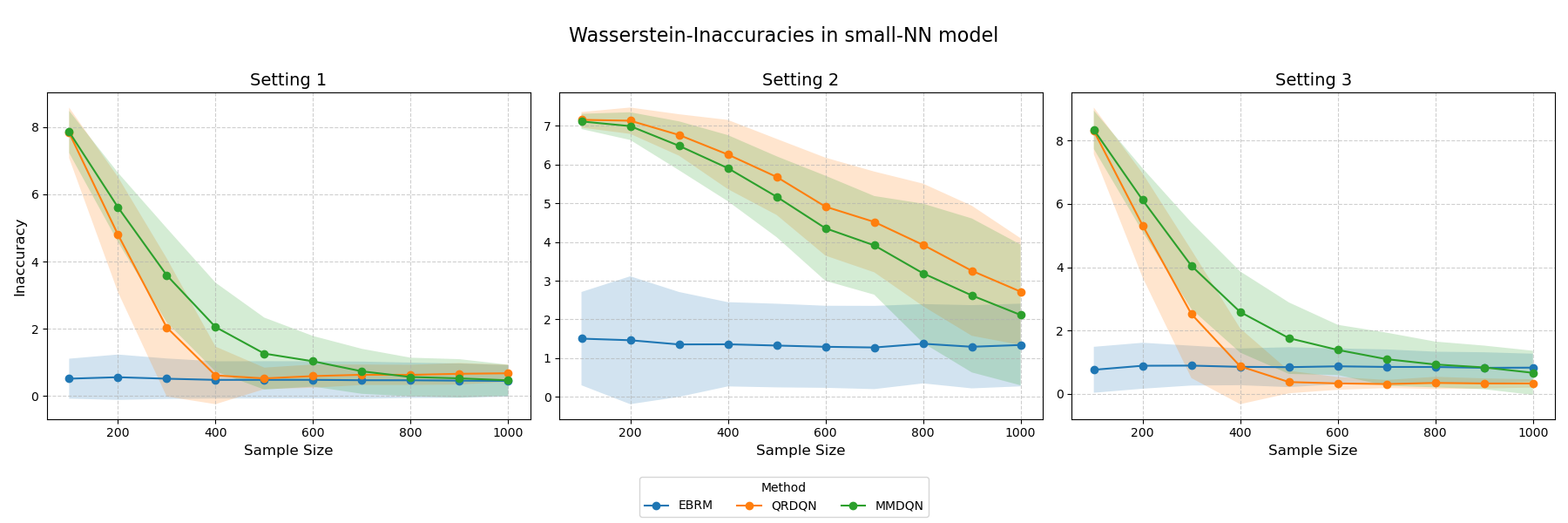}
		\includegraphics[width=0.90\linewidth]{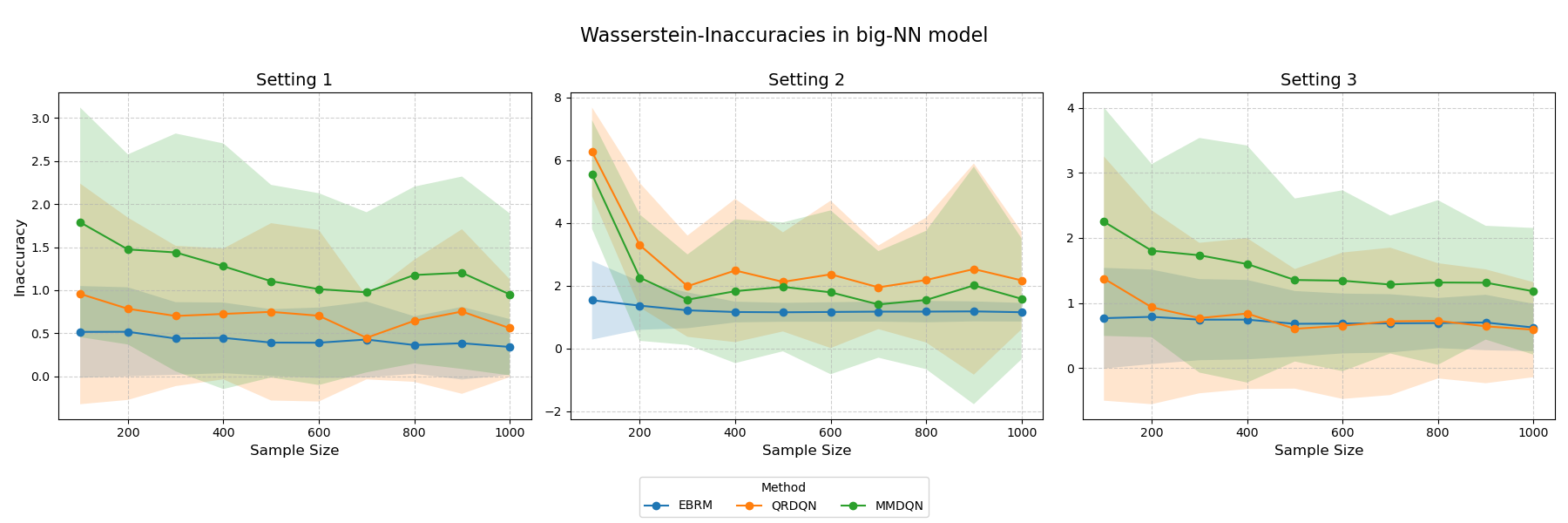}
		\caption{$\energyone$ and $\Wasserone$-inaccuracy based on small and big neural network models. Lines represent mean inaccuracy values and shaded regions represent the interval $(\text{Mean}\pm 2 \cdot \text{STD})$ (blue: EBRM, orange: QRDQN, green: MMDQN). }
		\label{fig:cartpole_energy} 
	\end{center}
	\vspace{.3in}
\end{figure}

Simulations are done in both neural networks shown in Table \ref{table:networks_cartpole}, whose results are shown in each of Tables \ref{table:Cartpole_energy_setting1}--\ref{table:Cartpole_wasserstein_setting3}, and visualized in Figure \ref{fig:cartpole_energy}. Outperformance of EBRM is evident in both accuracy and stability, particularly when equipped with the larger network model. This aligns with Theorem \ref{continuous_deterministic_realizable}, since larger models will lead to smaller extent of non-realizability.

\begin{table}[ht]
	\caption{Algorithmic details} 
	\label{table:networks_cartpole}
	\begin{center}
		\begin{tabular}{lc}
			\toprule
			\multicolumn{1}{c}{\bf Tuning parameters}  &\multicolumn{1}{c}{ Value }
			\\ \midrule 
			Neural network 1 (small)  &  FC ($\text{dim}(\mathcal{S}) \rightarrow 10$), ReLU, FC ($10 \rightarrow |\mathcal{A}|\times 5$)  \\ 
			Neural network 2 (big)  &  FC ($\text{dim}(\mathcal{S}) \rightarrow 64$), ReLU, FC ($64 \rightarrow 64$), ReLU, FC ($64 \rightarrow |\mathcal{A}|\times 5$)  \\ 
			Learning Rate  & 0.005 \\ 
			optimizer  & Adam \\ \bottomrule  
		\end{tabular}
	\end{center}
\end{table}

\begin{table}[ht]
	\caption{Details of experimental setting (Cartpole)} 
	\label{table:cartpole_three_settings}
	\begin{center}
		\begin{tabular}{lccc}
			\toprule
			\multicolumn{1}{c}{\bf }  &\multicolumn{1}{c}{Setting 1} &\multicolumn{1}{c}{Setting 2} &\multicolumn{1}{c}{Setting 3}
			\\ \midrule 
			reward at non-terminal state  & 1    & 1    & 1   \\ 
			reward at terminal state      & -5    & -5   & -5 \\ 
			epsilon                       & 0.9  & 0.9  & 0.7  \\ 
			repetition of each action     & 1   & 5    & 1 \\ \bottomrule  
		\end{tabular}
	\end{center}
\end{table}

\begin{table}[ht]
	\caption{Mean $\energyone$-inaccuracy (standard deviation in parenthesis) under Setting 1 over 30 simulations under small NN model (Network 1: Top) and big NN model (Network 2: Bottom). Smallest inaccuracy values in boldface.}  
	\label{table:Cartpole_energy_setting1}
	\begin{center}
		\begin{tabular}{lccccccccccc}
			\toprule
			\multicolumn{1}{c}{\bf $N$}  &\multicolumn{1}{c}{$100$} &\multicolumn{1}{c}{$200$} &\multicolumn{1}{c}{$300$}  &\multicolumn{1}{c}{$400$}  &\multicolumn{1}{c}{$500$}  &\multicolumn{1}{c}{$600$}  &\multicolumn{1}{c}{$700$}  &\multicolumn{1}{c}{$800$}  &\multicolumn{1}{c}{$900$}  &\multicolumn{1}{c}{$1000$} 
			\\ \midrule 
			EBRM & {\bf 0.275} & {\bf 0.316} & {\bf 0.272} & {\bf 0.238} & {\bf 0.238} & {\bf 0.231} & {\bf 0.222} & 0.218 & 0.208 & 0.195 \\
			& (0.256) & (0.311) & (0.254) & (0.225) & (0.231) & (0.224) & (0.232) & (0.217) & (0.202) & (0.180) \\ \hline
			QRDQN & 14.428 & 7.986 & 2.526 & 0.399 & 0.315 & 0.379 & 0.436 & 0.441 & 0.485 & 0.501 \\
			& (0.796) & (1.739) & (1.745) & (0.486) & (0.178) & (0.190) & (0.193) & (0.195) & (0.218) & (0.173) \\ \hline
			MMDQN & 14.248 & 8.948 & 4.345 & 1.552 & 0.701 & 0.477 & 0.285 & {\bf 0.195} & {\bf 0.187} & {\bf 0.155} \\
			& (0.710) & (1.181) & (1.394) & (0.927) & (0.478) & (0.288) & (0.217) & (0.157) & (0.142) & (0.115) \\ \bottomrule
		\end{tabular} 
	\end{center}
	\begin{center}
		\begin{tabular}{lccccccccccc}
			\toprule
			\multicolumn{1}{c}{\bf $N$}  &\multicolumn{1}{c}{$100$} &\multicolumn{1}{c}{$200$} &\multicolumn{1}{c}{$300$}  &\multicolumn{1}{c}{$400$}  &\multicolumn{1}{c}{$500$}  &\multicolumn{1}{c}{$600$}  &\multicolumn{1}{c}{$700$}  &\multicolumn{1}{c}{$800$}  &\multicolumn{1}{c}{$900$}  &\multicolumn{1}{c}{$1000$} 
			\\ \midrule 
			EBRM & {\bf 0.270} & {\bf 0.276} & {\bf 0.207} & {\bf 0.212} & {\bf 0.172} & {\bf 0.171} & {\bf 0.202} & {\bf 0.145} & {\bf 0.173} & {\bf 0.131} \\
			& (0.250) & (0.247) & (0.165) & (0.159) & (0.143) & (0.157) & (0.180) & (0.129) & (0.170) & (0.116) \\ \hline
			QRDQN & 0.787 & 0.542 & 0.430 & 0.475 & 0.590 & 0.487 & 0.207 & 0.411 & 0.547 & 0.316 \\
			& (0.824) & (0.603) & (0.452) & (0.367) & (0.727) & (0.597) & (0.214) & (0.393) & (0.654) & (0.280) \\ \hline
			MMDQN & 1.190 & 0.841 & 0.890 & 0.808 & 0.693 & 0.580 & 0.543 & 0.738 & 0.769 & 0.535 \\
			& (0.862) & (0.510) & (0.725) & (0.749) & (0.650) & (0.560) & (0.421) & (0.580) & (0.627) & (0.504) \\ \bottomrule
		\end{tabular} 
	\end{center}
\end{table}

\begin{table}[ht]
	\caption{Mean $\energyone$-inaccuracy (standard deviation in parenthesis) under Setting 2 over 30 simulations under small NN model (Network 1: Top) and big NN model (Network 2: Bottom). Smallest inaccuracy values in boldface.} 
	\begin{center}
		\begin{tabular}{lccccccccccc}
			\toprule
			\multicolumn{1}{c}{\bf $N$}  &\multicolumn{1}{c}{$100$} &\multicolumn{1}{c}{$200$} &\multicolumn{1}{c}{$300$}  &\multicolumn{1}{c}{$400$}  &\multicolumn{1}{c}{$500$}  &\multicolumn{1}{c}{$600$}  &\multicolumn{1}{c}{$700$}  &\multicolumn{1}{c}{$800$}  &\multicolumn{1}{c}{$900$}  &\multicolumn{1}{c}{$1000$} 
			\\ \midrule 
			EBRM & {\bf 0.373} & {\bf 0.374} & {\bf 0.320} & {\bf 0.305} & {\bf 0.296} & {\bf 0.280} & {\bf 0.275} & {\bf 0.304} & {\bf 0.275} & {\bf 0.295} \\ 
			& (0.281) & (0.399) & (0.300) & (0.234) & (0.232) & (0.218) & (0.239) & (0.200) & (0.194) & (0.200) \\ \hline
			QRDQN & 9.311 & 8.891 & 7.809 & 6.656 & 5.553 & 4.252 & 3.652 & 2.860 & 2.050 & 1.439 \\ 
			& (0.179) & (0.341) & (0.548) & (0.834) & (0.890) & (0.970) & (0.924) & (1.083) & (0.938) & (0.715) \\ \hline
			MMDQN & 9.014 & 8.206 & 6.755 & 5.395 & 4.045 & 2.839 & 2.256 & 1.553 & 1.102 & 0.765 \\
			& (0.198) & (0.436) & (0.678) & (0.848) & (0.840) & (0.932) & (0.743) & (0.815) & (0.738) & (0.567) \\ \bottomrule
		\end{tabular} 
	\end{center}
	\begin{center}
		\begin{tabular}{lccccccccccc}
			\toprule
			\multicolumn{1}{c}{\bf $N$}  &\multicolumn{1}{c}{$100$} &\multicolumn{1}{c}{$200$} &\multicolumn{1}{c}{$300$}  &\multicolumn{1}{c}{$400$}  &\multicolumn{1}{c}{$500$}  &\multicolumn{1}{c}{$600$}  &\multicolumn{1}{c}{$700$}  &\multicolumn{1}{c}{$800$}  &\multicolumn{1}{c}{$900$}  &\multicolumn{1}{c}{$1000$} 
			\\ \midrule 
			EBRM & {\bf 0.376} & {\bf 0.303} & {\bf 0.237} & {\bf 0.216} & {\bf 0.211} & {\bf 0.213} & {\bf 0.216} & {\bf 0.218} & {\bf 0.219} & {\bf 0.211} \\
			& (0.276) & (0.171) & (0.114) & (0.063) & (0.060) & (0.062) & (0.067) & (0.072) & (0.066) & (0.063) \\ \hline
			QRDQN & 6.338 & 2.022 & 0.783 & 1.239 & 0.881 & 1.097 & 0.745 & 0.981 & 1.372 & 0.924 \\
			& (1.248) & (1.081) & (0.636) & (1.151) & (0.668) & (1.064) & (0.600) & (1.147) & (2.015) & (0.700) \\ \hline
			MMDQN & 4.142 & 0.833 & 0.412 & 0.636 & 0.651 & 0.577 & 0.372 & 0.491 & 0.904 & 0.471 \\
			& (1.290) & (0.689) & (0.397) & (0.726) & (0.555) & (0.754) & (0.428) & (0.784) & (1.678) & (0.533) \\ \bottomrule
		\end{tabular} 
	\end{center}
\end{table}

\begin{table}[ht]
	\caption{Mean $\energyone$-inaccuracy (standard deviation in parenthesis) under Setting 3 over 30 simulations under small NN model (Network 1: Top) and big NN model (Network 2: Bottom). Smallest inaccuracy values in boldface.} 
	\begin{center}
		\begin{tabular}{lccccccccccc}
			\toprule
			\multicolumn{1}{c}{\bf $N$}  &\multicolumn{1}{c}{$100$} &\multicolumn{1}{c}{$200$} &\multicolumn{1}{c}{$300$}  &\multicolumn{1}{c}{$400$}  &\multicolumn{1}{c}{$500$}  &\multicolumn{1}{c}{$600$}  &\multicolumn{1}{c}{$700$}  &\multicolumn{1}{c}{$800$}  &\multicolumn{1}{c}{$900$}  &\multicolumn{1}{c}{$1000$} 
			\\ \midrule 
			EBRM & {\bf 0.611} & {\bf 0.728} & {\bf 0.729} & {\bf 0.665} & 0.654 & 0.675 & 0.644 & 0.634 & 0.604 & 0.603 \\
			& (0.476) & (0.512) & (0.429) & (0.391) & (0.417) & (0.386) & (0.381) & (0.325) & (0.339) & (0.286) \\ \hline
			QRDQN & 15.695 & 9.322 & 3.680 & 0.909 & {\bf 0.247} & {\bf 0.173} & {\bf 0.144} & {\bf 0.181} & {\bf 0.155} & {\bf 0.140} \\
			& (0.784) & (1.675) & (1.802) & (0.824) & (0.249) & (0.129) & (0.084) & (0.146) & (0.107) & (0.060) \\ \hline
			MMDQN & 15.533 & 10.283 & 5.528 & 2.521 & 1.327 & 0.868 & 0.629 & 0.506 & 0.466 & 0.330 \\
			& (0.691) & (1.144) & (1.379) & (1.083) & (0.714) & (0.404) & (0.409) & (0.343) & (0.307) & (0.246) \\ \bottomrule
		\end{tabular} 
	\end{center}
	\begin{center}
		\begin{tabular}{lccccccccccc}
			\toprule
			\multicolumn{1}{c}{\bf $N$}  &\multicolumn{1}{c}{$100$} &\multicolumn{1}{c}{$200$} &\multicolumn{1}{c}{$300$}  &\multicolumn{1}{c}{$400$}  &\multicolumn{1}{c}{$500$}  &\multicolumn{1}{c}{$600$}  &\multicolumn{1}{c}{$700$}  &\multicolumn{1}{c}{$800$}  &\multicolumn{1}{c}{$900$}  &\multicolumn{1}{c}{$1000$} 
			\\ \midrule 
			EBRM & {\bf 0.657} & {\bf 0.679} & {\bf 0.609} & {\bf 0.612} & {\bf 0.513} & {\bf 0.509} & {\bf 0.501} & {\bf 0.500} & {\bf 0.521} & {\bf 0.417} \\
			& (0.539) & (0.465) & (0.388) & (0.372) & (0.299) & (0.275) & (0.271) & (0.233) & (0.276) & (0.217) \\ \hline
			QRDQN & 1.638 & 0.900 & 0.679 & 0.833 & 0.518 & 0.562 & 0.685 & 0.657 & 0.525 & 0.479 \\
			& (1.605) & (1.025) & (0.908) & (0.894) & (0.668) & (0.839) & (0.822) & (0.643) & (0.594) & (0.558) \\ \hline
			MMDQN & 1.957 & 1.337 & 1.423 & 1.264 & 1.005 & 1.054 & 0.923 & 1.023 & 0.909 & 0.842 \\
			& (1.480) & (0.938) & (1.305) & (1.207) & (0.847) & (0.946) & (0.589) & (0.879) & (0.460) & (0.613) \\ \bottomrule
		\end{tabular} 
	\end{center}
\end{table}

\begin{table}[ht]
	\caption{Mean $\Wasserone$-inaccuracy (standard deviation in parenthesis) under Setting 1 over 30 simulations under small NN model (Network 1: Top) and big NN model (Network 2: Bottom). Smallest inaccuracy values in boldface.} 
	\begin{center}
		\begin{tabular}{lccccccccccc}
			\toprule
			\multicolumn{1}{c}{\bf $N$}  &\multicolumn{1}{c}{$100$} &\multicolumn{1}{c}{$200$} &\multicolumn{1}{c}{$300$}  &\multicolumn{1}{c}{$400$}  &\multicolumn{1}{c}{$500$}  &\multicolumn{1}{c}{$600$}  &\multicolumn{1}{c}{$700$}  &\multicolumn{1}{c}{$800$}  &\multicolumn{1}{c}{$900$}  &\multicolumn{1}{c}{$1000$} 
			\\ \midrule 
			EBRM & {\bf 0.519} & {\bf 0.563} & {\bf 0.518} & {\bf 0.483} & {\bf 0.484} & {\bf 0.484} & {\bf 0.473} & {\bf 0.472} & {\bf 0.461} & {\bf 0.453} \\
			& (0.297) & (0.337) & (0.301) & (0.276) & (0.277) & (0.277) & (0.277) & (0.263) & (0.255) & (0.234) \\ \hline
			QRDQN & 7.830 & 4.805 & 2.043 & 0.616 & 0.528 & 0.596 & 0.635 & 0.633 & 0.665 & 0.680 \\
			& (0.376) & (0.856) & (1.027) & (0.428) & (0.160) & (0.173) & (0.152) & (0.150) & (0.160) & (0.129) \\ \hline
			MMDQN & 7.859 & 5.617 & 3.595 & 2.054 & 1.265 & 1.036 & 0.738 & 0.569 & 0.529 & 0.471 \\
			& (0.313) & (0.514) & (0.701) & (0.660) & (0.535) & (0.376) & (0.334) & (0.287) & (0.285) & (0.232) \\ \bottomrule
		\end{tabular} 
	\end{center}
	\begin{center}
		\begin{tabular}{lccccccccccc}
			\toprule
			\multicolumn{1}{c}{\bf $N$}  &\multicolumn{1}{c}{$100$} &\multicolumn{1}{c}{$200$} &\multicolumn{1}{c}{$300$}  &\multicolumn{1}{c}{$400$}  &\multicolumn{1}{c}{$500$}  &\multicolumn{1}{c}{$600$}  &\multicolumn{1}{c}{$700$}  &\multicolumn{1}{c}{$800$}  &\multicolumn{1}{c}{$900$}  &\multicolumn{1}{c}{$1000$} 
			\\ \midrule 
			EBRM & {\bf 0.516} & {\bf 0.517} & {\bf 0.439} & {\bf 0.447} & {\bf 0.392} & {\bf 0.390} & {\bf 0.427} & {\bf 0.363} & {\bf 0.384} & {\bf 0.341} \\
			& (0.266) & (0.258) & (0.211) & (0.205) & (0.194) & (0.204) & (0.221) & (0.167) & (0.211) & (0.162) \\ \hline
			QRDQN & 0.957 & 0.784 & 0.701 & 0.724 & 0.749 & 0.704 & 0.447 & 0.645 & 0.753 & 0.559 \\
			& (0.641) & (0.529) & (0.408) & (0.380) & (0.515) & (0.498) & (0.242) & (0.355) & (0.478) & (0.284) \\ \hline
			MMDQN & 1.789 & 1.474 & 1.439 & 1.278 & 1.105 & 1.013 & 0.976 & 1.177 & 1.203 & 0.952 \\
			& (0.667) & (0.552) & (0.691) & (0.714) & (0.560) & (0.557) & (0.465) & (0.513) & (0.559) & (0.471) \\ \bottomrule
		\end{tabular} 
	\end{center}
\end{table}

\begin{table}[ht]
	\caption{Mean $\Wasserone$-inaccuracy (standard deviation in parenthesis) under Setting 2 over 30 simulations under small NN model (Network 1: Top) and big NN model (Network 2: Bottom). Smallest inaccuracy values in boldface.} 
	\begin{center}
		\begin{tabular}{lccccccccccc}
			\toprule
			\multicolumn{1}{c}{\bf $N$}  &\multicolumn{1}{c}{$100$} &\multicolumn{1}{c}{$200$} &\multicolumn{1}{c}{$300$}  &\multicolumn{1}{c}{$400$}  &\multicolumn{1}{c}{$500$}  &\multicolumn{1}{c}{$600$}  &\multicolumn{1}{c}{$700$}  &\multicolumn{1}{c}{$800$}  &\multicolumn{1}{c}{$900$}  &\multicolumn{1}{c}{$1000$} 
			\\ \midrule 
			EBRM & {\bf 1.505} & {\bf 1.463} & {\bf 1.355} & {\bf 1.359} & {\bf 1.326} & {\bf 1.294} & {\bf 1.276} & {\bf 1.373} & {\bf 1.295} & {\bf 1.340} \\
			& (0.602) & (0.826) & (0.677) & (0.544) & (0.540) & (0.530) & (0.537) & (0.512) & (0.537) & (0.534) \\ \hline
			QRDQN & 7.155 & 7.132 & 6.763 & 6.259 & 5.680 & 4.913 & 4.518 & 3.925 & 3.257 & 2.718 \\
			& (0.101) & (0.170) & (0.268) & (0.447) & (0.491) & (0.632) & (0.650) & (0.791) & (0.840) & (0.689) \\ \hline
			MMDQN & 7.114 & 6.992 & 6.484 & 5.907 & 5.169 & 4.354 & 3.914 & 3.192 & 2.620 & 2.116 \\
			& (0.099) & (0.179) & (0.316) & (0.427) & (0.522) & (0.679) & (0.637) & (0.899) & (0.994) & (0.908) \\ \bottomrule
		\end{tabular} 
	\end{center}
	\begin{center}
		\begin{tabular}{lccccccccccc}
			\toprule
			\multicolumn{1}{c}{\bf $N$}  &\multicolumn{1}{c}{$100$} &\multicolumn{1}{c}{$200$} &\multicolumn{1}{c}{$300$}  &\multicolumn{1}{c}{$400$}  &\multicolumn{1}{c}{$500$}  &\multicolumn{1}{c}{$600$}  &\multicolumn{1}{c}{$700$}  &\multicolumn{1}{c}{$800$}  &\multicolumn{1}{c}{$900$}  &\multicolumn{1}{c}{$1000$} 
			\\ \midrule 
			EBRM & {\bf 1.535} & {\bf 1.363} & {\bf 1.215} & {\bf 1.160} & {\bf 1.151} & {\bf 1.162} & {\bf 1.172} & {\bf 1.173} & {\bf 1.180} & {\bf 1.153} \\
			& (0.626) & (0.385) & (0.286) & (0.165) & (0.150) & (0.155) & (0.159) & (0.171) & (0.159) & (0.148) \\ \hline
			QRDQN & 6.263 & 3.290 & 1.983 & 2.483 & 2.120 & 2.360 & 1.947 & 2.180 & 2.528 & 2.169 \\
			& (0.707) & (0.991) & (0.807) & (1.141) & (0.790) & (1.174) & (0.666) & (0.995) & (1.682) & (0.779) \\ \hline
			MMDQN & 5.542 & 2.255 & 1.553 & 1.824 & 1.962 & 1.787 & 1.402 & 1.544 & 2.007 & 1.580 \\
			& (0.866) & (1.007) & (0.721) & (1.146) & (1.027) & (1.305) & (0.849) & (1.103) & (1.893) & (0.963) \\ \bottomrule
		\end{tabular} 
	\end{center}
\end{table}

\begin{table}[ht]
	\caption{Mean $\Wasserone$-inaccuracy (standard deviation in parenthesis) under Setting 3 over 30 simulations under small NN model (Network 1: Top) and big NN model (Network 2: Bottom). Smallest inaccuracy values in boldface.} 
	\label{table:Cartpole_wasserstein_setting3}
	\begin{center}
		\begin{tabular}{lccccccccccc}
			\toprule
			\multicolumn{1}{c}{\bf $N$}  &\multicolumn{1}{c}{$100$} &\multicolumn{1}{c}{$200$} &\multicolumn{1}{c}{$300$}  &\multicolumn{1}{c}{$400$}  &\multicolumn{1}{c}{$500$}  &\multicolumn{1}{c}{$600$}  &\multicolumn{1}{c}{$700$}  &\multicolumn{1}{c}{$800$}  &\multicolumn{1}{c}{$900$}  &\multicolumn{1}{c}{$1000$} 
			\\ \midrule 
			EBRM & {\bf 0.755} & {\bf 0.883} & {\bf 0.887} & {\bf 0.848} & 0.838 & 0.865 & 0.845 & 0.843 & 0.817 & 0.822 \\
			& (0.363) & (0.363) & (0.315) & (0.290) & (0.316) & (0.282) & (0.277) & (0.243) & (0.246) & (0.220) \\ \hline
			QRDQN & 8.312 & 5.322 & 2.518 & 0.864 & {\bf 0.367} & {\bf 0.324} & {\bf 0.304} & {\bf 0.339} & {\bf 0.322} & {\bf 0.323} \\
			& (0.370) & (0.823) & (1.013) & (0.599) & (0.178) & (0.101) & (0.065) & (0.097) & (0.080) & (0.065) \\ \hline
			MMDQN & 8.348 & 6.125 & 4.045 & 2.582 & 1.754 & 1.383 & 1.090 & 0.923 & 0.829 & 0.659 \\
			& (0.304) & (0.501) & (0.684) & (0.642) & (0.565) & (0.397) & (0.420) & (0.362) & (0.345) & (0.349) \\ \bottomrule
		\end{tabular} 
	\end{center}
	\begin{center}
		\begin{tabular}{lccccccccccc}
			\toprule
			\multicolumn{1}{c}{\bf $N$}  &\multicolumn{1}{c}{$100$} &\multicolumn{1}{c}{$200$} &\multicolumn{1}{c}{$300$}  &\multicolumn{1}{c}{$400$}  &\multicolumn{1}{c}{$500$}  &\multicolumn{1}{c}{$600$}  &\multicolumn{1}{c}{$700$}  &\multicolumn{1}{c}{$800$}  &\multicolumn{1}{c}{$900$}  &\multicolumn{1}{c}{$1000$} 
			\\ \midrule 
			EBRM & {\bf 0.768} & {\bf 0.788} & {\bf 0.744} & {\bf 0.745} & 0.681 & 0.686 & {\bf 0.688} & {\bf 0.692} & 0.701 & 0.624 \\
			& (0.386) & (0.363) & (0.311) & (0.305) & (0.252) & (0.230) & (0.224) & (0.193) & (0.213) & (0.182) \\ \hline
			QRDQN & 1.376 & 0.935 & 0.769 & 0.838 & {\bf 0.604} & {\bf 0.651} & 0.718 & 0.727 & {\bf 0.642} & {\bf 0.591} \\
			& (0.938) & (0.745) & (0.578) & (0.580) & (0.459) & (0.562) & (0.566) & (0.443) & (0.437) & (0.365) \\ \hline
			MMDQN & 2.250 & 1.805 & 1.734 & 1.599 & 1.354 & 1.343 & 1.284 & 1.316 & 1.313 & 1.180 \\
			& (0.877) & (0.665) & (0.901) & (0.911) & (0.626) & (0.695) & (0.530) & (0.632) & (0.437) & (0.487) \\ \bottomrule
		\end{tabular} 
	\end{center}
\end{table}

\clearpage

\subsection{Realizable And Nonrealizable Models} \label{realizable_and_nonrealizable_models}

\subsubsection{Introduction Of Simulation Settings} \label{given_environment}

The initial state distribution and behavior / target policies are
\begin{gather}
	S\sim \text{Unif}\big\{ 1,2,\cdots, 30 \big\} \quad \mbox{and} \quad A\sim b(\cdot|S), \label{policy_explanation} \\
	b(a|s)=1/2 \quad \mbox{and} \quad \pi(1|s)=1 \quad \mbox{for} \quad  \forall s,a\in\SAspace. \nonumber
\end{gather}
Let us assume that $s,a$ is given. The agent moves by one in the direction of $a\in\{-1,1\}$, that is $s'=s+a$ (value of which fully determines the reward distribution \eqref{environment_explanation}). If the agent is already blocked by the direction, that is $(s,a)=(1,-1)$ or $(s,a)=(30,1)$, it stays at the same position $s'=s$. With given values of $A_0>0, p_0\in(0,1), \sigma_0^2>0$, our transition $p(r,s'|s,a)$ is characterized by:
\begin{align}
	\text{Conditioned on } S+A=k, \ &R\sim N(\mu_k, \sigma_0^2) \quad \text{where} \quad \mu_k = \begin{cases}
		A_0\cdot p_0^k & (k=0,\cdots,30) \\
		0 & (k=31)
	\end{cases} \label{environment_explanation} \\
	\mbox{and } & S' = k \ \text{if } k\in\{1,\cdots,30\}, \quad S'=30 \ \text{if }k=31, \quad S'=1 \ \text{if } k=0. \nonumber
\end{align}
We assume infinite-horizontal setting. Following the environment \eqref{environment_explanation} and target policy \eqref{policy_explanation}, we have $Z_\pi(s,a)\sim \Upsilon_\pi(s,a)$ to be normal distributions, with expectation and variance as follows,
\begin{gather} \label{realizable_example_equations}  
	\mathbb{E}\big\{ Z_\pi(i,-1)  \big\} = A_0\cdot p_0^{i-1} \cdot \frac{1-(\gamma p_0)^{32-i}}{1-\gamma p_0} \ \ (i\geq 2), \quad \mathbb{V}\big\{ Z_\pi(i,\pm 1)  \big\} = \frac{\sigma_0^2}{1-\gamma^2} \ \ (i\geq 1) ,   \\ 
	\mathbb{E}\big\{ Z_\pi(i,1)  \big\} = A_0 \cdot p_0^{i+1} \cdot \frac{1-(\gamma p_0)^{30-i}}{1-\gamma p_0}  \ \ (i\geq1) , \quad  \mathbb{E}\big\{ Z_\pi(1,-1) \big\} = A_0 + \gamma\cdot \mathbb{E}\big\{ Z_\pi(1,1) \big\}. \nonumber
\end{gather}
We always let $A_0=100$, $p_0=0.9$ throughout the simulations. 

First, we assumed a realizable scenario where the correct model \eqref{realizable_example_equations} is known (Appendix \ref{realizable_model_plot}), only not knowing the values of $A_0$, $p_0$. Here, we always assumed $\gamma=0.99$ and tried two settings with $\sigma_0^2=20$ and $\sigma_0^2=5000$. Second, we always tried the non-realizable scenario where there is a model misspecification \eqref{nonrealizability_misspecified_model}, as will be demonstrated in Appendix \ref{nonrealizable_model_plot}. Here, we always assumed $\sigma_0^2=20$, trying $\gamma=0.50$ and $\gamma=0.99$.

\subsubsection{Realizable Scenario} \label{realizable_model_plot}

\begin{figure}[ht] 
	\vspace{.3in}
	\begin{center}
		\includegraphics[width=0.50\linewidth]{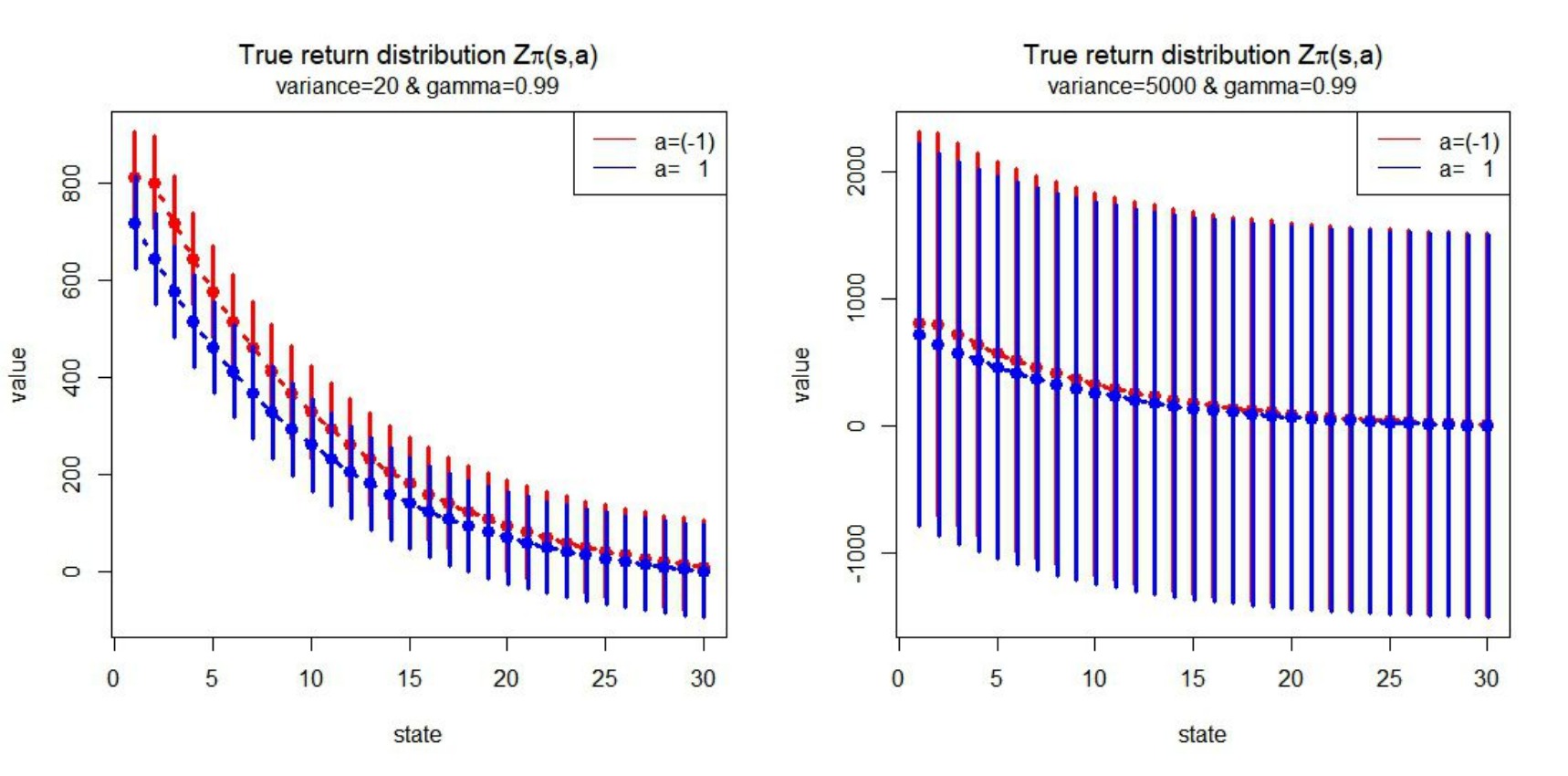}
		\caption{Red and blue represent the distributions of $Z_\pi(s,-1)$ and $Z_\pi(s,1)$ respectively. The dots indicate the expectation values and the vertical bars include (Mean$\pm$ 3$\cdot$SD).}
		\label{true_distributions} 
	\end{center}
	\vspace{.3in}
\end{figure}

In the realizable scenario, we assume that Equations (\ref{realizable_example_equations}) are known, except the values of $A_0=100$ and $p_0=0.9$. The distributions can be plotted as Figure \ref{true_distributions}, each for $\sigma_0^2=20$ and $\sigma_0^2=5000$.

\subsubsection{Non-realizable Scenario} \label{nonrealizable_model_plot}

On the other hand, in the non-realizable scenario, we assume that we are not aware of the true model (\ref{realizable_example_equations}). Instead, we assume that we are only aware of the decreasing trend demonstrated in Figure \ref{true_distributions}. That being said, we apply the following linear model that holds for all $1\leq i \leq 30$. with four different parameters $\beta_L,\beta_R,\beta_1\in\mathbb{R}$, $\sigma^2>0$,
\begin{gather}
	\mathbb{E}\big( Z_\pi(i,-1)  \big) = \beta_L + \beta_1\cdot i,  \ \  \mathbb{E}\big( Z_\pi(i,1)  \big) = \beta_R + \beta_1\cdot i, \ \ \mathbb{V}\big( Z_\pi(i,\pm 1)  \big) = \frac{\sigma^2}{1-\gamma^2}.  \label{nonrealizability_misspecified_model}
\end{gather}
This means that the distributions (conditioned on each $s,a$) have common variance, common slope in expectations, but different $y$-intercepts in expectations. 

\begin{figure}[h] 
	\vspace{.3in}
	\begin{center}
		\includegraphics[width=0.50\linewidth]{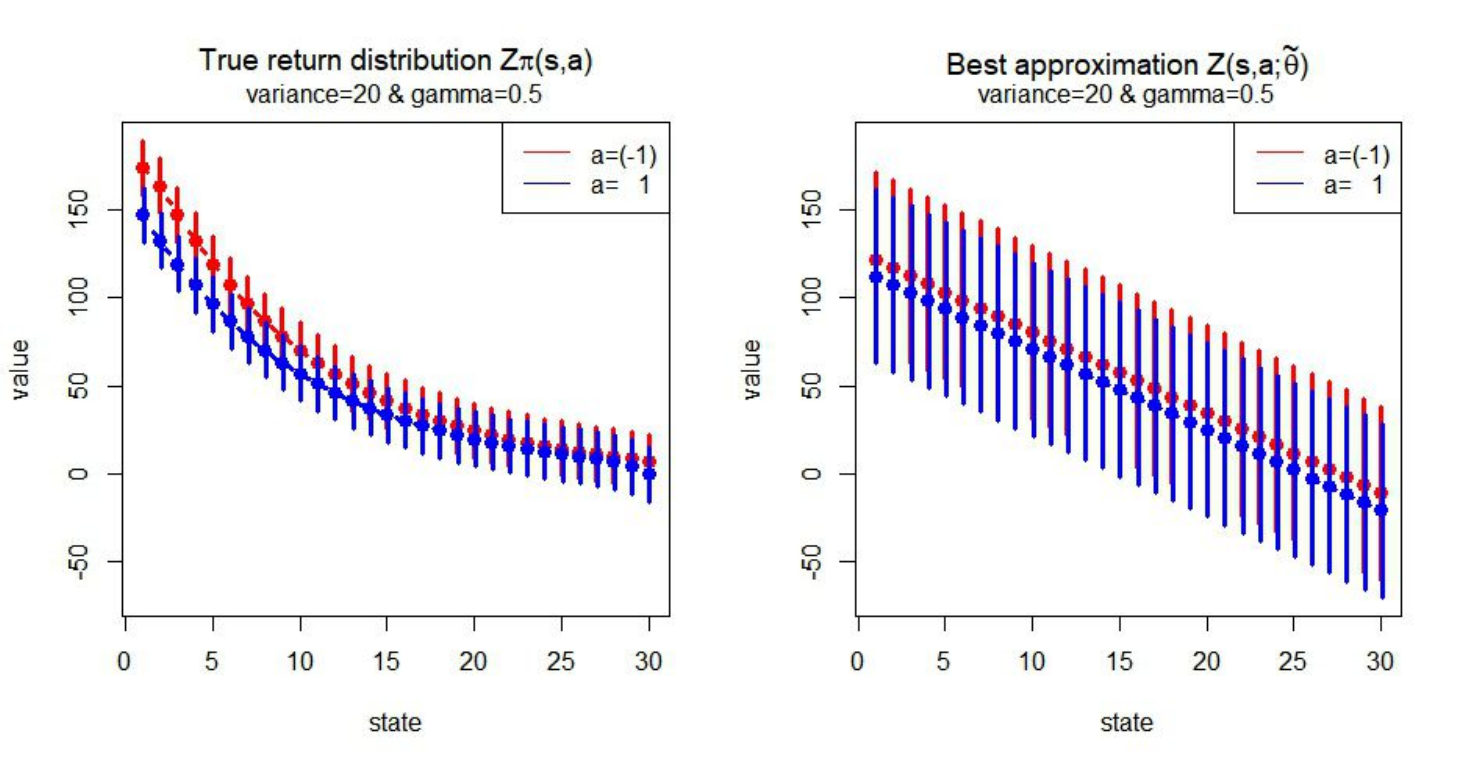}
		\includegraphics[width=0.50\linewidth]{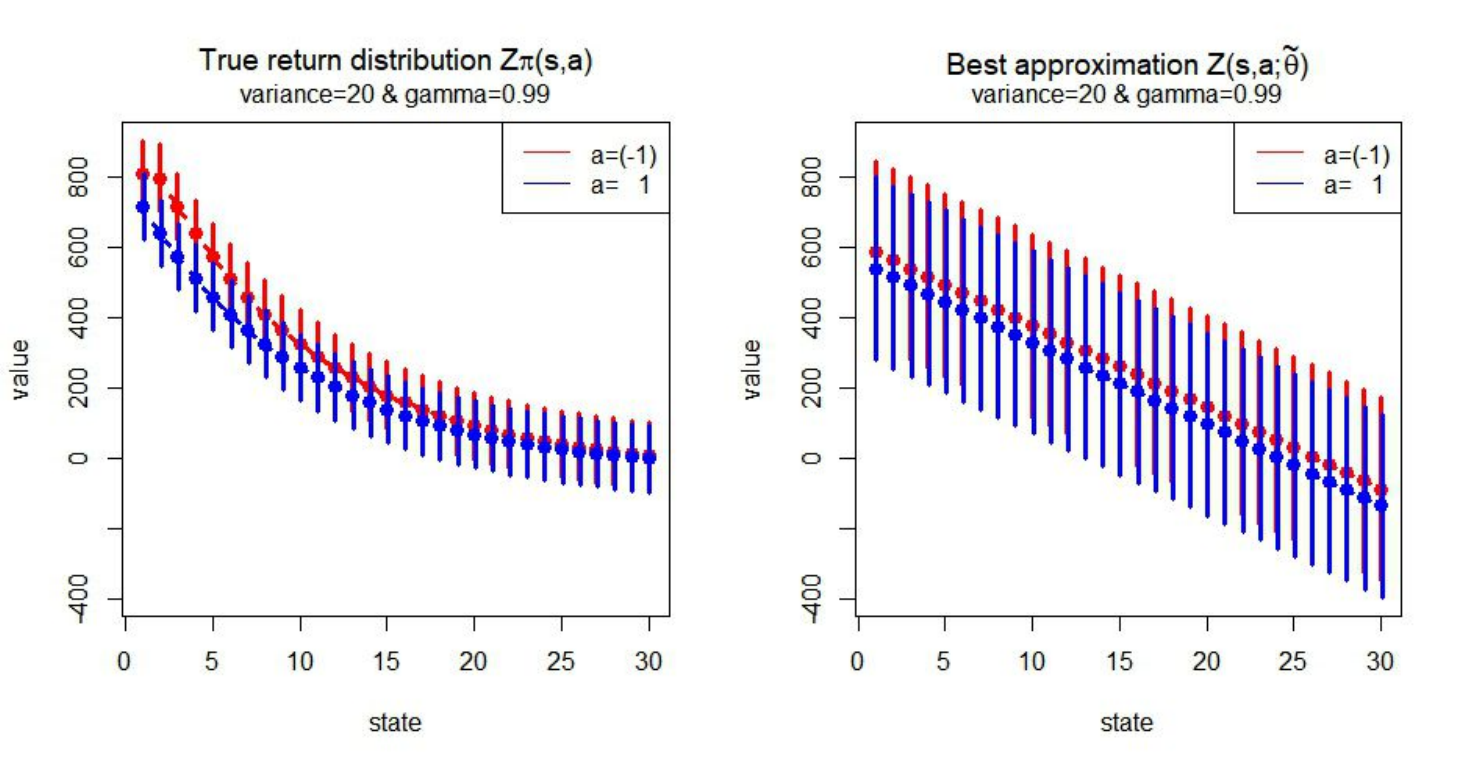}
		\caption{Red and blue represent the distributions of $Z_\pi(s,-1)$ and $Z_\pi(s,1)$, or $Z(s,-1;\tilde{\theta})$ and $Z(s,1;\tilde{\theta})$, respectively. The dots indicate the expectation values and the vertical bars include (Mean$\pm$ 3$\cdot$SD).}
		\label{true_and_pseudotrue_distributions} 
	\end{center}
	\vspace{.3in}
\end{figure}

We always assumed $\sigma_0^2=20$ and tried two different settings, $\gamma=0.50$ and $\gamma=0.99$. Denoting the parameter as $\theta=(\beta_L , \beta_R, \beta_1, \sigma^2)$ and candidate space as $\Theta=\mathbb{R} \times \mathbb{R} \times \mathbb{R}^{-} \times \mathbb{R}^{+}$, the best approximation value $\tilde{\theta}$ that minimizes the inaccuracy $\energyonebar(\Upsilon_\theta, \Upsilon_\pi)$ are calculated in Table \ref{pseudotrue_table}. They are visualized in Figure \ref{true_and_pseudotrue_distributions} (true distributions $\Upsilon_\pi$ on left and best approximations $\Upsilon_{\tilde{\theta}}$ on right). 

\begin{table}[ht]
	\caption{Best approximation values and minimum inaccuracy} \label{pseudotrue_table}
	\begin{center}
		\begin{tabular}{lccccc}
			\toprule
			\multicolumn{1}{c}{\bf Scenario}  &\multicolumn{1}{c}{ $\beta_L$} &\multicolumn{1}{c}{$\beta_R$} &\multicolumn{1}{c}{$\beta_0$} &\multicolumn{1}{c}{$\sigma^2$}  &\multicolumn{1}{c}{\bf Minimum $\energyonebar$-inaccuracy} 
			\\ \midrule 
			$\gamma=0.50$   & 126.216 & 116.614  & -4.571 & 203.099 & 13.238   \\ %
			$\gamma=0.99$   & 610.970 & 562.782 & -23.246 & 149.866 & 63.216 \\ \bottomrule  
		\end{tabular}
	\end{center}
\end{table}

\subsection{Tuning Parameters Of Each Method} \label{Tuning_parameters_in_simulations}

\subsubsection{EBRM} \label{Tuning_parameters_in_simulations_EBRM}

Energy distance \eqref{Energy_Distance} is calculated via numerical integration given the densities of the probability measures. Here is the algorithm of choosing the step level $m$, solely based on the observed data. The basic skeleton is based on SLOPE suggested by \citet{su2020adaptive} based on Lepski's rule \cite{lepskii1991problem}. Starting from large enough $m$, we can decrease it until the intersection of the confidence intervals (formed with multiple bootstraps) becomes a null set.

\begin{algorithm}
	\caption{Lepski's rule of selecting step level $m$}
	\label{Lepski_m_choosing}
	\begin{algorithmic}
		\Require $1=m_0<m_1<\cdots m_K$ \\
		\textbf{Input: } $\mathcal{D}=\{(s_i,a_i,r_i,s_i')\}_{i=1}^{N}$, $J$, $M$, $(m_1, m_2, \cdots , m_K)$ \\
		\textbf{Output: } $m_k$
		\State Estimate $\hat{\theta}$ with single-step estimation (\ref{Biased_EStimator_onestep}).
		\State $k \leftarrow K+1$, $O_{K+1} \leftarrow [-\infty, \infty]$.
		\While{$O_k \neq \varnothing$}
		\State $k\leftarrow k-1$.
		\If{$k\neq 0$}
		\For{$j=1,\cdots,J$}
		\State Estimate $\hat{\theta}_{m_k,j}^{(B)}$ with multi-step estimation (\ref{Bootstrap_objective_function}) of step level $m=m_k$.
		\State Calculate $\hat{e}_{k,j}:=\energyonebarhat(\Upsilon_{\hat{\theta}}, \Upsilon_{\hat{\theta}_{m_k,j}^{(B)}})$.
		\EndFor
		\State Calculate the sample mean ($\hat{\mu}_k$) and variance ($\hat{s}_k$) of $\hat{e}_{k,j}$ ($j=1,\cdots,J$).
		\State Calculate $I_k:=[\hat{\mu}_k \pm 1.96\cdot \hat{s}_k]$.
		\State $O_k \leftarrow O_{k+1}\cap I_k$.  
		\Else 
		\State $O_k\leftarrow \varnothing$.
		\EndIf
		\EndWhile
	\end{algorithmic}
\end{algorithm}

\begin{figure}[ht] 
	\vspace{.3in}
	\begin{center}
		\includegraphics[width=1.00\linewidth]{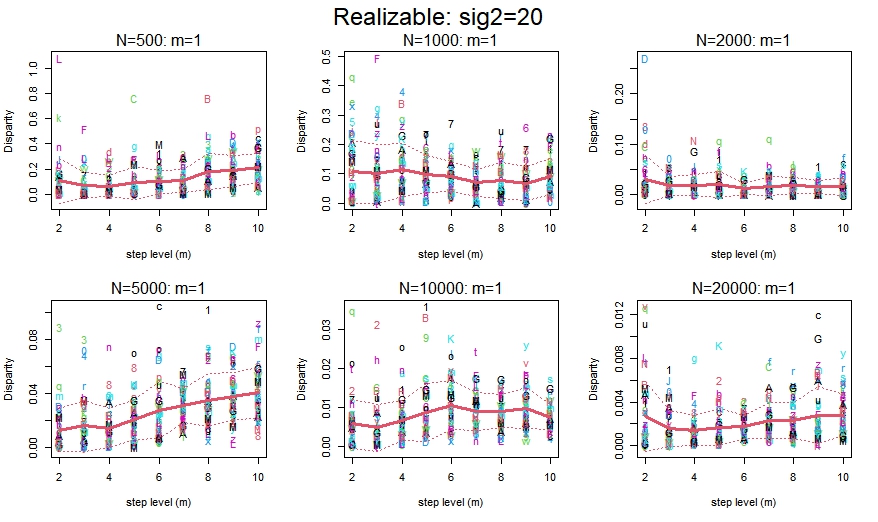}
		\includegraphics[width=1.00\linewidth]{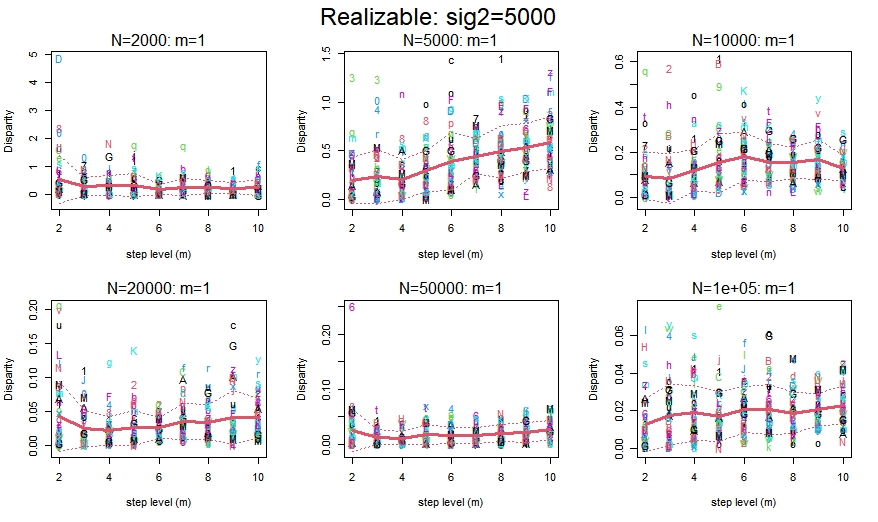}
		\caption{Realizable, $\sigma_0^2=20$ (top 6 figures) $\sigma_0^2=5000$ (bottom 6 figures).}
		\label{optimalm_real_var5000} 
	\end{center}
	\vspace{.3in}
\end{figure}

\begin{figure}[t] 
	\vspace{.3in}
	\begin{center}
		\includegraphics[width=1.00\linewidth]{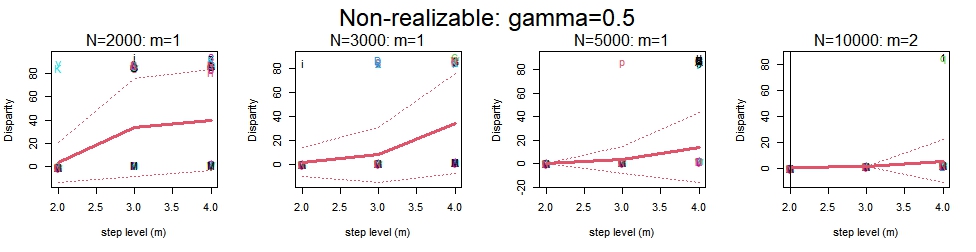}
		\includegraphics[width=1.00\linewidth]{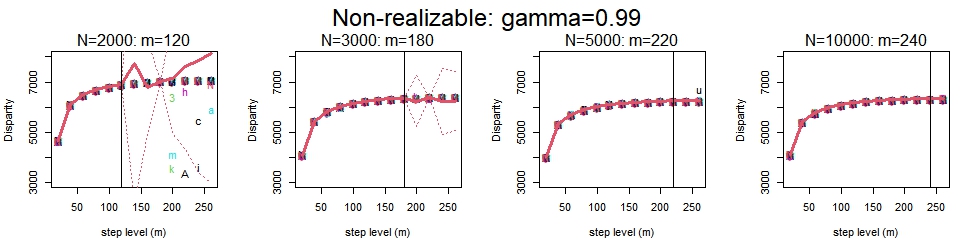}
		\caption{Non-realizable, $\gamma=0.50$ (top 4 figures) and $\gamma=0.99$ (bottom 4 figures). }
		\label{optimalm_nonreal} 
	\end{center}
	\vspace{.3in}
\end{figure}

Throughout multiple simulations in each setting (\ref{realizable_model_plot} and \ref{nonrealizable_model_plot}) for each sample size $N$ (demonstrated in \ref{energyonebar_inaccuracies}), it is rigorous to pick its own optimal step level $m$. However, since they do not differ significantly, we picked the step level $m$ via Algorithm \ref{Lepski_m_choosing} based on the first simulated data, and used the same value of $m$ throughout the remaining simulations, in order to save computational burden. We applied the algorithm with $M=N$ and $J=50$. Obviously, we had to try larger values of $(m_1,\cdots, m_K)$ for non-realizable scenario with $\gamma=0.99$ than $\gamma=0.50$. However, to avoid numerical issues in integration caused by extremely small $\gamma^m$, we limited the choice of step levels into $m\leq 4$ ($\gamma=0.50$) and $m\leq 250$ ($\gamma=0.99$). The corresponding intervals $I_k$ are visualized in Figures \ref{optimalm_real_var5000} and \ref{optimalm_nonreal}, and the selected step level $m_*$ is specified in Table \ref{optimalm_selection_table}.

\begin{table}[ht]
	\caption{Selected step level $m_*$} 
	\label{optimalm_selection_table}
	\begin{center}
		\begin{tabular}{lcccccc}
			\toprule
			\multicolumn{1}{c}{\bf Realizable}  &\multicolumn{1}{c}{$N=500$} &\multicolumn{1}{c}{$N=1000$} &\multicolumn{1}{c}{$N=2000$} &\multicolumn{1}{c}{$N=5000$}  &\multicolumn{1}{c}{$N=10000$}  &\multicolumn{1}{c}{$N=20000$} 
			\\ \midrule 
			$\sigma_0^2=20$   & 1 & 1 & 1 & 1 & 1 & 1   \\ \bottomrule
		\end{tabular}
	\end{center}
	\begin{center}
		\begin{tabular}{lcccccc}
			\toprule
			\multicolumn{1}{c}{\bf Realizable}  &\multicolumn{1}{c}{$N=2000$} &\multicolumn{1}{c}{$N=5000$} &\multicolumn{1}{c}{$N=10000$} &\multicolumn{1}{c}{$N=20000$}  &\multicolumn{1}{c}{$N=50000$}  &\multicolumn{1}{c}{$N=100000$} 
			\\ \midrule 
			$\sigma_0^2=5000$ & 1 & 1 & 1 & 1 & 1 & 1   \\ \bottomrule  
		\end{tabular}
	\end{center}
	\begin{center}
		\begin{tabular}{lcccc}
			\toprule
			\multicolumn{1}{c}{\bf Non-realizable}  &\multicolumn{1}{c}{$N=2000$} &\multicolumn{1}{c}{$N=3000$} &\multicolumn{1}{c}{$N=5000$} &\multicolumn{1}{c}{$N=10000$}
			\\ \midrule 
			$\gamma=0.50$   & 1 & 1 & 1 & 2   \\ %
			$\gamma=0.99$ & 120 & 180 & 220 & 240   \\ \bottomrule  
		\end{tabular}
	\end{center}
\end{table}

\clearpage

\subsubsection{FLE} \label{FLE_tuning_parameters_section}

\citet{wu2023distributional} did not officially suggest a rule of selecting the number of partitions $T$ and their sizes $|\mathcal{D}_1|,\cdots,|\mathcal{D}_T|$, so we utilized its asymptotic result (Corollary 4.14 by \citet{wu2023distributional}) to construct the following heuristic rule based on pre-determined values of $N_0$, $T_0$, and $l>0$,
\begin{align} %
	T(N) \buildrel let \over = \log_{\big({\frac{1}{\gamma}}\big)^{1-\frac{1}{2l}}}\bigg\{ \frac{1}{C_0}\cdot \bigg(\frac{N}{\log N}\bigg)^{\frac{1}{2l}}  \bigg\}, \quad \mbox{where } C_0>0 \text{ satisfies } T(N_0)=T_0. \nonumber
\end{align}
Note that larger value of $l$ slows down the increasing speed of $T(N)$. In addition, we prevent the number of partition $T$ from becoming too small, we put a lower bound $\tilde{T}$, so that we have $ T^*(N)=\max \{ \tilde{T} , \lfloor T(N) \rfloor \}$.
Then each partition has $|\mathcal{D}_t|=\lfloor N/T^*(N)\rfloor$, but may have some remaining observations when $N$ is not divisible by the chosen $T$. In this case, we included all the remaining observations into the last partition $\mathcal{D}_T$. The tuning parameters are chosen after multiple numerical experiments, and the following choice (Table \ref{FLE_tuning_parameters}) seemed to work best.

\begin{table}[ht]
	\caption{Tuning parameters of FLE}
	\label{FLE_tuning_parameters}
	\begin{center}
		\begin{tabular}{lccccc}
			\toprule
			\multicolumn{1}{c}{\bf Sample size}  &\multicolumn{1}{c}{$l$} &\multicolumn{1}{c}{$N_0$} &\multicolumn{1}{c}{$T_0$}  &\multicolumn{1}{c}{$\tilde{T}$} 
			\\ \midrule 
			realizable, $\sigma_0^2=50$         &  10  &  2000  &  25  &  15   \\  %
			realizable, $\sigma_0^2=5000$       &  10  &  20000 &  25  &  15   \\ %
			non-realizable, $\gamma=0.50$   &  0.7 &  3000  &  10  &  10   \\ %
			non-realizable, $\gamma=0.99$   &  10  &  2000  &  25  &  15 \\ \bottomrule  
		\end{tabular} 
	\end{center}
\end{table}

\subsubsection{QRTD} \label{QRTD_tuning_parameters}

QRTD \citep{dabney2018distributional} was originally designed for updating value distributions only conditioned on the state $s\in\mathcal{S}$, not the action $a\in\mathcal{A}$ (refer to their Equation (12)). However, we could readily develop it towards the action-value distributions conditioned on $s,a\in\SAspace$. That is, with $\tau_i\in[0,1]$ ($1\leq i\leq N_\tau$) being cdf values and $\theta_i(s,a)$ being the corresponding quantile value, they model the distribution to be uniform across $\theta_i(s,a)$,
\begin{align*}
	Z_\theta(s,a) =  \frac{1}{N_\tau}\sum_{i=1}^{N_\tau} \delta_{\theta_i(s,a)}.
\end{align*}
Analogously to their original version, we update it as follows, whenever a single new observation $(s,a,r,s')$ is collected,
\begin{gather} %
	\text{Sample }  i \sim \text{Unif}\big\{1,\cdots, N_\tau \big\}  \nonumber  \\
	\theta_{i}(s,a) \leftarrow \theta_i(s,a) + \alpha_0 \cdot \{ {\tau}_i - \mathbf{1}(r+\gamma z' < \theta_i(s,a)) \} \quad \text{where } z'\sim Z_\theta (s',a'), \ a'\sim\pi(\cdot|s') . \nonumber
\end{gather}
The number of quantiles is chosen to be 99, that is $\tau_i=i/100$ $(1\leq i\leq 99)$. As \citet{dabney2018distributional} mentioned, we repeat the above procedure
for multiple times within each iteration, which we let to be the same as $N_\tau=99$ times in our case. We chose $\alpha_0=5$ for $\sigma_0^2=20$ and $\alpha_0=2$ for $\sigma_0^2=5000$ of realizable cases, since they worked fine empirically.

\subsection{Simulation Results} \label{energyonebar_inaccuracies}

As was mentioned in Section \ref{Simulations}, we included three different inaccuracy measures (Tables \ref{realizable99_energy_inaccuracy}--\ref{realizable99_wassermix_inaccuracy}). Expectation-extended energy distance $\energyonebar$ is the one that we have used in our main theoretical results (Theorems \ref{Realizable_Final_Theorem} and \ref{thetaBhat_convergence_generalized}). For the sake of fairness, we also included expectation-extended \eqref{expectation_distance} Wasserstein-1 metric $\Wasserbar$, based on that Wasserstein-1 metric plays an important role in DRL literature (Table \ref{Comparison_Methods}). Lastly, we also used $\Wasserone$ to compare the marginal distributions $\Upsilon^{marginal}$ which is defined as the mixture of $\{\Upsilon(s,a): s,a\in\SAspace\}$ with weights $\{\probsa:s,a\in\SAspace\}$, which was used in Corollary 4.14 by \citet{wu2023distributional}. Of course, \cite{wu2023distributional} used $\pi_\mu(s,a)$ as weights, however we will use $\probsa$ for consistency with other inaccuracy measures. Measurement of the second and third inaccuracies using $\Wasserbar$ and $\Wasserone$ was approximated via R package \texttt{transport} using randomly generated samples.

\subsubsection{Realizable Scenario} \label{realizable_simulations}

\begin{table}[ht]
	\caption{Mean $\energyonebar$-inaccuracy (standard deviation in parenthesis) over 100 simulations under realizability ($\gamma=0.99$) for $\sigma_0^2=20$ (top) versus $\sigma_0^2=5000$ (bottom). Smallest inaccuracy values are in boldface.}
	\label{realizable99_energy_inaccuracy}
	\begin{center}
		\begin{tabular}{lccccccc}
			\toprule
			\multicolumn{1}{c}{\bf Sample size}  &\multicolumn{1}{c}{$500$} &\multicolumn{1}{c}{$1000$} &\multicolumn{1}{c}{$2000$}  &\multicolumn{1}{c}{$5000$}  &\multicolumn{1}{c}{$10000$}  &\multicolumn{1}{c}{$20000$} 
			\\ \midrule 
			EBRM   & {\bf 0.164}  &  {\bf 0.066}  & {\bf 0.046}  & {\bf 0.019}  & {\bf 0.008}  &  {\bf 0.005}   \\ 
			& (0.227)  &  (0.087)  &  (0.060)  &  (0.022)  &  (0.010)  &  (0.007)\\ \hline  
			FLE    & 17.729  &  8.802  &  5.533  &  2.385  &  1.220  &  0.761   \\ 
			& (15.438)  &  (9.175)  &  (6.448)  &  (2.883)  &  (1.618)  &  (0.888) \\ \hline  
			QRTD   & 149.338  &  64.259  &  48.679  &  46.032  &  49.402  &  49.965  \\ 
			& (25.221)  &  (23.160)  &  (34.323)  &  (30.909)  &  (34.617)  &  (31.458) \\ \bottomrule
		\end{tabular} 
	\end{center}
	\begin{center}
		\begin{tabular}{lccccccc}
			\toprule
			\multicolumn{1}{c}{\bf Sample size}  &\multicolumn{1}{c}{$2000$} &\multicolumn{1}{c}{$5000$} &\multicolumn{1}{c}{$10000$}  &\multicolumn{1}{c}{$20000$}  &\multicolumn{1}{c}{$50000$}  &\multicolumn{1}{c}{$100000$} 
			\\ \midrule 
			EBRM   & {\bf 0.728}  & {\bf 0.301}  & {\bf 0.128}  & {\bf 0.074}  & {\bf 0.028}  &  {\bf 0.018}   \\ 
			& (0.920)  &  (0.354)  &  (0.167)  &  (0.105)  &  (0.034)  &  (0.022)  \\ \hline  
			FLE    & 24.603  &  14.482  &  6.528  &  5.062  &  2.662  &  1.522   \\ 
			& (25.768)  &  (16.101)  &  (7.814)  &  (6.007)  &  (3.386)  &  (1.985) \\ \hline  
			QRTD   & 105.274  &  75.173  &  70.483  &  74.398  &  73.533  &  77.358   \\ 
			& (11.728)  &  (21.515)  &  (33.965)  &  (52.039)  &  (70.004)  &  (62.997)  \\ \bottomrule  
		\end{tabular} 
	\end{center}
\end{table}

\clearpage

\begin{table}[ht]
	\caption{Mean $\Wasserbar$-inaccuracy (standard deviation in parenthesis) over 100 simulations under realizability ($\gamma=0.99$) for $\sigma_0^2=20$ (top) versus $\sigma_0^2=5000$ (bottom). Smallest inaccuracy values are in boldface.}
	\label{realizable99_wasser_inaccuracy}
	\begin{center}
		\begin{tabular}{lccccccc}
			\toprule
			\multicolumn{1}{c}{\bf Sample size}  &\multicolumn{1}{c}{$500$} &\multicolumn{1}{c}{$1000$} &\multicolumn{1}{c}{$2000$}  &\multicolumn{1}{c}{$5000$}  &\multicolumn{1}{c}{$10000$}  &\multicolumn{1}{c}{$20000$} 
			\\ \midrule 
			EBRM   & {\bf 2.176}  & {\bf 1.523}  & {\bf 1.339}  & {\bf 0.985}  & {\bf 0.782}  & {\bf 0.706}   \\ 
			& (1.442)  & (0.864)  & (0.651)  & (0.388)  & (0.227)  & (0.171)   \\ \hline  
			FLE    & 22.912  & 15.755  & 12.374  & 8.036  & 5.694  & 4.590   \\ 
			& (12.774)  & (9.229)  & (7.843)  & (5.091)  & (3.773)  & (2.856)  \\ \hline 
			QRTD   & 105.561  & 64.290  & 56.739  & 54.397  & 57.145  & 57.953   \\ 
			& (12.418)  & (13.936)  & (23.716)  & (22.259)  & (24.314)  & (22.252) \\ \bottomrule 
		\end{tabular} 
	\end{center}
	\begin{center}
		\begin{tabular}{lccccccc}
			\toprule
			\multicolumn{1}{c}{\bf Sample size}  &\multicolumn{1}{c}{$2000$} &\multicolumn{1}{c}{$5000$} &\multicolumn{1}{c}{$10000$}  &\multicolumn{1}{c}{$20000$}  &\multicolumn{1}{c}{$50000$}  &\multicolumn{1}{c}{$100000$} 
			\\ \midrule 
			EBRM   & {\bf 21.221}  & {\bf 15.532}  & {\bf 12.371}  & {\bf 11.178}  & {\bf 9.971}  & {\bf 9.694}   \\ 
			& (10.337)  & (6.117)  & (3.595)  & (2.717)  & (1.143)  & (0.802)  \\ \hline  
			FLE    & 101.232  & 79.628  & 53.745  & 49.426  & 35.453  & 27.493  \\ 
			& (58.586)  & (46.772)  & (33.948)  & (29.198)  & (21.370)  & (16.038)   \\ \hline  
			QRTD   & 274.405  & 236.383  & 223.537  & 223.399  & 218.028  & 224.539    \\ 
			& (11.003)  & (22.376)  & (38.935)  & (63.145)  & (82.134)  & (76.002)  \\ \bottomrule  
		\end{tabular} 
	\end{center}
\end{table}

\begin{table}[ht]
	\caption{Mean $\Wasserone$-inaccuracy of marginal distributions (standard deviation in parenthesis) over 100 simulations under realizability ($\gamma=0.99$) for $\sigma_0^2=20$ (top) versus $\sigma_0^2=5000$ (bottom). Smallest inaccuracy values are in boldface.}
	\label{realizable99_wassermix_inaccuracy}
	\begin{center}
		\begin{tabular}{lccccccc}
			\toprule
			\multicolumn{1}{c}{\bf Sample size}  &\multicolumn{1}{c}{$500$} &\multicolumn{1}{c}{$1000$} &\multicolumn{1}{c}{$2000$}  &\multicolumn{1}{c}{$5000$}  &\multicolumn{1}{c}{$10000$}  &\multicolumn{1}{c}{$20000$} 
			\\ \midrule 
			EBRM   & {\bf 2.052}  & {\bf 1.406}  & {\bf 1.231}  & {\bf 0.843}  & {\bf 0.629}  & {\bf 0.508}   \\ 
			& (1.524)  & (1.005)  & (0.778)  & (0.556)  & (0.350)  & (0.255)   \\ \hline  
			FLE    & 22.835  & 15.694  & 12.328  & 8.013  & 5.596  & 4.591   \\ 
			& (12.810)  & (9.272)  & (7.859)  & (5.113)  & (3.786)  & (2.945)  \\ \hline 
			QRTD   &  97.856 &  56.694 & 47.738  & 45.764  & 49.851  & 49.877   \\ 
			& (13.235)  & (14.905)  & (25.251)  & (25.656)  & (27.463)  & (25.945) \\ \bottomrule 
		\end{tabular} 
	\end{center}
	\begin{center}
		\begin{tabular}{lccccccc}
			\toprule
			\multicolumn{1}{c}{\bf Sample size}  &\multicolumn{1}{c}{$2000$} &\multicolumn{1}{c}{$5000$} &\multicolumn{1}{c}{$10000$}  &\multicolumn{1}{c}{$20000$}  &\multicolumn{1}{c}{$50000$}  &\multicolumn{1}{c}{$100000$} 
			\\ \midrule 
			EBRM   & {\bf 18.021}  & {\bf 11.613}  & {\bf 7.528}  & {\bf 5.441 }  & {\bf 3.607 }  & {\bf 3.062}   \\ 
			& (12.227)  & (8.077)  & (5.306)  & (4.184)  & (2.487)  & (1.981)   \\ \hline  
			FLE    & 94.556  & 71.430  &  46.740 & 44.915  & 31.774  &  23.378  \\ 
			& (62.630)  & (51.205)  & (36.986)  & (31.905)  & (23.190)  & (18.076)  \\ \hline 
			QRTD   & 247.308  &  198.257 &  191.908 & 195.787  & 194.908  & 202.076   \\ 
			& (16.843)  & (29.911)  & (48.057)  & (73.533)  & (91.526)  & (86.031) \\ \bottomrule 
		\end{tabular} 
	\end{center}
\end{table}

\subsubsection{Non-realizable Scenario} \label{nonrealizable_simulations}

Now we tried non-realizable settings with the misspecified model \eqref{nonrealizability_misspecified_model} of Section \ref{nonrealizable_model_plot}, based on tuning parameters determined in Tables \ref{optimalm_selection_table} and \ref{FLE_tuning_parameters}. Since the point of this experiment is to see how each method can perform with a misspecified model. we excluded QRTD, which does not have any model assumption on the return distributions. We could see in Table \ref{nonrealizable_energy_inaccuracy} that EBRM approached the minimum possible level of Energy Distance (13.327 for $\gamma=0.50$ and 63.216 for $\gamma=0.99$) as we increased the sample size $N$. This forms contrast with FLE that even deteriorated as sample size grows, which we can supposedly attribute to huge violation of completeness that FLE is based upon. Supposedly, the malfunctioning of FLE in Tables \ref{nonrealizable_energy_inaccuracy}--\ref{nonrealizable_wassermix_inaccuracy} (under $\gamma=0.99$) can be attributed to its completeness assumption (Assumption 4.12 of FLE \citep{wu2023distributional}).

\begin{table}[t]
	\caption{Mean $\energyonebar$-inaccuracy (standard deviation in parenthesis) under non-realizability for $\gamma=0.50$ (top) VS $\gamma=0.99$ (bottom). Smallest inaccuracy values are in boldface. Minimum possible $\energyonebar$-inaccuracy values are 13.237 ($\gamma=0.50$) and 63.216 ($\gamma=0.99$).} \label{nonrealizable_energy_inaccuracy}
	\begin{center}
		\begin{tabular}{lccccc}
			\toprule
			\multicolumn{1}{c}{\bf Sample size}  &\multicolumn{1}{c}{$2000$} &\multicolumn{1}{c}{$3000$} &\multicolumn{1}{c}{$5000$}  &\multicolumn{1}{c}{$10000$} 
			\\ \midrule 
			EBRM   &  {\bf 14.323}  &  {\bf 14.306}  & {\bf 14.299}  & {\bf 13.544}   \\ 
			&  (0.209)  &  (0.152)  &  (0.128)  &  (0.065)  \\ \hline  
			FLE    &  15.199  &  15.206  &  15.171  &  15.171   \\ 
			&  (0.490)  &  (0.374)  &  (0.306)  &  (0.227) \\ \bottomrule  
		\end{tabular} 
	\end{center}
	\begin{center}
		\begin{tabular}{lccccc}
			\toprule
			\multicolumn{1}{c}{\bf Sample size}  &\multicolumn{1}{c}{$2000$} &\multicolumn{1}{c}{$3000$} &\multicolumn{1}{c}{$5000$}  &\multicolumn{1}{c}{$10000$} 
			\\ \midrule 
			EBRM   & {\bf 107.464}  & {\bf 94.525}  & {\bf 81.097}  & {\bf 70.572}   \\ 
			& (40.038)  &  (47.712)  &  (37.623)  &  (9.791)   \\ \hline  
			FLE    & 453.045  &  488.791  &  620.049  &  789.298   \\ 
			& (41.320)  &  (37.872)  &  (37.668)  &  (39.350)   \\ \bottomrule  
		\end{tabular} 
	\end{center}
\end{table}

\begin{table}[ht]
	\caption{Mean $\Wasserbar$-inaccuracy (standard deviation in parenthesis) under non-realizability for $\gamma=0.50$ (top) VS $\gamma=0.99$ (bottom). Smallest inaccuracy values are in boldface.} 
	\label{nonrealizable_wasserbar_inaccuracy}
	\begin{center}
		\begin{tabular}{lccccc}
			\toprule
			\multicolumn{1}{c}{\bf Sample size}  &\multicolumn{1}{c}{$2000$} &\multicolumn{1}{c}{$3000$} &\multicolumn{1}{c}{$5000$}  &\multicolumn{1}{c}{$10000$} 
			\\ \midrule 
			EBRM   &  19.245  &  19.232  &  19.202  &  17.589    \\ 
			&    (0.405)  &  (0.308)  &  (0.258)  &  (0.240)  \\ \hline  
			FLE    &  {\bf 15.036}  &  {\bf 15.049}  &  {\bf 15.047}  &  {\bf 15.037}     \\ 
			&    (0.392)  &  (0.330)  &  (0.256)  &  (0.197)  \\ \bottomrule  
		\end{tabular} 
	\end{center}
	\begin{center}
		\begin{tabular}{lccccc}
			\toprule
			\multicolumn{1}{c}{\bf Sample size}  &\multicolumn{1}{c}{$2000$} &\multicolumn{1}{c}{$3000$} &\multicolumn{1}{c}{$5000$}  &\multicolumn{1}{c}{$10000$} 
			\\ \midrule 
			EBRM   & {\bf 138.195}  &  {\bf 102.317}   & {\bf 89.389}   &  {\bf 82.698}     \\ 
			&  (23.291)  &  (24.515)  &  (17.663)  &  (5.661)    \\ \hline  
			FLE    & 260.564   &  279.848  &   348.004  &   437.266    \\ 
			&  (20.451)  &  (18.896)  &  (19.126)  &  (20.810)   \\ \bottomrule   
		\end{tabular} 
	\end{center}
\end{table}

\begin{table}[ht]
	\caption{Mean $\Wasserone$-inaccuracy of marginal distributions (standard deviation in parenthesis) under non-realizability for $\gamma=0.50$ (top) VS $\gamma=0.99$ (bottom). Smallest inaccuracy values are in boldface.} \label{nonrealizable_wassermix_inaccuracy}
	\begin{center}
		\begin{tabular}{lccccc}
			\toprule
			\multicolumn{1}{c}{\bf Sample size}  &\multicolumn{1}{c}{$2000$} &\multicolumn{1}{c}{$3000$} &\multicolumn{1}{c}{$5000$}  &\multicolumn{1}{c}{$10000$} 
			\\ \midrule 
			EBRM   &  14.098  &  14.088  &  14.087  &  {\bf 13.218}    \\ 
			&    (0.227)  &  (0.167)  &  (0.142)  &  (0.093)  \\ \hline  
			FLE    &  {\bf 13.954}  &  {\bf 13.977}  &  {\bf 13.968}  &  13.970     \\ 
			&    (0.359)  &  (0.292)  &  (0.234)  &  (0.184)  \\ \bottomrule  
		\end{tabular} 
	\end{center}
	\begin{center}
		\begin{tabular}{lccccc}
			\toprule
			\multicolumn{1}{c}{\bf Sample size}  &\multicolumn{1}{c}{$2000$} &\multicolumn{1}{c}{$3000$} &\multicolumn{1}{c}{$5000$}  &\multicolumn{1}{c}{$10000$} 
			\\ \midrule 
			EBRM   & {\bf 94.235   }  &  {\bf 83.430  }   & {\bf 72.561    }   &  {\bf 66.634  }     \\ 
			&  (33.942)  &  (28.586)  &  (17.119)  &  (5.605)    \\ \hline  
			FLE    & 259.938     &  279.249    &   347.143    &   436.259      \\ 
			&  (20.479)  &  (18.934)  &  (19.162)  &  (21.147)   \\ \bottomrule  
		\end{tabular} 
	\end{center}
\end{table}

\clearpage

\subsection{Ablation Studies} \label{sec:ablation}

In order to check the significant impact of choosing the hyperparameter $m$ (step-level), we did comparison between multi-step EBRM (where the step level is chosen based on Algorithm \ref{Lepski_m_choosing}) and single-step EBRM (which always uses $m=1$). Since Algorithm \ref{Lepski_m_choosing} gave us $m=1$ for most other cases (see Table \ref{optimalm_selection_table}), we will only do the comparison for non-realizable setting with $\gamma=0.99$. In Table \ref{table:ablation}, we could see big difference in performances, which thereby demonstrates the importance of choosing the step-level.

\begin{table}[ht]
	\caption{Ablation studies: comparison between multi-step EBRM and single-step EBRM. Mean $\energyonebar$-inaccuracy (standard deviation in parenthesis) under non-realizability for $\gamma=0.99$} 
	\label{table:ablation}
	\begin{center}
		\begin{tabular}{lccccc}
			\toprule
			\multicolumn{1}{c}{\bf Sample size}  &\multicolumn{1}{c}{$2000$} &\multicolumn{1}{c}{$3000$} &\multicolumn{1}{c}{$5000$}  &\multicolumn{1}{c}{$10000$} 
			\\ \midrule 
			multi-step EBRM   &  107.464  &  94.525  &  81.097  &  70.572    \\ 
			&    (40.038)  &  (47.712)  &  (37.623)  &  (9.791)  \\ \hline  
			single-step EBRM    &  6442.613  &  6289.079  &  6198.352  &  6256.774     \\ 
			&    (961.878)  &  (810.606)  &  (756.493)  &  (654.959)  \\ \bottomrule  
		\end{tabular} 
	\end{center}
\end{table}

\subsection{Convergence Under Non-completeness}\label{sec:FLE_wrongconvergence} 

As shown in Section \ref{nonrealizable_simulations}, for non-realizable setting with $\gamma=0.99$, performance of FLE was not only inaccurate, but also deteriorating as we increase the sample size. This is because FLE is based upon a strong assumption called ``completeness'' that already implies realizability. If realizability is violated, then completeness is of course violated, since completeness is stronger than realizability. For larger discount rate $\gamma$, the extent of non-realizability grows even larger due to the large quantity of return $Z_\pi(s,a):=\sum_{t\geq1} \gamma^{t-1}R_t$. This further takes a huge toll in completeness, and can lead to convergence towards a very distant target. This commonly happens for fitted-q-based or TD-based algorithms when completeness is not satisfied (e.g. \cite{kolter2011fixed}, \cite{tsitsiklis1996analysis}, \cite{munos2008finite}).

\begin{figure}[h] 
	\vspace{.3in}
	\begin{center}
		\includegraphics[width=0.90\linewidth]{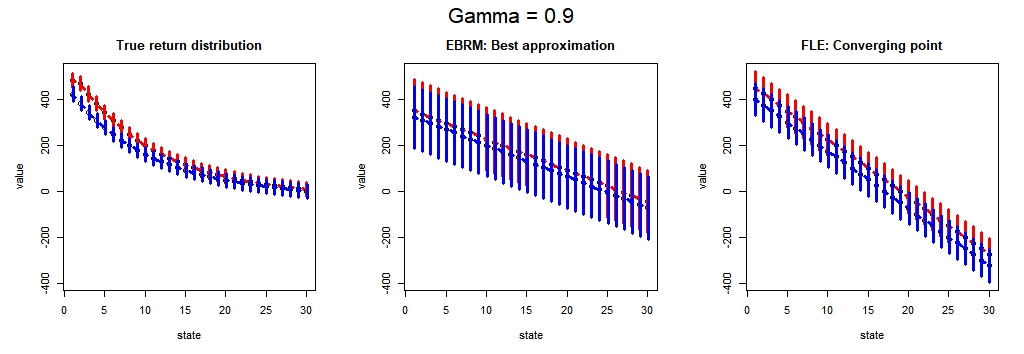}
		\includegraphics[width=0.90\linewidth]{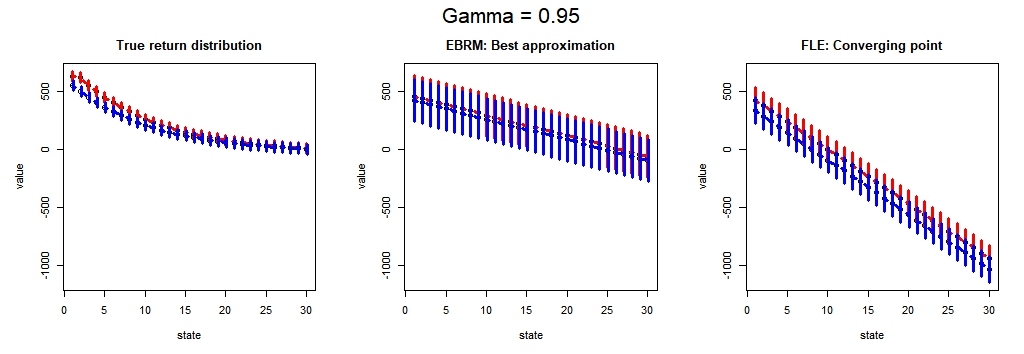}
		\includegraphics[width=0.90\linewidth]{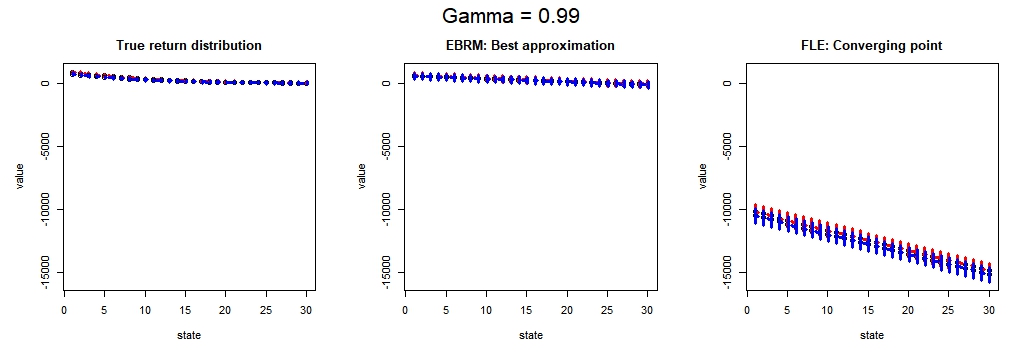}
		\caption{Red and blue represent the distributions of $Z(s,-1)$ and $Z(s,1)$, with dots indicating the expectation values and the vertical bars include (Mean$\pm$ 3$\cdot$SD). The left column represents true distributions. The middle and right columns represent the final converging distributions of EBRM and FLE.}
		\label{FLE_wrongconvergence} 
	\end{center}
	\vspace{.3in}
\end{figure}

The following are the visualizations of the distributions that EBRM and FLE finally converge to, under non-realizable modeling \eqref{nonrealizability_misspecified_model}. In the figures below, the left column shows the true distributions $Z_\pi(s,a)$. The middle shows the best $\energyonebar$-approximations achieved by $\thetatilde$ defined in \eqref{pseudotrue}, which EBRM-multi-step converges to. The right shows the final converging distributions of FLE, which is approximated by using sufficiently large amount of samples and iterations ($T=1000$ iterations, with each iteration using $10^6$ samples). We can observe that EBRM stays close to true distribution of $Z_\pi(s,a)$, whereas FLE converges towards a very distant distribution from $Z_\pi(s,a)$ as $\gamma$ grows larger (Figure \ref{FLE_wrongconvergence}).

\subsection{Computational Details} \label{computational_details}

Throughout all simulations in our paper, we only used CPU, and all simulations did not use over 1GB of memory. Average computation time for a single simulation is as follows.

\end{document}